\documentclass[final]{cvpr}

\usepackage{times}
\usepackage{epsfig}
\usepackage{graphicx}
\usepackage{amsmath}
\usepackage{amssymb}

\usepackage{cite}
\usepackage{booktabs}
\usepackage{tabulary}
\usepackage[dvipsnames]{xcolor}
\usepackage[caption=false]{subfig}
\makeatletter
\@namedef{ver@everyshi.sty}{}
\makeatother
\usepackage{tikz}
\usepackage{enumitem}

\usepackage[title]{appendix}

\definecolor{HighlightG}{HTML}{39b54a}  
\definecolor{HighlightB}{HTML}{0071bc}  
\newcommand{\hl}[1]{\textcolor{HighlightG}{#1}}
\newcommand{\gain}[1]{\fontsize{7pt}{8pt}\selectfont{\hl{\textbf{#1}}}}

\newcolumntype{x}[1]{>{\centering\arraybackslash}p{#1pt}}
\newcolumntype{y}[1]{>{\raggedright\arraybackslash}p{#1pt}}
\newcolumntype{z}[1]{>{\raggedleft\arraybackslash}p{#1pt}}
\newlength\savewidth\newcommand\shline{\noalign{\global\savewidth\arrayrulewidth
  \global\arrayrulewidth 1pt}\hline\noalign{\global\arrayrulewidth\savewidth}}
\newcommand{\tablestyle}[2]{\setlength{\tabcolsep}{#1}\renewcommand{\arraystretch}{#2}\centering\footnotesize}

\newlength{\Oldarrayrulewidth}

\newcommand{\app}{\raise.17ex\hbox{$\scriptstyle\sim$}}
\newcommand{\x}{\times}

\definecolor{demphcolor}{RGB}{100,100,100}
\newcommand{\demph}[1]{\textcolor{demphcolor}{#1}}
\newcommand{\mypm}[1]{{\tiny{{\demph{{$\pm$#1}}}}}}
\newcommand{\mynpm}[1]{{\tiny{{\textcolor{white}{{$\pm$#1}}}}}}

\newcommand{\bbox}{\mathbf{s}}
\newcommand{\feat}{\mathbf{z}}
\newcommand{\track}{\tau}
\newcommand{\shot}{c}

\makeatletter\renewcommand\paragraph{\@startsection{paragraph}{4}{\z@}
  {.5em \@plus1ex \@minus.2ex}{-.5em}{\normalfont\normalsize\bfseries}}\makeatother

\newcommand{\rbr}[1]{\left(#1\right)}
\newcommand{\sbr}[1]{\left[#1\right]}
\newcommand{\cbr}[1]{\left\{#1\right\}}

\usepackage[pagebackref=true,breaklinks=true,colorlinks,bookmarks=false]{hyperref}

\def\cvprPaperID{433} 
\def\confYear{CVPR 2021}

\begin{document}

\title{Towards Long-Form Video Understanding}

\author{
Chao-Yuan Wu \quad Philipp Kr\"ahenb\"uhl\vspace{3mm}\\
The University of Texas at Austin
}

\maketitle

\begin{abstract}
Our world offers a never-ending stream of visual stimuli,
yet today's vision systems only accurately recognize patterns within a few seconds.
These systems understand the present, but fail to contextualize it in past or future events.
In this paper, we study \emph{long-form} video understanding.
We introduce a framework for modeling long-form videos and develop evaluation protocols on large-scale datasets.
We show that existing state-of-the-art short-term models are limited for long-form tasks.
A novel object-centric transformer-based video recognition architecture performs significantly better on \textbf{7} diverse tasks.
It also outperforms comparable state-of-the-art on the AVA dataset.\vspace{-4mm}
\end{abstract}

\section{Introduction}
Our world tells an endless story of people, objects, and their interactions, each person with its own goals, desires, and intentions.
Video recognition aims to understand this story from a stream of moving pictures.
Yet, top-performing recognition models focus exclusively on short video clips, and learn primarily about the \emph{present} --- objects, places, shapes, \etc.
They fail to capture how this present connects to the \emph{past} or \emph{future}, and only snapshot a very limited version of our world's story.
They reason about the `\emph{what}', `\emph{who}', and `\emph{where}'
but struggle to connect these elements to form a full picture.
The reasons for this are two fold:
First, short-term models derived from powerful image-based architectures benefit from years of progress in static image recognition~\cite{tran2015learning,carreira2017quo}.
Second, many current video recognition tasks require little long-term temporal reasoning~\cite{ucf101,hmdb,kay2017kinetics}.

\begin{figure}[t]
\centering
\includegraphics[width=\linewidth]{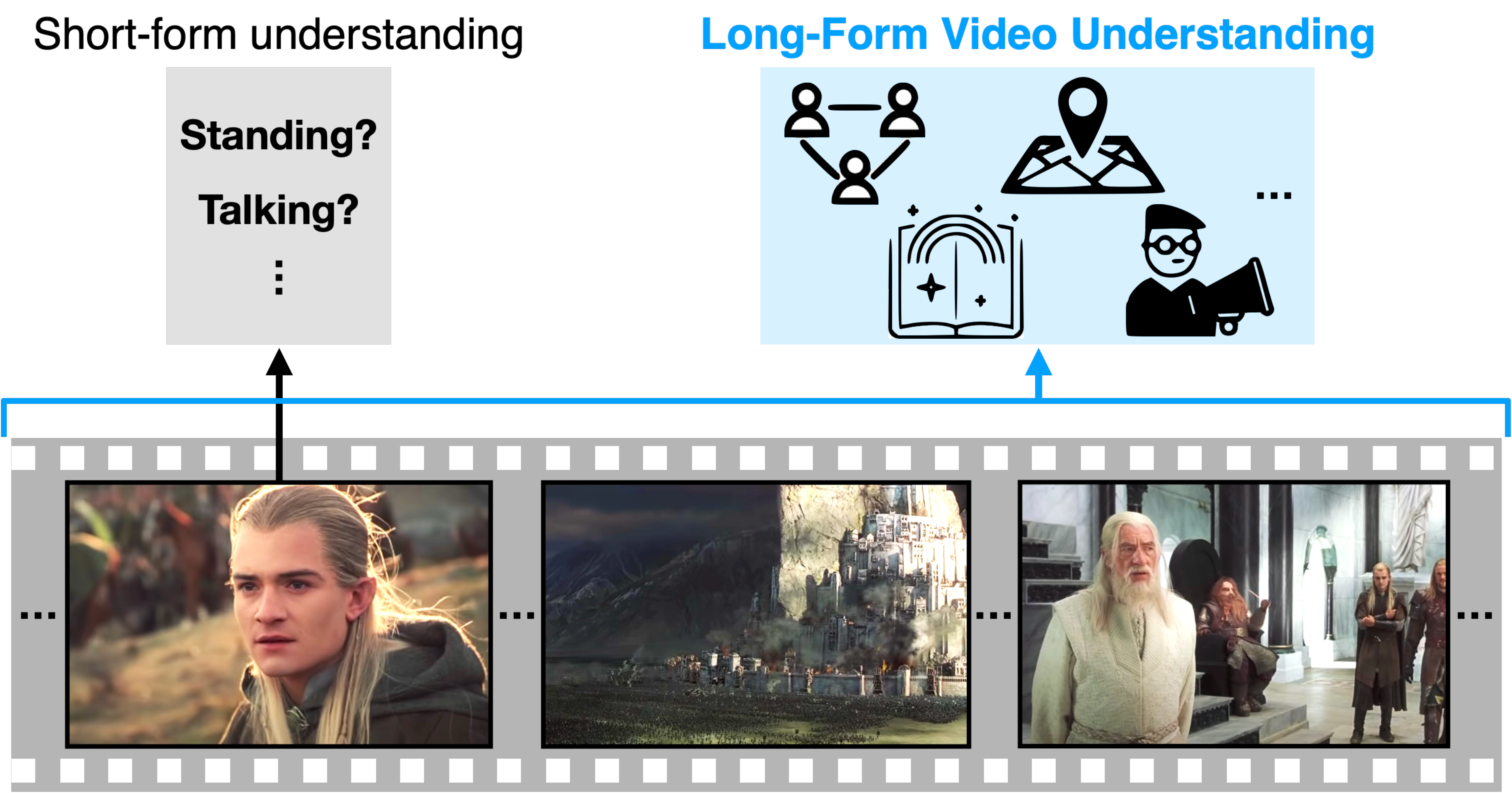}
\caption{\textbf{Long-Form Video Understanding} 
aims at 
understanding the ``full picture" of a long-form video.
Examples include understanding the storyline of a movie,
the relationships among the characters, the message conveyed by their creators, the aesthetic styles, \etc.
It is in contrast to `short-form video understanding', 
which models short-term patterns to infer local properties.\vspace{-2mm}
}
\label{fig:teaser}
\end{figure}

In this paper, we take a step towards leveling the playing field between short-term and long-term models,
and study \emph{long-form video understanding} problems (\figref{teaser}).
First, we design a novel object-centric long-term video recognition model.
Our model takes full advantage of current image-based recognition architectures to detect and track all objects, including people, throughout a video, but additionally captures the complex synergies among objects across time in a transformer-based architecture~\cite{vaswani2017attention}, called \mbox{\emph{Object Transformers}}.
Tracked instances of arbitrary length along with their visual features form basic semantic elements.
A transformer architecture then models arbitrary interactions between these elements.
This object-centric design takes inspiration from early work that
builds space-time instance representations~\cite{gorelick2007actions,yilmaz2005actions,bobick2001recognition,Efros03},
but further considers more complex inter-instance interactions over a long span of time.
The model can be trained directly for a specific end-task or pre-trained in a self-supervised fashion similar to models in image recognition~\cite{pathak2016context,noroozi2016unsupervised,moco,chen2020simple,tian2019contrastive} and language understanding~\cite{bert,roberta,Lan2020ALBERT,yang2019xlnet}.

Second, we introduce a large-scale benchmark, which comprises of 9 diverse tasks on more than 1,000 hours of video.
Tasks range from content analysis to predicting user engagement and higher-level movie metadata.
On these long-form tasks, current short-term approaches
fail to perform well,
even with strong (Kinetics-600~\cite{k600}, AVA~\cite{gu2018ava}) pre-training and various aggregation methods.

Our experiments show that \emph{Object Transformers} outperform existing state-of-the-art methods on most of the long-form tasks, and significantly outperform the current state-of-the-art on existing datasets, such as AVA 2.2.
The videos we use are publicly available and free.
Code is available at: \url{https://github.com/chaoyuaw/lvu}.

\section{Related Work}
\paragraph{Short-form video understanding}
has seen tremendous progress in both efficiency~\cite{feichtenhofer2020x3d,multigrid,tran2019video,zolfaghari2018eco} and accuracy~\cite{simonyan2014two,slowfast,carreira2017quo,nonlocal} in recent years.
Most state-of-the-art models are based on 2D or 3D CNNs
operating on short videos of less than five seconds~\cite{simonyan2014two,slowfast,multigrid,feichtenhofer2020x3d,carreira2017quo,nonlocal,tran2019video,zhou2018temporal,zolfaghari2018eco}.
A few works explore long-term patterns
for improving local pattern recognition~\cite{lfb,shvets2019leveraging},
but not long-form understanding.

\paragraph{Long-form video understanding} is less explored.
It aims to understand the full picture of a much longer video (\eg, minutes or longer).
Tapaswi~\etal~\cite{tapaswi2016movieqa} introduce a movie question answering dataset based on both text and video data.
The benchmark, however, is dominated by language-only approaches~\cite{tapaswi2016movieqa}, making it less ideal for evaluating progress of computer vision.
Vicol~\etal~\cite{vicol2018moviegraphs}, Xiong~\etal~\cite{xiong2019graph}, and Huang~\etal~\cite{huang2020movie} use vision-only movie understanding datasets,
but their videos are not publicly accessible due to copyright issues.
Bain~\etal~\cite{bain2020condensed} and Zellers~\etal~\cite{zellers2019recognition} propose joint vision-language benchmarks for text-to-video retrieval and question answering, respectively.

In this paper, we introduce a new long-form video understanding benchmark of 9 vision-only tasks on more than 30K freely accessible videos. 
Our evaluation is relatively simple compared to prior work that involves language components in evaluation protocols.

Some studies propose efficient architectures~\cite{hussein2019timeception,zhou2018temporal,karpathy2014large}
 or pooling-based methods~\cite{wang2016temporal,fernando2016rank,girdhar2017actionvlad} that may operate on long-form videos.
 These methods primarily focus on the interactions between adjacent frames, while our model captures the long-range interactions between tracked objects.

\paragraph{Representing instances}
as space-time trajectory
has a long history in computer vision~\cite{gkioxari2015finding,prest2012explicit,Efros03,gorelick2007actions}.
Our work takes inspiration from these concepts, but further considers inter-instance relationships in our methods.

\paragraph{Interaction modeling} for images
is widely studied for improving, \eg, object detection~\cite{chen2017spatial}, human action recognition~\cite{gkioxari2018detecting}, or 3D recognition~\cite{zhang2020phosa}.
For videos, a growing line of work models interactions among objects or features for improving short-term recognition~\cite{wang2018videos,baradel2018object,zhang2019structured,ma2018attend,nagarajan2020ego,tekin2019h+,girdhar2019video,ji2020action,mavroudi2020representation}.
They mainly leverage spatial but not temporal structures of a video.

\paragraph{Self-supervised learning}
drives the success of natural language processing models~\cite{bert,roberta,Lan2020ALBERT}, visual pattern learning~\cite{moco,chen2020simple,noroozi2016unsupervised,pathak2016context,jabri2020space,wang2019learning,goroshin2015unsupervised,misra2016shuffle},
and image-language joint representation learning~\cite{chen2020uniter,lu2019vilbert,tan2019lxmert,li2020unicoder,vlbert}.
Some of these methods are video-based like ours, but aim at learning robust \emph{spatial} rather than temporal features~\cite{jabri2020space,wang2019learning,goroshin2015unsupervised,misra2016shuffle}.
For example, Jabri~\etal~\cite{jabri2020space} track spatial features across frames to learn viewpoint-, scale-, or occlusion-invariant features for each instance.
Instead, our goal is to learn long-term and high-level interaction patterns.
Several other papers leverage multiple modalities for learning joint concepts~\cite{arandjelovic2017look,korbar2018cooperative,vlbert,nagrani2020speech2action,sun2019learning}.
Our method requires only visual data.
Sun~\etal~\cite{videobert} recently propose a joint language-vision model for learning long-term concepts on cooking videos.
It shares a similar goal to our approach.
The main difference is that they use a `frame-as-word', `video-as-sentence' analogy, while we build object-centric representations.
Our model captures interactions between objects, while a `frame-as-word' approach captures the interactions between adjacent video frames.
We will show the significance of this design in experiments.

\section{Preliminaries}\label{sec:prelim}
Existing short-term models parse many aspects of a video. 
They detect objects, track boxes, \etc.
This local understanding forms a useful building block for our Object Transformers.
Instead of ``re-learning" these short-term concept from scratch, our method builds on these short-term recognition modules.
We briefly review these methods and introduce notations below.

\paragraph{Action and Object Detection.}
The states and properties of humans and objects take a central role in the story told by the visual world.
In this paper, we use an action detection model~\cite{slowfast} to recognize the atomic actions~\cite{gu2018ava} of humans, and an object detector~\cite{ren2015faster} to find objects with their categories.
We denote the bounding box of a detected person or object $i$ at frame $t$ by $\bbox_{t,i} \in \mathbb{R}^4$,
and the associated feature representation by $\feat_{t,i}$.

\begin{figure}[t]
\centering
\begin{minipage}[c]{0.65\linewidth}
\includegraphics[width=\linewidth]{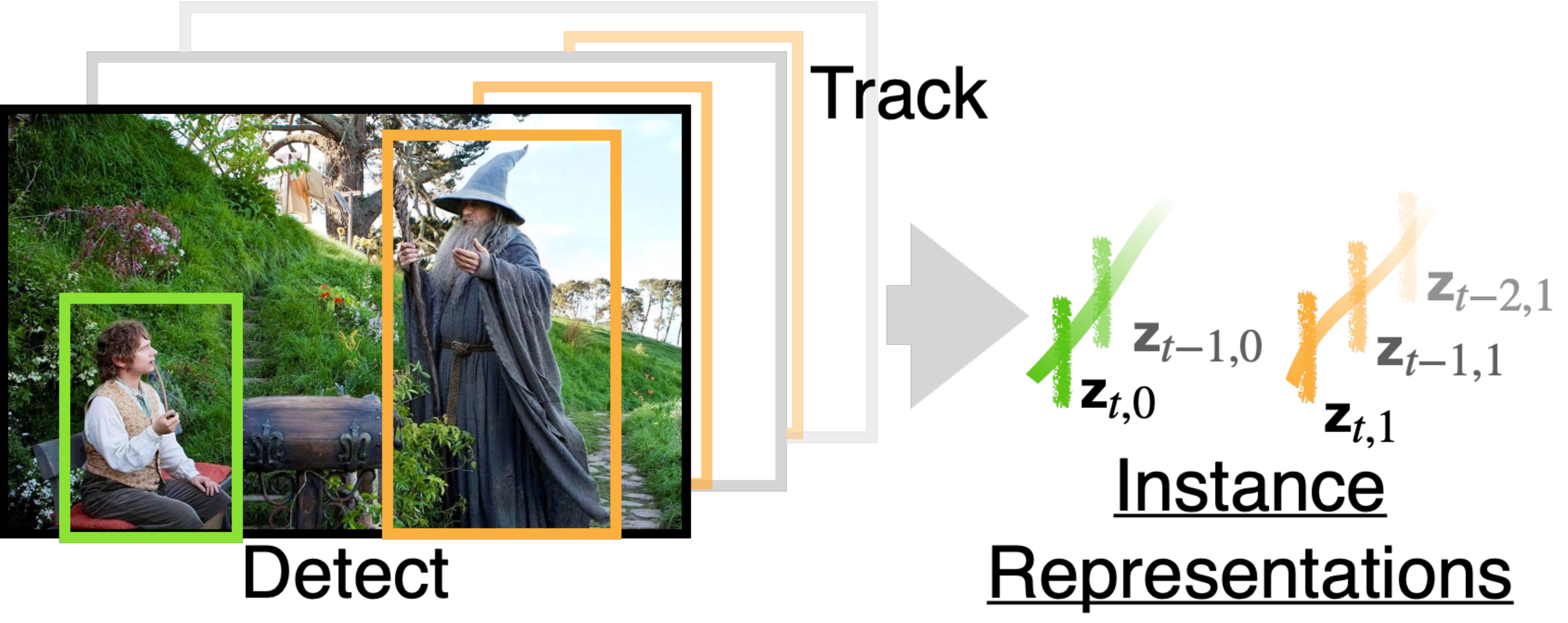}
\end{minipage}\hfill
\begin{minipage}[c]{0.312\linewidth}
\caption{We leverage short-term detection and tracking to form instance representations.
}\label{fig:instance}
\end{minipage}
\end{figure}

\begin{figure*}[t]
\vspace{-3mm}
\centering
\null\hfill
\subfloat[\textbf{Long-Form Understanding}\label{fig:method:longterm}]{%
\includegraphics[height=3.2cm]{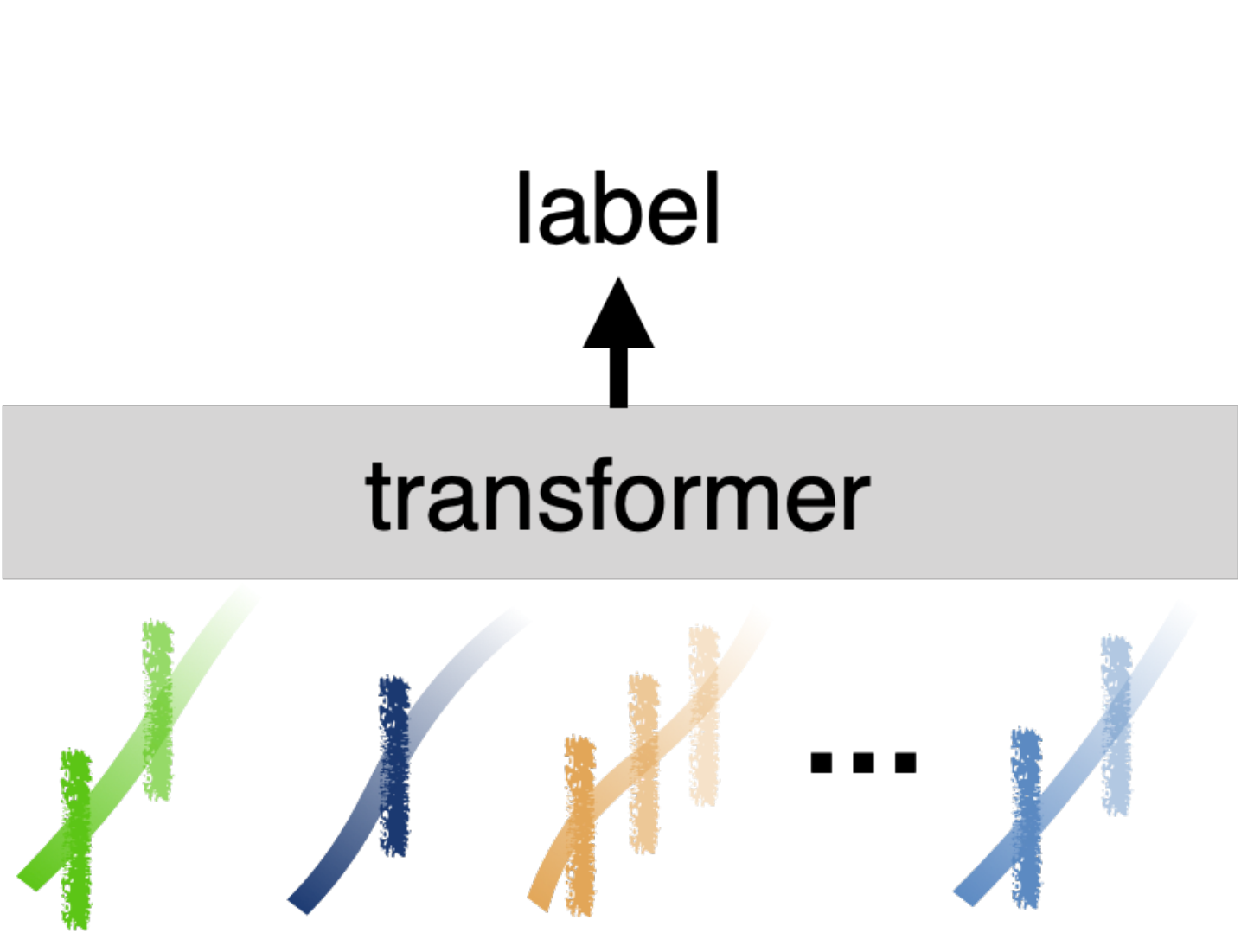}}
\hfill
\subfloat[\textbf{Masked-Instance Pre-Training}\label{fig:method:mask}]{%
\includegraphics[height=3.2cm]{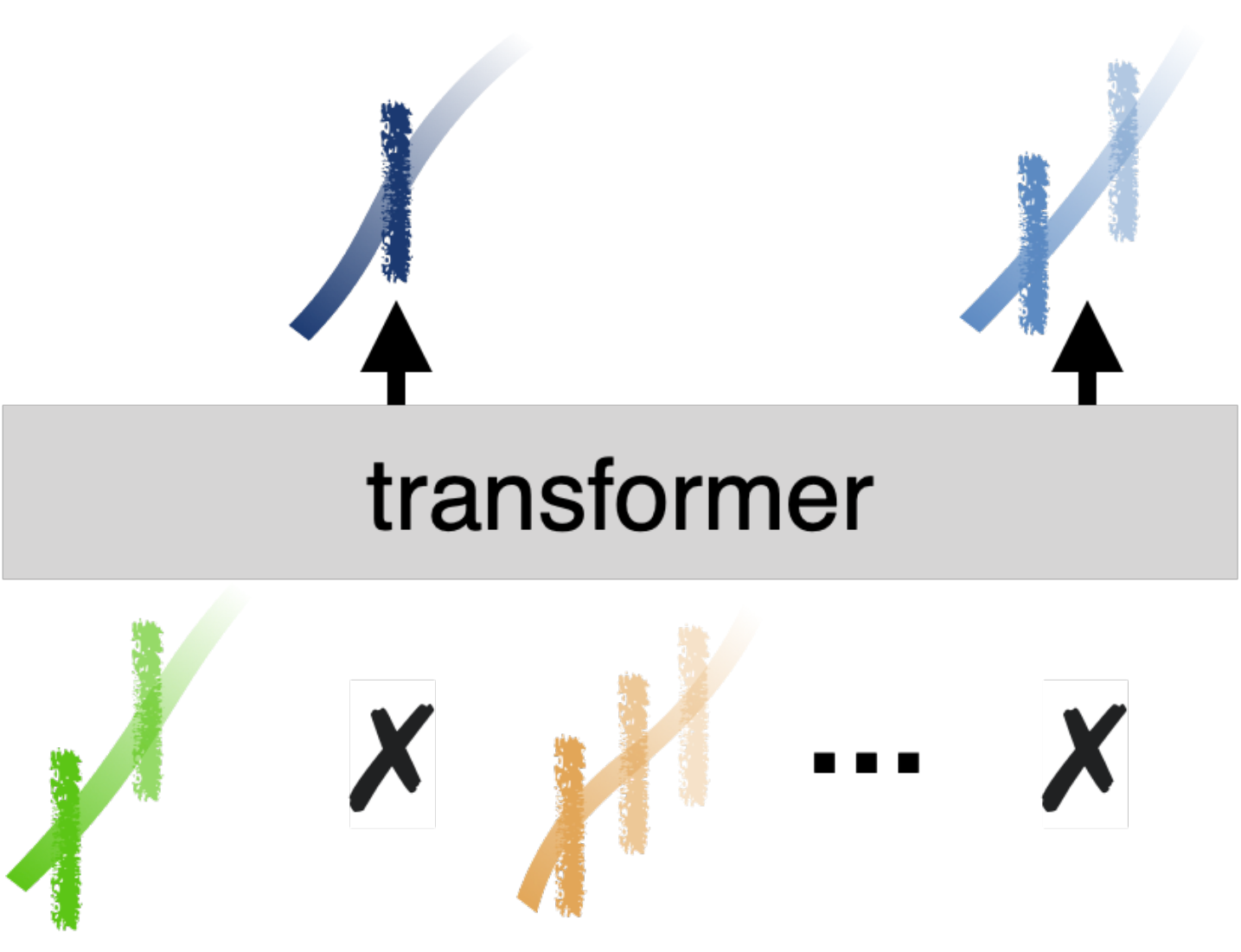}}
\hfill
\subfloat[\textbf{Compatibility Pre-Training}\label{fig:method:compat}]{%
\includegraphics[height=3.2cm]{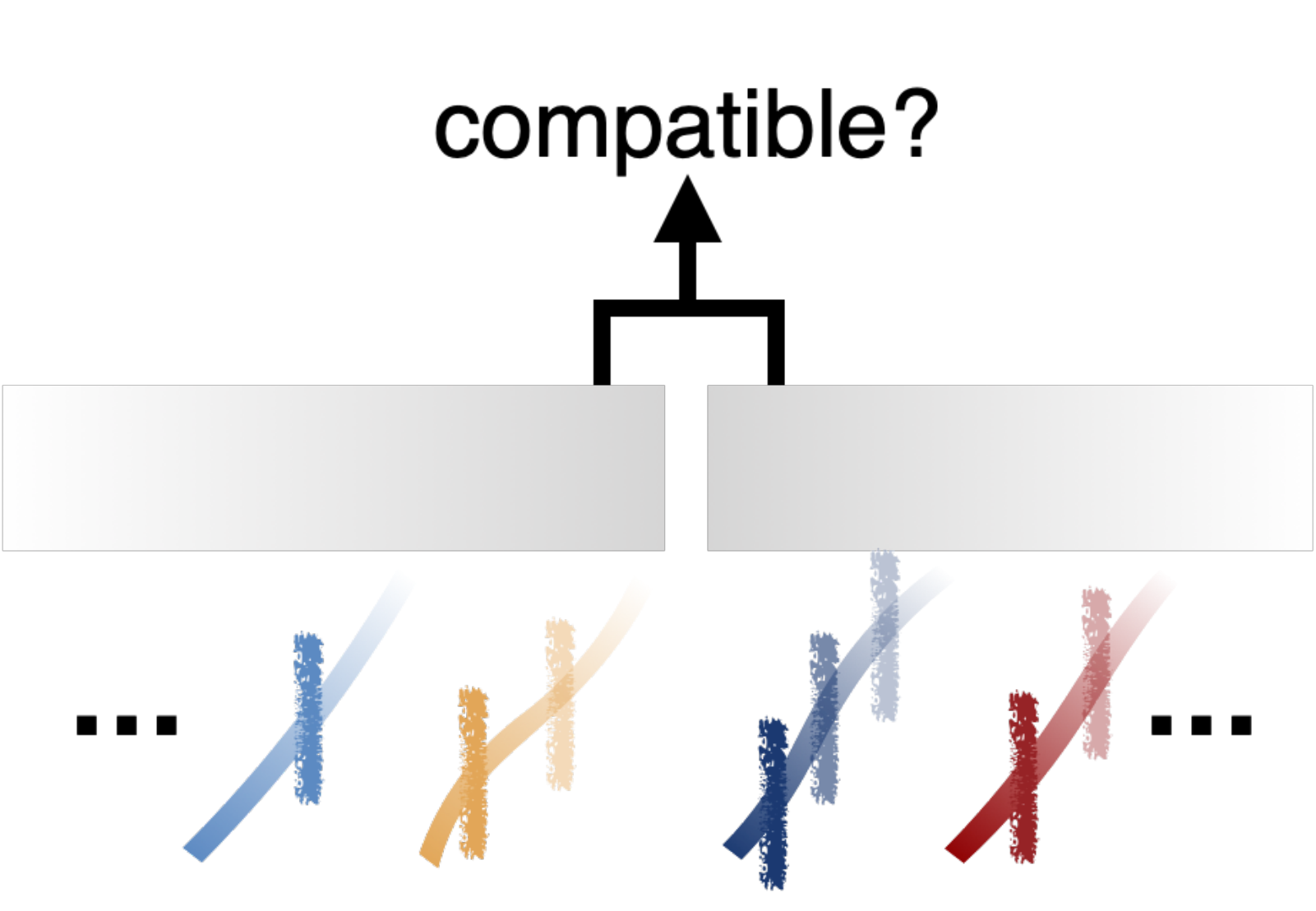}}
\hfill\null
\vspace{2mm}
\caption{\textbf{Long-Form Video Understanding with Object Transformers.}
Object Transformers take in instance-level representations and model the synergy among them for long-form video tasks (\ref{fig:method:longterm}).
To address the sparsity in supervising signals, we pre-train the model to predict the semantic representations of randomly masked instances (\ref{fig:method:mask}) and/or
predict ``compatibility'' between two videos (\ref{fig:method:compat}).
Both pre-training tasks encourage Object Transformers to learn long-term semantics, commonsense, or human social behaviors.
}
\label{fig:method}
\end{figure*}

\paragraph{Tracking.}
An instance often appear in multiple frames.
Tracking algorithms track these appearances over time and associate them to their identity~\cite{tomasi1991detection,nam2016learning}.
We use $\track_{t, i}$ to denote the associated instance index of detection $i$ at time $t$.\footnote{%
In instance segmentation literature~\cite{he2017mask}, the term `instance' often refers to one
appearance in one frame. We extend the definition and use `instance' to refer to one appearance in a space-time region, which may comprise multiple frames.
}

\paragraph{Shot Transition Detection.}
Shot transitions or ``cuts" segment a video into shots.
They form natural semantic boundaries.
A rule-based thresholding strategy typically suffices for shot transition detection~\cite{castellano2018pyscenedetect}.
$\shot_u$ denotes the shot an instance $u$ is in.

The methods above parse local properties of a video
but do not connect them to form a more complete picture of the whole video. 
We tackle this issue next.

\section{Long-Form Video Understanding}
We propose \emph{Object Transformers} for long-form video understanding.
It builds on two key ideas:
1) An object-centric design,
and 2) a self-supervised learning approach.

\subsection{Object-Centric Design}

Instead of modeling videos as width$\x$height$\x$time volume of pixels,
we take a more structured approach and model \emph{how each instance evolves in space and time} and the \emph{synergy between the instances}.

Consider a set of instances $\mathcal{U}$
(people, tables, cars, \ldots) found and tracked by short-term models (\secref{prelim}).
Each instance $u\in\mathcal{U}$
is associated with features in space-time $\cbr{\rbr{t, \bbox_{t,i}, \feat_{t,i}} \mid \track_{t,i} = u, \forall t, i}$ (\figref{instance}),
where $t$, $\bbox_{t,i}$, $\feat_{t,i}$, and $\track_{t,i}$ denote
the time stamp, spatial locations, short-term features, and the tracked identity, respectively.

We build a transformer-based architecture~\cite{vaswani2017attention} to model both how each instance $u \in \mathcal{U}$ evolves and interacts with other instances.
The transformer takes a set of representation vectors as input.
In our case, each vector correspond to a box-level representation together with its position, link and shot information.
Namely, for each $(t', \bbox', \feat')$ associated with $u$, we construct one input vector
\begin{align}
\mathbf{y}' :=& \mathbf{W}^{(\mathrm{feat})} \feat' + \mathbf{W}^{(\mathrm{spatial})} \bbox' + \mathbf{E}^{(\mathrm{temporal})}_{t'}\nonumber\\
& + \mathbf{E}^{(\mathrm{instance})}_u + \mathbf{E}^{(\mathrm{shot})}_{\shot_u} + \mathbf{b},
\end{align}
where the matrices $\mathbf{W}^{(\mathrm{feat})}$ and $\mathbf{W}^{(\mathrm{spatial})}$
project $\feat'$ and $\bbox'$ into a shared 768-dimensional vector space, and
$\mathbf{b}$ is a bias term.
$\mathbf{E}^{(\mathrm{temporal})}$ and $\mathbf{E}^{(\mathrm{shot})}$ are
position embeddings~\cite{bert} indexed by `time stamp' and `shot index', respectively.
We additionally add a learned instance-level embedding vector $\mathbf{E}^{(\mathrm{instance})}$ so that the model knows what inputs belong to the same instance.
However, learning instance-specific embeddings cannot generalize to new videos with unseen instances.
We thus randomly assign instance indices at each forward pass.
This encourages the model to leverage only ``instance distinctiveness" rather than memorizing instance-specific information.
The exact model specification is given in the Appendix.

We use a learned vector $\mathbf{E}^{\mathrm{[CLS]}}$ to be the first token of each example (similar to the ``[CLS]" special token in Devlin~\etal~\cite{bert}), and use the output vector corresponding to that position, $\mathbf{v}^{\mathrm{[CLS]}}$, as the video-level representation.
We use a linear output head $h^\mathrm{(task)}(\mathbf{v}^{\mathrm{[CLS]}})$ to perform each video-level end-task.
\figref{method:longterm} illustrates our model.
In \secref{pretrain}, we will introduce additional output heads, $h^\mathrm{(mask)}$ and $h^\mathrm{(compat)}$
along with the associated loss functions $\ell^{(\mathrm{mask})}$ and $\ell^{(\mathrm{compat})}$ for pre-training object transformers in a self-supervised manner.

\paragraph{Discussion: Object-Centric \vs Frame-Centric \vs Pixel-Volume.}
Most existing methods either view a video as a list of 2D images (\eg, \cite{karpathy2014large,videobert}) or a width$\x$height$\x$time pixel volume (\eg, \cite{tran2015learning,carreira2017quo,slowfast}).
While these views are convenient, we argue that they are unnatural ways to look at the signals, possibly leading to difficulties in learning.
After all, a video frame is simply a projection of (constantly changing) objects and scenes in a 3D world snapshotted at a particular point of time.
Modeling videos through modeling the interactions among a list of 2D images likely suffers from model misspecification, because the projected 2D images do not interact with each other ---
It is the objects in our 3D world that interact with each other.
Object Transformers directly model these interactions.

Modeling a video as a width$\x$height$\x$time pixel volume~\cite{tran2015learning} amplifies the problem even more, especially for long videos,
because the pixel volume is simply a stack of the 2D projections,
with arbitrary camera positions.
The semantic of the viewed world is however, invariant to these artifacts introduced by observers.
The pixel-volume view ignores this invariance, and thus likely hurts data efficiency, let alone the prohibitive cost to scale 3D CNNs to long-form videos.
Object Transformers leverage tracking and avoid these issues.
In \secref{exp}, we will empirically demonstrate the advantage of the object-centric design over existing frame-list or pixel-volume designs.

\subsection{Self-Supervision}\label{sec:pretrain}
Long-form video understanding also brings challenges in supervision.
Intuitively, long-form videos could enjoy less `supervising signal' per pixel, given its potentially larger number of pixels per annotation.
In \secref{exp}, we will see that a long-form video model trained from scratch indeed suffers generalization challenges in many tasks.
One way to alleviate the supervision issue is to first pre-train our model  in a self-supervised fashion on unlabeled videos, before fine-tuning on the end-tasks.\footnote{%
These pre-training methods are self-supervised, because they do not require additional annotations to perform.
Our full approach is not self-supervised, because the short-term features are learned from labeled data.
}
We present two pre-training strategies below.
\paragraph{1) Masked-Instance Prediction.}
One natural choice of the pretext task is a masked instance prediction task,
similar to Context Encoders~\cite{pathak2016context} or BERT~\cite{bert}.
Namely, we mask out features of randomly selected instances $\mathcal{M} \subset \mathcal{U}$,
and train our Object Transformers to predict the ``semantics" (\eg, object categories, person actions) of the masked instances.
Note that we mask out only the feature vector $\feat$, but retain time stamp $t$, spatial position $\bbox$, instance embedding $\mathbf{E}^{(\mathrm{instance})}$, and shot embedding $\mathbf{E}^{(\mathrm{shot})}$ for specifying `where in a video' to predict.
Following standard practice~\cite{bert}, masked feature $\feat$ is replaced by a learned embedding $\feat^{(\mathrm{mask})}$ with 80\% probability, replaced by a randomly sampled feature with 10\% probability, and stays unchanged with the remaining 10\% probability.
At each output position corresponding to the masked inputs,
we use a output head $h^\mathrm{(mask)}$ to predict a probability vector $\hat{\mathbf{p}} \in \Delta^{d-1}$ that regresses the pseudo-label\footnote{%
We infer pseudo-labels using the same short-term model that is used to compute the feature vector $\feat$.}
$\mathbf{p} \in \Delta^{d-1}$ using distillation loss~\cite{hinton2015distilling} (with temperature $T=1$),
\begin{align}
\ell^{(\mathrm{mask})}\rbr{\mathbf{p}, \hat{\mathbf{p}}} := \sum_{k=0}^{d-1} -\mathbf{p}_{k}\log\rbr{\hat{\mathbf{p}}_{k}}.
\end{align}

\figref{method:mask} presents a visual illustration.
Intuitively, this task asks `\emph{What object might it be?}' or `\emph{What a person might be doing?}' in the masked regions, given the context.
We humans can perform this task very well, given our social knowledge and commonsense.
We train Object Transformers to do the same.

\paragraph{Discussion: Masked-Instance Prediction \vs Masked-Frame Prediction.}
Our method share similar spirits with Sun~\etal~\cite{videobert}'s `Masked-Frame Prediction' pretext task (if removing their language components),
but with important distinctions.
`Masked-Frame Prediction' is in fact quite easy, as linear interpolation using the 2 adjacent frames already provides
a strong solution in most cases due to continuity of physics.
This observation is consistent with the observations in Sun~\etal~\cite{videobert} that such pre-training method is data hungry and that their `visual-only' variant (without using language) is not effective.
Masked-Instance Prediction, on the other hand, does not suffer from the trivial solution.
It directly models how objects interact with each other,
rather than learning to interpolate in the projected 2D space.

\paragraph{Discussion: Masked-Instance Prediction \vs Spatial-Feature Learning Methods.}
Also note that our goal of pre-training is different from the goal of most prior work on
self-supervised method on videos~\cite{arandjelovic2017look,korbar2018cooperative,jabri2020space,wang2019learning,goroshin2015unsupervised,misra2016shuffle}.
They typically involve tracking an object or an interest point over time to learn (\eg, view-point, scale, occlusion, lighting) invariance on one instance.
Their goal is learning robust \emph{spatial} representations.
In this paper, we aim at learning longer-term patterns in videos.

\paragraph{2) Span Compatibility Prediction.}
The second pre-training pretext task we use is to classify whether two spans of video are ``\emph{compatible}''.
For example, we may define two spans to be compatible when they come from the same scene
or happen one after another.
To solve this task, the model is encouraged to learn high-level semantics concepts, \eg,
`\emph{wedding ceremony}' should be more compatible with `\emph{party}' and `\emph{dinner}'
than `\emph{camping}' or `\emph{wrestling}'.
\figref{method:compat} illustrates this method.
We use an output head $h^\mathrm{(compat)}$ to obtain $\mathbf{v} = h^\mathrm{(compat)}\rbr{\mathbf{v}^{\mathrm{[CLS]}}}$
and use the InfoNCE loss~\cite{oord2018representation} for compatibility training:
\begin{align}
\ell^{(\mathrm{compat})}\rbr{\mathbf{v}, \mathbf{v}^+, \mathbf{v}^-} = -\log\frac{e^{\rbr{\mathbf{v} \cdot \mathbf{v}^+}}}{e^{\rbr{\mathbf{v} \cdot \mathbf{v}^+}} + \sum_{n=0}^{N-1}e^{\rbr{\mathbf{v} \cdot \mathbf{v}^{-}_n}}},
\end{align}
where $\mathbf{v}^+$ and $\mathbf{v}^{-}_n$ correspond to the spans compatible and incompatible with $\mathbf{v}$, respectively.

\paragraph{Discussion: Comparison to Next-Sentence Prediction.}
Compatibility prediction is a modified version of the ``next-sentence prediction'' task commonly used in NLP~\cite{bert}.
One distinction is that while languages typically have strict grammar and rich structures,
videos are more flexible in structure.
For example, an event of `dinner' can take place with arbitrary number of people for arbitrarily long
and potentially presented in multiple shots in a video.
We thus relax the requirement of predicting immediate adjacency,
and enforce a more relaxed ``compatibility" objective.
We will describe our exact instantiation in \secref{implement}.

\newcommand{\imgwithbox}[2]{%
\begin{minipage}{\linewidth}
\begin{tikzpicture}
    \node (img) {\includegraphics[width=\linewidth]{#2}};
    \node [above left=1.32mm, fill=white, opacity=0.7] at (img.south east){\color{black}{#1}};
\end{tikzpicture}\vspace{-1.5mm}\end{minipage}}
\begin{figure*}[t]
\vspace{-4.3mm}
\null\hfill
\subfloat[\textbf{Relationship}]{%
\tablestyle{0.0pt}{0.0}
\begin{tabular}{@{}x{90}@{}}
\imgwithbox{wife \& husband}{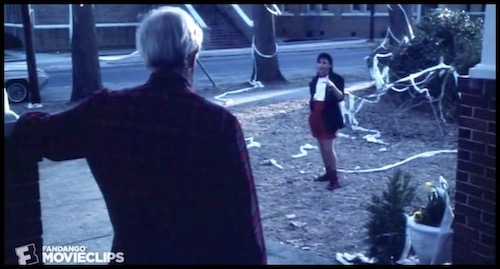}\\
\imgwithbox{friends}{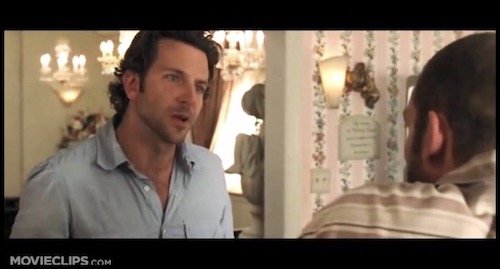}\\
\imgwithbox{boyfriend \& girlfriend}{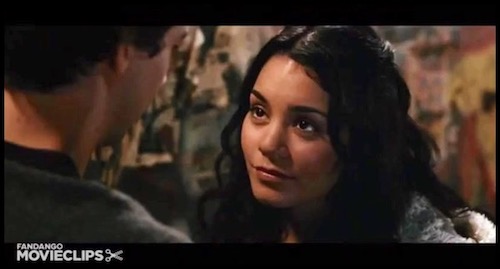}
\end{tabular}}
\hfill
\subfloat[\textbf{Way of Speaking}]{%
\tablestyle{0.0pt}{0.0}
\begin{tabular}{@{}x{90}@{}}
\imgwithbox{confront}{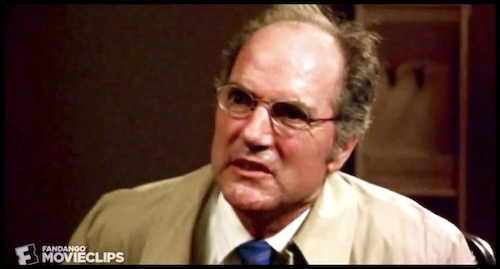}\\
\imgwithbox{explain}{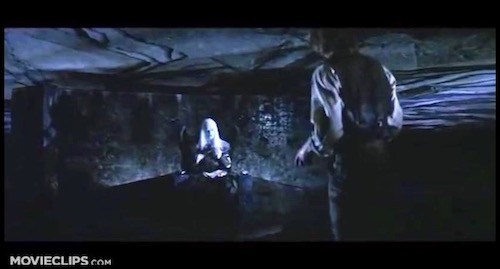}\\
\imgwithbox{discuss}{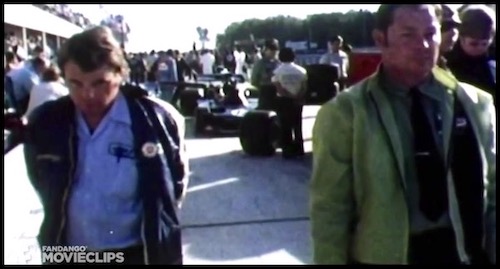}
\end{tabular}}
\hfill
\subfloat[\textbf{Scene/Place}]{%
\tablestyle{0.0pt}{0.0}
\begin{tabular}{@{}x{90}@{}}
\imgwithbox{airport}{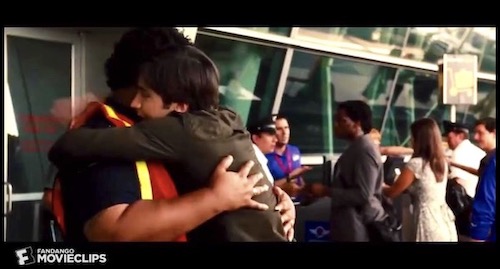}\\
\imgwithbox{school}{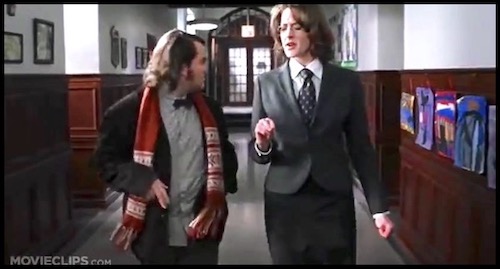}\\
\imgwithbox{office}{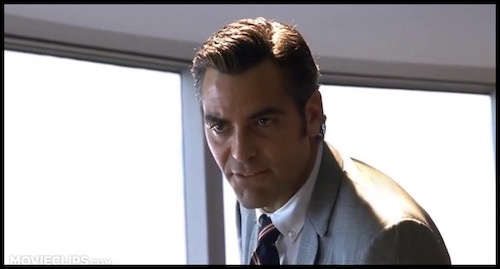}
\end{tabular}}
\hfill
\subfloat[\textbf{`Like' ratio}]{%
\tablestyle{0.0pt}{0.0}
\begin{tabular}{@{}x{90}@{}}
\imgwithbox{98.5\%}{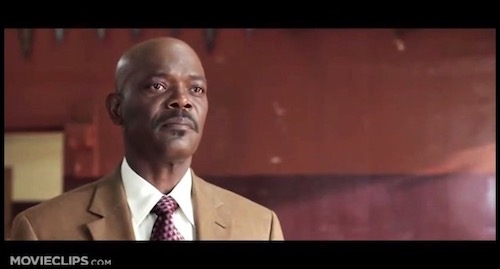}\\
\imgwithbox{84.9\%}{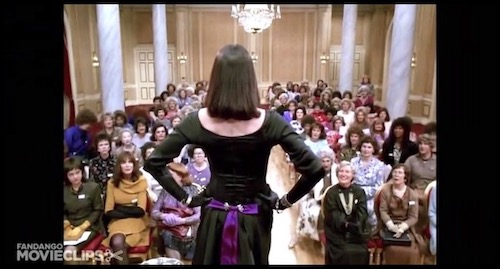}\\
\imgwithbox{73.3\%}{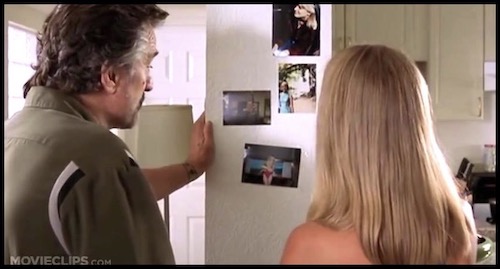}
\end{tabular}}
\hfill
\subfloat[\textbf{View Count}]{%
\tablestyle{0.0pt}{0.0}
\begin{tabular}{@{}x{90}@{}}
\imgwithbox{1.4K}{{figs/tasks/jpegs/view_1.4k}.jpg}\\
\imgwithbox{104.5K}{{figs/tasks/jpegs/view_104.5k}.jpg}\\
\imgwithbox{3.6M}{{figs/tasks/jpegs/view_3.6m}.jpg}
\end{tabular}}
\hfill\null
\vspace{-3mm}\\
\null\hfill
\subfloat[\textbf{Director}]{%
\tablestyle{0.0pt}{0.0}
\begin{tabular}{@{}x{90}@{}}
\imgwithbox{Quentin Tarantino}{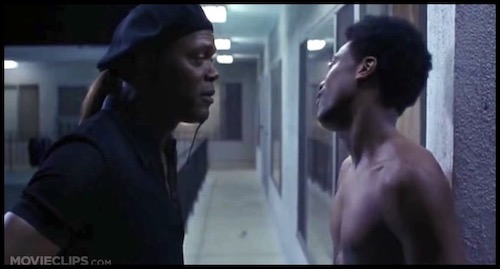}\\
\imgwithbox{Ron Howard}{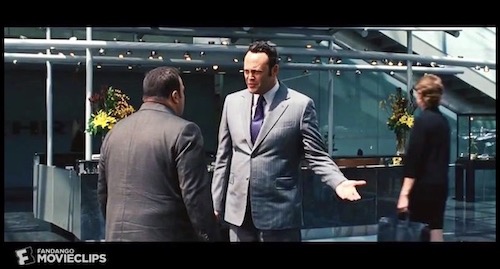}\\
\imgwithbox{Peter Jackson}{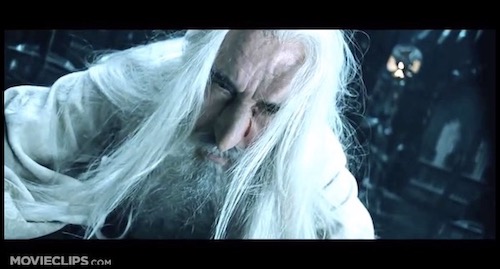}
\end{tabular}}
\hspace{0.8mm}
\subfloat[\textbf{Genre}]{%
\tablestyle{0.0pt}{0.0}
\begin{tabular}{@{}x{90}@{}}
\imgwithbox{romance}{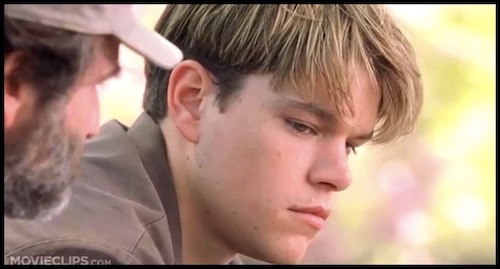}\\
\imgwithbox{horror}{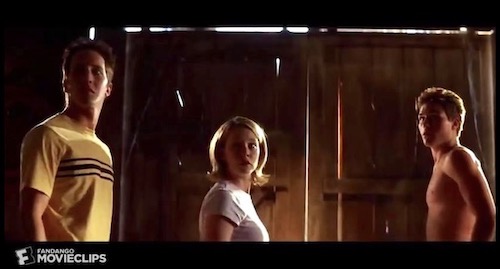}\\
\imgwithbox{comedy}{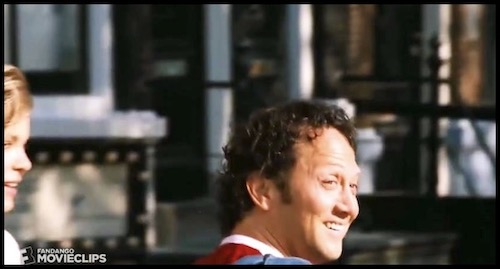}
\end{tabular}}
\hspace{0.8mm}
\subfloat[\textbf{Writer}]{%
\tablestyle{0.0pt}{0.0}
\begin{tabular}{@{}x{90}@{}}
\imgwithbox{John Hughes}{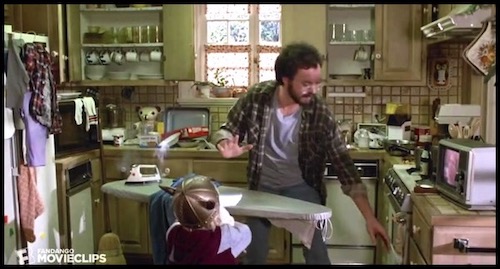}\\
\imgwithbox{Stephen King}{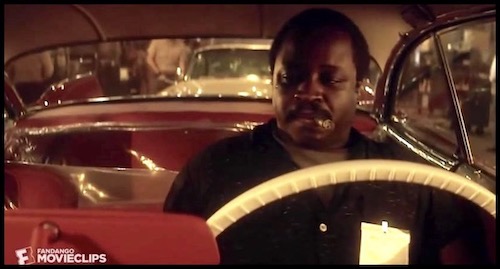}\\
\imgwithbox{David Koepp}{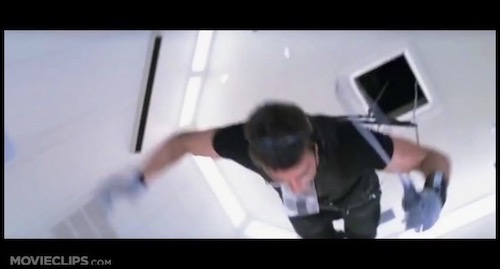}
\end{tabular}}
\hspace{0.8mm}
\subfloat[\textbf{Year}\label{fig:year}]{%
\tablestyle{0.0pt}{0.0}
\begin{tabular}{@{}x{90}@{}}
\imgwithbox{1950s}{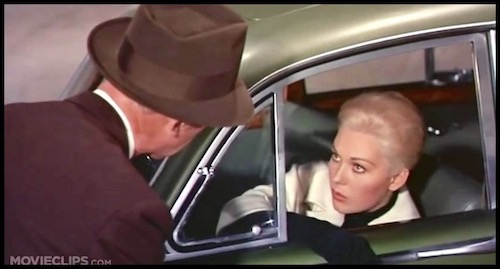}\\
\imgwithbox{1980s}{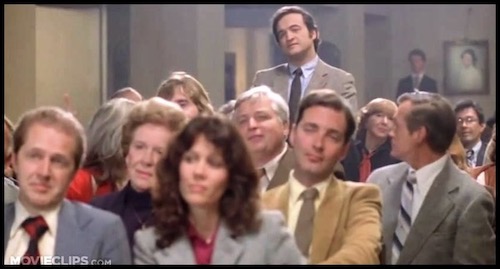}\\
\imgwithbox{2010s}{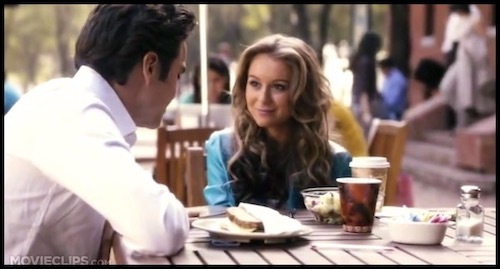}
\end{tabular}}
\hfill\null
\vspace{1mm}
\caption{\textbf{The Long-Form Video Understanding (LVU) Benchmark.}
Here we present three examples with their annotations for each task.
LVU contains a wide range of tasks for probing different aspects of video understanding research and model design.
The full list of classes for each task, more details, and more examples are available in the Appendix.
\vspace{-1mm}}\label{fig:tasks}
\end{figure*}

\subsection{Implementation Details}\label{sec:implement}
\paragraph{Instance Representations.}
We use a Faster R-CNN~\cite{ren2015faster} with ResNet-101~\cite{resnet} backbone and FPN~\cite{lin2017feature} pre-trained on COCO~\cite{coco}
to find objects other than humans.
The model obtains 42.0 box AP on COCO\@.
We use the RoIAlign~\cite{he2017mask} pooled feature vector and the end of the Faster-RCNN as feature vector $\feat$.
For person action detection, we adopt a
Faster-R-CNN-based person detector~\cite{ren2015faster} ($\app$93.9 AP@50)
commonly used in prior work~\cite{lfb,slowfast} to detect people first,
and use a ResNet-101~\cite{resnet} SlowFast network~\cite{slowfast} with non-local blocks~\cite{nonlocal} to compute RoIAlign~\cite{he2017mask}
pooled features as $\feat$ for each person box.
The model is pre-trained on AVA~\cite{gu2018ava} and achieves 29.4\% mAP on the AVA validation set.
We represent $s_{i,j}$ as the positions of the four corners $\rbr{s_{i,j}^\mathrm{(top)}, s_{i,j}^\mathrm{(bottom)}, s_{i,j}^\mathrm{(left)}, s_{i,j}^\mathrm{(right)}}$, where each of the values are normalized in $\sbr{0, 1}$.
For tracking, we adopt the algorithm described in Gu~\etal~\cite{gu2018ava}.
We use PySceneDetect~\cite{castellano2018pyscenedetect} for shot transition detection.

\paragraph{Compatibility Prediction.}
The MovieClips dataset~\cite{movieclips} we use contain (typically one-to-three-minute-long) segments of movies.
In this paper, we define two spans to be compatible if they come from the same segment.

When training with compatibility prediction, each mini-batch of size $n$ comprises $n / 2$ pairs of positive examples $(\mathbf{v}, \mathbf{v}^+)$. 
Each pair uses all other examples in the same mini-batch as negative examples $\mathbf{v}^-$.

\paragraph{Output Heads.}
Following prior work~\cite{bert},
$h^\mathrm{(mask)}$ is a 2-layer MLP\@.
$h^\mathrm{(compat)}$ and all end tasks use dropout with rate 0.1 followed by a linear layer.

\section{The Long-Form Video Understanding (LVU) Benchmark}\label{sec:benchmark}

\begin{table*}[t]
\centering
\tablestyle{1.0pt}{1.05}
\begin{tabular}{@{\extracolsep{3pt}}
l
x{30}|x{36}x{34}x{34}x{34}x{34}
x{38}x{34}x{34}x{34}
@{}}
& average & \multicolumn{3}{c}{content} & \multicolumn{2}{c}{user engagement} & \multicolumn{4}{c}{metadata}\\
\cline{3-5}  \cline{6-7}  \cline{8-11}
& rank &
relation ($\uparrow$) &
speak ($\uparrow$) &
scene ($\uparrow$) &
like ($\downarrow$) &
views ($\downarrow$) &
director ($\uparrow$) &
genre ($\uparrow$) &
writer ($\uparrow$) &
year ($\uparrow$)\\
\shline
R101-SlowFast+NL~\cite{resnet,slowfast,nonlocal}& 2.44 & 52.4\mynpm{0.0} & 35.8\mynpm{0.0} & 54.7\mynpm{0.0} & 0.386\mynpm{0.000} & 3.77\mynpm{0.00} & 44.9\mynpm{0.0} & 53.0\mynpm{0.0} & 36.3\mynpm{0.0} & \textbf{52.5}\mynpm{0.0}\\
VideoBERT~\cite{videobert} & 2.22 & 52.8\mypm{1.0} & 37.9\mypm{0.9} & 54.9\mypm{1.0} & 0.320\mypm{0.016} & 4.46\mypm{0.07} & 47.3\mypm{1.7} & 51.9\mypm{0.6} & \textbf{38.5}\mypm{1.1} & 36.1\mypm{1.4}\\
\textbf{Object Transformer} & \textbf{1.33} & \textbf{53.1}\mypm{1.4} & \textbf{39.4}\mypm{1.2} & \textbf{56.9}\mypm{1.0} & \textbf{0.230}\mypm{0.005} & \textbf{3.55}\mypm{0.05} & \textbf{51.2}\mypm{0.8} & \textbf{54.6}\mypm{0.6} & 34.5\mypm{0.9} & 39.1\mypm{1.2}\\
\end{tabular}
\vspace{2mm}
\caption{\textbf{Comparison to Prior Work.} 
Our Object Transformer outperforms both baselines by a clear margin in terms of the overall ranking.
The results support that modeling the synergy across people and objects is important for understanding a long-form video.
Interestingly, short-term models suffice to work well for \emph{year prediction},
which matches our expectation, since the year can often be recognized through solely the picture resolution/quality (\figref{year}).
{
We report the average over 5 runs with standard error for VideoBERT and Object Transformer.}\vspace{-2mm}
}\label{tab:mainresults}
\end{table*}

We introduce a new benchmark that contains 9 tasks for evaluating long-form video understanding.
The benchmark is constructed on the publicly available MovieClips dataset~\cite{movieclips}, which contains $\app$30K videos from $\app$3K movies.\footnote{Videos are accessed on February 26, 2020. Animations are excluded.
Outros are removed for all videos.}
We resize all videos such that their height is 480 pixels.
Each video is typically one-to-three-minute long.

\paragraph{Tasks.}
Our tasks cover a wide range of aspects of long-form videos, including \textbf{content understanding}
(\emph{`relationship'},
\emph{`speaking style'},
\emph{`scene/place'}),
\textbf{user engagement prediction}
(\emph{`YouTube like ratio'},
\emph{`YouTube popularity'}), and
\textbf{movie metadata prediction}
(\emph{`director'},
\emph{`genre'},
\emph{`writer'},
\emph{`movie release year'}).
\figref{tasks} presents examples of each task.
For content understanding tasks, 
we parse the description associated with each video and use the most common discovered categories (\eg, `friends', `wife \& husband', \etc.) to form a task (\eg, `relationship' prediction).
We use YouTube statistics for user engagement prediction tasks.
For metadata prediction tasks, we obtain the metadata from the corresponding IMDb entries\footnote{\url{https://www.imdb.com/}}.
Task construction details, statistics, and more examples are available in the Appendix.

\paragraph{Evaluation Protocol.}
Content understanding and metadata prediction tasks are single-label classification tasks, evaluated by top-1 classification accuracy.
User engagement prediction tasks are single-valued regression tasks, evaluated by mean-squared-error (MSE).
Compared to existing tasks~\cite{zellers2019recognition,bain2020condensed} on this dataset,
the output space and evaluation protocol of LVU is relatively simple.
We hope this choice makes result interpretation easier.
Each task is split into 70\% for training, 15\% for validation, and 15\% for testing.
Since we predict ``movie" specific metadata for metadata prediction tasks, 
we make sure the three splits contain mutually exclusive sets of movies.
We select hyperparameters based on validation results,
and report all results on test sets.

\section{Experiments}\label{sec:exp}
\paragraph{Pre-Training Details.}
We pre-train our models on the MovieClip videos for
308,000 iterations with a batch size of 16 (2 epochs of all possible, overlapping spans) using Adam~\cite{kingma2014adam},
with a weight decay of 0.01 and a base learning rate of 10$^{-4}$.
We use linear learning rate decay and linear warm-up~\cite{resnet,goyal2017accurate} for the first 10\% of the schedule, following prior work~\cite{roberta}.
We sample 60-second video spans for training our models.\footnote{%
In preliminary experiments, we do not see advantages with a longer training schedule or using spans longer than 60 seconds.}
Since each example contains a different number of instances of different lengths, we perform attention masking as typically implemented in standard frameworks~\cite{Wolf2019HuggingFacesTS}.

\paragraph{End-Task Fine-Tuning Details.}
Following prior work~\cite{roberta},
we perform grid search on training epochs and batch size $\in \cbr{16, 32}$ on validation sets.
We report the average performance over 5 runs in \secref{mainresults}.
We use a base learning rate of 2e-5 (the same as what is used in BERT~\cite{bert}), which we find to work well for all tasks.
Other hyperparameters are the same as pre-training.

\subsection{Main Results}\label{sec:mainresults}
We start with evaluating different state-of-the-art existing methods on long-form tasks,
and comparing them with the proposed Object Transformers.

\paragraph{Compared Methods.}
The most prominent class of video understanding methods today is probably 3D CNNs with late fusion~\cite{tran2015learning,carreira2017quo,nonlocal,slowfast,multigrid,tran2019video},
which has been widely used for a wide range of tasks~\cite{kay2017kinetics,ucf101,hmdb,sigurdsson2016hollywood}.
To compare with this category of methods,
we use a large state-of-the-art model, a ResNet-101~\cite{resnet} SlowFast network~\cite{slowfast} with non-local blocks~\cite{nonlocal} running on 128 frames,
pre-trained on Kinetics-600~\cite{k600} and AVA~\cite{gu2018ava} as a baseline method.
We train the network using SGD with cosine learning rate schedule, linear warmup~\cite{goyal2017accurate}, and a weight decay of 10$^{-4}$, following standard practice~\cite{slowfast}.
We select the base learning rate and the number of training epochs on validation set for each task; More details are in the Appendix.

Another promising baseline we compare to is the recently proposed frame-based long-term models,
VideoBERT~\cite{videobert}.
We compare with its vision-only variant, since language is beyond the scope of this paper.\footnote{We reached out to the authors of VideoBERT~\cite{videobert},
but they were not able to share the code with us.
We thus present results based on our re-implementation.
We select hyperparameters for VideoBERT~\cite{videobert} with the same grid-search protocol as our method for fair comparison.
More implementation details are in the Appendix.}

\begin{table*}[t]
\vspace{-1.em}
\small
\hspace{-1.5em}
\resizebox{1.05\linewidth}{!}{%
\subfloat[\textbf{Pre-training}\label{tab:abl:pretrain}]{%
\tablestyle{1.0pt}{1.05}
\begin{tabular}{@{}
l
x{23}x{23}x{23}x{23}x{23}
x{23}x{23}x{23}x{23}
@{}}
pre-train&
relation &
speak &
scene &
like$\downarrow$ &
views$\downarrow$ &
director &
genre &
writer &
year\\
\shline
None & 46.9 & 39.8 & 53.8 & 0.262 & 3.44 & 43.0 & 55.8 & 34.5 & 35.0\\
\textbf{Mask} & \textbf{54.7} & \textbf{40.3} & {58.0} & 0.238 & 3.71 & 53.3 & \textbf{56.1} & \textbf{35.1} & \textbf{40.6}\\
\textbf{Mask+Compat} & 50.0 & 32.8 & \textbf{60.0} & \textbf{0.234} & \textbf{3.37} & \textbf{58.9} & 49.3 & 32.7 & {39.9}\\
\hline
$\Delta$&\gain{(+7.8)}&\gain{(+0.5)}&\gain{(+6.2)}&\gain{(-.028)}&\gain{(-.07)}&\gain{(+15.9)}&\gain{(+0.3)}&\gain{(+0.6)}&\gain{(+5.6)}\\
\end{tabular}
}  

\subfloat[\textbf{Long-term module}\label{tab:abl:module}]{%
\tablestyle{1.0pt}{1.05}
\begin{tabular}{@{}
l
x{23}x{23}x{23}x{23}x{23}
x{23}x{23}x{23}x{23}
@{}}
&
relation &
speak &
scene &
like$\downarrow$ &
views$\downarrow$ &
director &
genre &
writer &
year\\
\shline
Short-term & 50.0 & 40.3 & 52.9 & 0.366 & {3.57} & 54.2 & 52.9 & 28.6 & 37.8\\
Avg pool & 37.5 & 36.8 & 57.1 & 0.496 & 3.82 & 40.2 & 54.4 & \textbf{37.5} & 32.9\\
Max pool & 50.0 & 37.8 & {58.8} & 0.284 & 3.78 & 52.3 & {55.8} & 32.7 & 34.3\\
\textbf{Transformer} & \textbf{54.7} & \textbf{40.3} & \textbf{60.0} & \textbf{0.234} & \textbf{3.37} & \textbf{58.9} & \textbf{56.1} & {35.1} & \textbf{40.6}
\end{tabular}
}  
}  

\hspace{-1.5em}
\resizebox{1.05\linewidth}{!}{%
\subfloat[\textbf{Modality}\label{tab:abl:modality}]{%
\tablestyle{1.0pt}{1.05}
\begin{tabular}{@{}
l
x{23}x{23}x{23}x{23}x{23}
x{23}x{23}x{23}x{23}
@{}}
&
relation &
speak &
scene &
like$\downarrow$ &
views$\downarrow$ &
director &
genre &
writer &
year\\
\shline
Person & \underline{54.7} & \textbf{40.3} & \textbf{60.0} & 0.234 & \textbf{3.37} & \textbf{58.9} & \textbf{56.1} & 35.1 & 40.6\\
Person+Obj.\ \ \ \ \ \  & \underline{54.7} & 37.8 & {58.8} & \textbf{0.223} & 3.67 & 48.6 & 55.8 & \textbf{36.3} & \textbf{42.0}\\
\end{tabular}
}  

\subfloat[\textbf{Number of pre-training videos}\label{tab:abl:size}]{%
\tablestyle{1.0pt}{1.05}
\begin{tabular}{%
@{}
l
x{23}x{23}x{23}x{23}x{23}
x{23}x{23}x{23}x{23}
@{}}
&
relation &
speak &
scene &
like$\downarrow$ &
views$\downarrow$ &
director &
genre &
writer &
year\\
\shline
10k & 50.0 & \textbf{40.8} & 58.0 & \textbf{0.230} & 3.42 & 53.3 & 53.2 & 32.7 & 37.8\\
\textbf{30k (all)}\ \ \ \quad\quad & \textbf{54.7} & {40.3} & \textbf{60.0} & {0.234} & \textbf{3.37} & \textbf{58.9} & \textbf{56.1} & \textbf{35.1} & \textbf{40.6}\\
\end{tabular}
}  
}  
\vspace{1.5mm}
\caption{\textbf{Ablation Experiments.}
Our results validate that self-supervised pre-training brings consistent gains across tasks (\ref{tab:abl:pretrain}).
We also observe that simpler pooling methods are not as effective as transformer,
supporting that object-level interaction modeling is beneficial (\ref{tab:abl:module}).
Modeling non-person objects is beneficial for a few tasks, but modeling humans along is already strong is most of the tasks (\ref{tab:abl:modality}).
Finally, pre-training on more data helps in most cases (\ref{tab:abl:size}), suggesting promising future work using even larger datasets.
($\downarrow$: lower is better)
}
\label{tab:abl}
\end{table*}

\paragraph{Results.}
\tabref{mainresults} shows that our Object Transformer outperforms both baselines by a clear margin in terms of the overall ranking.
The short-term model (`R101-SlowFast+NL'), 
is not able to perform well even with a large backbone and strong pre-training (Kinetics-600~\cite{k600} and AVA~\cite{gu2018ava}).
This validates the importance of long-term modeling.
We also observe that object-centric modeling (Object Transformers) is advantageous compared with frame-centric modeling (`VideoBERT'~\cite{videobert}).
Interestingly, short-term models suffice to work well for \emph{year prediction}. 
This should not come as a surprise, since local statistics such as image quality or color style already capture a lot about the `\emph{year}' of a video (\eg, as shown in \figref{year}).
VideoBERT~\cite{videobert} works well for \emph{writer} prediction,
suggesting that this task might not require too much detailed interaction modeling.

In short, a long-term and object-centric design is important for a wide range of LVU tasks.

\subsection{Ablation Experiments}

\paragraph{Pre-Training.}
We first evaluate the impact of the proposed pre-training methods.
\tabref{abl:pretrain} shows that on all tasks we evaluate,
pre-training is beneficial.\footnote{%
In ablation experiments, we report the results without averaging over 5 runs due to computation resource constraints.
Thus the results are slightly different from the results reported in \tabref{mainresults}.
}
In particular, Masked Instance Pre-Training alone works well in almost all tasks,
while adding Compatibility Pre-Training helps in 4 out of the 9 tasks.
Interestingly, our results are similar to observations in NLP research, where
the `masked-language model' alone works well on some tasks (\eg, \cite{joshi2020spanbert}),
while additionally using `next-sentence-prediction' helps on others (\eg, \cite{bert}).
In other parts of this paper, we use the best performing pre-training method (selected based on validation results) as the default for each task.

\paragraph{Long-Term Module.}
Most existing methods perform either pooling-based aggregation~\cite{wang2016temporal} or no aggregation at all (late fusion only)~\cite{nonlocal,slowfast} when it comes to long-term modeling.
\tabref{abl:module} compares our Object Transformer with these approaches.
All methods in \tabref{abl:module} build on the same input features. 
The only difference is the module built on top of these features.
Object Transformer works better on 8 out of the 9 tasks,
showing that for long-form video understanding,
a more powerful object-level interaction modeling is advantageous.
Interestingly, for `\emph{movie writer}' prediction,
a transformer does not outperform even average pooling.
We conjecture that writer prediction might require a higher level of cognition or abstraction ability, that is beyond what transformers can do.
We think studying this task is interesting future work.

\paragraph{Modality.}
While humans are arguably the most central elements for understanding a video,
we study the benefit of including other objects.
\tabref{abl:modality} shows that
adding objects brings only mild improvement on three tasks.
This suggests that human behavior understanding plays an crucial role in
most long-form tasks.

\paragraph{Number of Pre-Training Videos.}
A great advantage of our pre-training methods is that they are self-supervised,
not requiring any human annotations.
It is thus relatively easy to pre-train on large-scale datasets.
\tabref{abl:size} shows that on most tasks, pre-training on more data helps,
suggesting promising future research that leverages even more data.

\paragraph{Model Complexity.}
\tabref{complexity} presents a breakdown analysis for complexity in terms of both
model size and FLOPs.
We see that our Object Transformer is small and efficient.
It is 2.2$\x$ smaller and takes only 0.7\% of the FLOPs compared to short-term action detection.
We thus expect future research on long-form video understanding to be accessible.

\paragraph{Example Predictions of Masked Instance Prediction.}
Finally, we present case study to understand what the model learns with Masked Instance Prediction.
\figref{viz} presents three examples from a hold-out dataset of AVA~\cite{gu2018ava} along with the model outputs.
We see that our model leverages the context, some of them not on the same frame, and make sensible predictions without seeing the actual content.
This validates that long-term human interactions can indeed be learned in a self-supervised way.

\begin{table}[t]
\centering
\tablestyle{6pt}{1.0}
\begin{tabular}{@{}lx{55}x{55}x{55}@{}}
& Action Detection & Object Detection & Transformer\\
\shline
params (M) & 59.2 & 60.6 & 27.0\\
FLOPs (G) & 242.0 & 88.6 & 1.8\\
\end{tabular}
\vspace{1mm}
\caption{\textbf{Inference Complexity Breakdown.}
Object Transformer is small and efficient --- taking only 0.7\% of the FLOPs and being 2.2$\x$ smaller compared to Action Detection.
}
\label{tab:complexity}
\end{table}

\subsection{Experiments on AVA}\label{sec:ava}
\begin{table}[t]
\centering
\tablestyle{4pt}{1.0}
\begin{tabular}{@{}lx{30}x{28}x{18}x{20}@{}}
& input& pre-train & mAP & FLOPs\\
\shline
\demph{\textbf{AVA V2.1}}\\
\quad \demph{AVA}~\cite{gu2018ava} & \demph{V+F} & \demph{K400} & \demph{15.6} & \demph{-}\\
\quad \demph{ACRN}~\cite{sun2018actor} & \demph{V+F} & \demph{K400} & \demph{17.4} & \demph{-}\\
\quad \demph{STEP}~\cite{yang2019step} & \demph{V} & \demph{K400} & \demph{18.6} & \demph{-}\\
\quad \demph{Zhang~\etal}~\cite{zhang2019structured} & \demph{V} & \demph{K400} & \demph{22.2} & \demph{-}\\
\quad \demph{RTPR}~\cite{li2018recurrent} & \demph{V+F} & \demph{ImageNet} & \demph{22.3} & \demph{-}\\
\quad \demph{Girdhar~\etal}~\cite{girdhar2019video} & \demph{V} & \demph{K400} & \demph{25.0} & \demph{-}\\
\quad \demph{LFB}~\cite{lfb} & \demph{V} & \demph{K400} & \demph{27.7} & \demph{-}\\
\quad \demph{AVSlowFast}~\cite{xiao2020audiovisual} & \demph{V+A} & \demph{K400} & \demph{27.8} & \demph{-}\\
\hline
\textbf{AVA V2.2}\\
\quad \demph{AVSlowFast}~\cite{xiao2020audiovisual} & \demph{V+A} & \demph{K400} & \demph{28.6} & \demph{-}\\
\quad \demph{AIA}~\cite{tang2020asynchronous} & \demph{V} & \demph{K700} & \demph{32.3} & \demph{-}\\
\quad X3D-XL~\cite{feichtenhofer2020x3d} & V & K600 & 27.4 & -\\
\quad SlowFast R101~\cite{slowfast} & V & K600 & 29.4 & 1.000$\x$\\
\quad Object Transformer (masked) & V & K600 & 29.3 & 1.007$\x$\\
\quad \textbf{Object Transformer} & V & K600 & \textbf{31.0} & 1.007$\x$
\end{tabular}
\vspace{1mm}
\caption{\textbf{Action Recognition Results on AVA\@.} 
Object Transformer outperforms prior work, which use only short-term information.
Our results suggest that long-term interaction and context are beneficial for short-form tasks as well.
(V: Visual; A: Audio; F: Flow; K400: \cite{kay2017kinetics}; K600: \cite{k600}; K700: \cite{k700}.)}
\label{tab:ava}
\end{table}

\begin{figure*}[t]
\centering
\begin{tabular}{@{}l@{}}
\includegraphics[width=\linewidth]{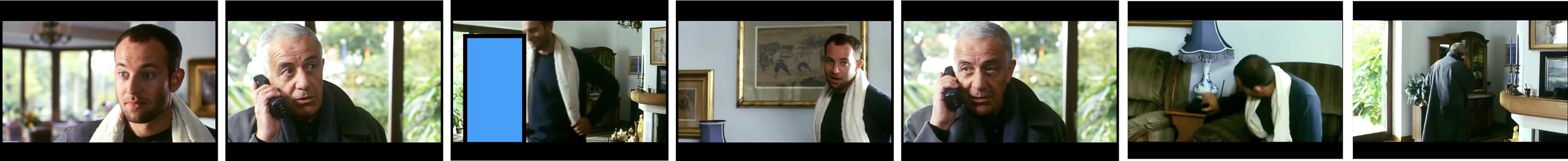}\vspace{-1.3mm}\\
Predictions: \emph{stand} (99.2\%), \emph{answer phone} (97.8\%)\vspace{1.3mm}\\
\includegraphics[width=\linewidth]{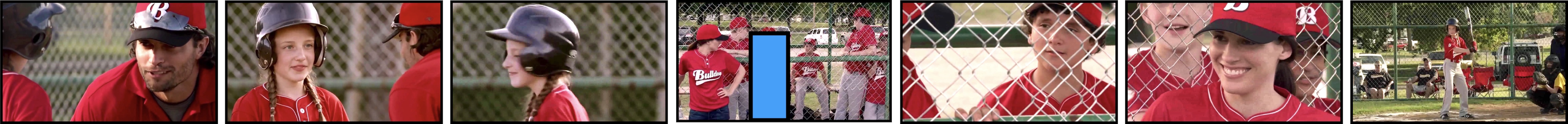}\vspace{-1.3mm}\\
Predictions: \emph{hand clap} (79.1\%), \emph{stand} (71.5\%), \emph{watch (a person)} (69.4\%)\vspace{1.3mm}\\
\includegraphics[width=\linewidth]{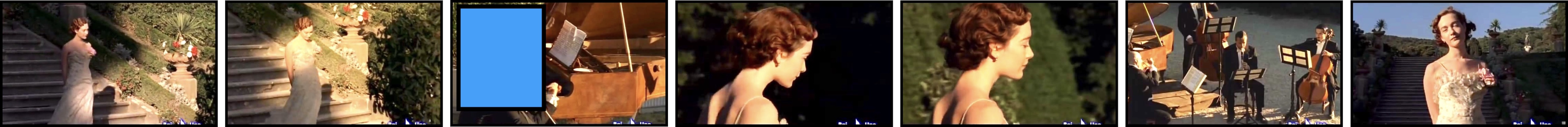}\vspace{-1.3mm}\\
Predictions: \emph{play musical instrument} (90.1\%), \emph{sit} (56.9\%)
\end{tabular}
\vspace{0.5mm}
\caption{\textbf{Masked Instance Prediction Examples.}
Here we present three examples along with their masked instances, and the actions of these instances predicted by our model.$^\dagger$
Without seeing the actual content, our model leverages long-term context and makes plausible predictions. Some of these (\eg, the top example) are not possible without modeling longer-term context. 
(Best viewed on screen.)
\newline
{\footnotesize $^\dagger$: Here we list predictions with $\geq$50\% probabilities.
We present only 7 frames for each example at 0.5 FPS due to the space constraint;
See the Appendix for the full examples.}\vspace{-3.5mm}
}
\label{fig:viz}
\end{figure*}

So far we have evaluated Object Transformers on long-form video tasks.
We next evaluate its ability of improving ``short-form" recognitions through incorporating long-term context.
Here evaluate our method on AVA~\cite{gu2018ava} V2.2 for spatial-temporal action detection.
Performance is measured by mAP\@, following standard practice~\cite{gu2018ava}.

\paragraph{Adaptation to AVA\@.}
Note that directly fine-tuning Object Transformers to predict the box-level AVA outputs would lead to a ``short-cut" solution, 
where the model directly looks at the corresponding input feature $\feat$ without long-term modeling.
We thus mask out the input features $\feat$ for the target instances (similar to masked-instance pre-training) when fine-tuning the AVA\@ model.
This, however, would put our model at a disadvantage, since prediction given only context is much harder than the original task.
We thus take a simple approach of late fusing the short-term prediction, and fine-tuning only the final linear layer (for
 2000 iterations using a base learning rate of 10$^{-4}$).
This procedure is efficient,
as no updating of the attention layers are involved.

\paragraph{Results.}
We evaluate and compare our Object Transformer with prior work in \tabref{ava}.
We see that without using optical-flow~\cite{gu2018ava,sun2018actor,li2018recurrent}, audio~\cite{xiao2020audiovisual}, or task-specific engineering, our Object Transformer outperforms state-of-the-art short-term models that use comparable (K600~\cite{k600}) feature pre-training by \textbf{1.6\%} absolute (29.4\% $\rightarrow$ \textbf{31.0}\%).
This shows that even for short-form tasks,
it is beneficial to consider long-term context to supplement or disambiguate cases where local patterns are insufficient.
Note that this improvement comes almost ``for free", as it only uses \textbf{0.7\%} of additional FLOPs and fine-tuning of a linear layer.
Interestingly, a ``masked" Object Transformer \emph{without} late fusion (denoted `Object Transformers (masked)' in \tabref{ava}), still achieves 29.3,
demonstrating that ``\emph{given context only}", our model is able to leverage context and predict the semantics of the masked parts of a video with state-of-the-art quality (also see \figref{viz} for qualitative results).

In short, plugging a Object Transformer into a short-form task is easy,
and it leads to a clear improvement.

\section{Conclusion}
In this paper, we take a step towards understanding long-form videos.
We build a new benchmark with 9 tasks on publicly available large datasets to evaluate a wide range of aspects of the problem.
We observe that existing short-term models or frame-based long-term models are limited in most of these tasks.
The proposed Object Transformers that model the synergy among people and objects work significantly better.
We hope this is a step towards deeper, more detailed, and more insightful understanding of our endlessly evolving visual world, with computer vision.

{{\noindent\textbf{Acknowledgments.}
This material is based upon work supported by the National Science Foundation under Grant No.\ IIS-1845485 and IIS-2006820, and the NSF Institute for Foundations of Machine Learning. Chao-Yuan was supported by the Facebook PhD Fellowship.}}

{\small
\bibliographystyle{ieee_fullname}
\bibliography{long_form_vu}
}
\clearpage

\setcounter{section}{0}
\begin{subappendices}
\renewcommand{\thesection}{\Alph{section}}%

\section{Supplementary Implementation Details}
\paragraph{Object Transformer Architecture.}
We develop our default architecture based on the BERT$_\texttt{BASE}$ architecture from Delvin~\etal~\cite{bert}.
Specifically, our architecture uses 12 attention heads per layer in a 3-layer\footnote{Original BERT$_\texttt{BASE}$ is 12-layer, but we found 3 layers suffice for Object Transformers to achieve good performance.} design, with 64-dimensional attention heads, 768-dimensional hidden layers, and 3072-dimensional feed-forward networks.
Since each example contains a different number of instances of different lengths, we perform attention masking as typically implemented in standard frameworks~\cite{Wolf2019HuggingFacesTS}.
Each example contains up to 256 tokens when using person features only (default) and up to 512 tokens when additionally using object features.
To construct positional embeddings, we first encode the position into a three-value vector, (distance from start, distance from end, distance from center), and use a linear transformation to map this vector to a 768-dimensional embedding vector.
We use attention dropout of 0.1 following standard practice~\cite{bert}.

\paragraph{End-Task Fine-Tuning Details.}
Similar to prior work in natural language processing~\cite{roberta,bert},
we select the batch size $\in \cbr{16, 32}$ and the number of training epochs from 
\{3, 5, 10, 20, 30, 50, 100, 200, 300, 500, 1000, 2000\}
for each task on the validation set.

\paragraph{`R101-SlowFast+NL' Baseline Implementation Details.}
We use the open-source PySlowFast codebase~\cite{fan2020pyslowfast} for this 3D CNN baseline.
The Kinetics-600~\cite{k600} and AVA~\cite{gu2018ava} pre-training model weights are from the PySlowFast Model Zoo\footnote{\url{https://github.com/facebookresearch/SlowFast/blob/master/MODEL_ZOO.md}}.
We observe that 3D CNNs are more sensitive to learning rates than transformers.
We thus additionally select learning rate $\in \cbr{0.001, 0.010, 0.025}$ for each task.
Linear warmup is used for the first 5\% of the training schedule followed by a cosine learning rate decay following standard practice~\cite{slowfast}.

\paragraph{VideoBERT Implementation Details.}
The main difference between Object Transformers and VideoBERT~\cite{videobert}
lies in the object-centric \vs frame-centric view in design.
In our experiments, we aim at comparing this central design choice, 
while controlling minor implementation details.
To this end, we use the same positional embedding formulation as ours for VideoBERT for fair comparison.
In addition, Sun~\etal~\cite{sun2019learning} notes that continuous input vectors is advantageous over discrete tokens.
We thus use dense input vectors with the same InfoNCE-style training objective for VideoBERT, instead of using discretized tokens as in their original paper,
for consistent comparison to Object Transformers.
We also use the same ResNet-101~\cite{resnet} SlowFast~\cite{slowfast} architecture with non-local blocks~\cite{nonlocal} as what Object Transformers use, and pre-train the model on Kinetics-600~\cite{k600}.
We select the best hyperparameters for each task for VideoBERT using the same
grid search protocol as Object Transformers.

\newcommand{\imgwithboxtwo}[2]{%
\parbox[t][13.3mm][t]{\linewidth}{
\begin{minipage}{\linewidth}
\begin{tikzpicture}
    \node (img) {\includegraphics[width=\linewidth]{#2}};
    \node [above left=1mm, fill=white, opacity=0.7] at (img.south east){\color{black}{\footnotesize #1}};
\end{tikzpicture}
\end{minipage}}}
\begin{figure*}[t]
\begin{minipage}[t]{68.6pt}
\tablestyle{0.0pt}{0.0}
\begin{tabular}[t]{@{}x{68.6}@{}}
\textbf{Relationship}
\imgwithboxtwo{wife \& husband}{figs/tasks/jpegs/rel_hw.jpg}\\
\imgwithboxtwo{friends}{figs/tasks/jpegs/rel_friend.jpg}\\
\imgwithboxtwo{boyfriend/girlfriend}{figs/tasks/jpegs/rel_bg.jpg}\\
\imgwithboxtwo{\begin{tabular}{r}ex-boyfriend/\\girlfriend\end{tabular}}{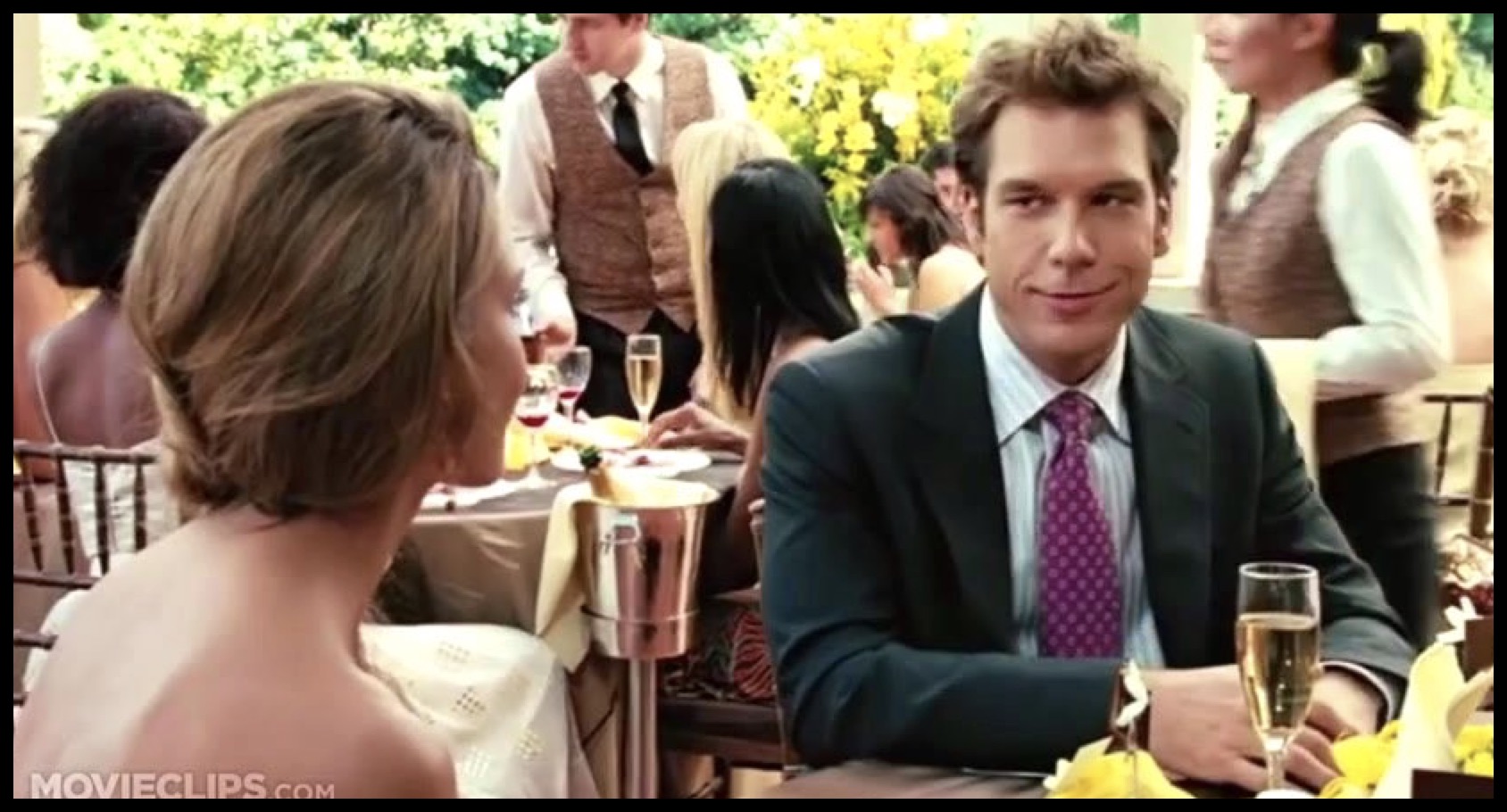}
\end{tabular}
\end{minipage}
\begin{minipage}[t]{68.6pt}
\tablestyle{0.0pt}{0.0}
\begin{tabular}[t]{@{}x{68.6}@{}}
\textbf{Way of Speaking}\\
\imgwithboxtwo{confront}{figs/tasks/jpegs/speak_confront.jpg}\\
\imgwithboxtwo{explain}{figs/tasks/jpegs/speak_exp.jpg}\\
\imgwithboxtwo{discuss}{figs/tasks/jpegs/speak_discuss.jpg}\\
\imgwithboxtwo{teach}{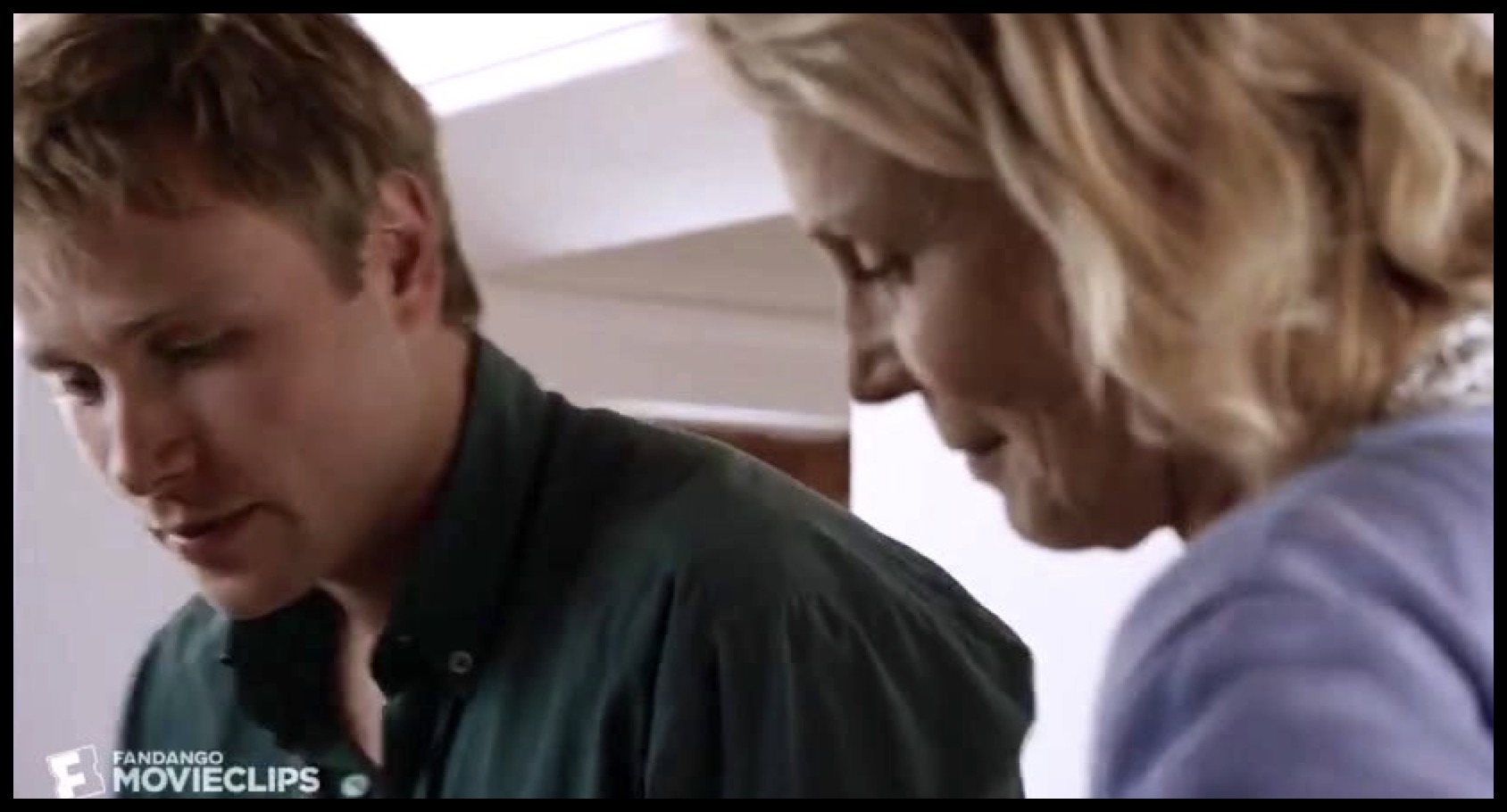}\\
\imgwithboxtwo{threaten}{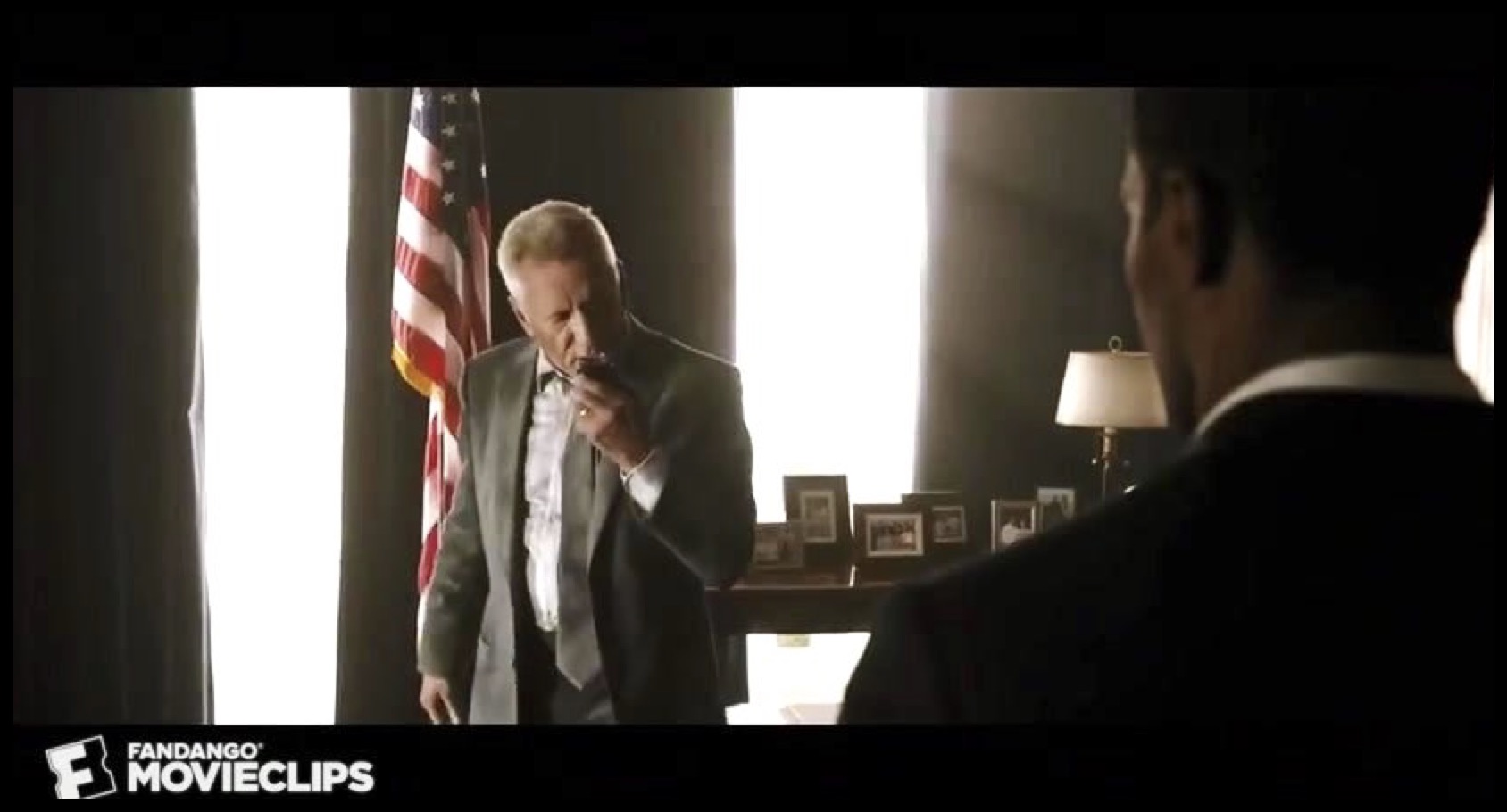}
\end{tabular}
\end{minipage}
\begin{minipage}[t]{68.6pt}
\tablestyle{0.0pt}{0.0}
\begin{tabular}[t]{@{}x{68.6}@{}}
\textbf{Scene/Place}\\
\imgwithboxtwo{airport}{figs/tasks/jpegs/place_airport.jpg}\\
\imgwithboxtwo{school}{figs/tasks/jpegs/place_school.jpg}\\
\imgwithboxtwo{office}{figs/tasks/jpegs/place_office.jpg}\\
\imgwithboxtwo{hotel}{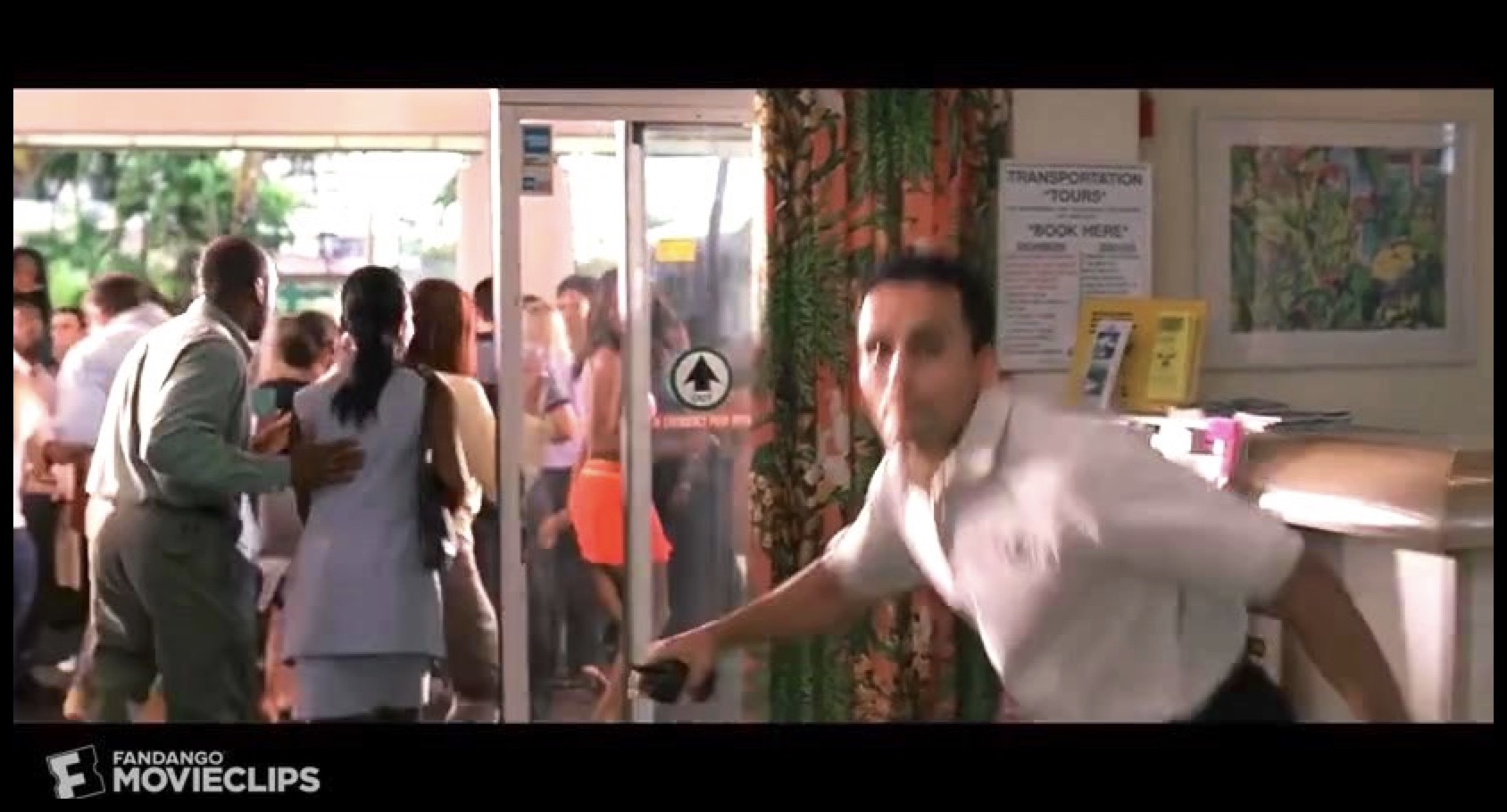}\\
\imgwithboxtwo{prison}{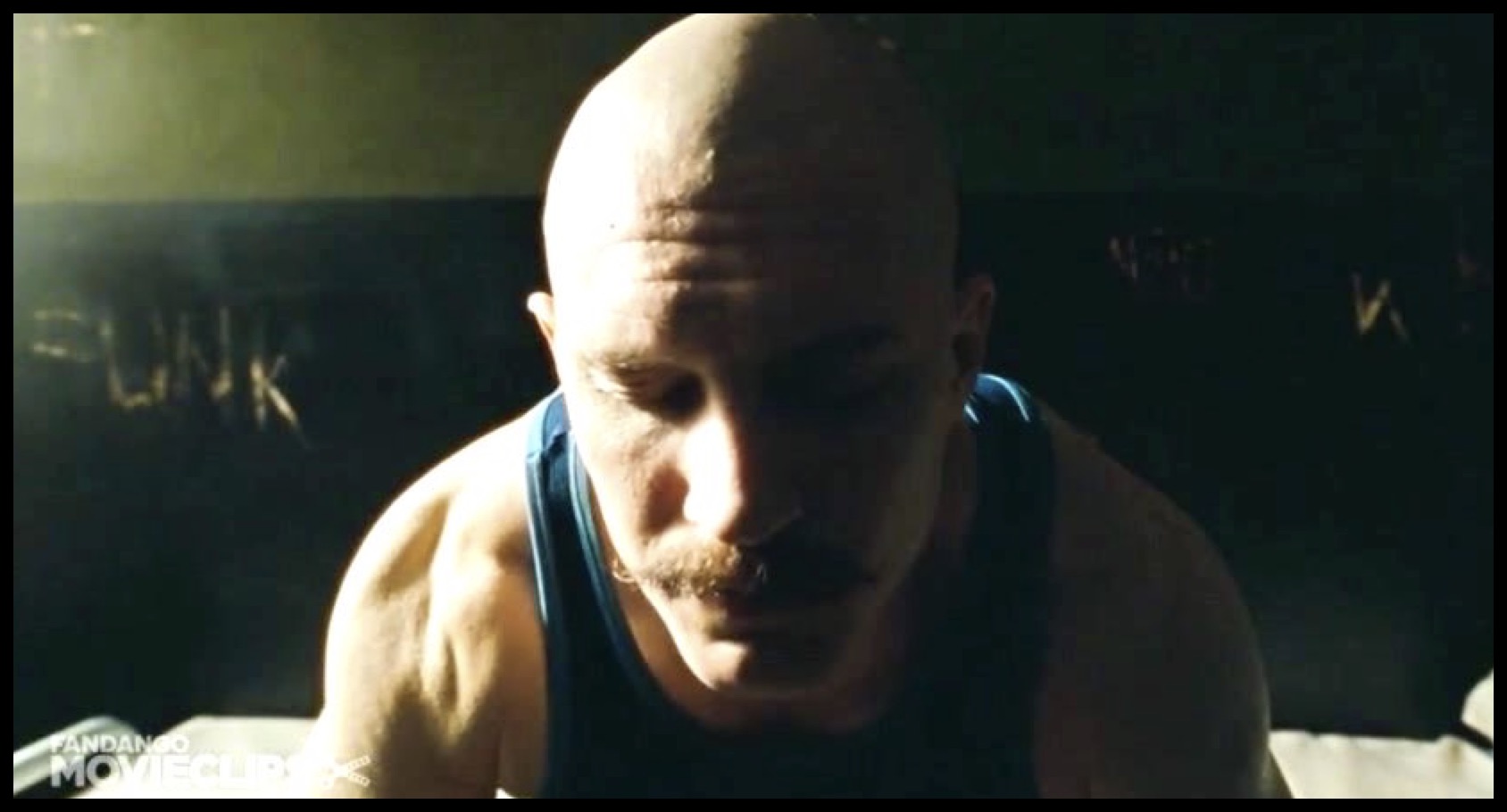}\\
\imgwithboxtwo{restaurant}{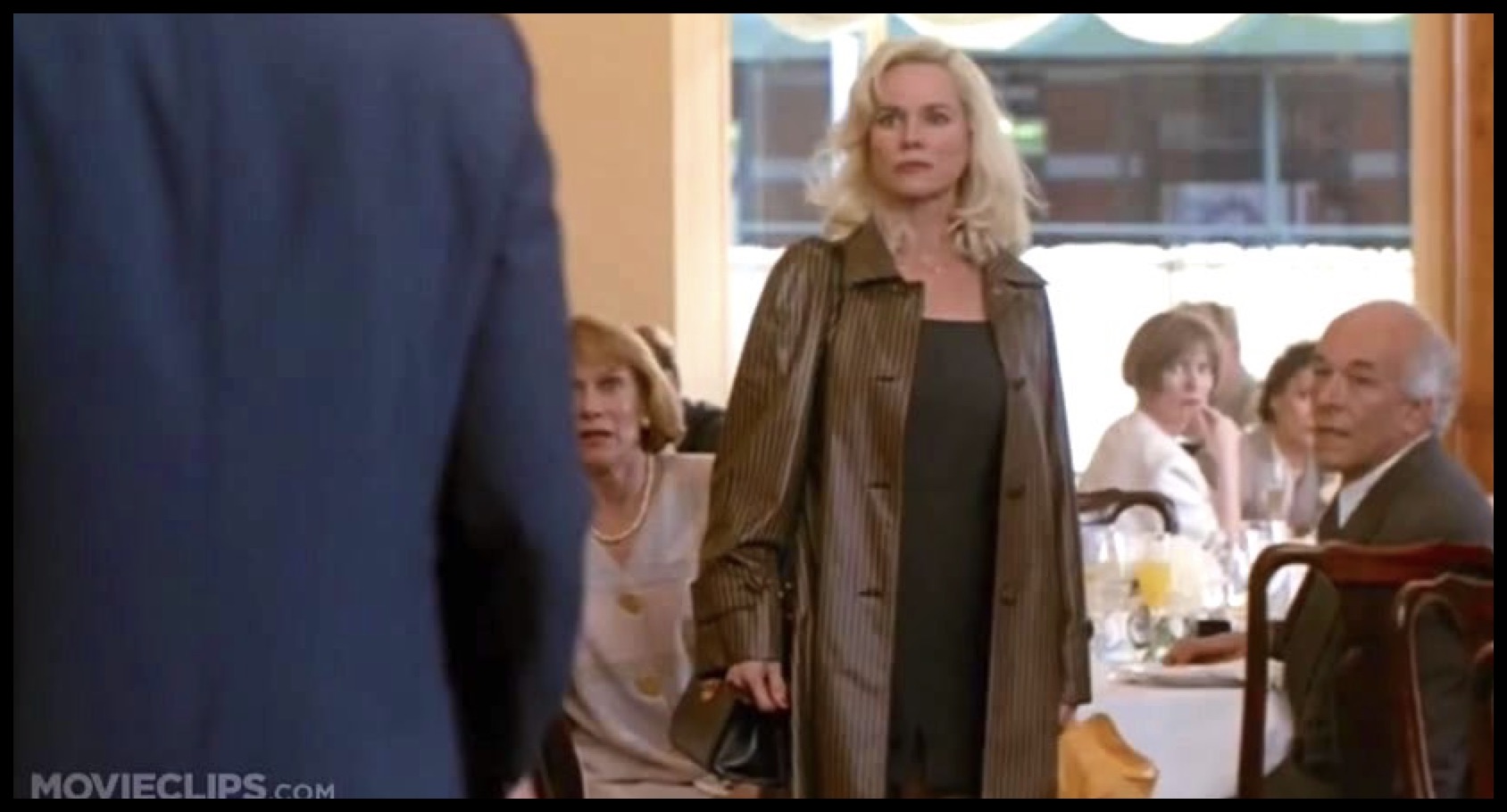}
\end{tabular}
\end{minipage}
\begin{minipage}[t]{68.6pt}
\tablestyle{0.0pt}{0.0}
\begin{tabular}[t]{@{}x{68.6}@{}}
\textbf{Director}\\
\imgwithboxtwo{Quentin Tarantino}{figs/tasks/jpegs/dir_qt.jpg}\\
\imgwithboxtwo{Ron Howard}{figs/tasks/jpegs/dir_ron.jpg}\\
\imgwithboxtwo{Peter Jackson}{figs/tasks/jpegs/dir_peter.jpg}\\
\imgwithboxtwo{Martin Scorsese}{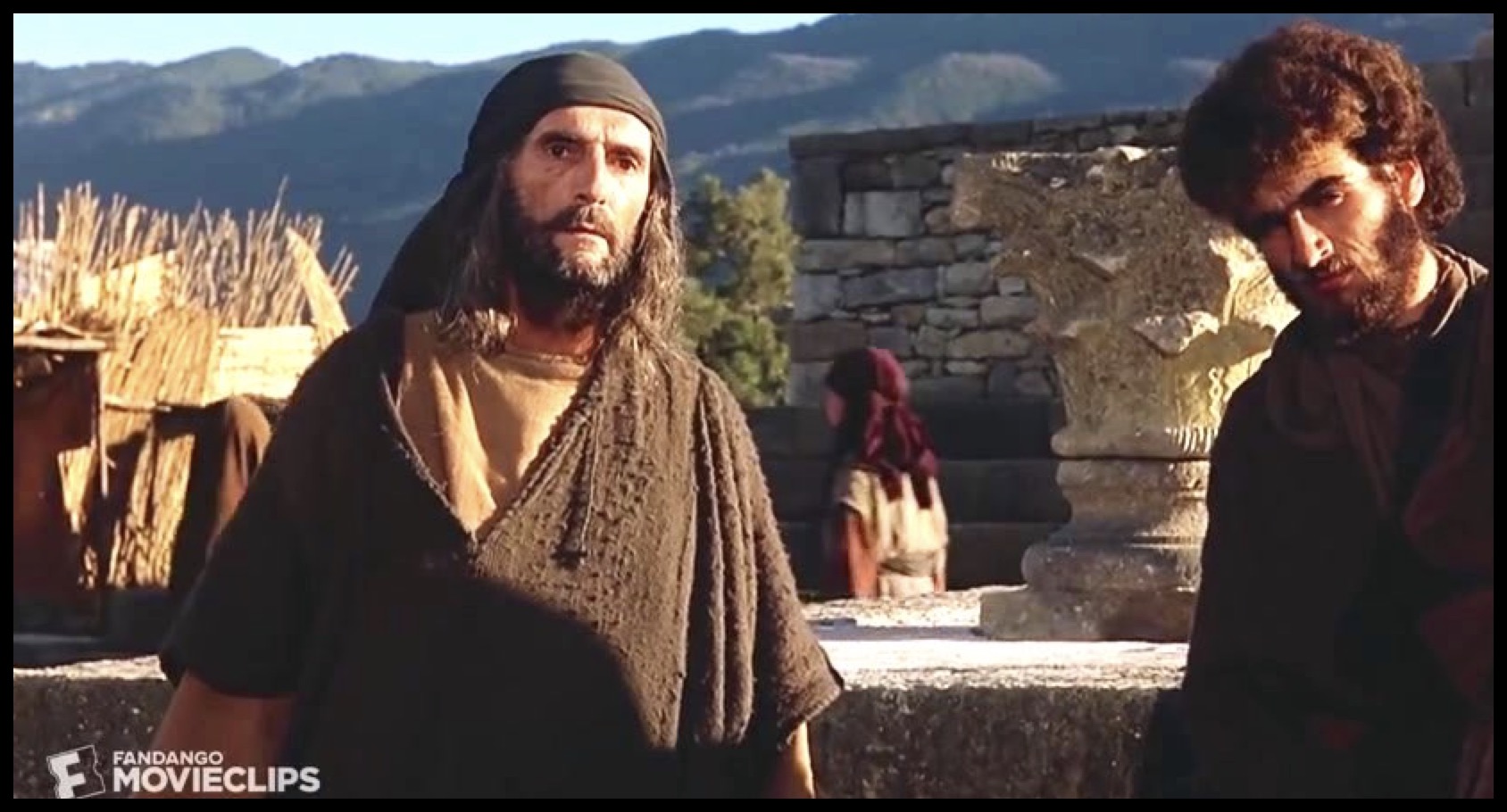}\\
\imgwithboxtwo{Steven Spielberg}{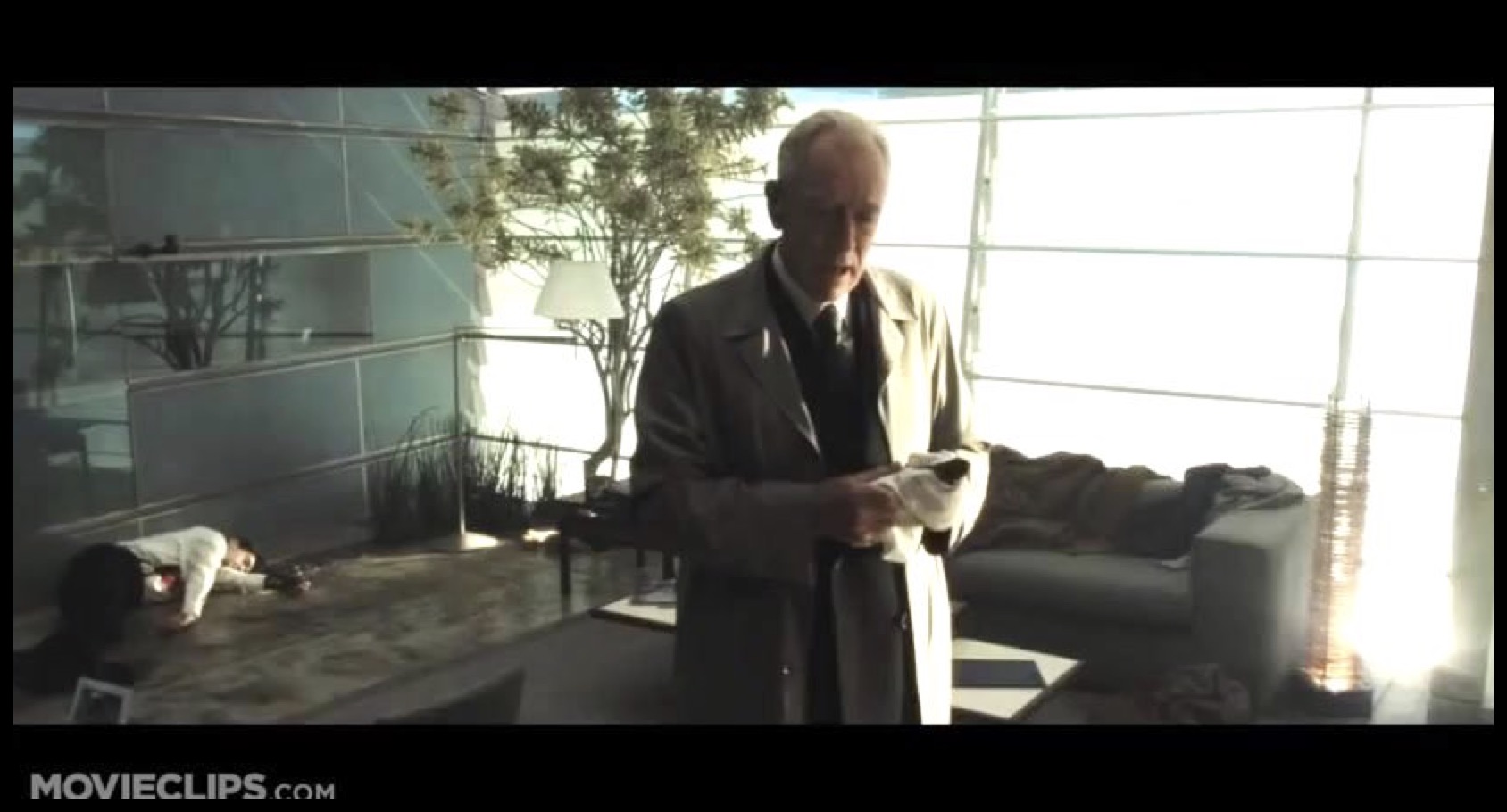}\\
\imgwithboxtwo{Ridley Scott}{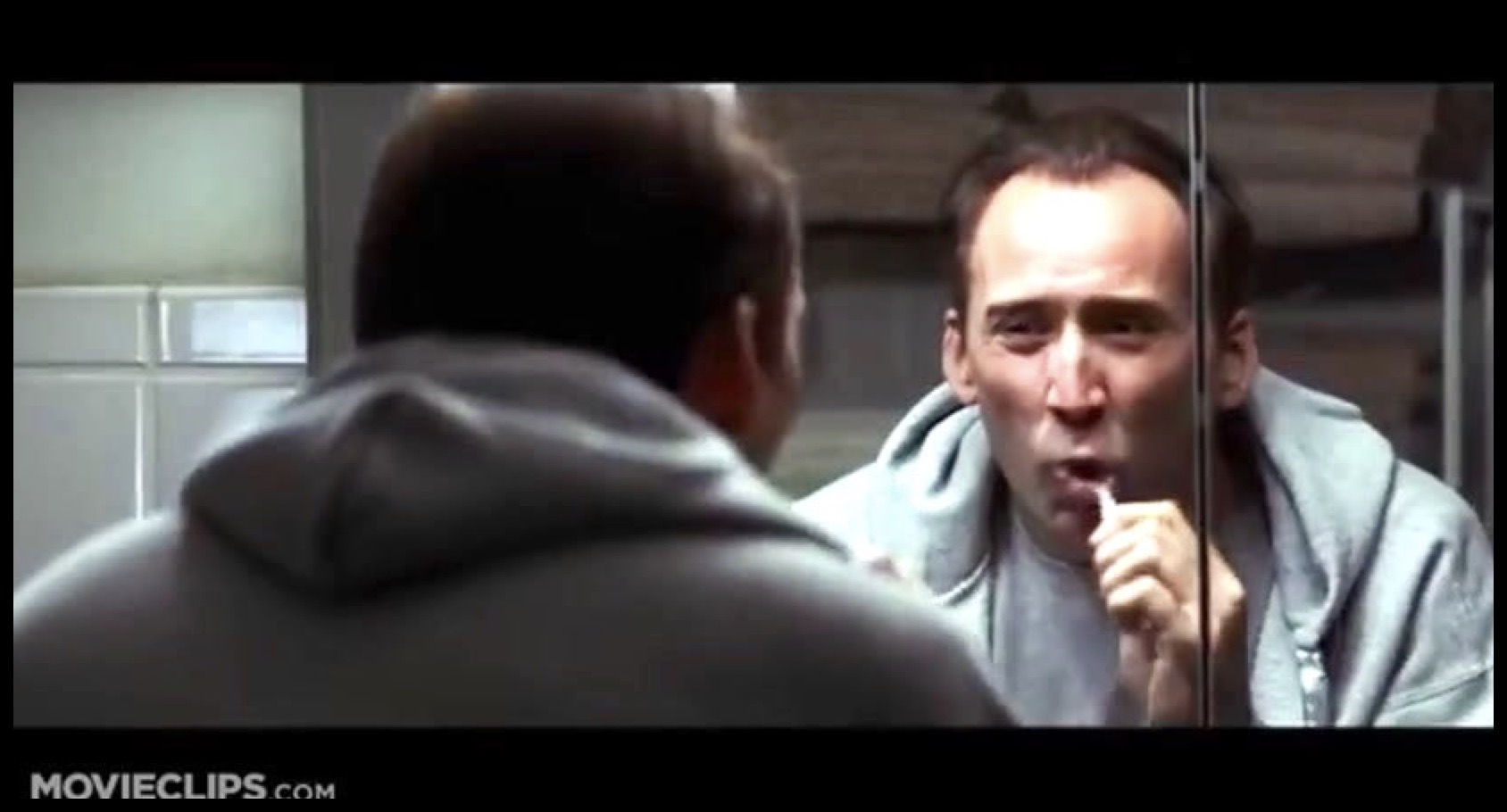}\\
\imgwithboxtwo{Robert Rodriguez}{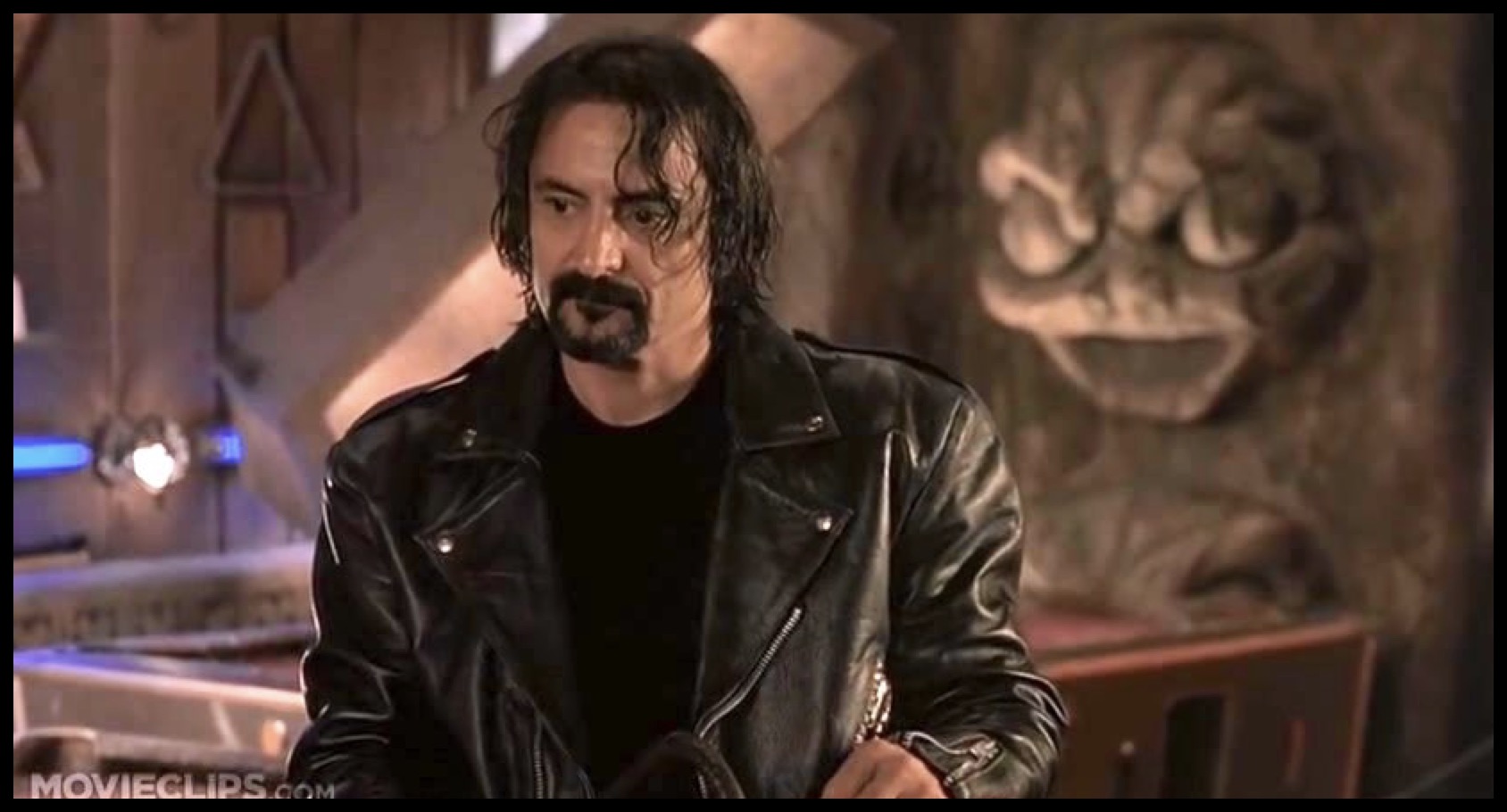}\\
\imgwithboxtwo{Mark Atkins}{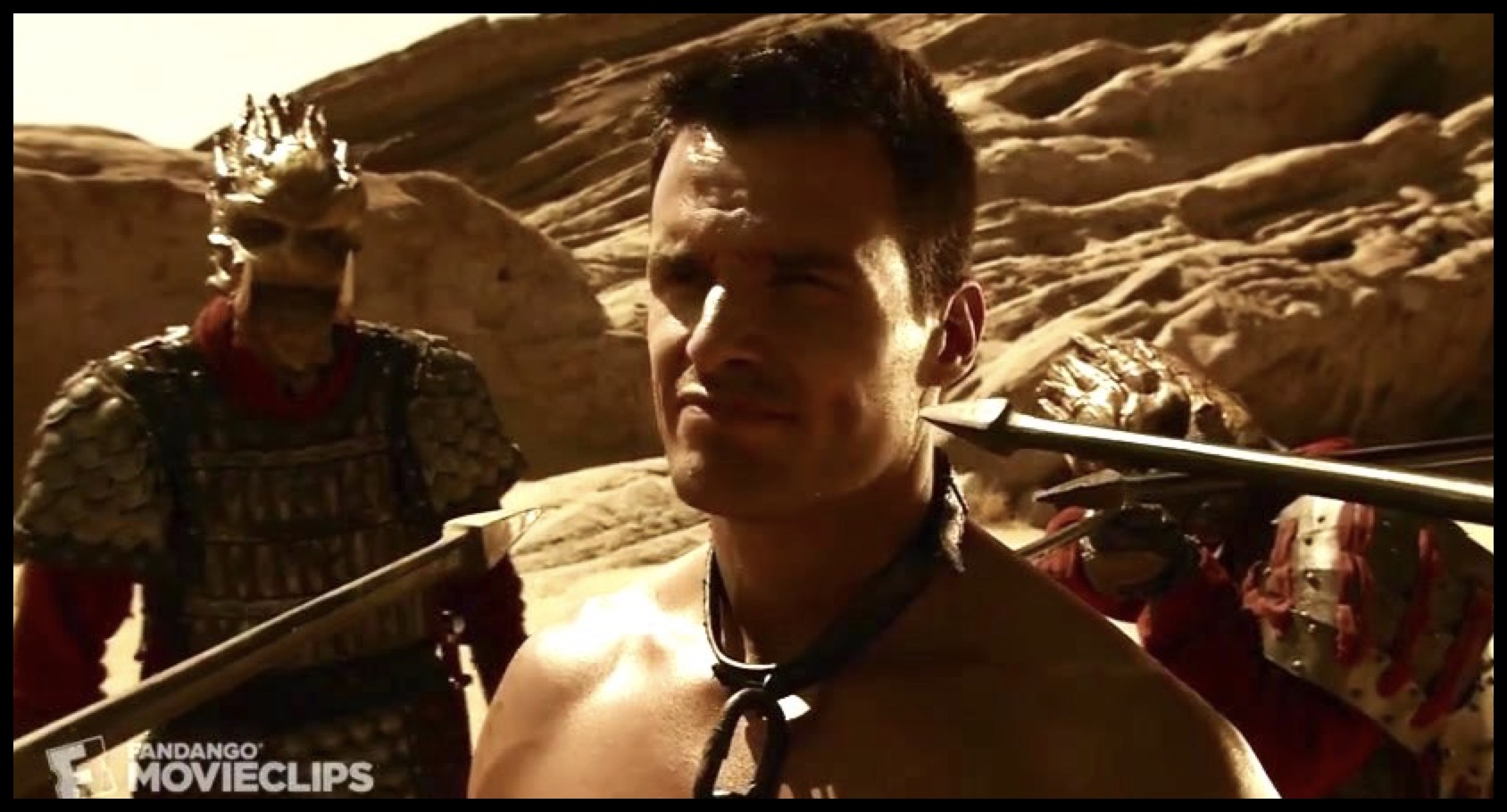}
\end{tabular}
\end{minipage}
\begin{minipage}[t]{68.6pt}
\tablestyle{0.0pt}{0.0}
\begin{tabular}[t]{@{}x{68.6}@{}}
\textbf{Genre}\\
\imgwithboxtwo{romance}{figs/tasks/jpegs/g_romance}\\
\imgwithboxtwo{horror}{figs/tasks/jpegs/g_horror}\\
\imgwithboxtwo{comedy}{figs/tasks/jpegs/g_comedy}\\
\imgwithboxtwo{action}{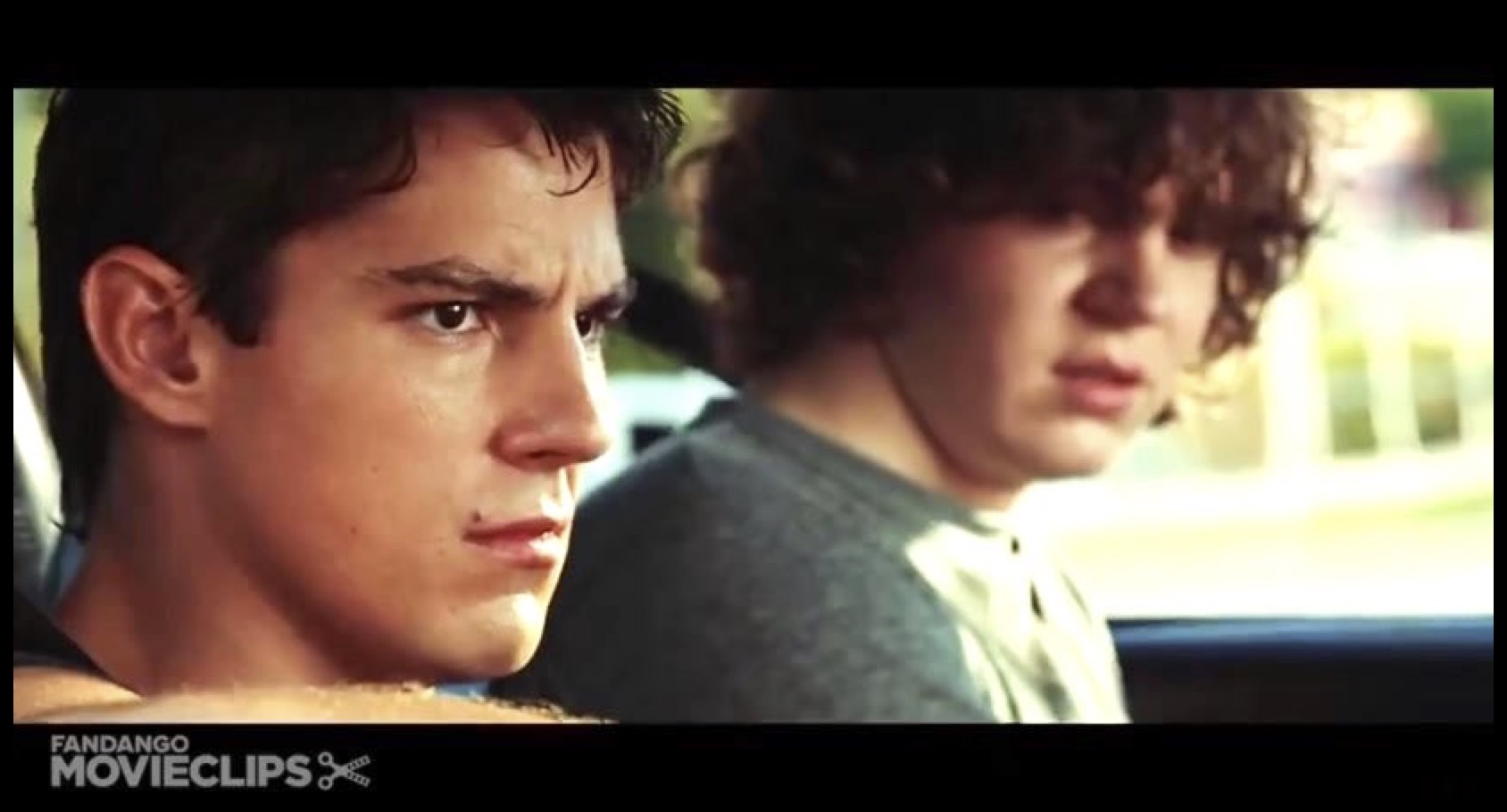}
\end{tabular}
\end{minipage}
\begin{minipage}[t]{68.6pt}
\tablestyle{0.0pt}{0.0}
\begin{tabular}[t]{@{}x{68.6}@{}}
\textbf{Writer}\\
\imgwithboxtwo{John Hughes}{figs/tasks/jpegs/w_hughes.jpg}\\
\imgwithboxtwo{Stephen King}{figs/tasks/jpegs/w_stev.jpg}\\
\imgwithboxtwo{David Koepp}{figs/tasks/jpegs/w_dave.jpg}
\imgwithboxtwo{Sylvester Stallone}{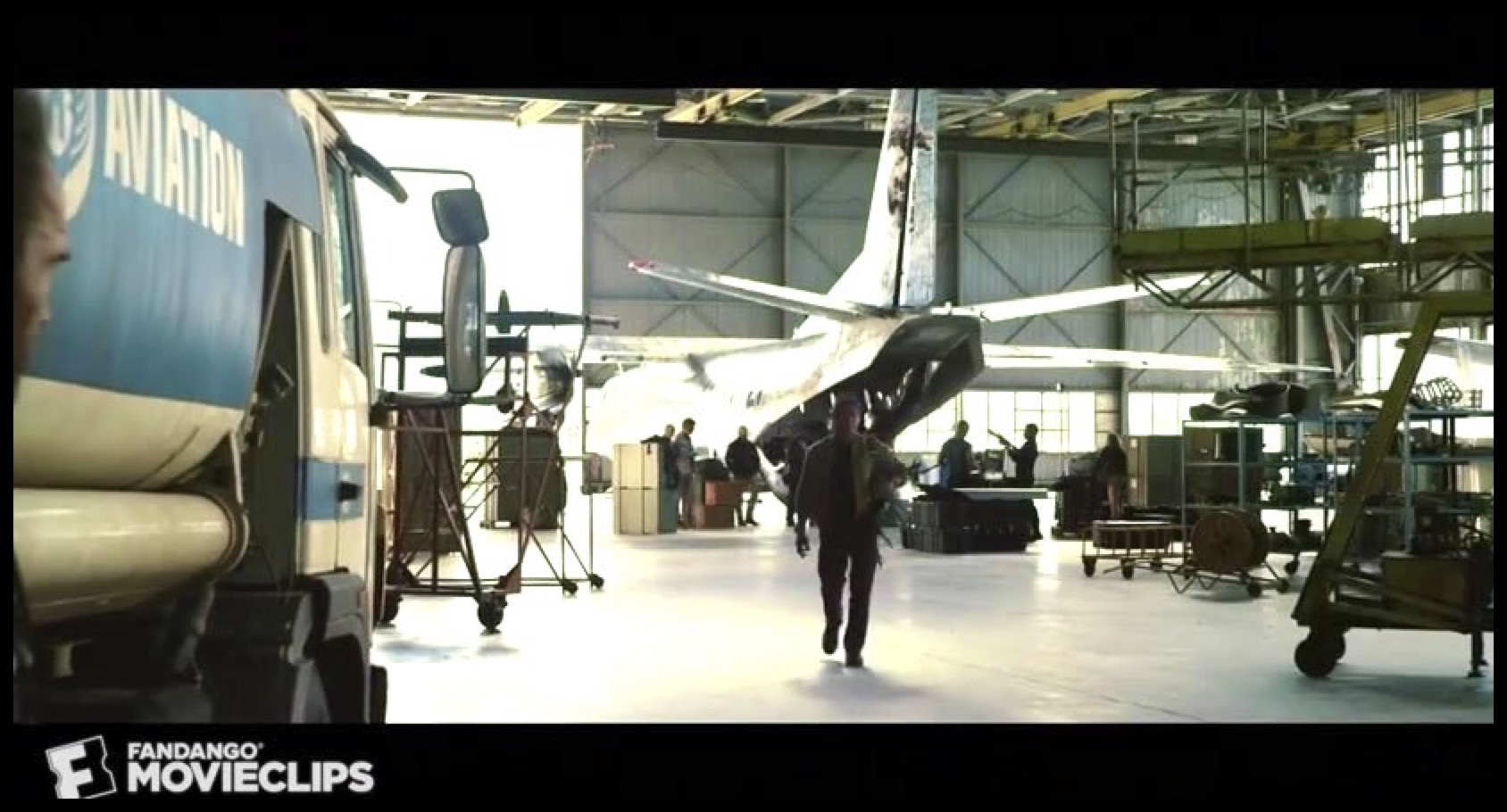}\\
\imgwithboxtwo{Ian Fleming}{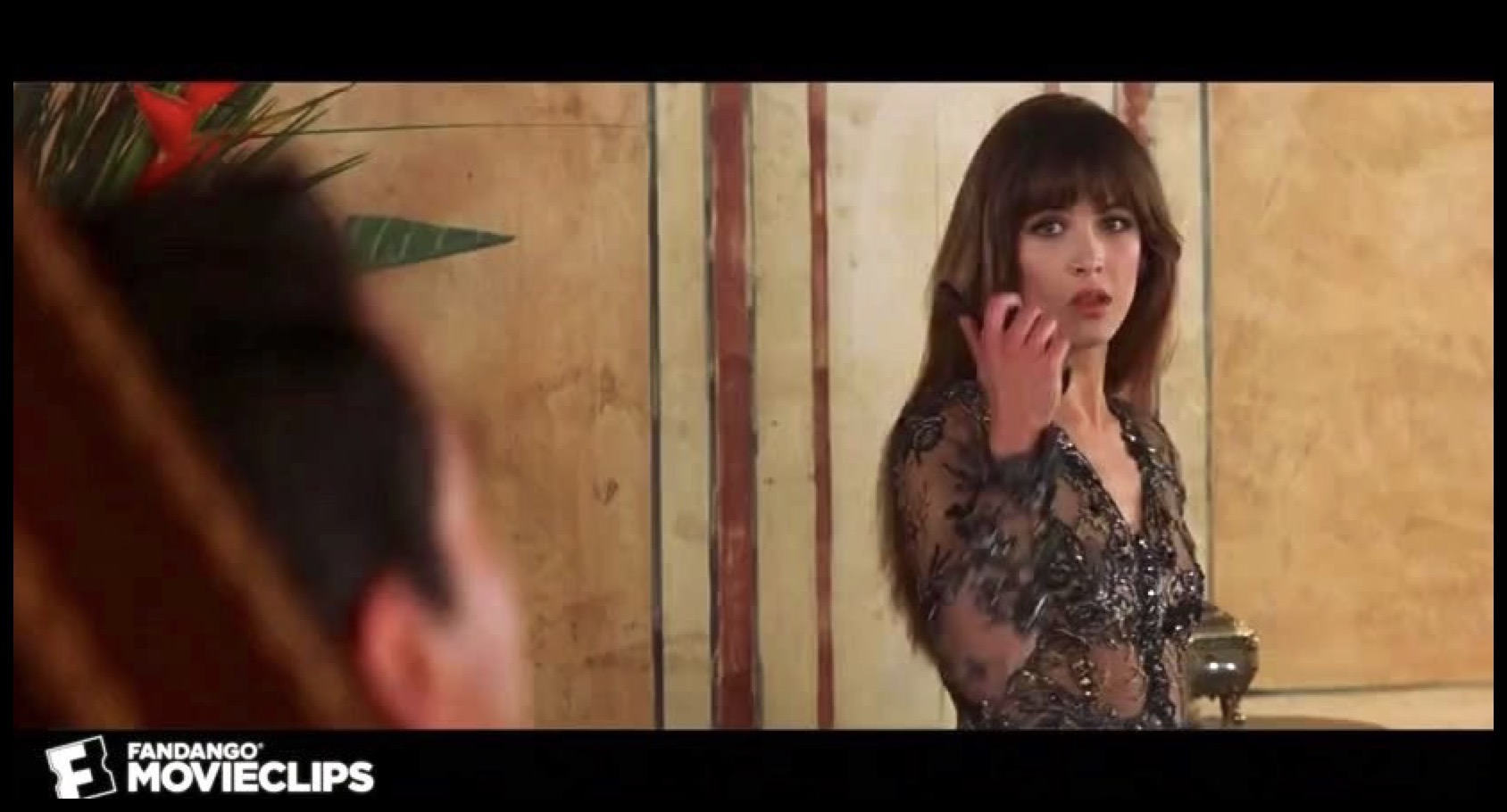}\\
\imgwithboxtwo{Akiva Goldsman}{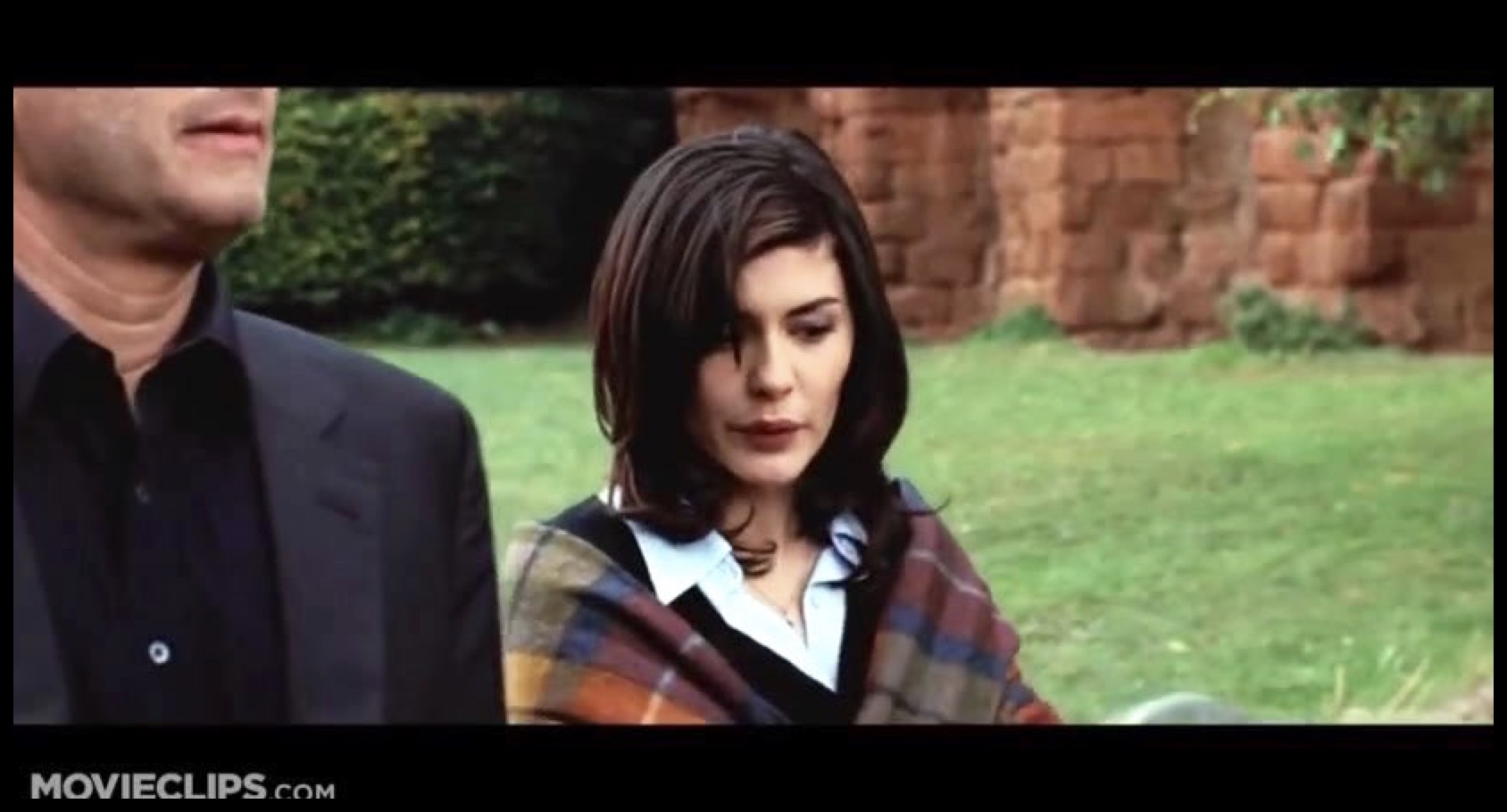}\\
\imgwithboxtwo{no writer}{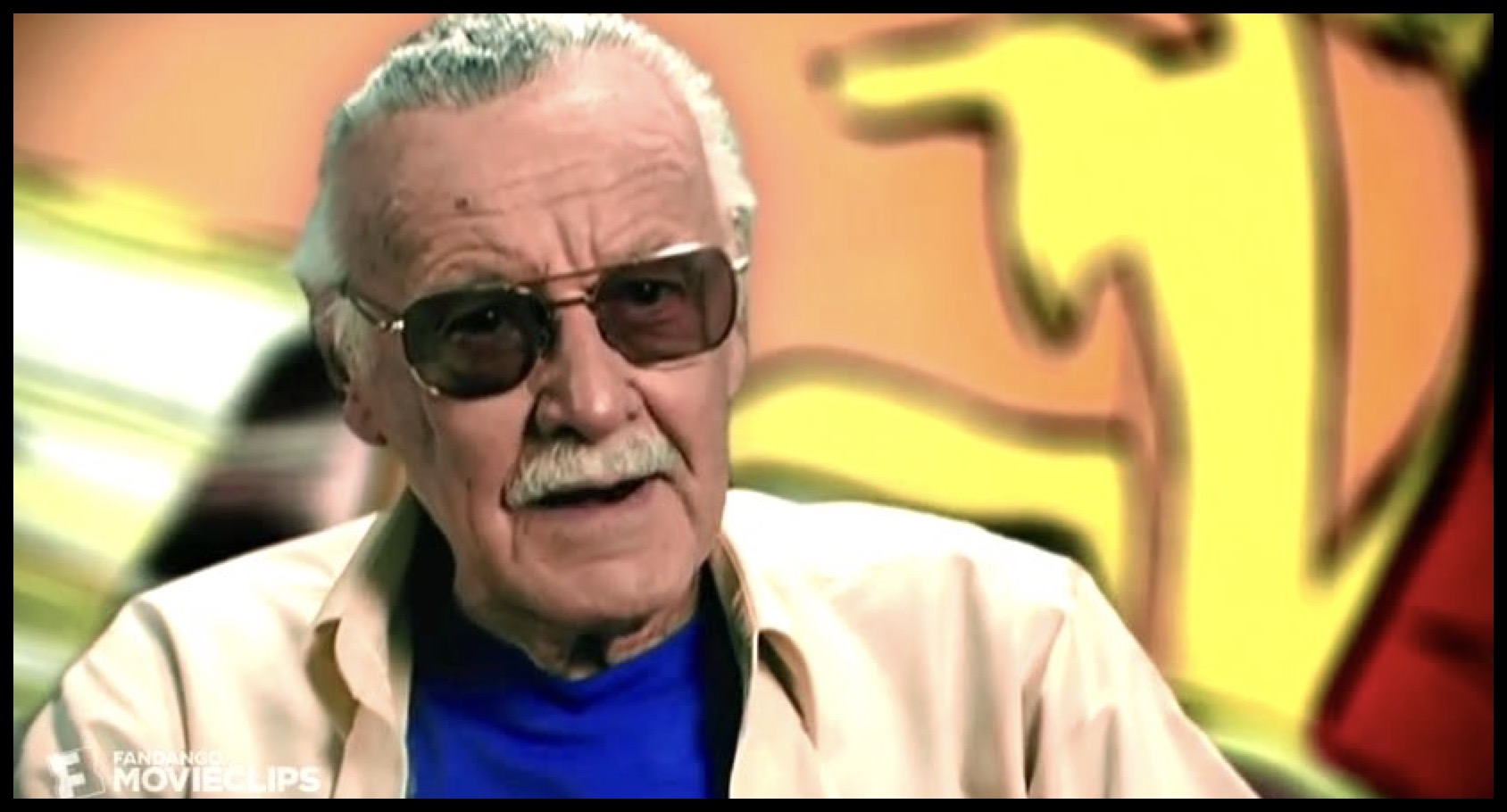}
\end{tabular}
\end{minipage}
\begin{minipage}[t]{68.6pt}
\tablestyle{0.0pt}{0.0}
\begin{tabular}[t]{@{}x{68.6}@{}}
\textbf{Year}\\
\imgwithboxtwo{1930s}{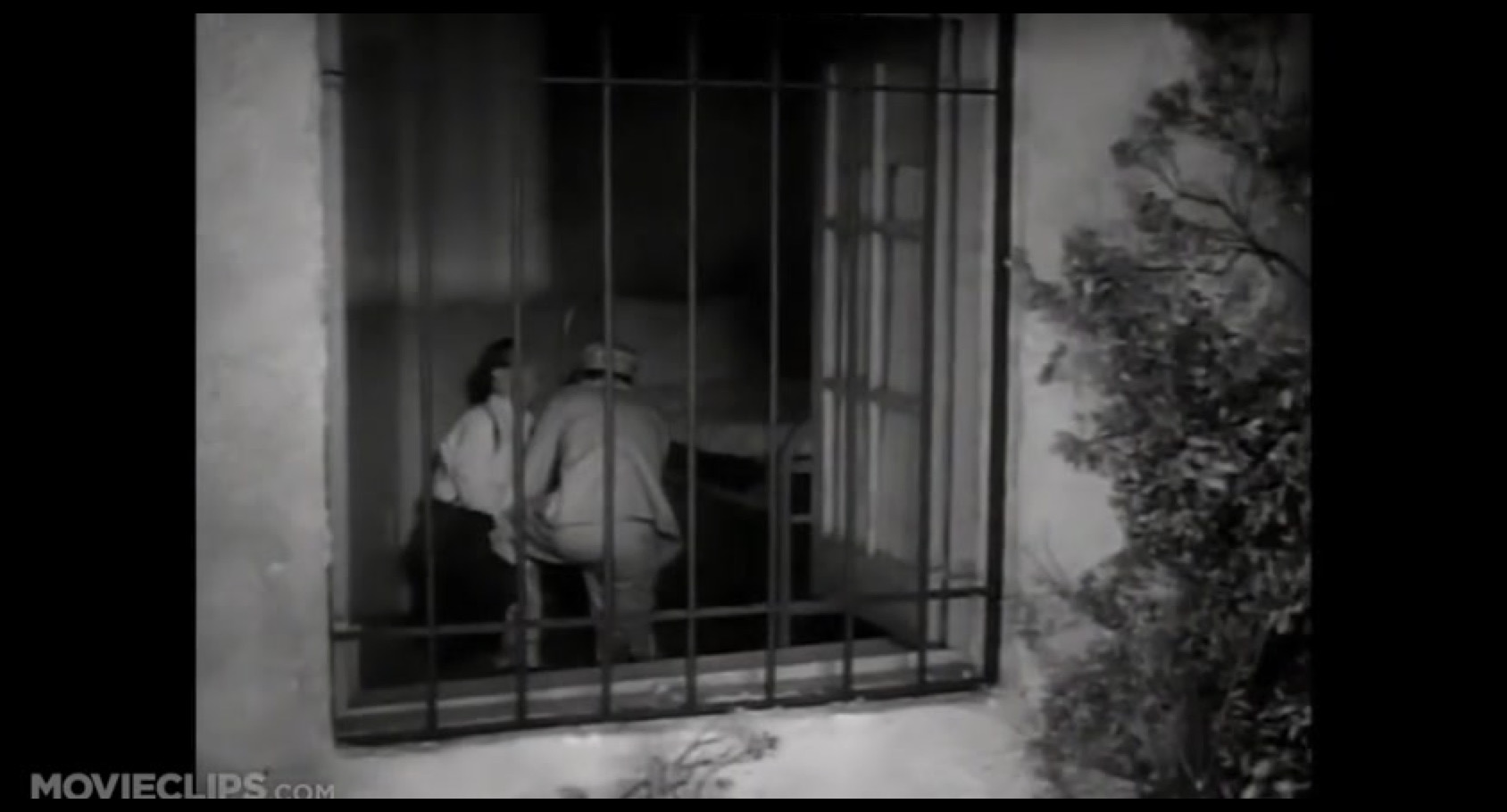}\\
\imgwithboxtwo{1940s}{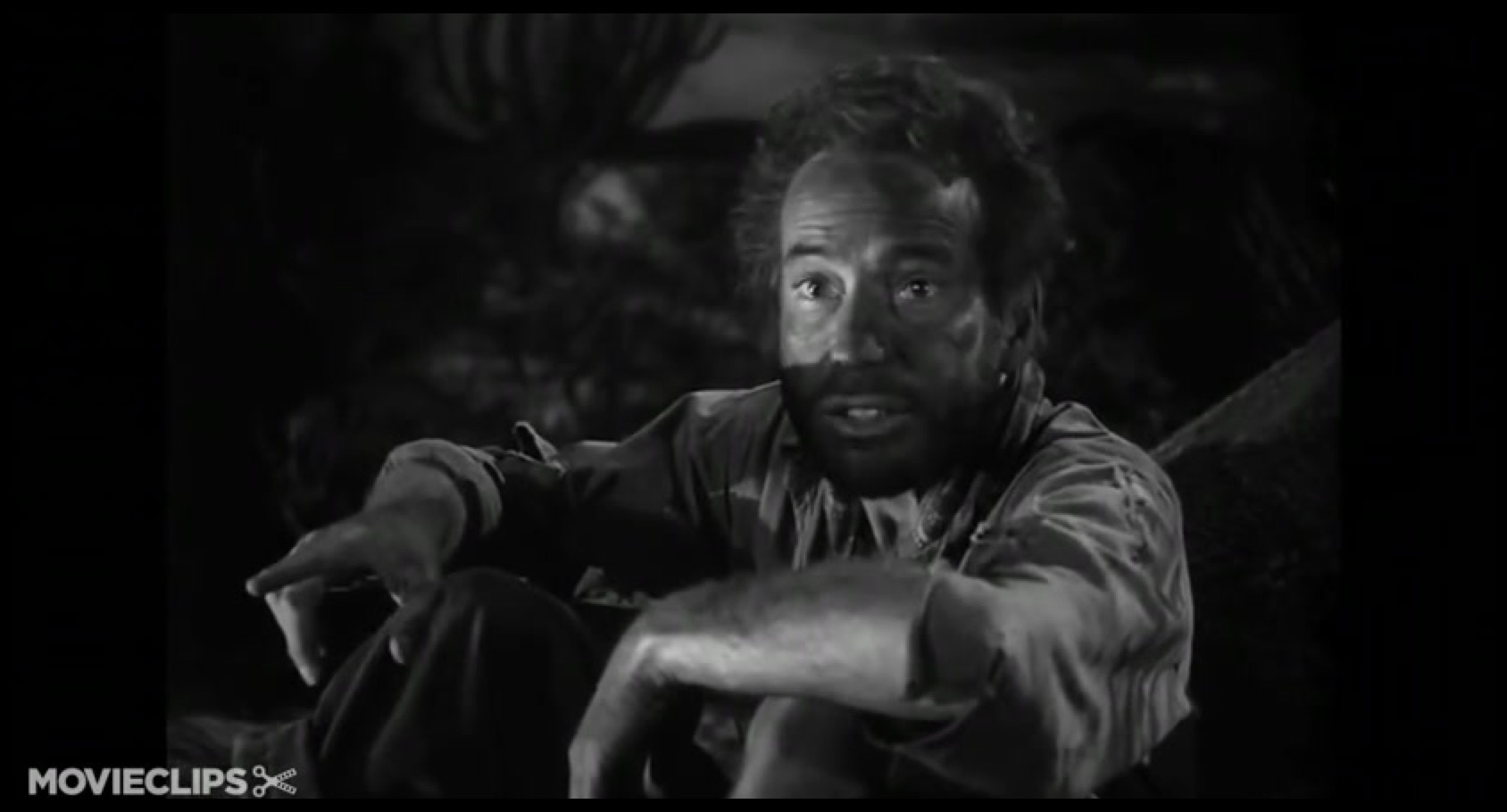}\\
\imgwithboxtwo{1950s}{figs/tasks/jpegs/y_1950.jpg}\\
\imgwithboxtwo{1960s}{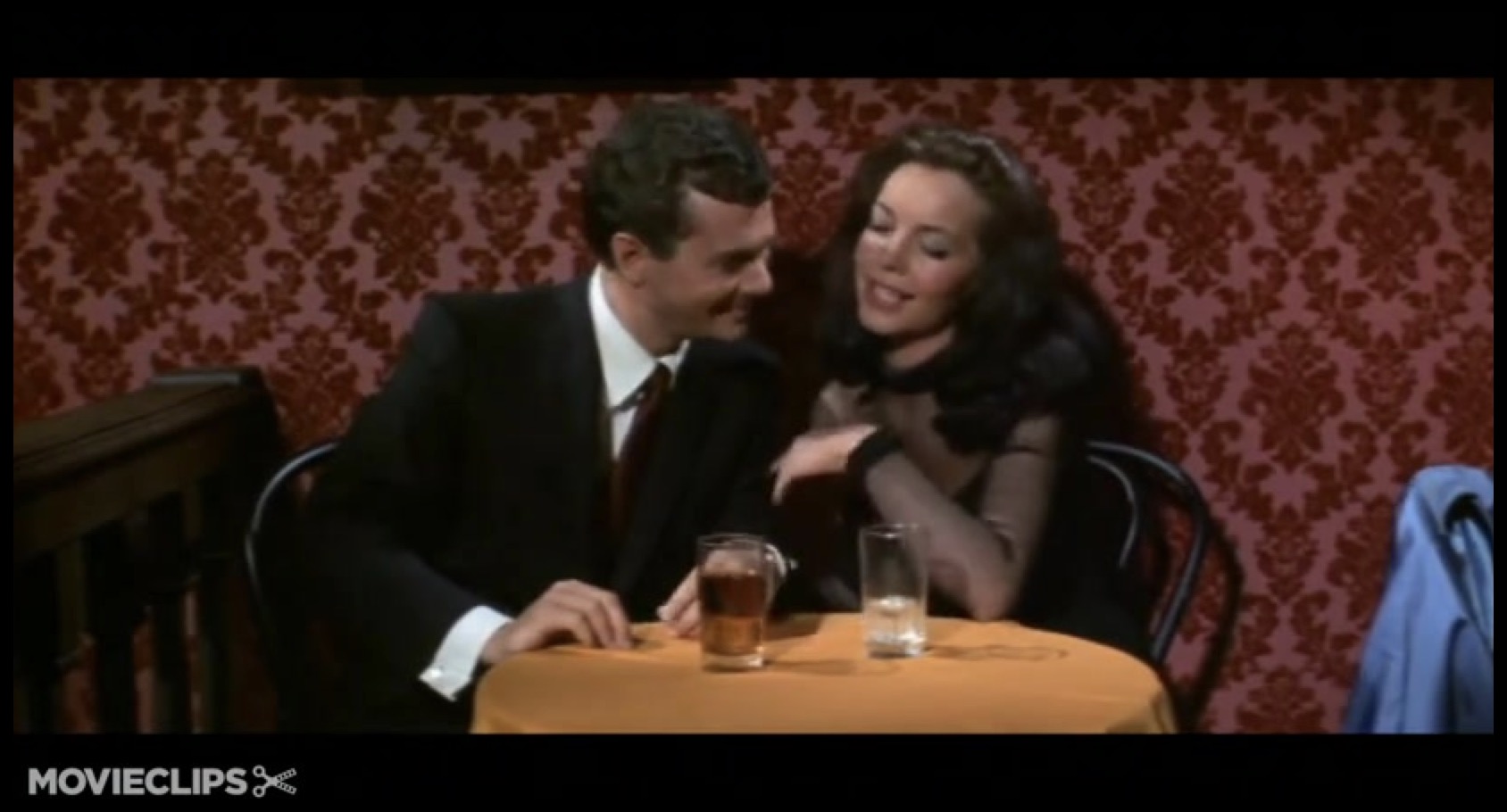}\\
\imgwithboxtwo{1970s}{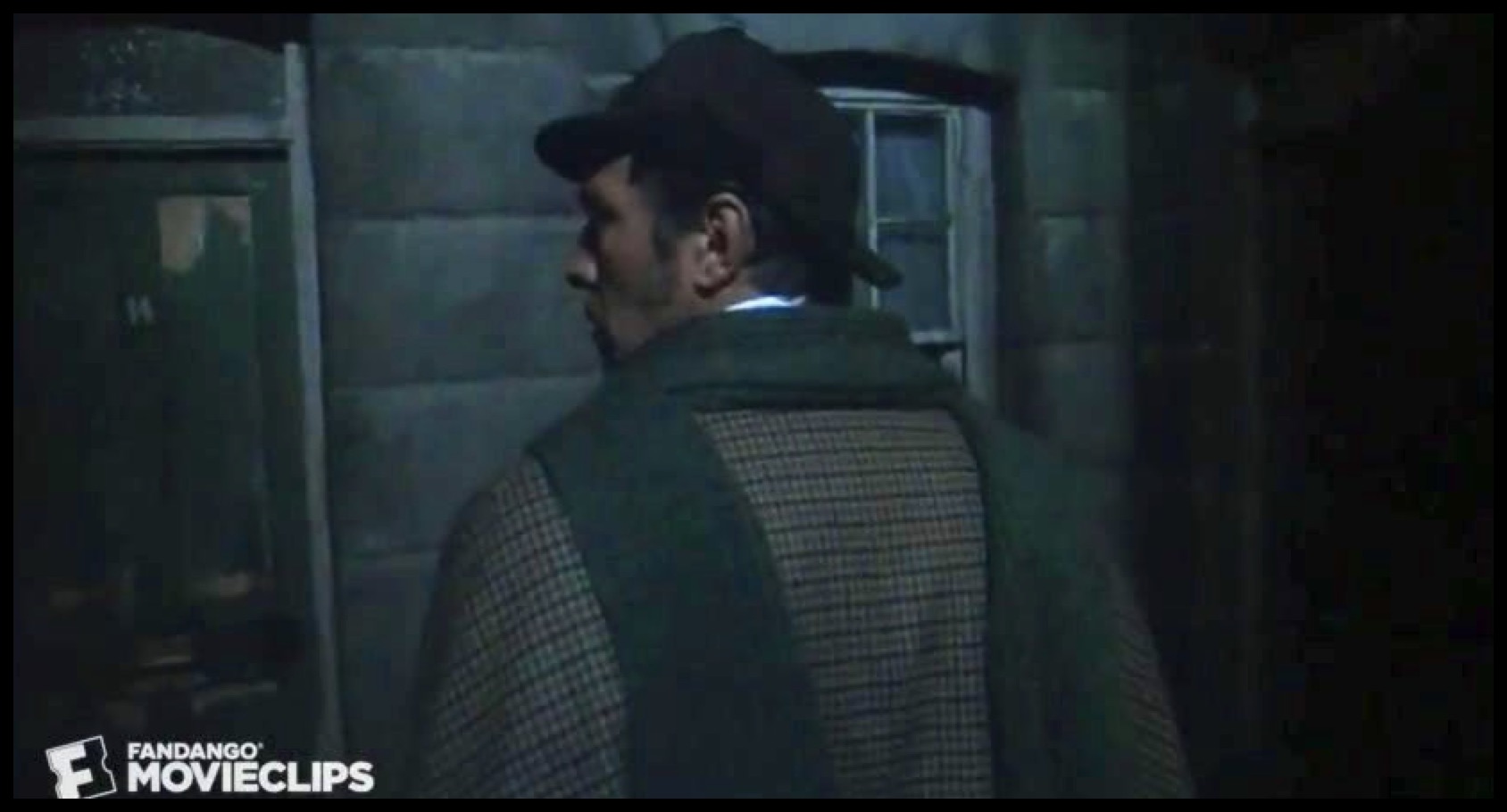}\\
\imgwithboxtwo{1980s}{figs/tasks/jpegs/y_1980.jpg}\\
\imgwithboxtwo{1990s}{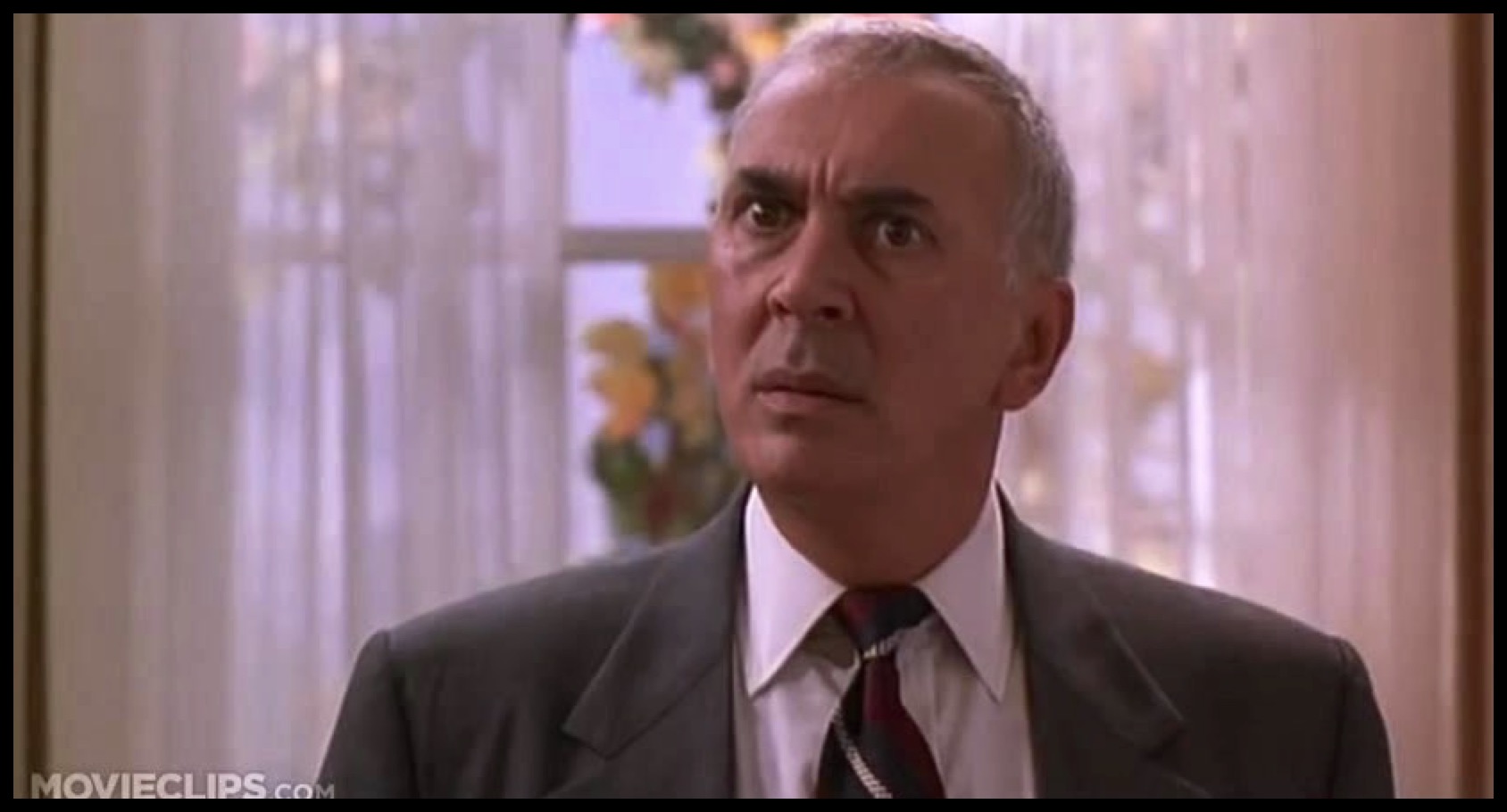}\\
\imgwithboxtwo{2000s}{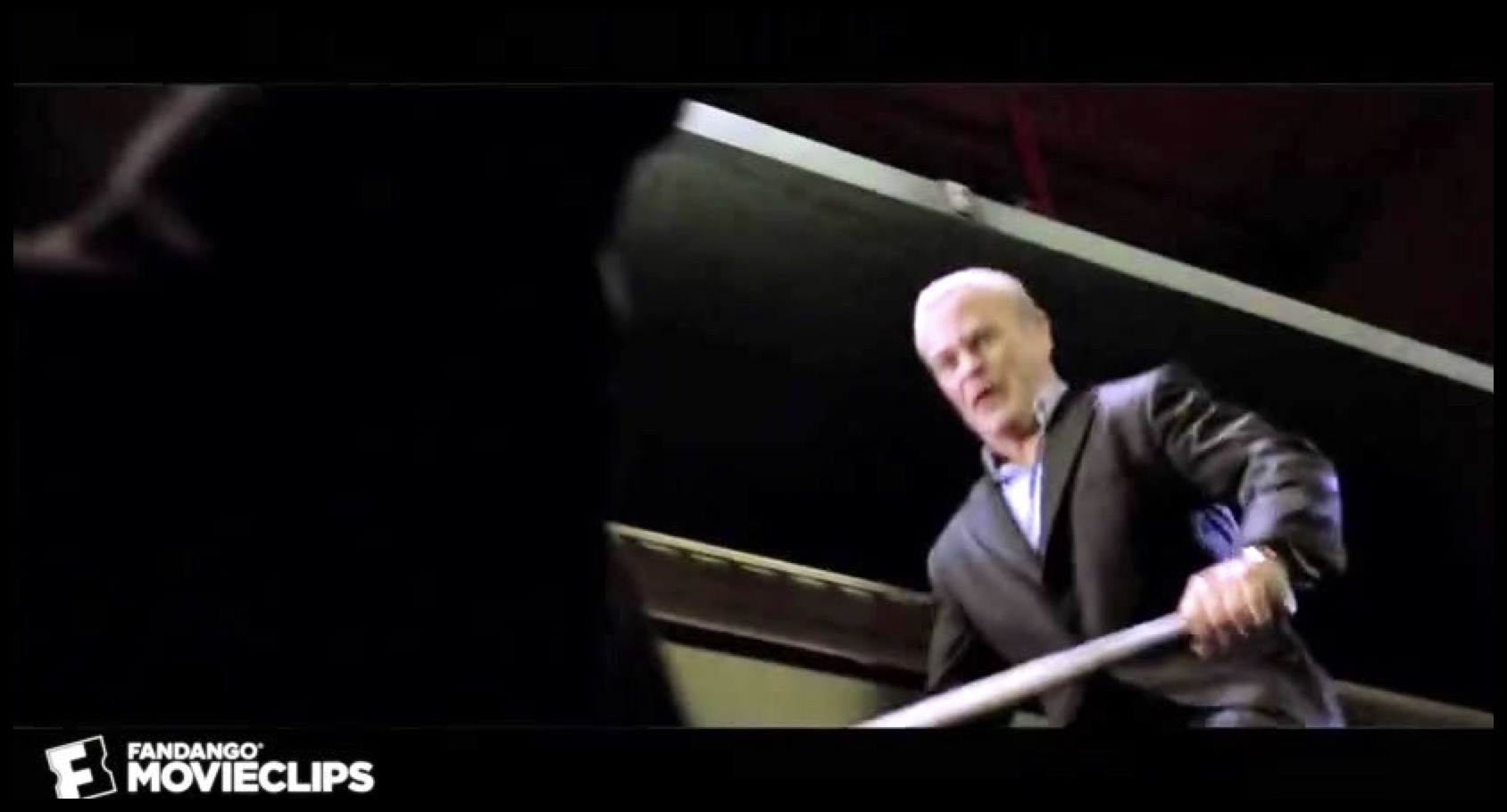}\\
\imgwithboxtwo{2010s}{figs/tasks/jpegs/y_2010.jpg}
\end{tabular}
\end{minipage}
\vspace{5mm}
\caption{\textbf{Additional Examples for Classification Tasks.}
Here we present example frames for all the classes in each classification task.
(Best viewed on screen.)\vspace{-2mm}}\label{fig:tasks}
\end{figure*}

\section{Supplementary Dataset Details}
Here we provide additional details for the 9 LVU tasks.
\begin{itemize}[topsep=1mm,partopsep=0px,leftmargin=*,]
	\setlength\itemsep{0.0em}
	\item \textbf{Relationship prediction} is a 4-way classification task over `\emph{friends}', `\emph{wife-and-husband}', `\emph{boyfriend-and-girlfriend}', and `\emph{ex-boyfriend-and-ex-girlfriend}'.
	The ground-truth labels are mined from the description associated with each video.
	For example, given the description, `\emph{Rosemary (Mia Farrow) and her husband (John Cassavetes) quarrel about doctors; she feels the baby kicking.}', we can infer the `\emph{wife-husband}' relationship for this video.
	This task contains 226 videos.
	\item \textbf{Way of speaking prediction} is a 5-way classification task over `\emph{explain}', `\emph{confront}', `\emph{discuss}', `\emph{teach}', and `\emph{threaten}'. 
	The labels are mined analogously to the relationship prediction task. 
	This task contains 1,345 videos.
	\item \textbf{Scene/place prediction} is a 6-way classification task over `\emph{office}', `\emph{airport}', `\emph{school}', `\emph{hotel}', `\emph{prison}', and `\emph{restaurant}'. 
	The labels are mined analogously to the relationship prediction task. 
	This task contains 723 videos.
	\item \textbf{Director prediction} is an 8-way classification task over `\emph{Ron Howard}', `\emph{Martin Scorsese}', `\emph{Steven Spielberg}', `\emph{Quentin Tarantino}', `\emph{Ridley Scott}', `\emph{Peter Jackson}', `\emph{Robert Rodriguez}', and `\emph{Mark Atkins}'.
	These classes correspond to the 10 most frequent directors in our dataset, excluding Ethan Coen and Joel Coen.
	The Coen brothers co-direct frequently; we remove them to set up a single-label task.
	This task contains 950 videos. 
	\item \textbf{Writer prediction} is 7-way classification task over `\emph{Stephen King}', `\emph{Sylvester Stallone}', `\emph{John Hughes}', `\emph{Ian Fleming}', `\emph{Akiva Goldsman}', `\emph{David Koepp}', and `\emph{no writer}' (\eg, documentary). 
	They correspond to the 10 most frequent writers in our dataset, excluding Ethan Coen and Joel Coen (due the same reason we discussed above) and Richard Maibaum, whose movies largely overlap with Ian Fleming movies.
	This task contains 1,111 videos in total.
	\item \textbf{Genre prediction} is a 4-way classification task over `\emph{Action/Crime/Adventure}', `\emph{Thriller/Horror}', `\emph{Romance}', and `\emph{Comedy}'. 
	The labels are obtained through IMDb.
	We exclude videos that belong to more than one of these genres.
	There are 4,307 videos in this task.
	\item \textbf{Year prediction} is a 9-way classification task over the movie release ``decades", in `\emph{1930s}', `\emph{1940s}', \ldots, `\emph{2010s}'. 
	The labels are obtained through IMDb.
	This task contains 1,078 videos.
	\item \textbf{YouTube like ratio prediction} is regression task to predict how much a video is ``liked", namely,
	$\frac{likes}{likes + dislikes} \cdot 10$.\footnote{We scale the target by 10 (thus the target is in [0, 10]) for consistency with recommendation system literatures (\eg, \cite{diao2014jointly}), where the scale of ratings are often in [0, 10].}
	We access the like and dislike counts using YouTube Data API V3\footnote{\url{https://developers.google.com/youtube/v3}} on August 4th, 2020.
	We use videos with at least 30,000 votes in this task.
	The most liked video is the `\emph{A Pocketful of Sunshine}' scene from movie `\emph{Easy A (2010)}'\footnote{\url{https://www.youtube.com/watch?v=ylvh800i85I}} with 32,967 likes and 242 dislikes.
	The least liked video is the `\emph{Seeing an Old Friend}' scene from movie `\emph{Enter the Ninja (1981)}'\footnote{\url{https://www.youtube.com/watch?v=G_1jQkCRF58}} with 18,595 likes and 15,432 dislikes.
	This task contains 940 videos in total.
	\item \textbf{YouTube view count prediction} is a regression task.
	The view counts are also accessed using YouTube API V3 on August 4th, 2020.
	Since the view counts follow a long-tail distribution,
	we predict $\log(views)$ in this task.
	To control the effect of video upload time, all videos in this task were uploaded to YouTube in the same year (2011).
	The most viewed video is the `\emph{Kong Battles the T-Rexes}' scene from `\emph{King Kong (2005)}'\footnote{\url{https://www.youtube.com/watch?v=ZYZsJYZVt5g}} with 132,862,771 views. 
	The least viewed video is the `\emph{Margo to the Rescue}' scene from `\emph{The Locksmith (2010)}'\footnote{\url{https://www.youtube.com/watch?v=oGKr8bdx_5E}} with 531 views.
	This task contains 827 videos.
\end{itemize}
For the three content tasks, we spot-checked the training set
and corrected any wrong labels in the validation and test sets ($\sim$1 in 9).
\figref{tasks} presents example frames for all the classes in each classification task.

\section{Qualitative Evaluation Full Sequences}
In Fig.\ 5 of the main paper, we present subsampled frames of three examples of Masked Instance Prediction.
In \figref{fullexamples}, we present the full 60-second frames.

\newcommand{\fullex}[1]{\includegraphics[width=0.1\linewidth]{figs/viz/full_examples/#1}\hspace{-0.3mm}}
\begin{figure*}[t]
\tablestyle{0.0pt}{0.0}
\resizebox{1.00\linewidth}{!}{%
\begin{tabular}[t]{@{}c@{}}
\fullex{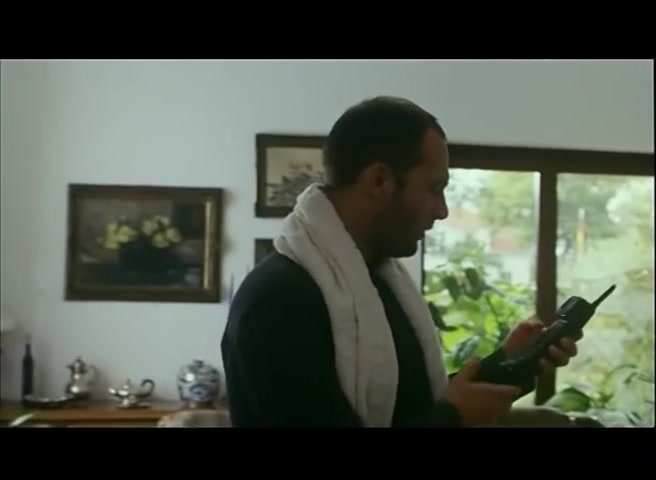}
\fullex{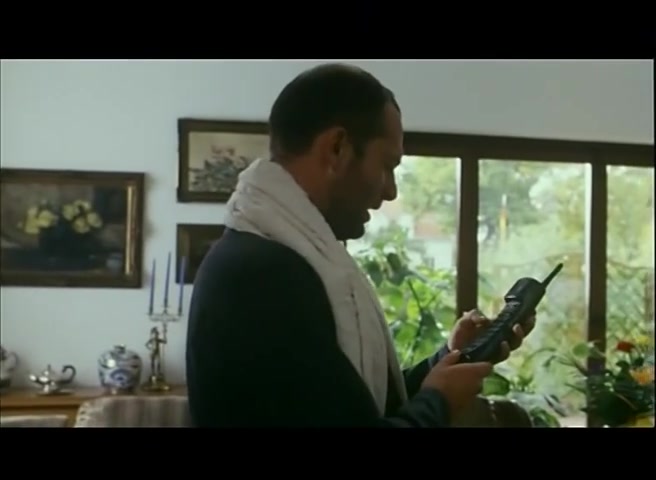}
\fullex{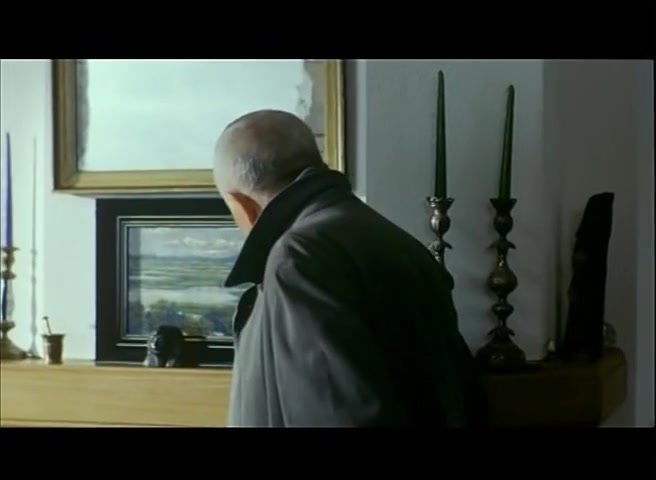}
\fullex{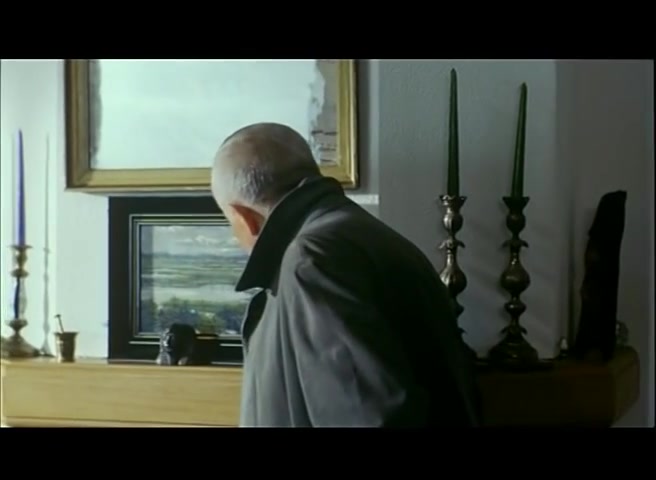}
\fullex{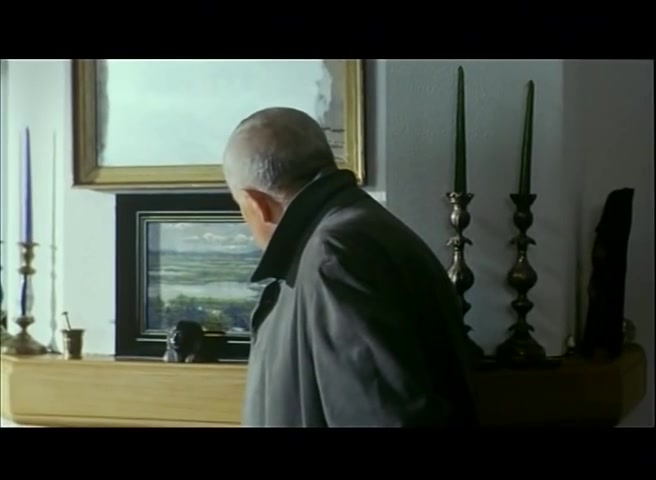}
\fullex{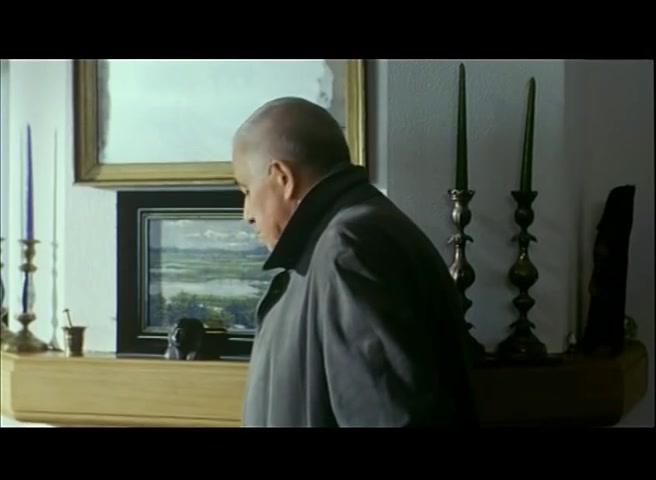}
\fullex{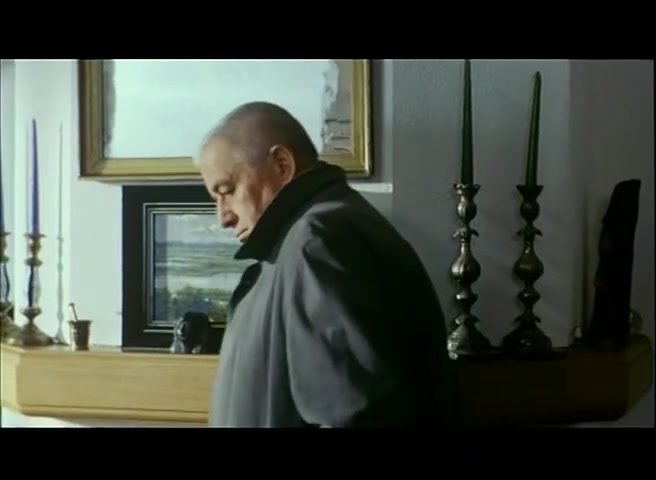}
\fullex{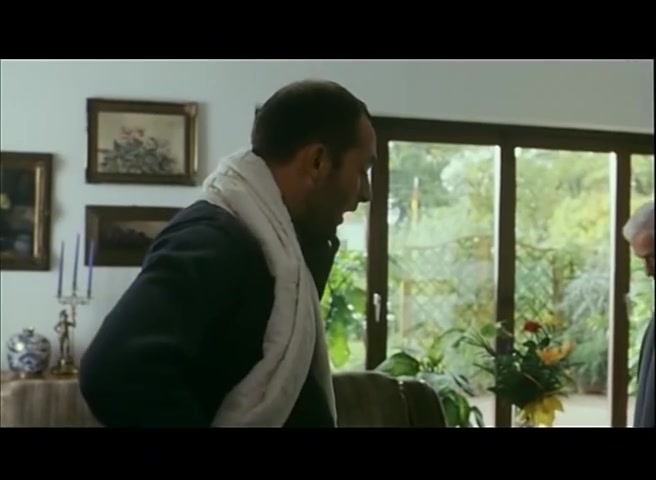}
\fullex{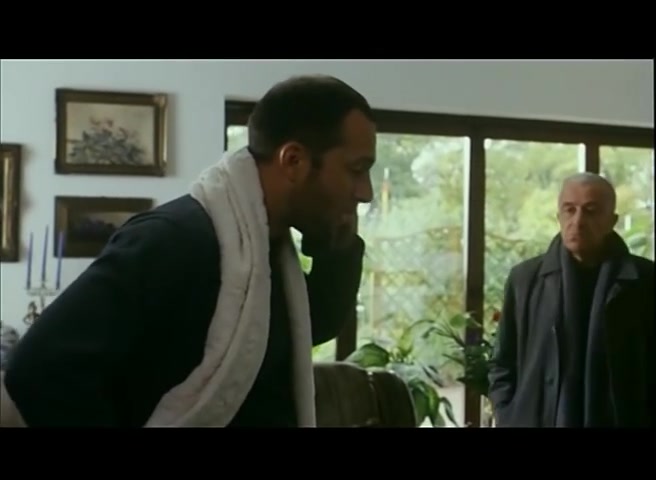}
\fullex{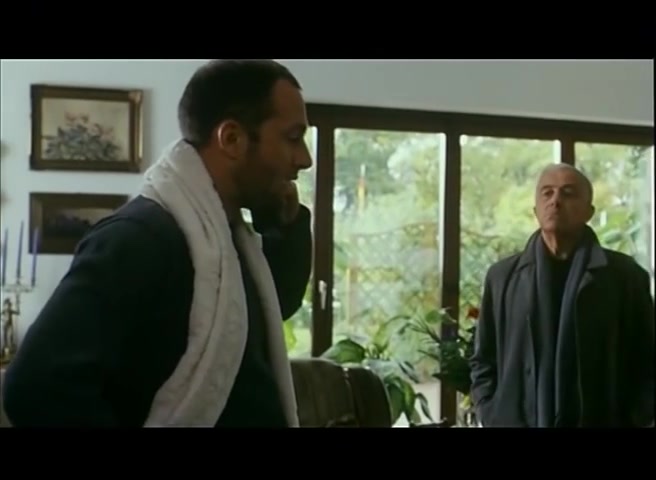}
\vspace{0.7mm}\\
\fullex{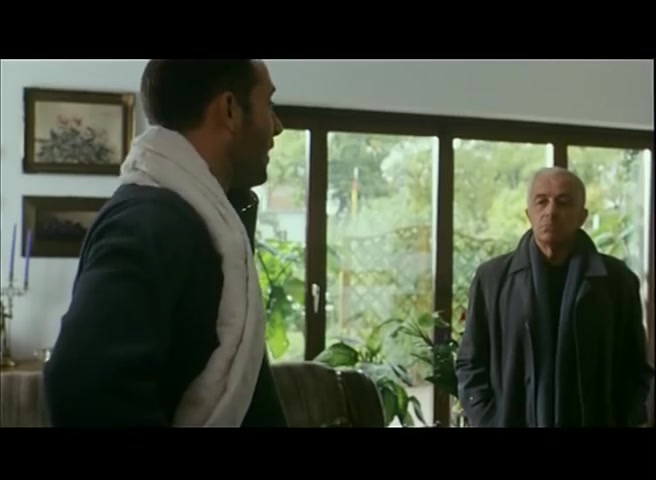}
\fullex{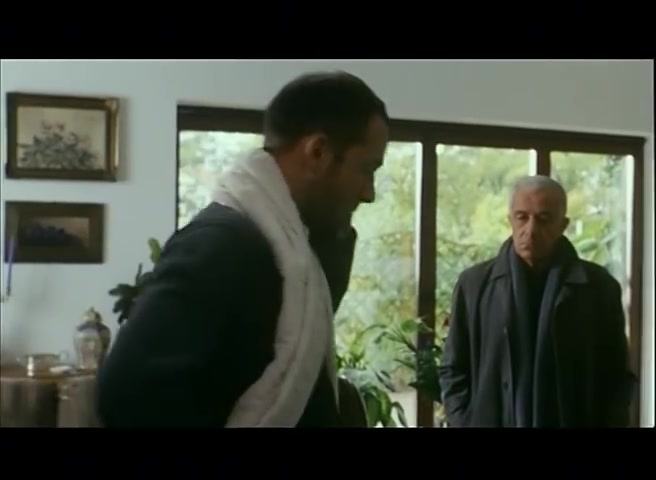}
\fullex{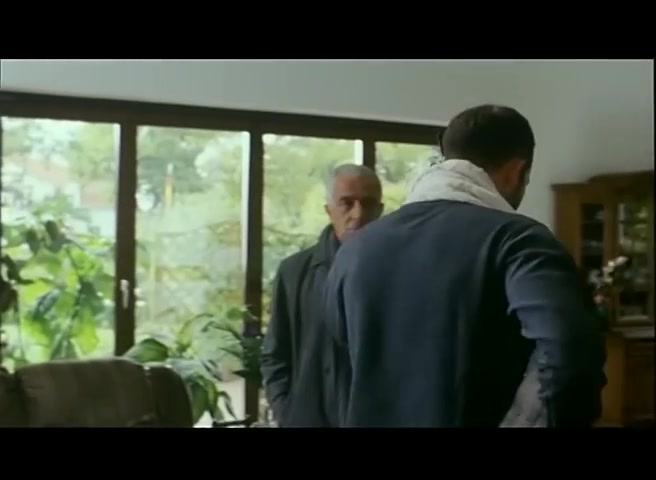}
\fullex{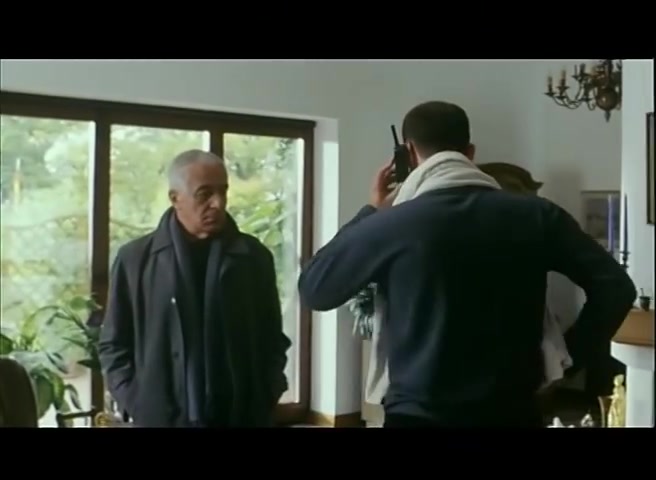}
\fullex{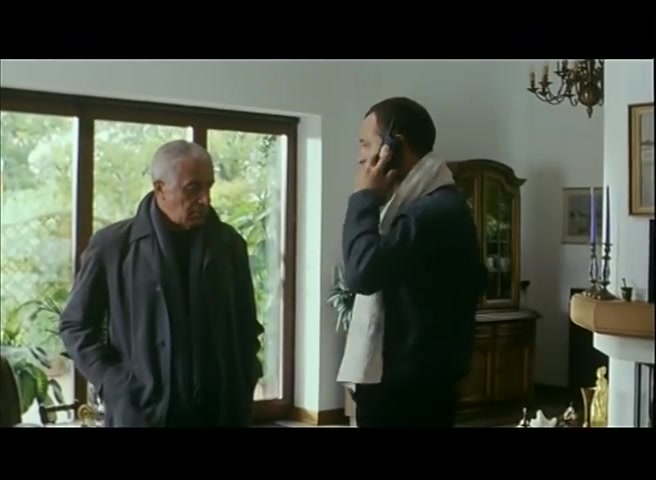}
\fullex{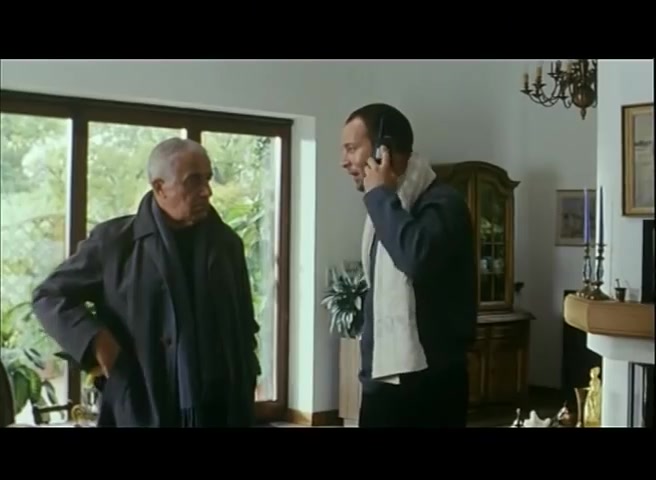}
\fullex{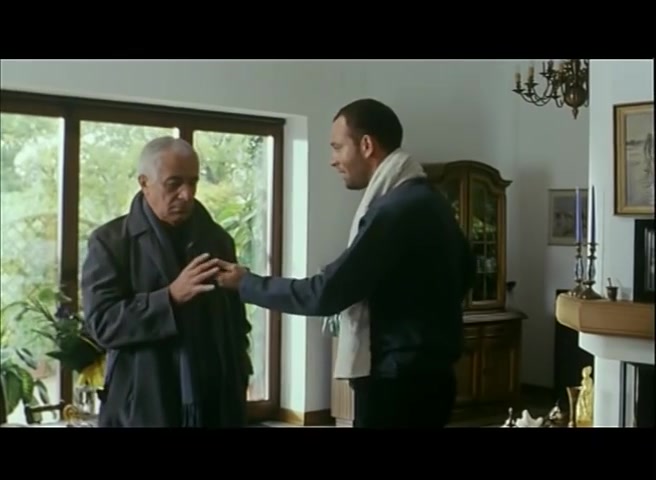}
\fullex{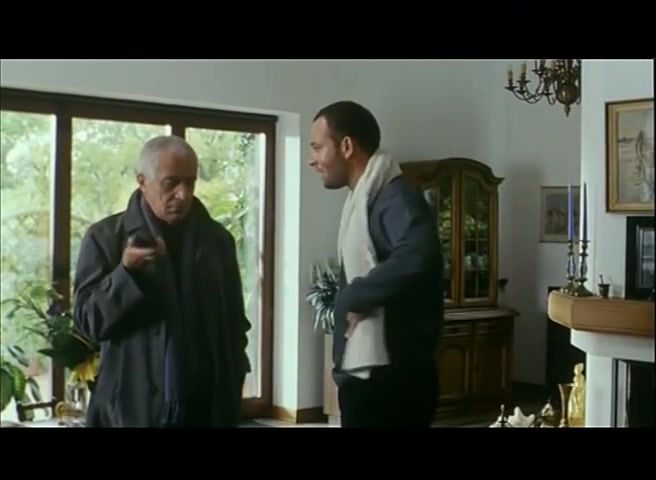}
\fullex{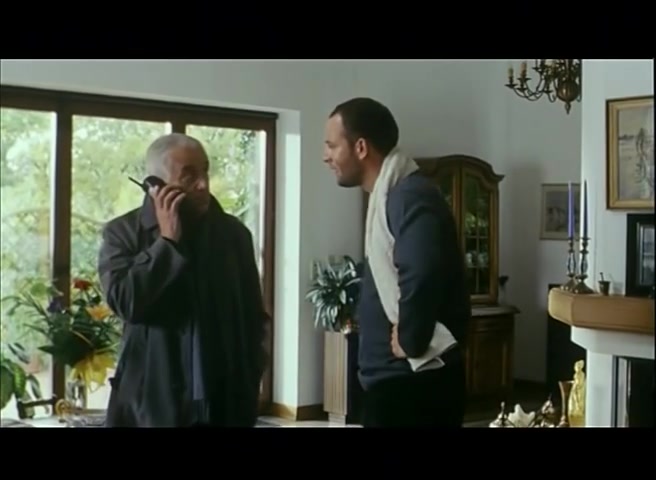}
\fullex{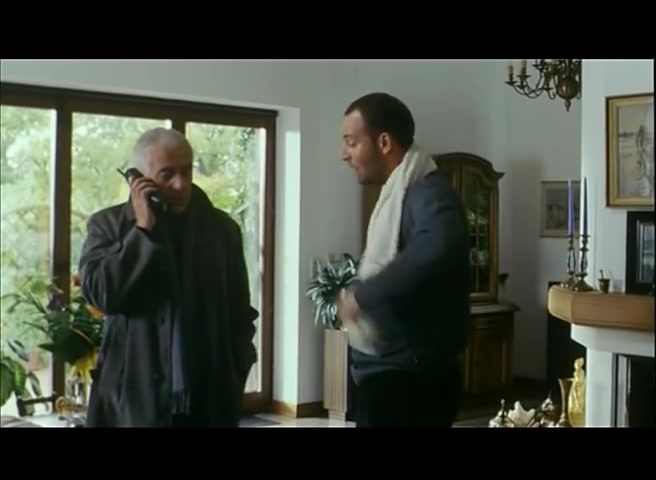}
\vspace{0.7mm}\\
\fullex{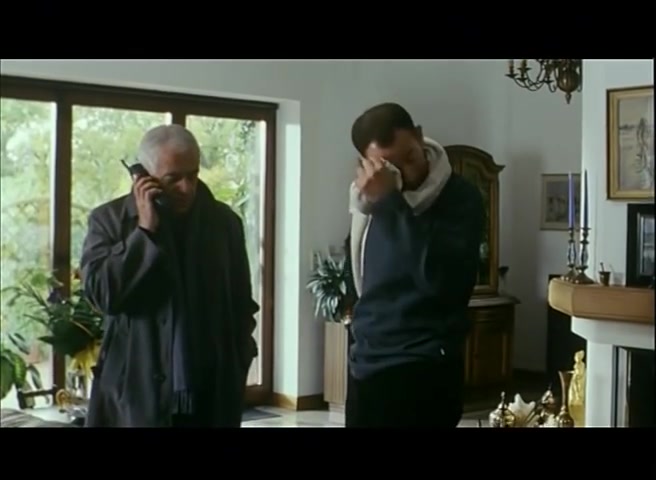}
\fullex{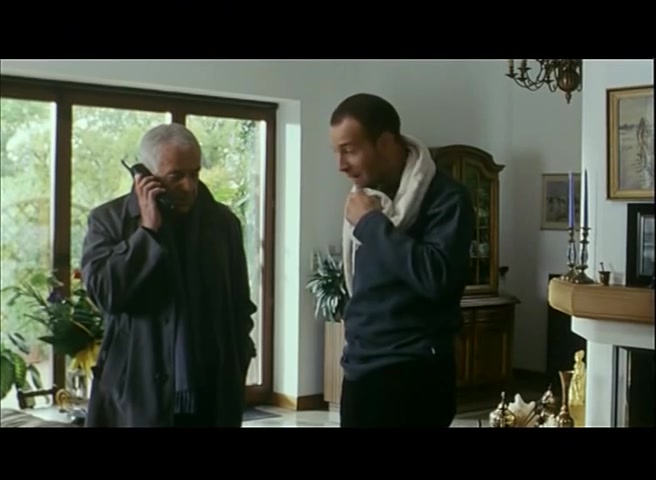}
\fullex{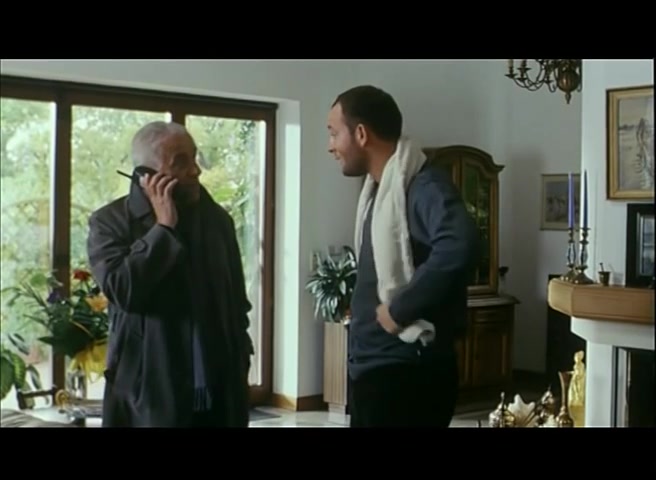}
\fullex{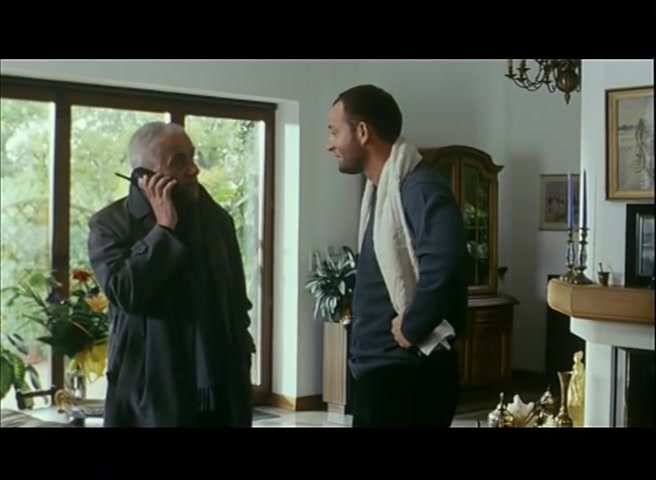}
\fullex{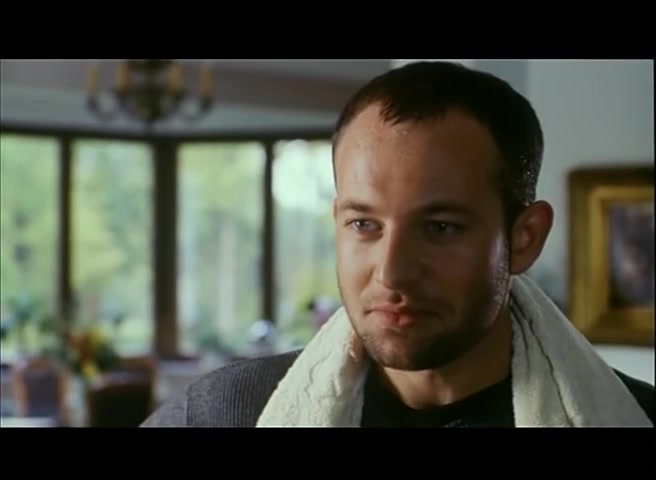}
\fullex{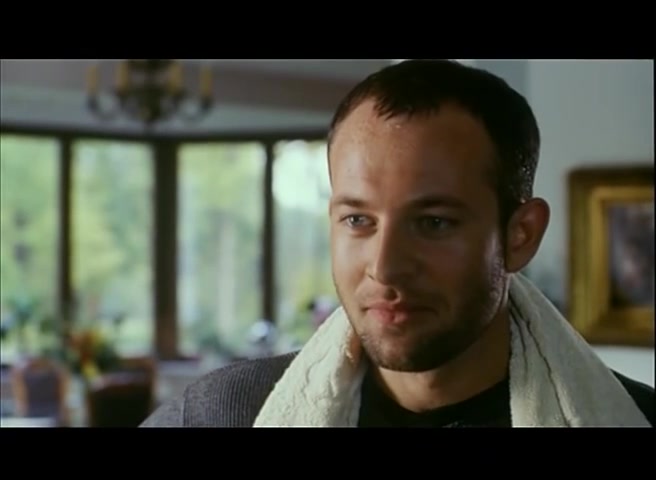}
\fullex{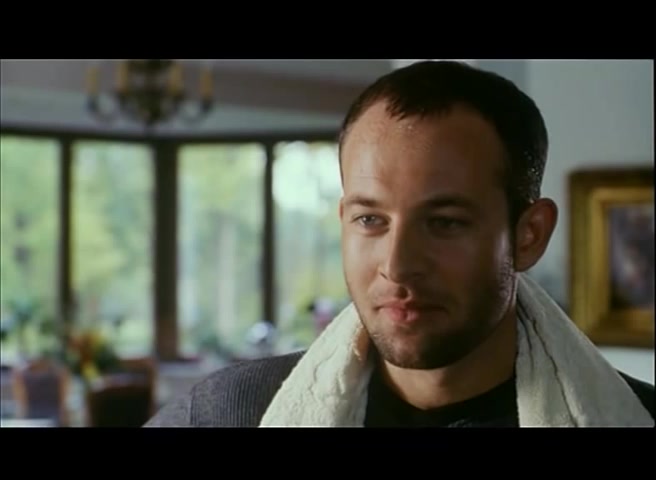}
\fullex{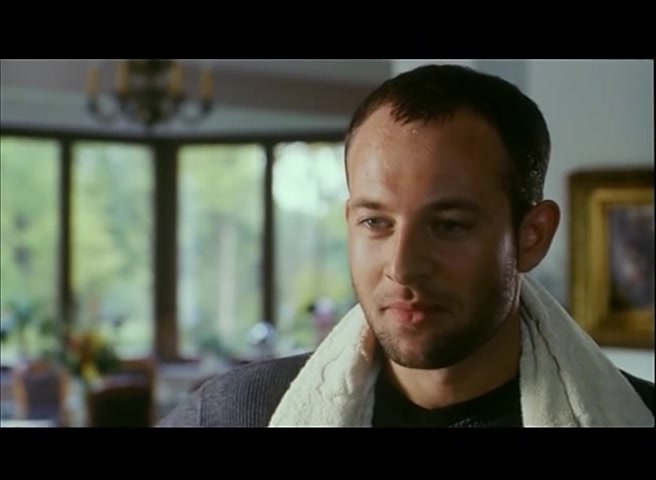}
\fullex{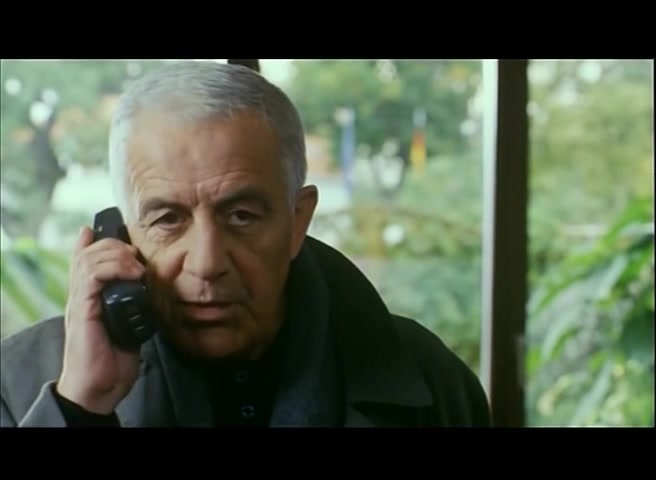}
\fullex{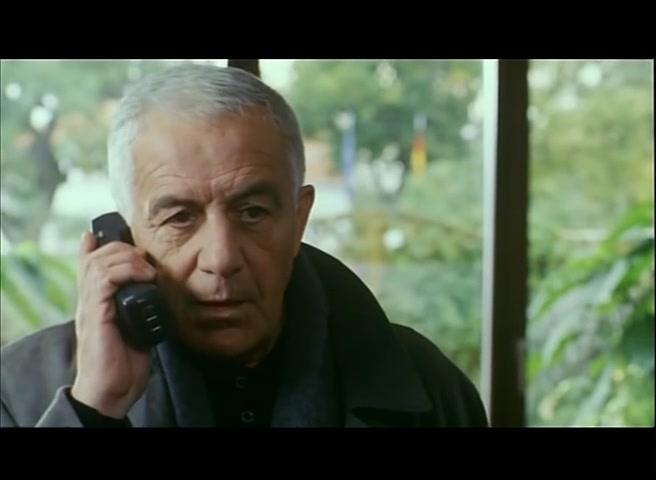}
\vspace{0.7mm}\\
\fullex{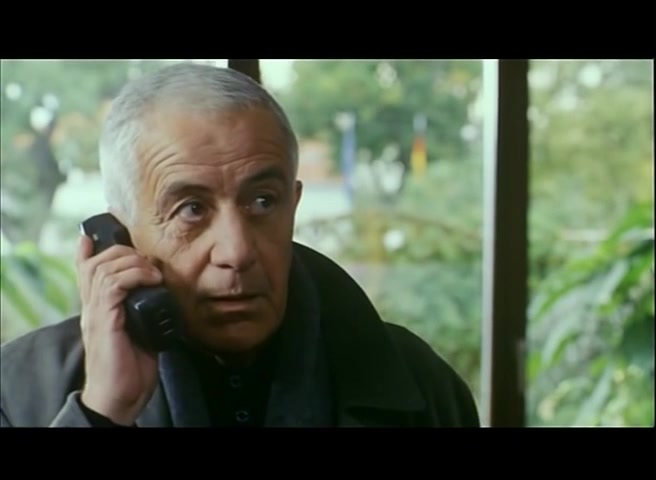}
\fullex{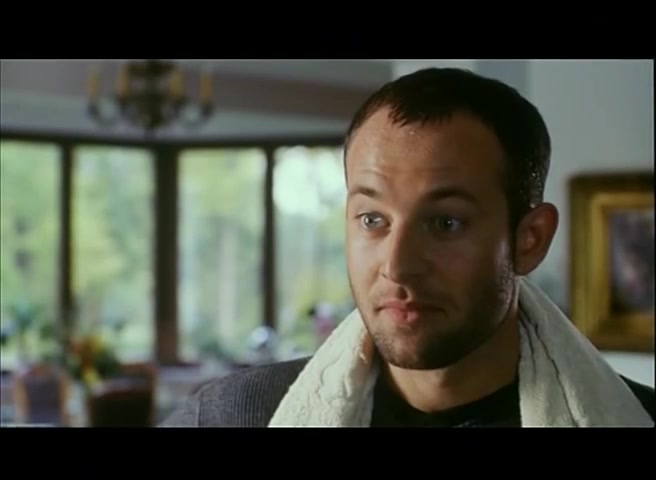}
\fullex{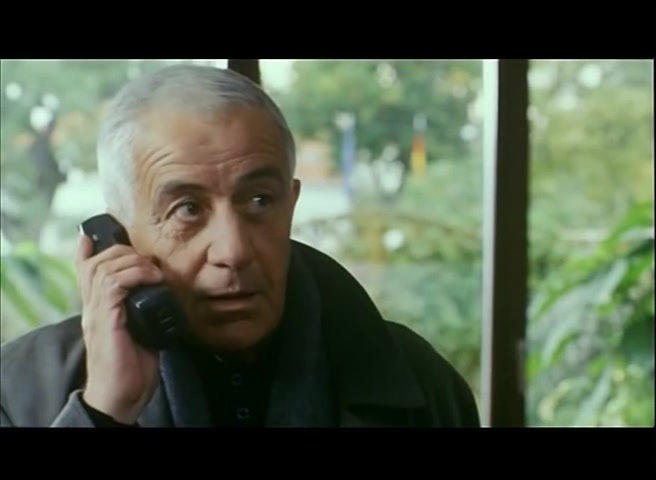}
\fullex{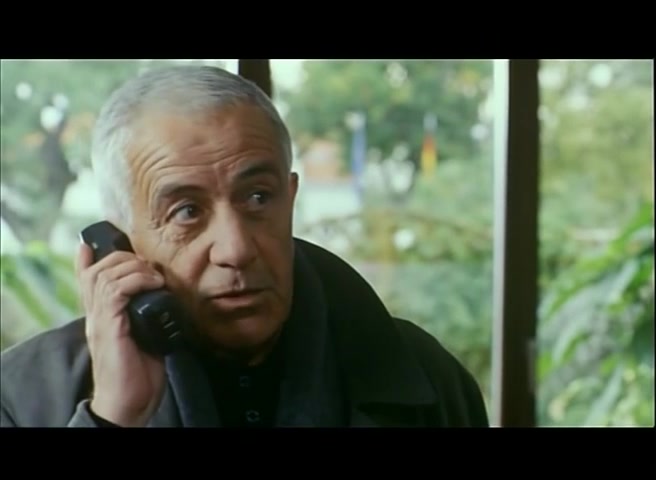}
\fullex{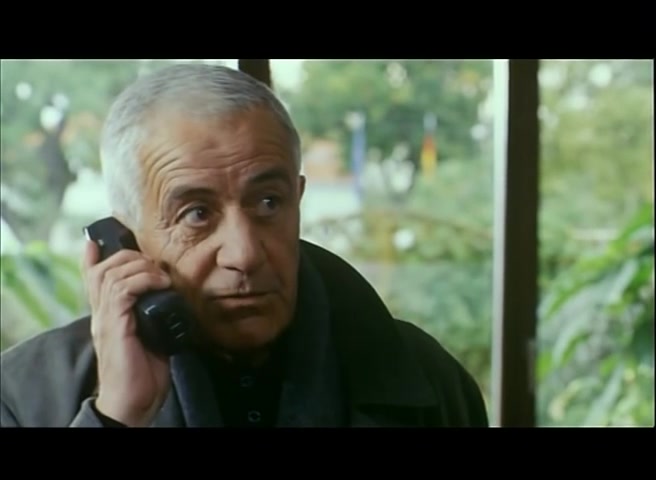}
\fullex{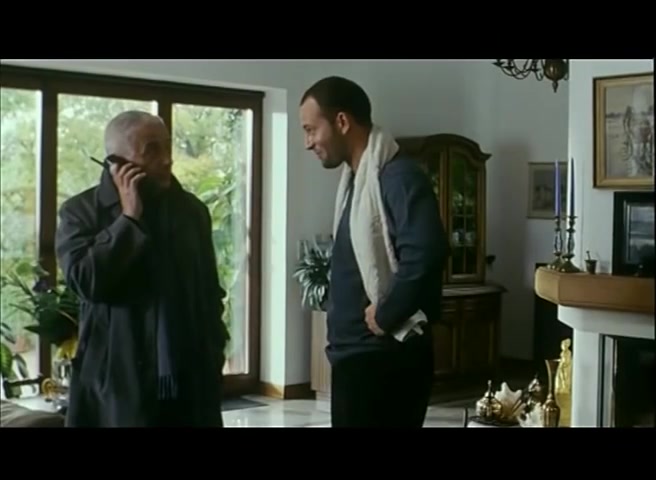}
\fullex{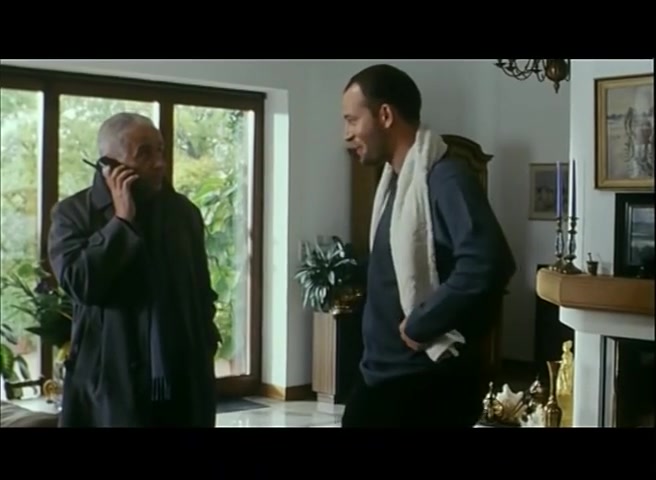}
\fullex{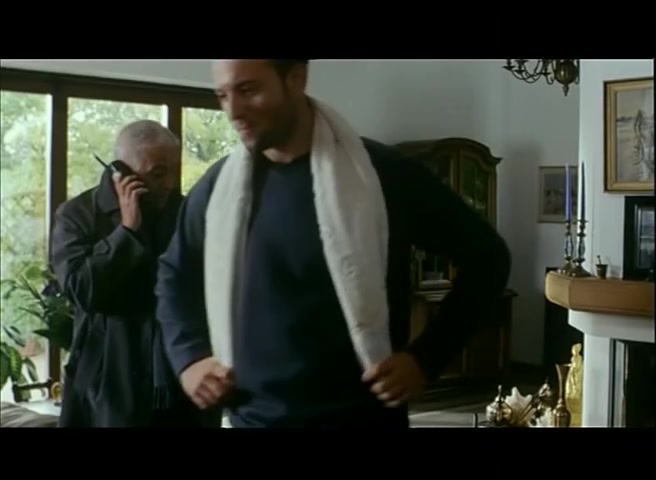}
\fullex{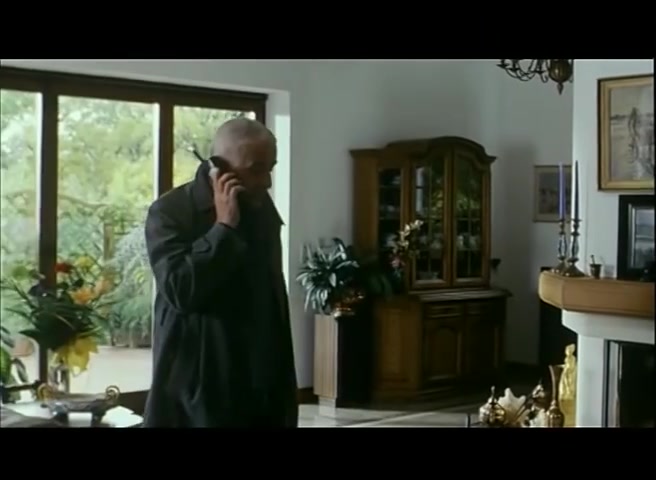}
\fullex{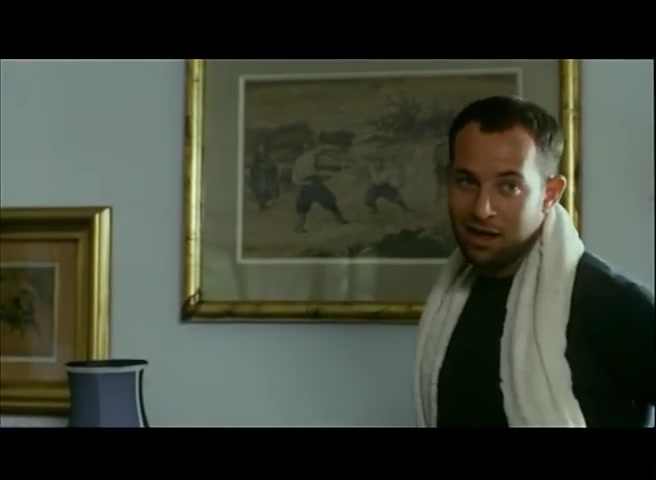}
\vspace{0.7mm}\\
\fullex{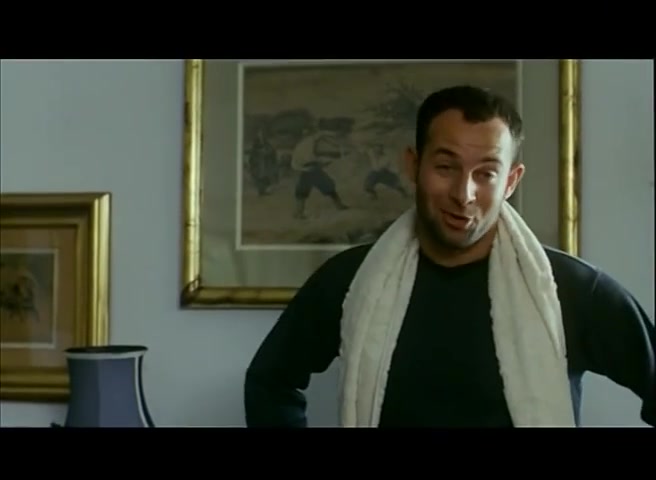}
\fullex{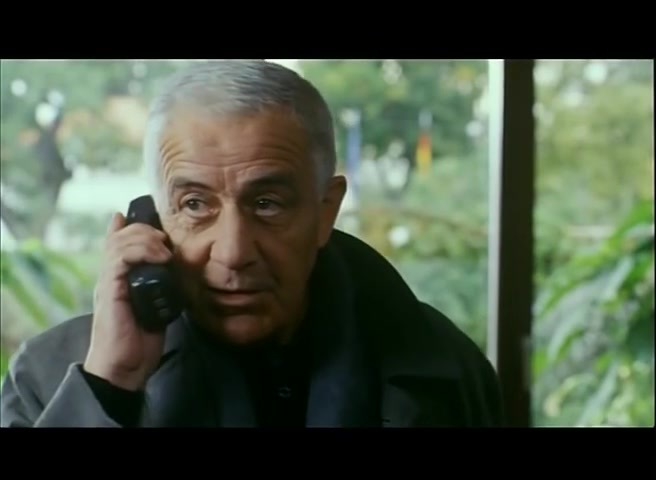}
\fullex{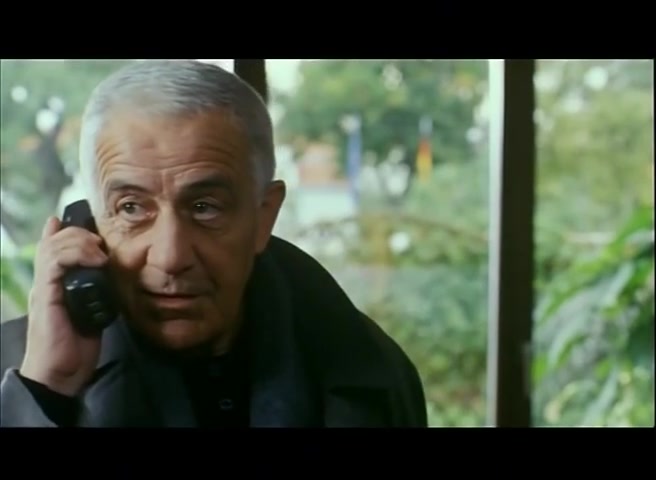}
\fullex{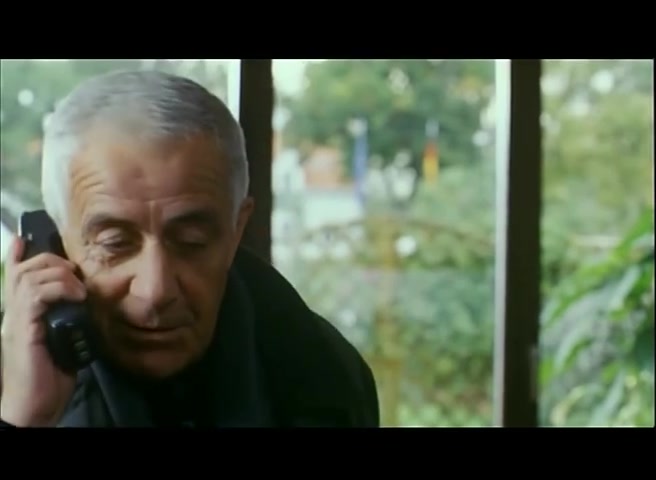}
\fullex{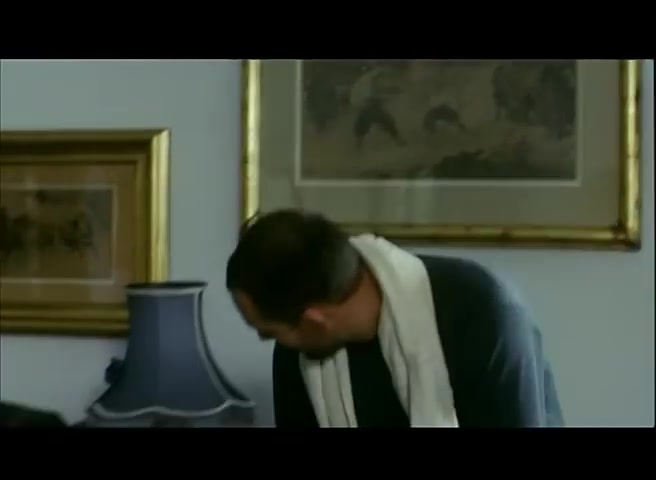}
\fullex{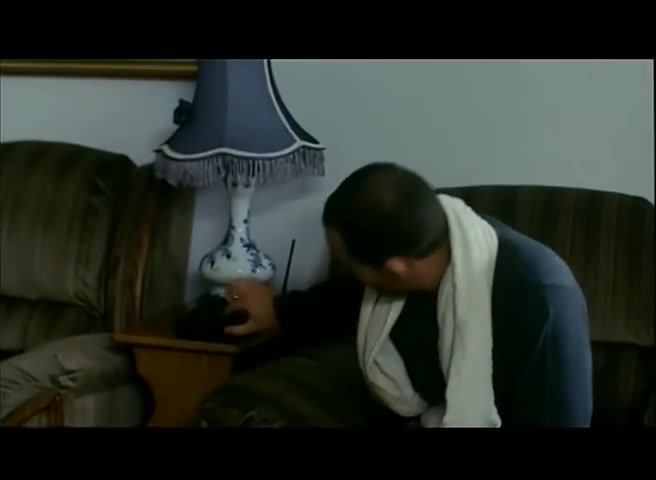}
\fullex{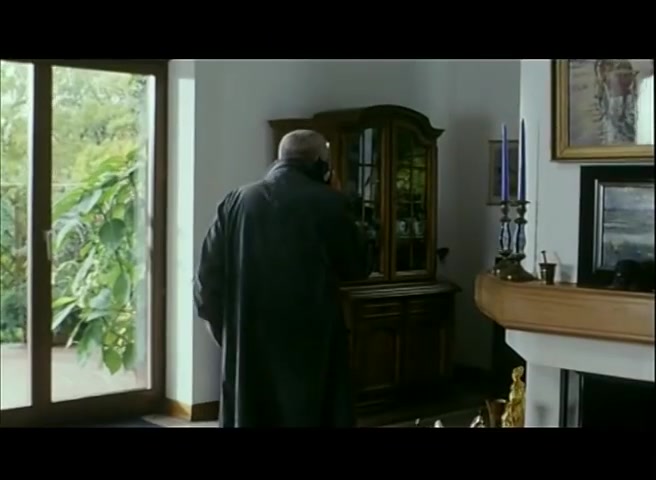}
\fullex{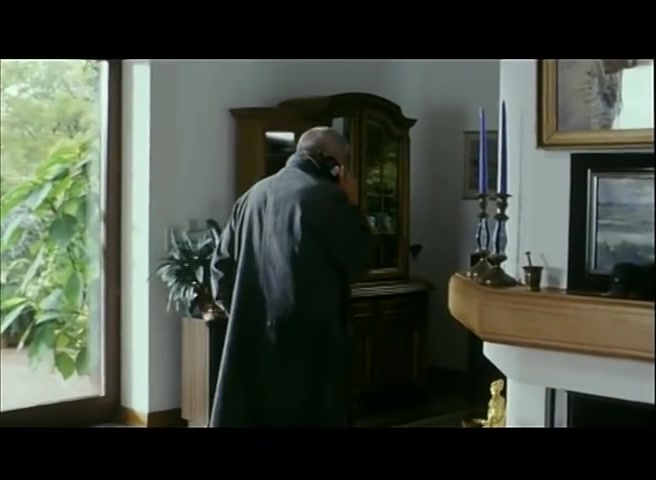}
\fullex{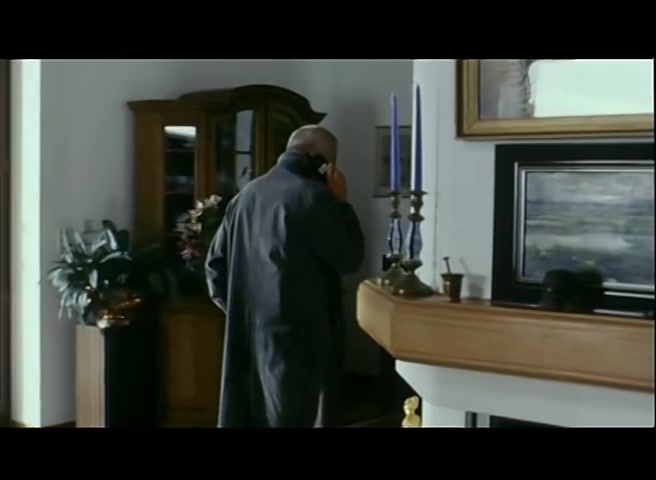}
\fullex{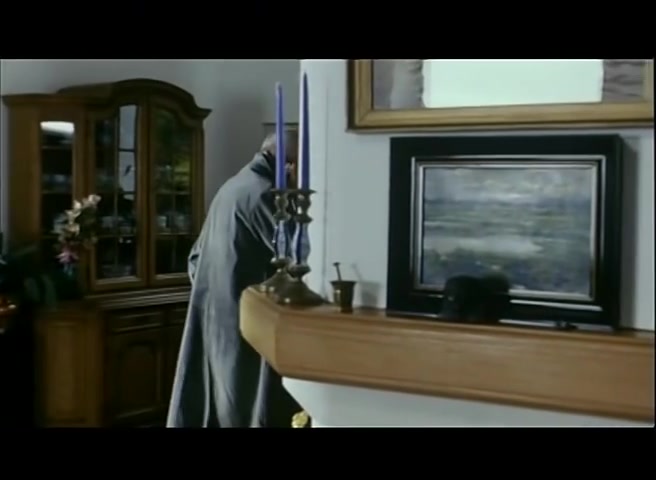}
\vspace{0.7mm}\\
\fullex{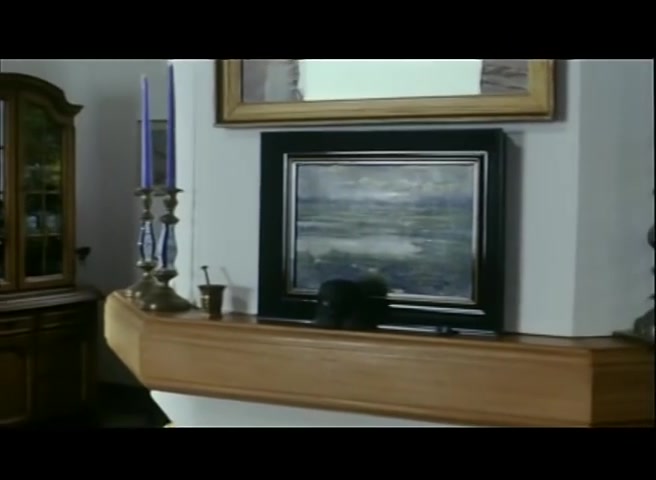}
\fullex{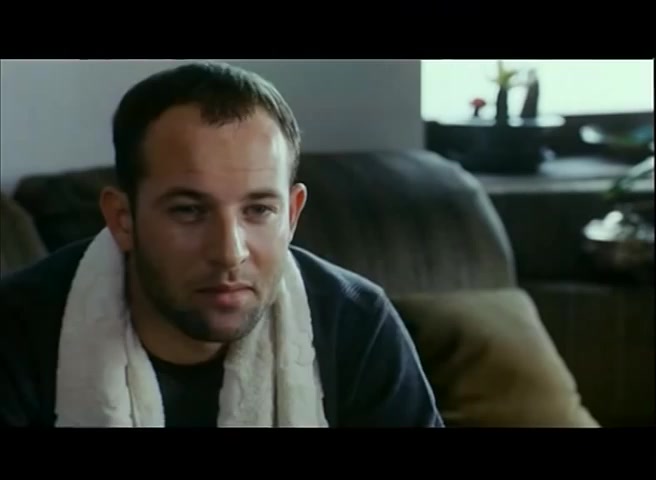}
\fullex{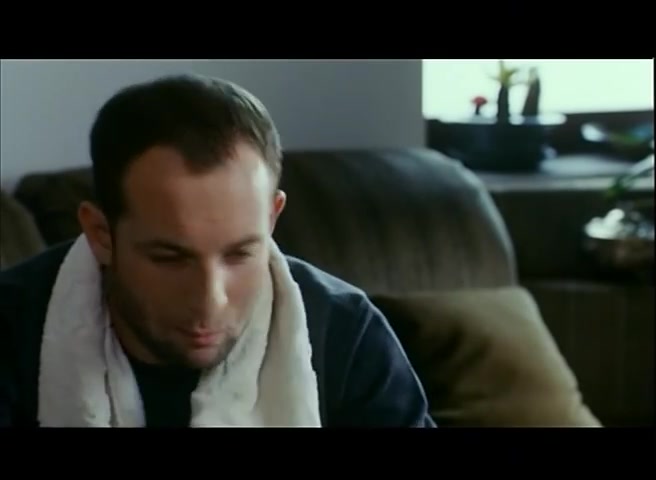}
\fullex{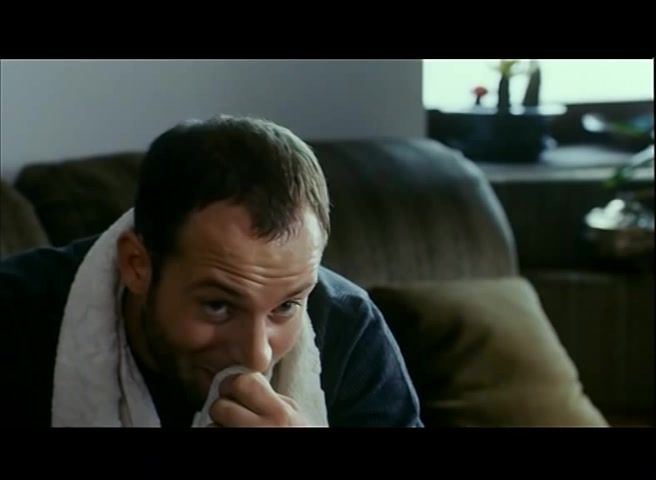}
\fullex{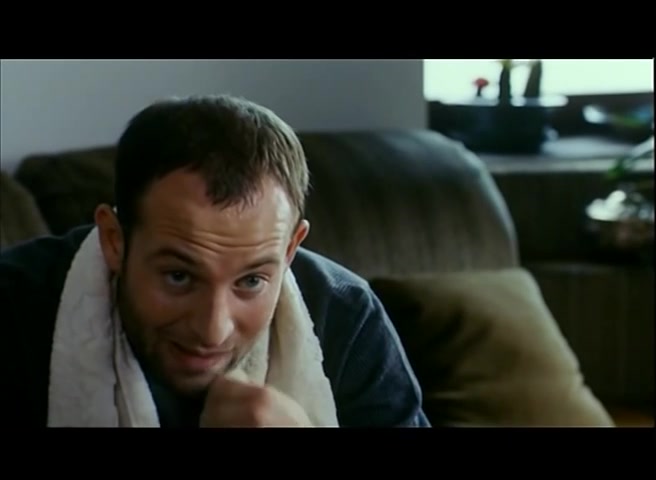}
\fullex{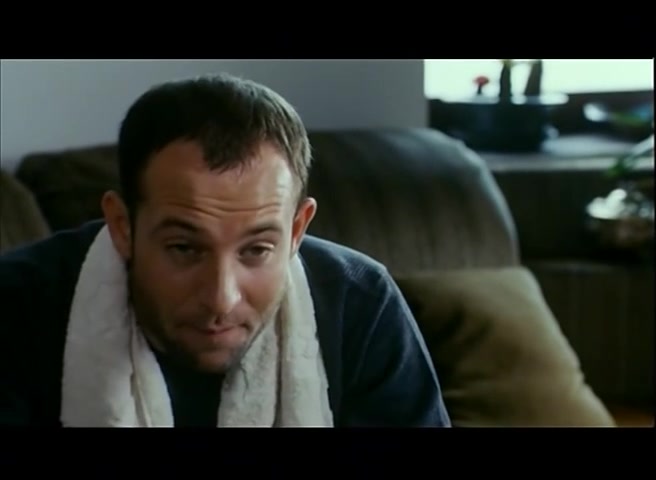}
\fullex{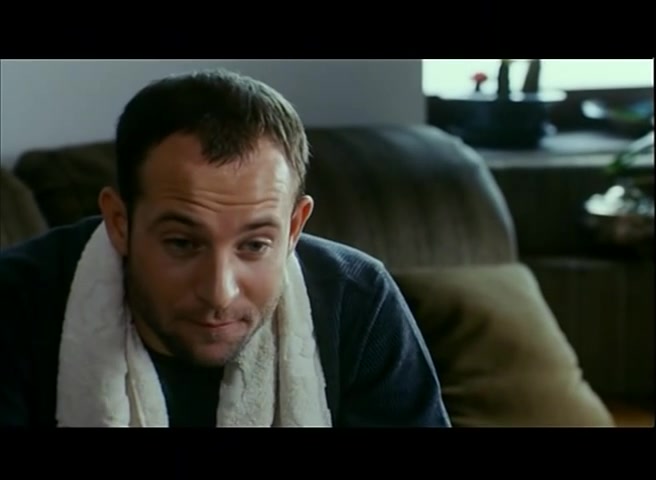}
\fullex{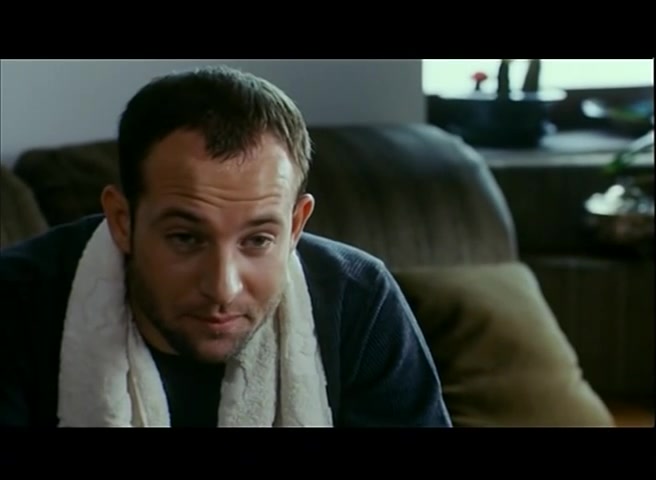}
\fullex{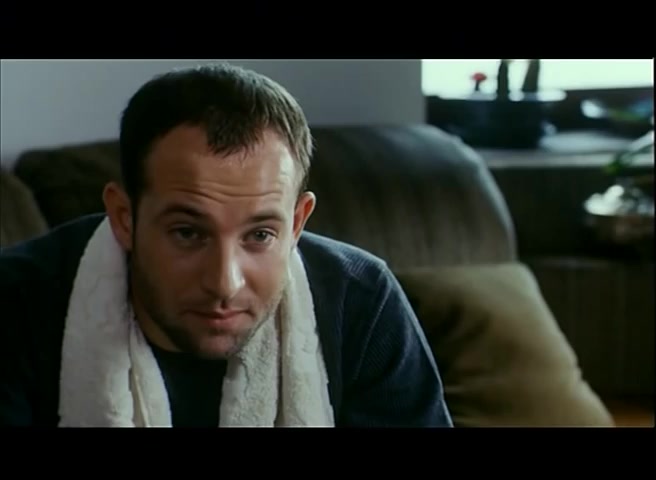}
\fullex{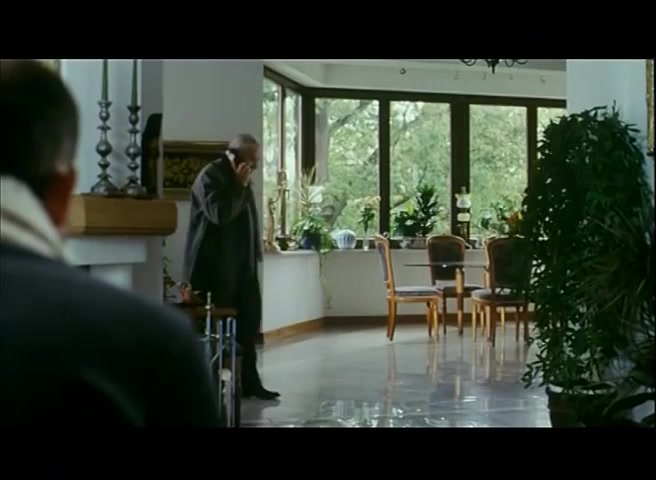}
\vspace{3mm}\\
\fullex{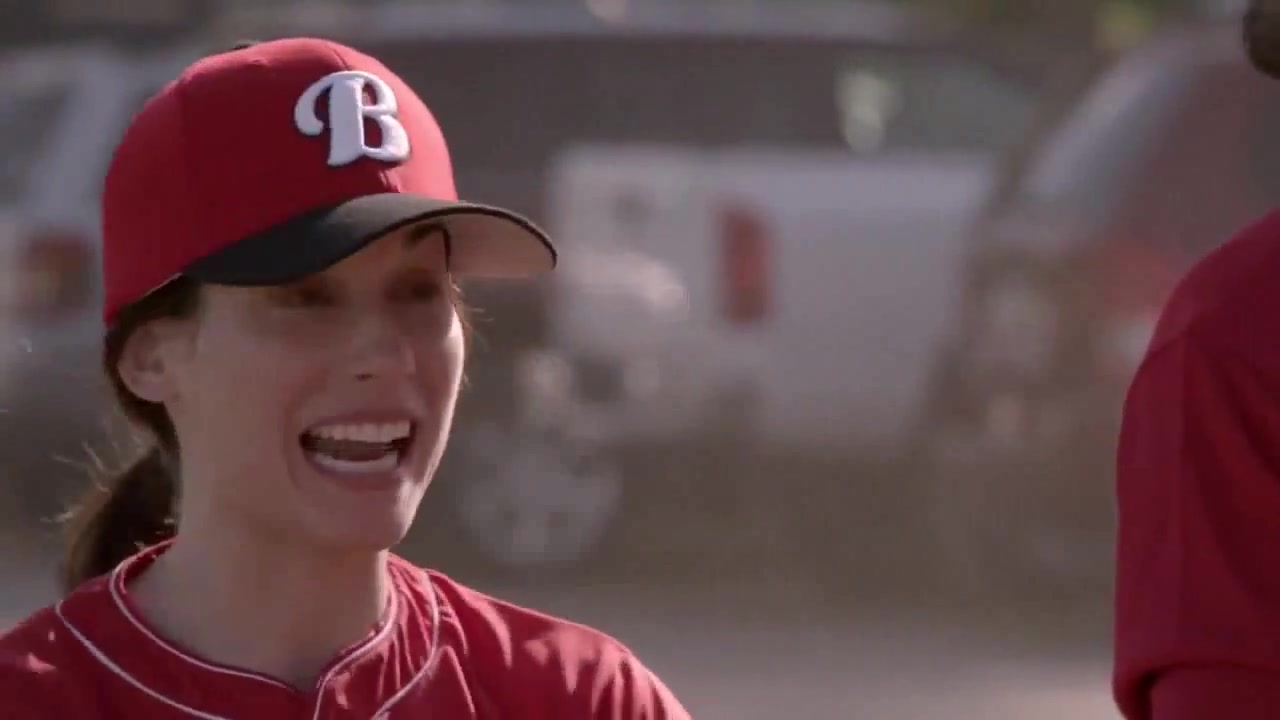}
\fullex{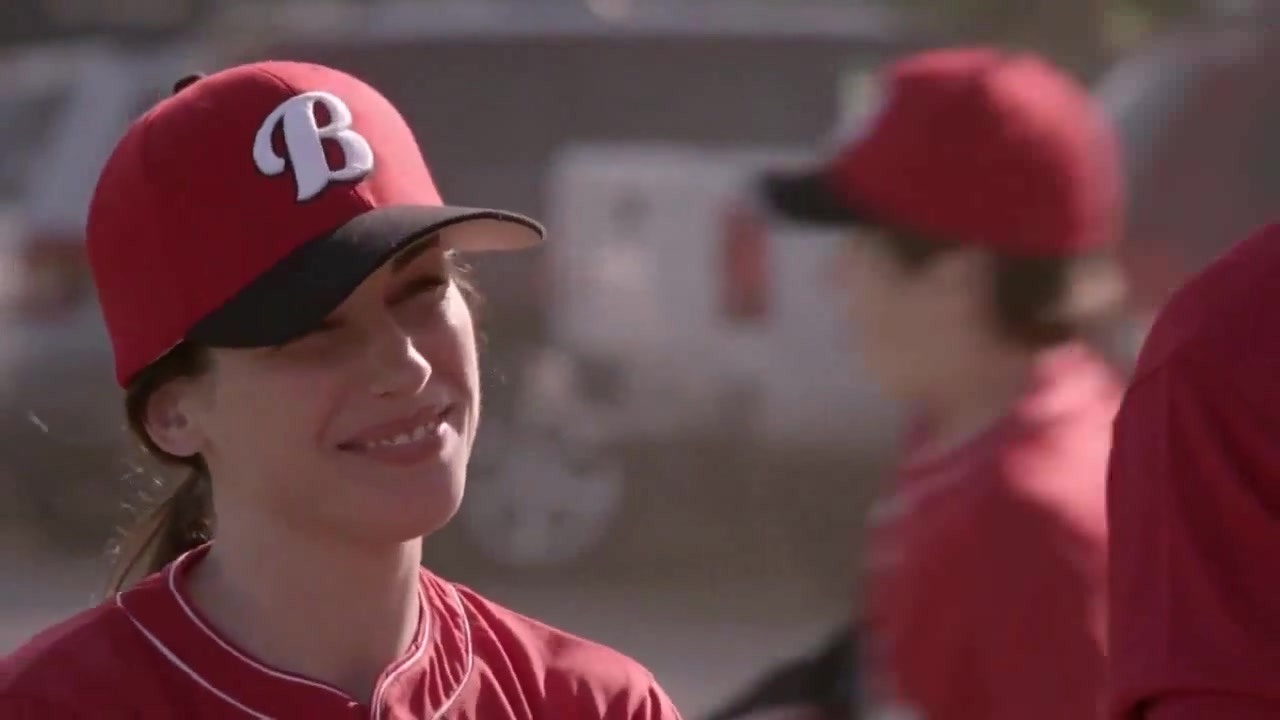}
\fullex{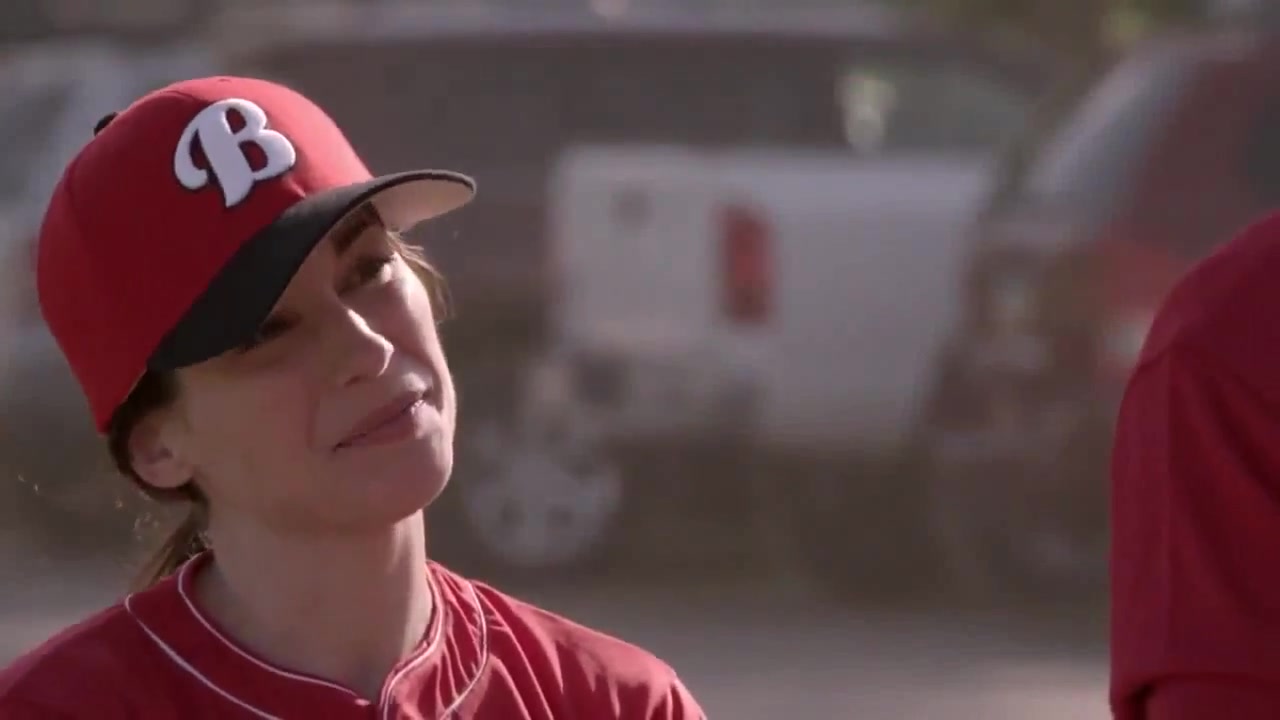}
\fullex{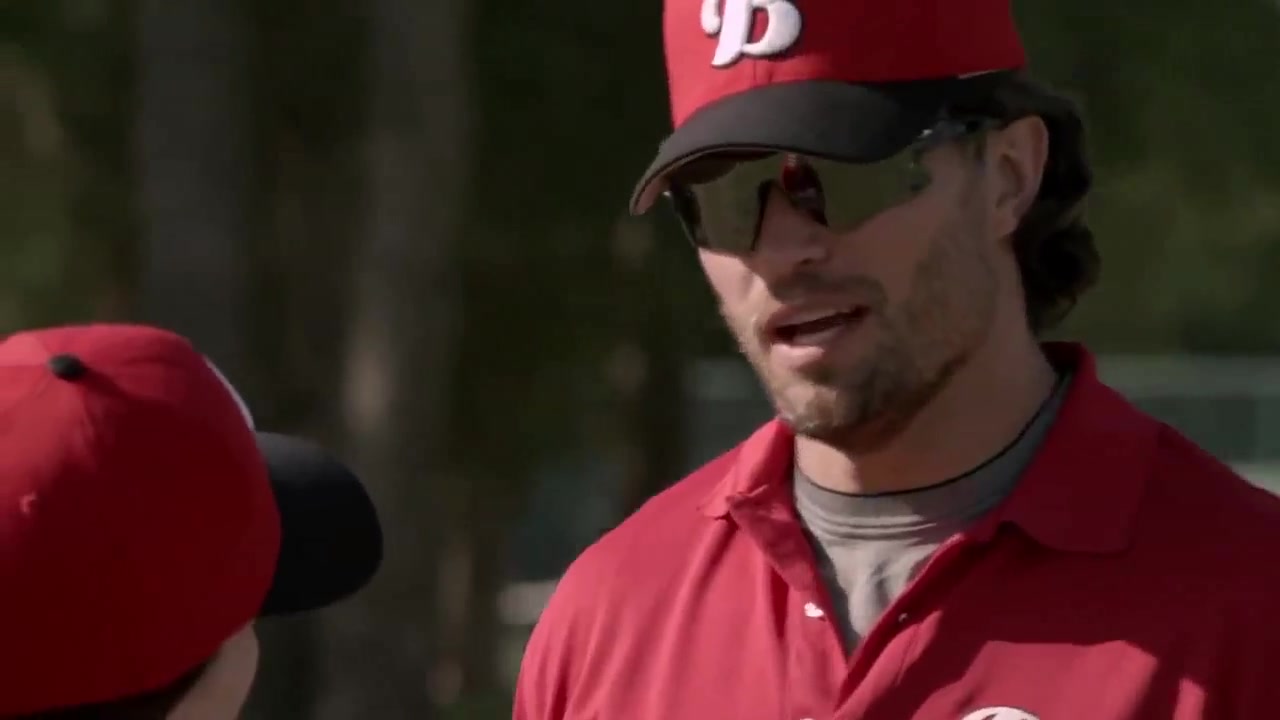}
\fullex{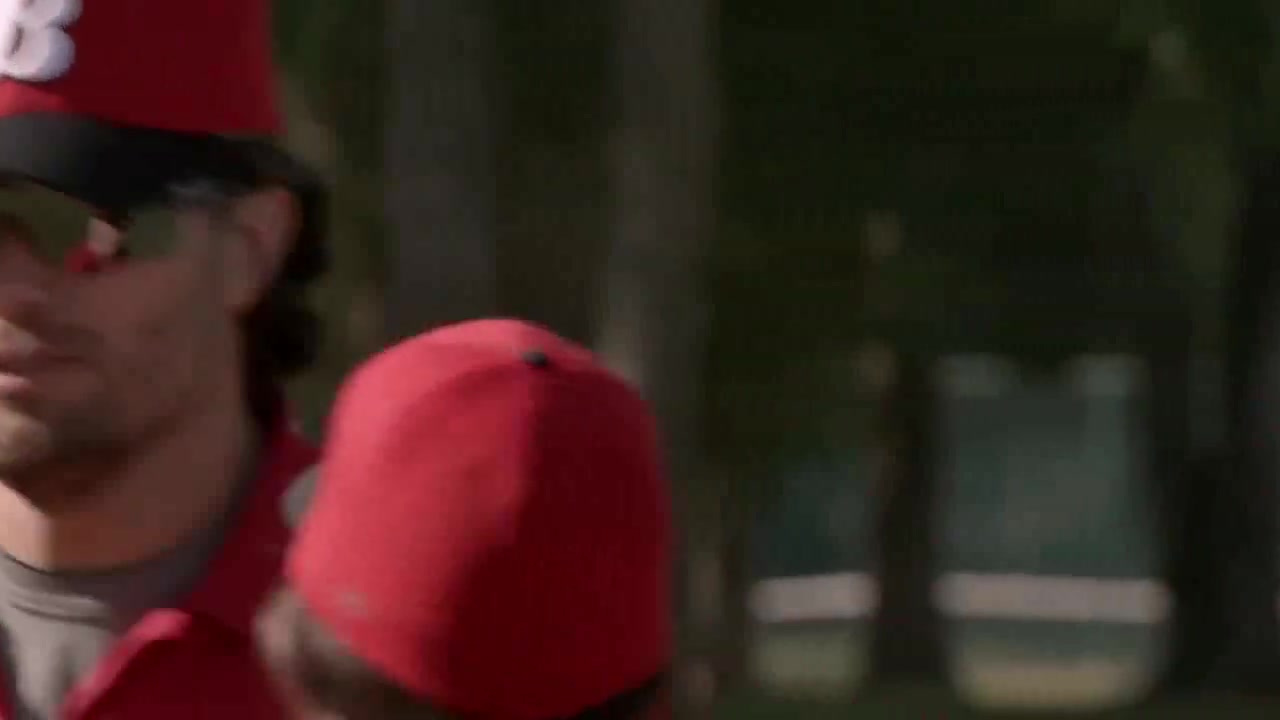}
\fullex{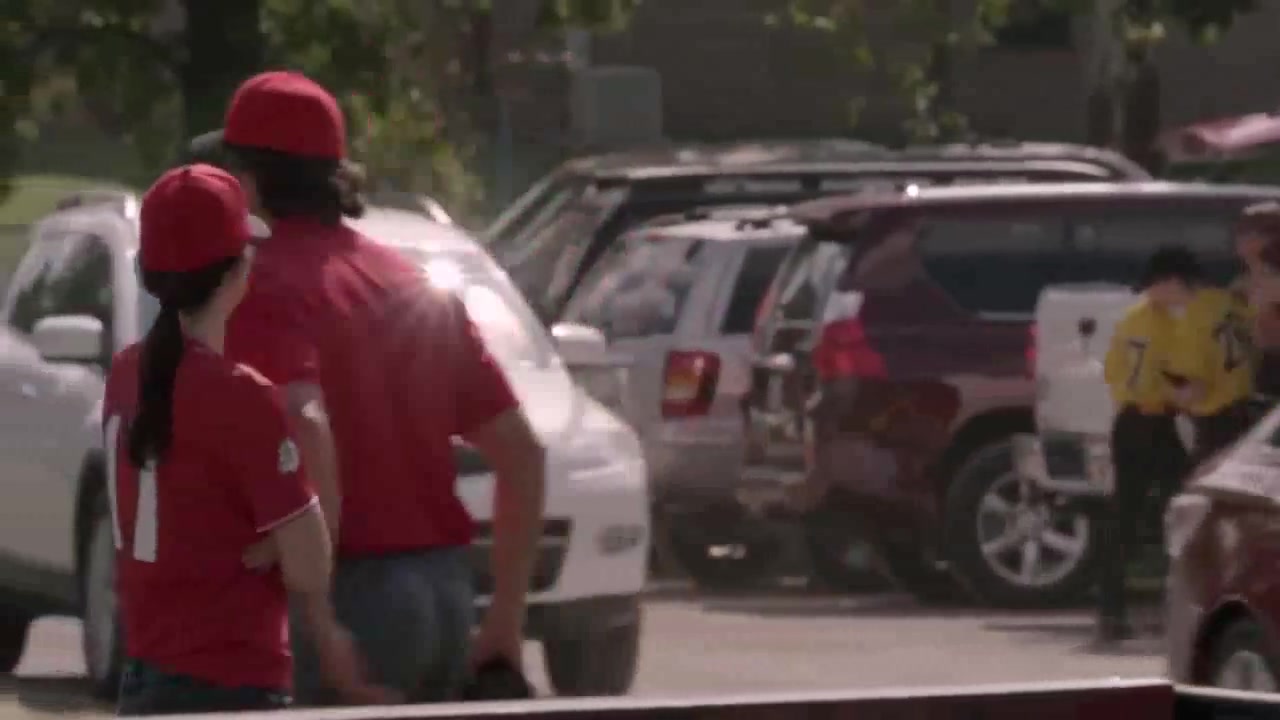}
\fullex{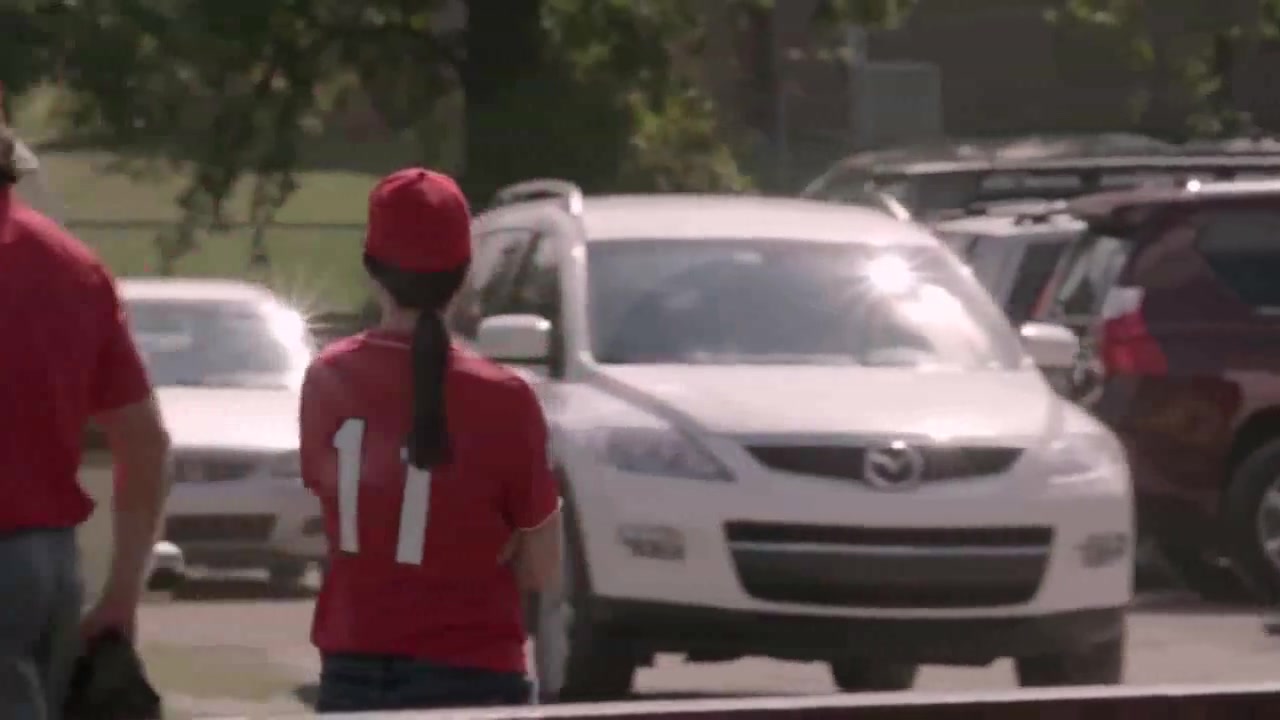}
\fullex{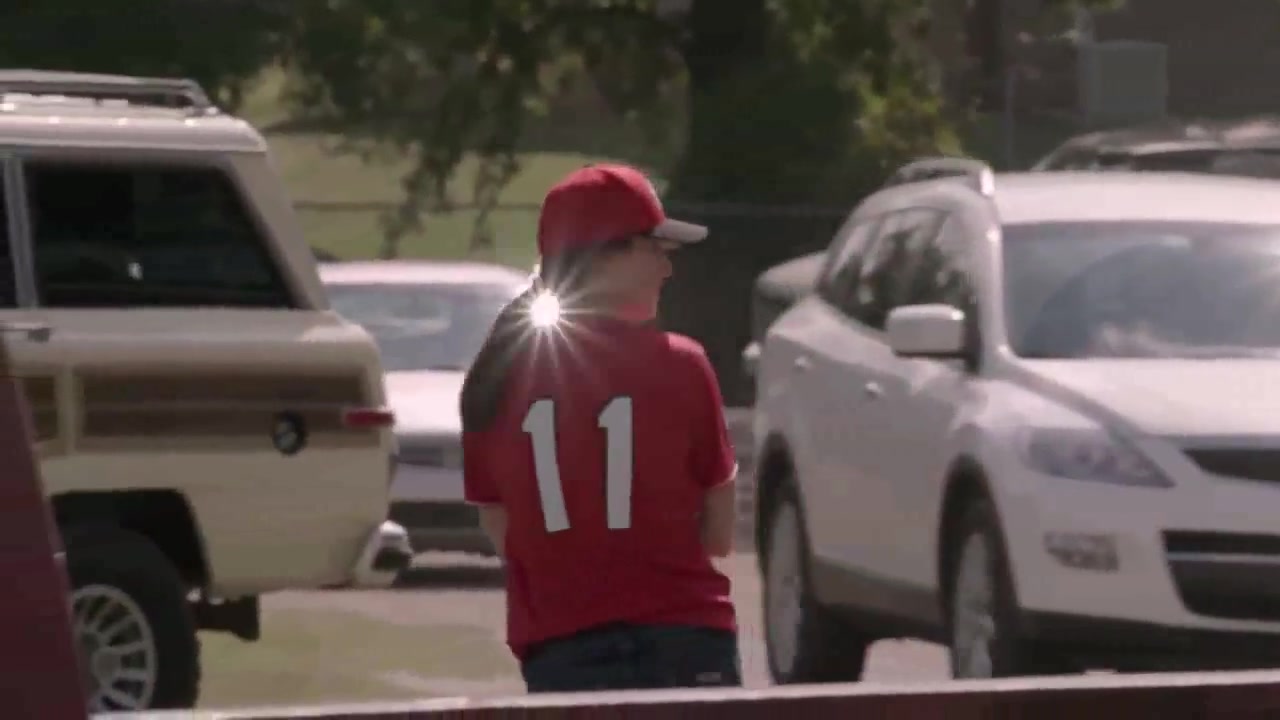}
\fullex{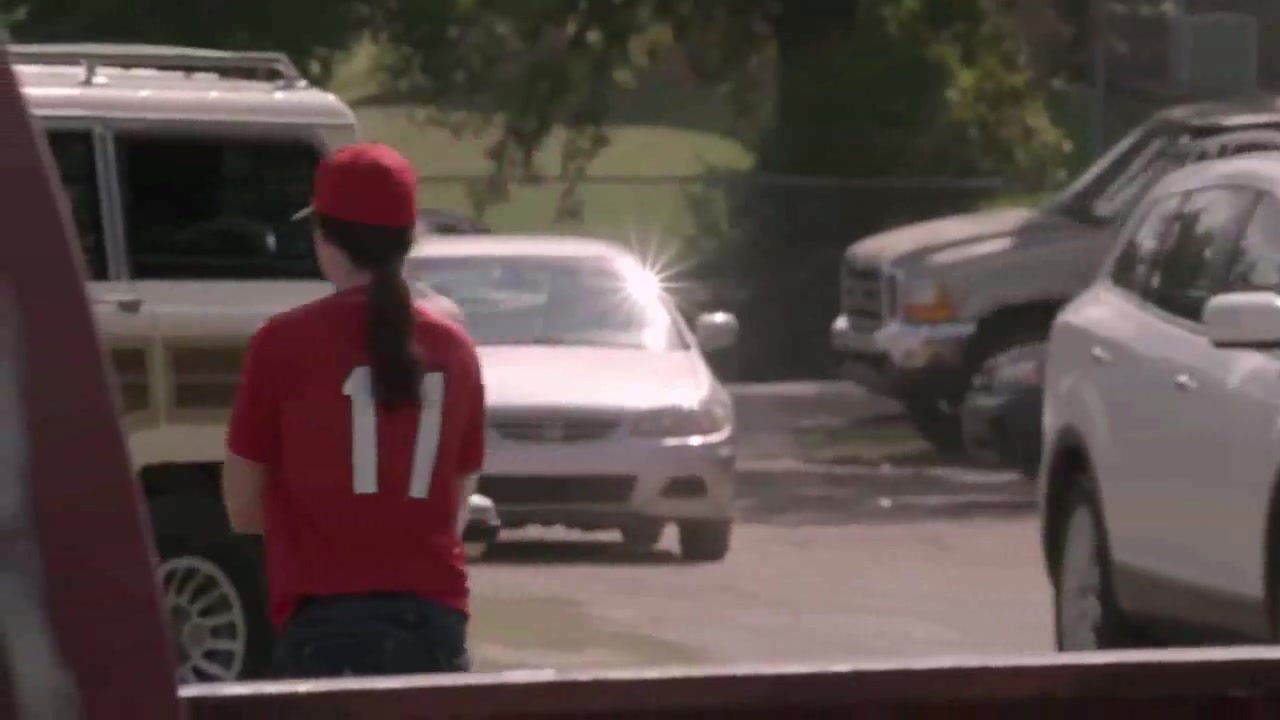}
\fullex{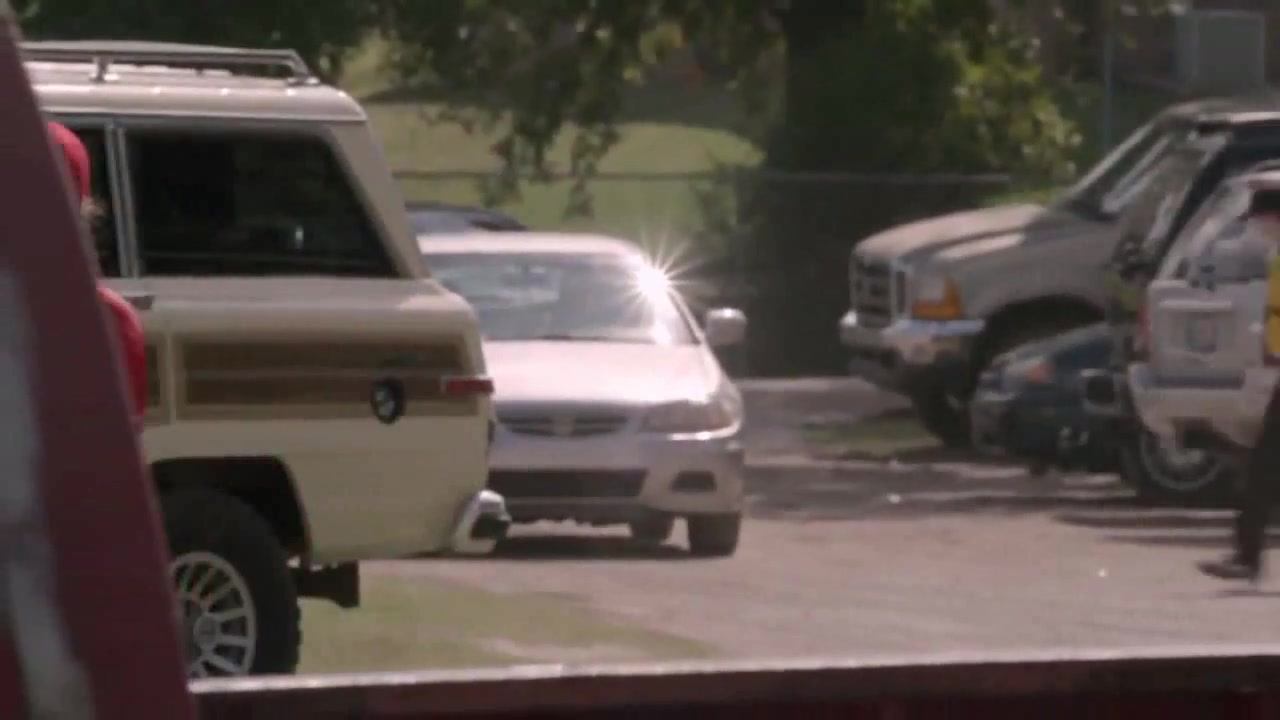}
\vspace{0.7mm}\\
\fullex{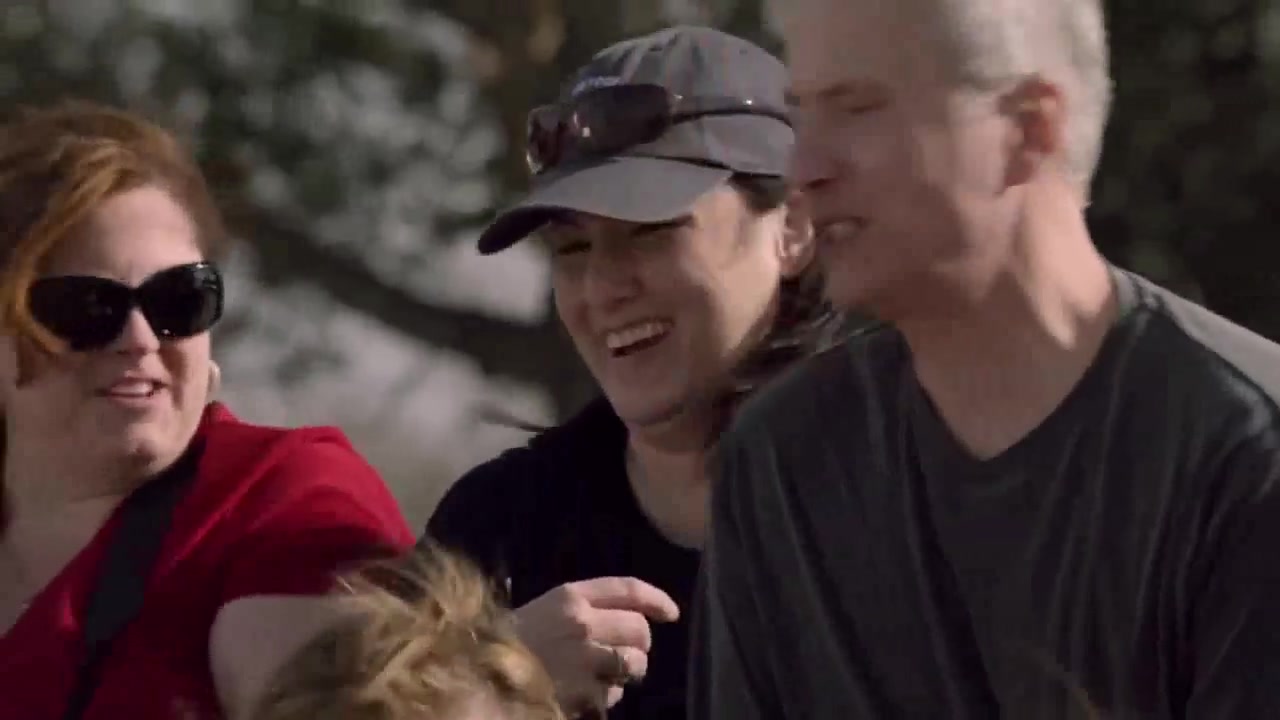}
\fullex{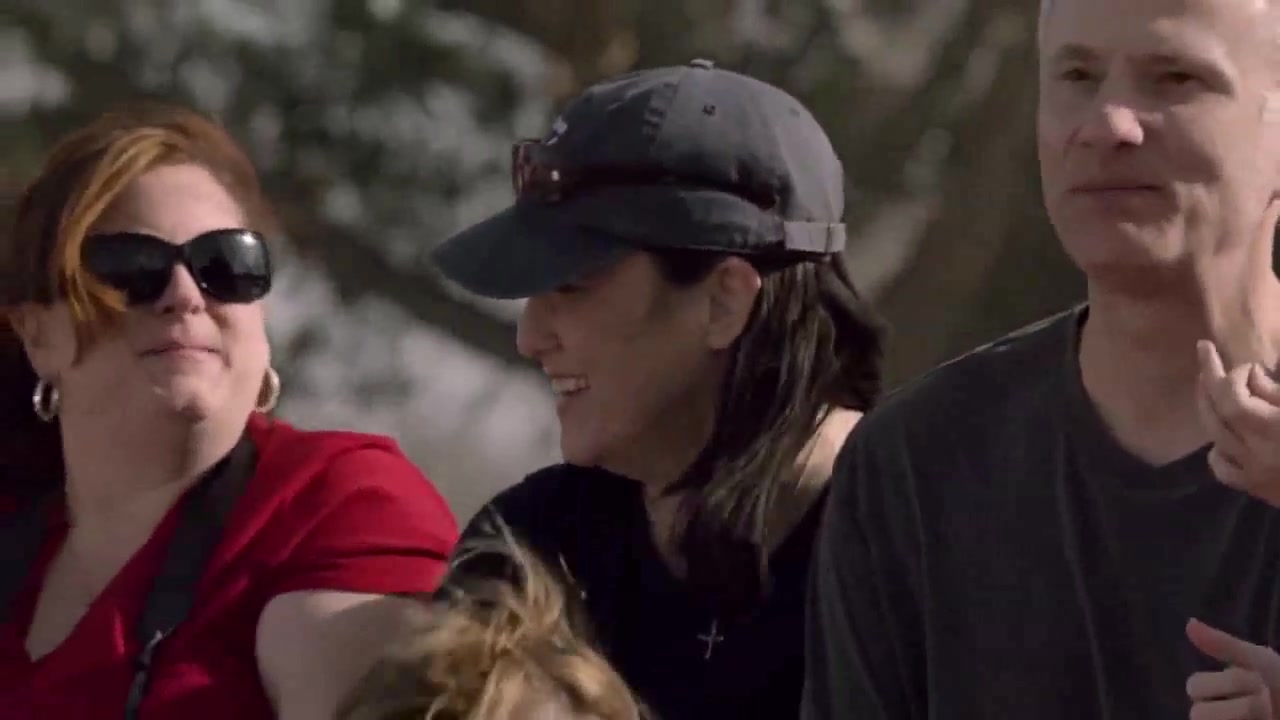}
\fullex{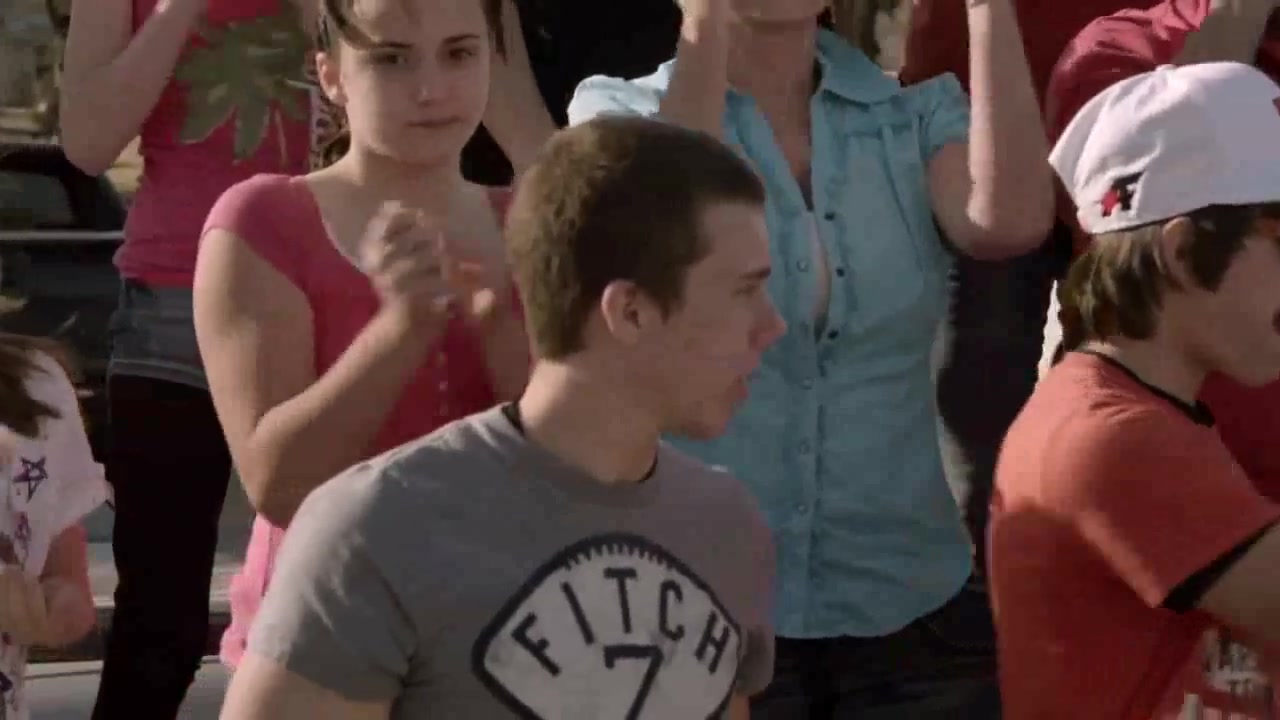}
\fullex{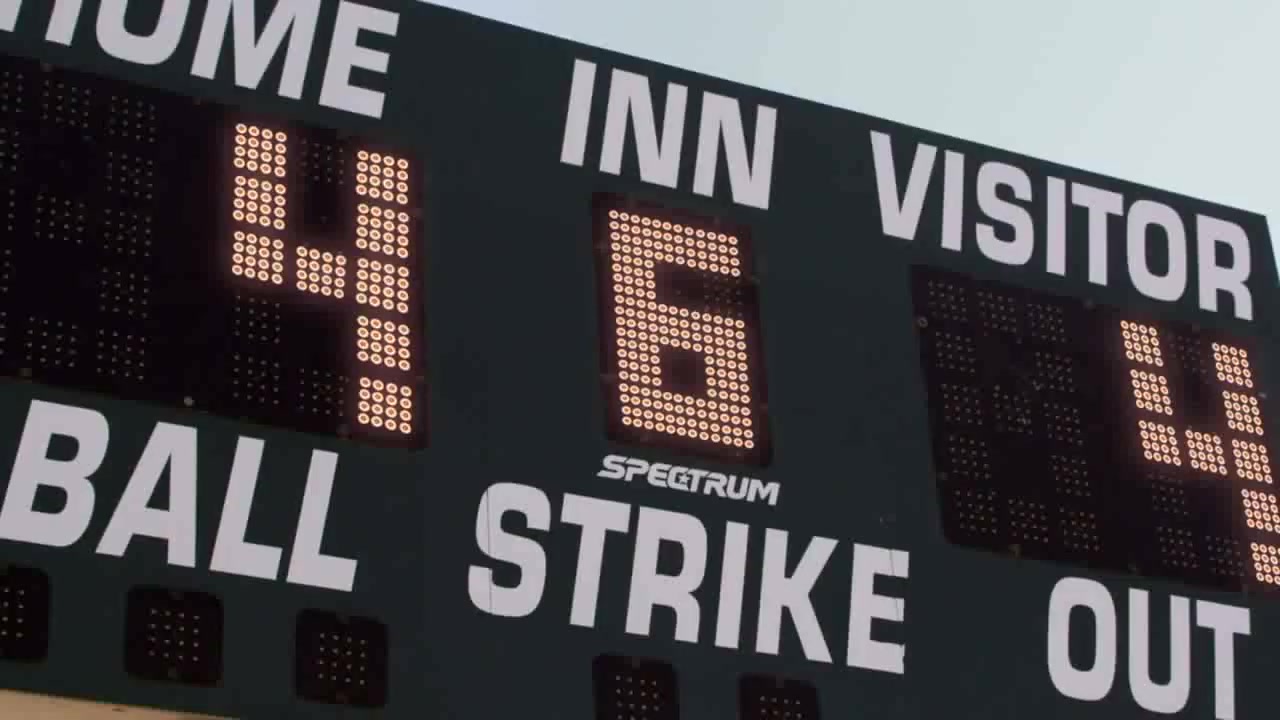}
\fullex{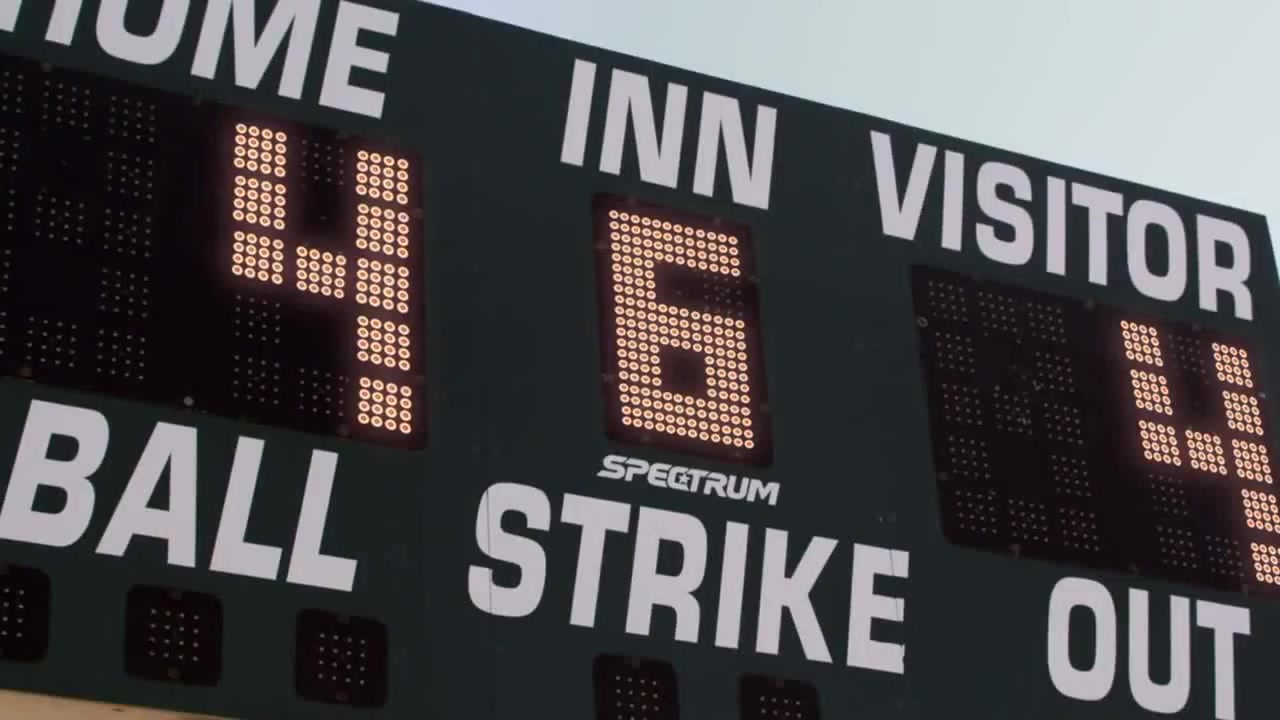}
\fullex{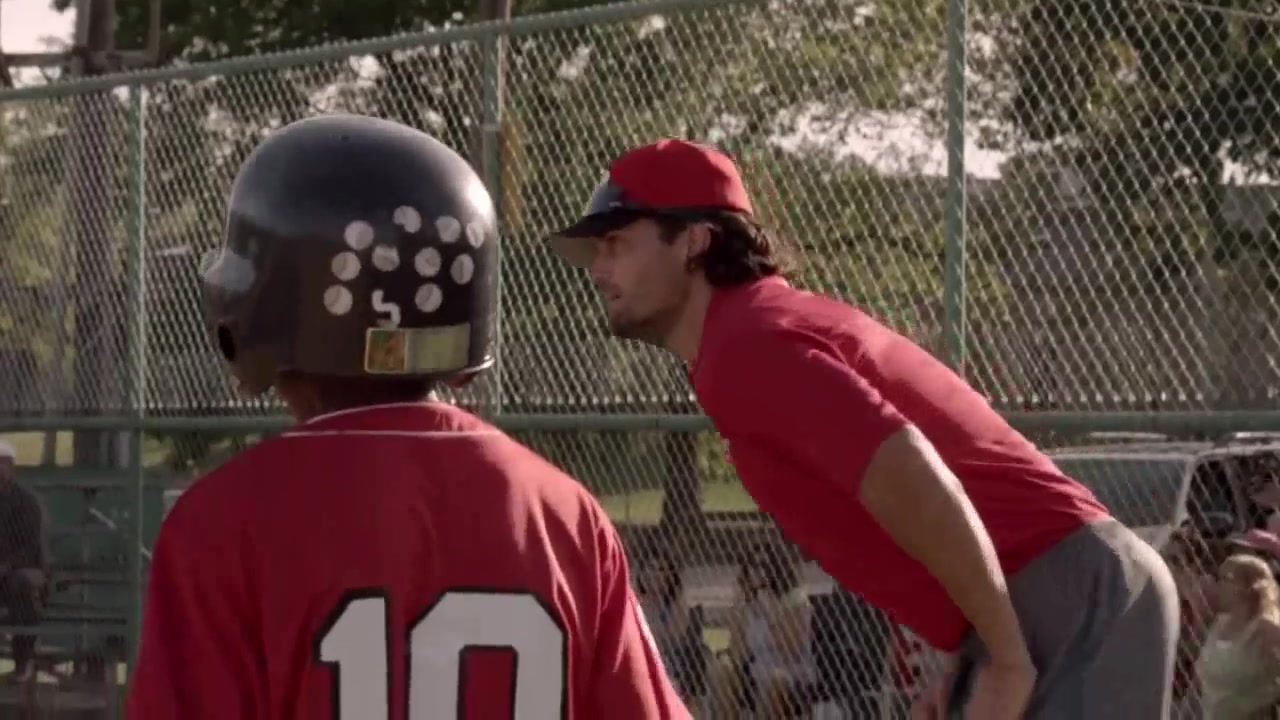}
\fullex{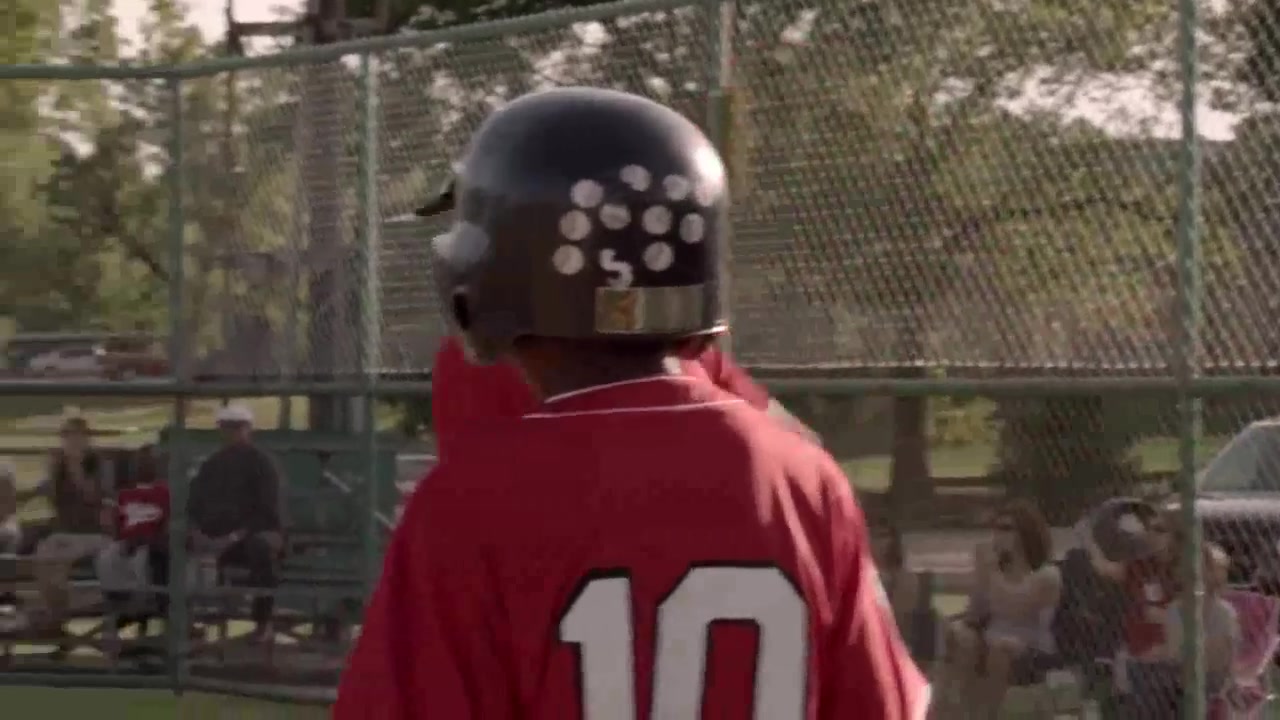}
\fullex{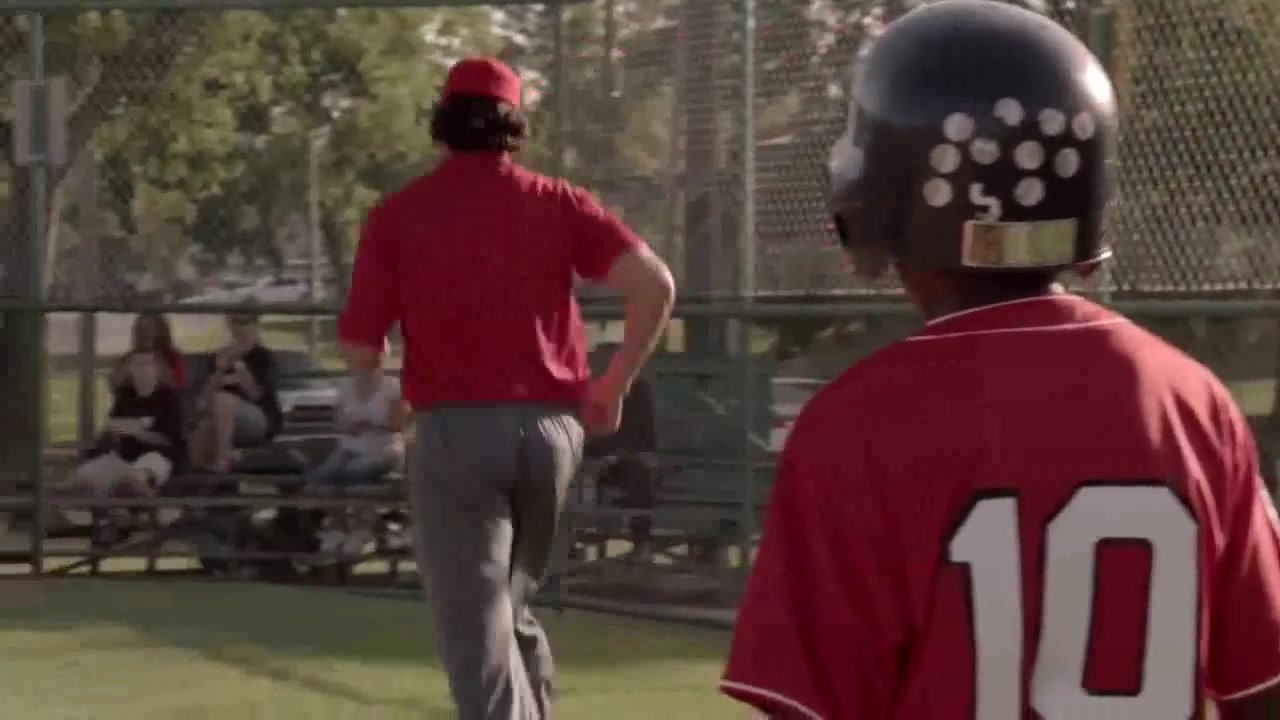}
\fullex{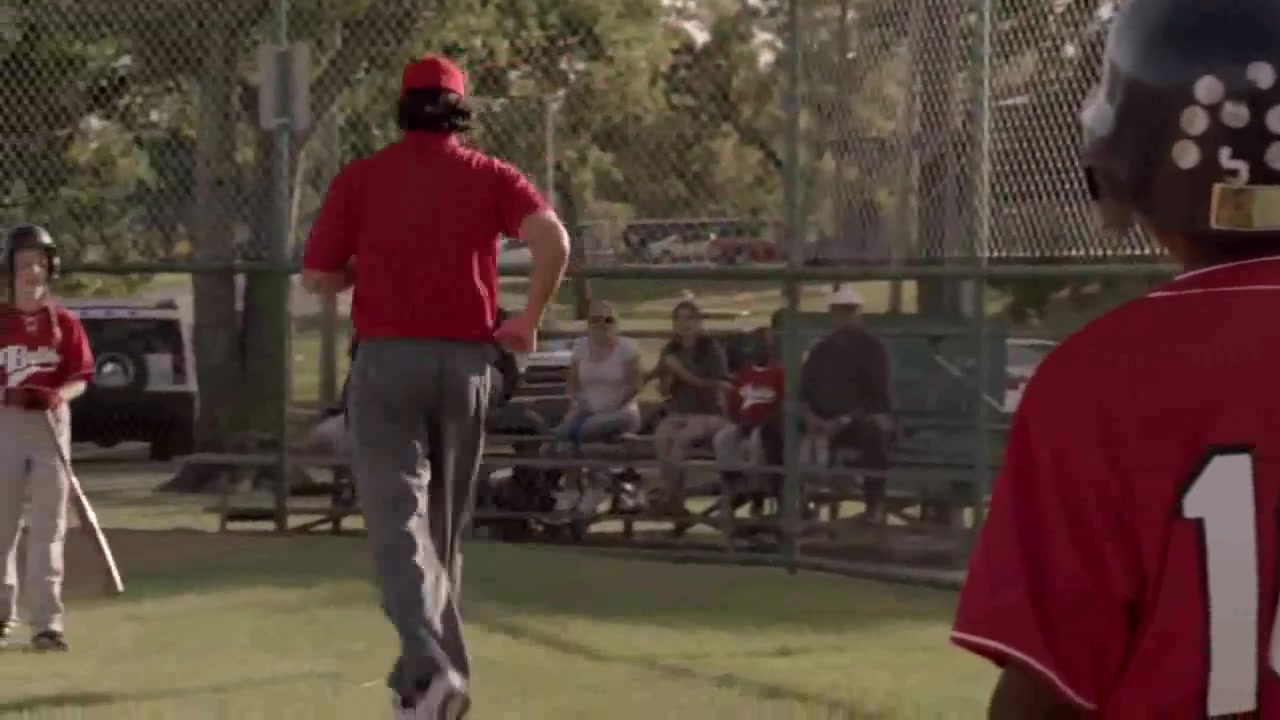}
\fullex{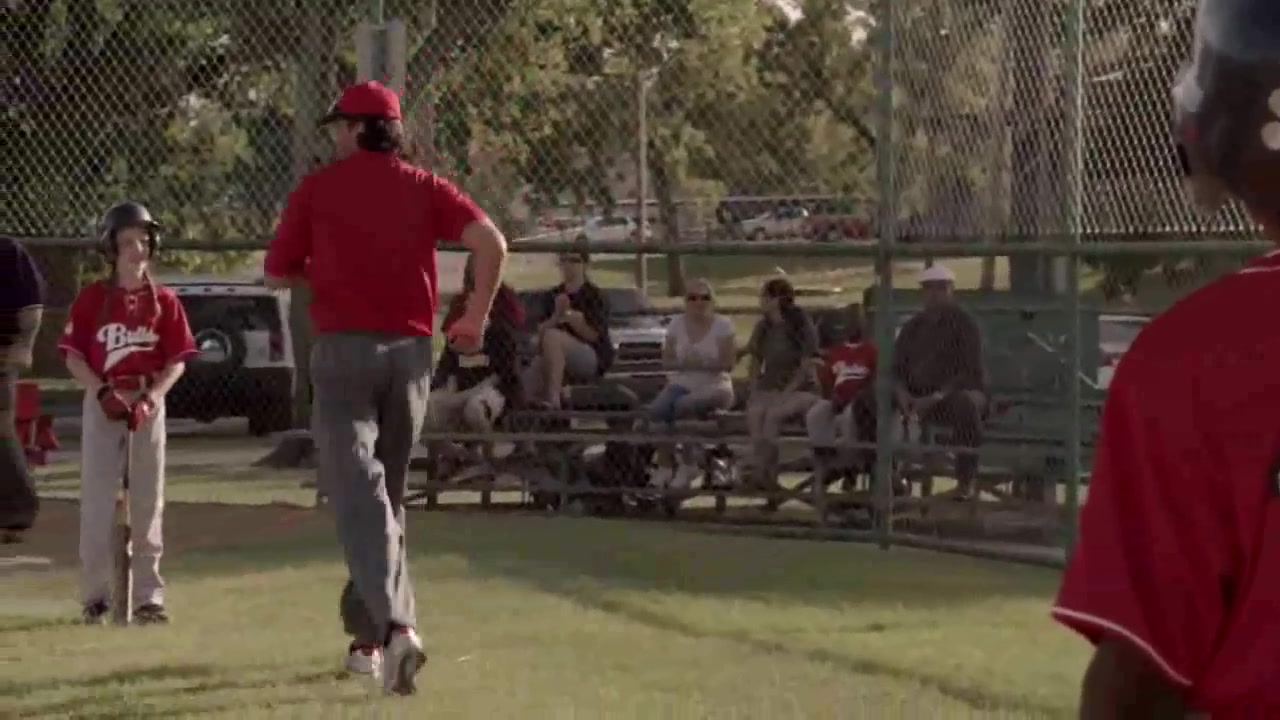}
\vspace{0.7mm}\\
\fullex{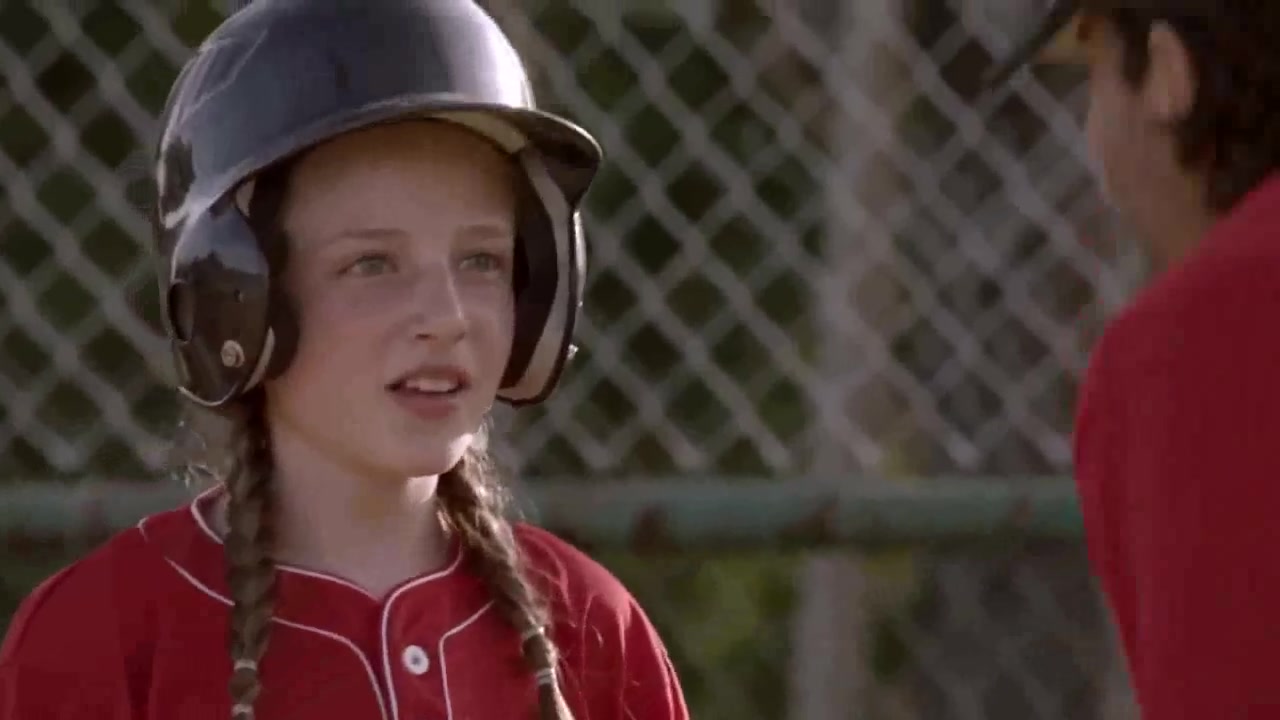}
\fullex{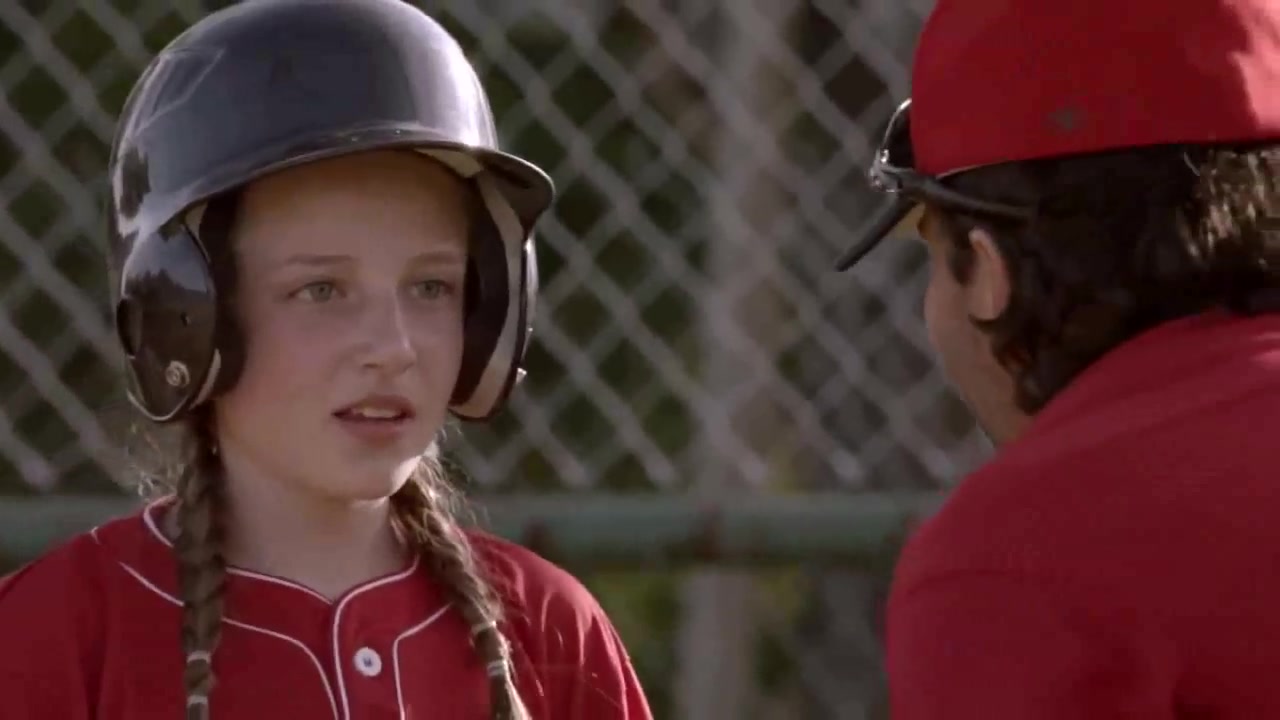}
\fullex{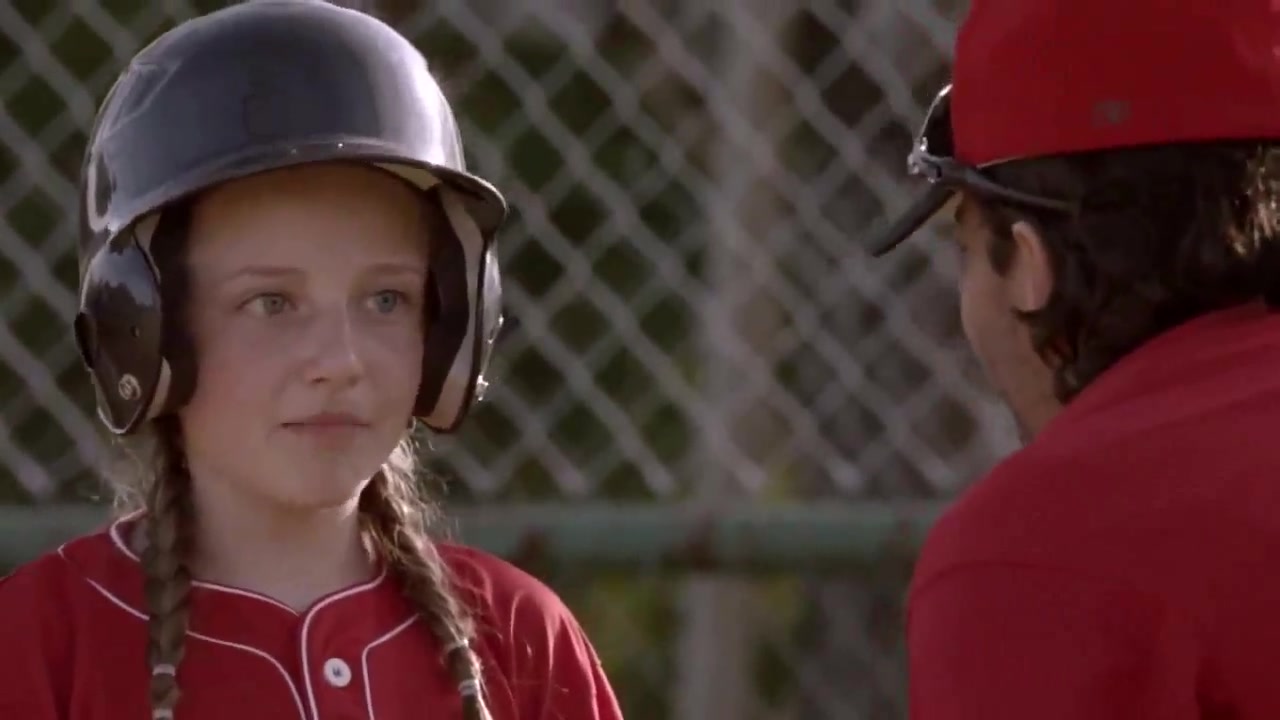}
\fullex{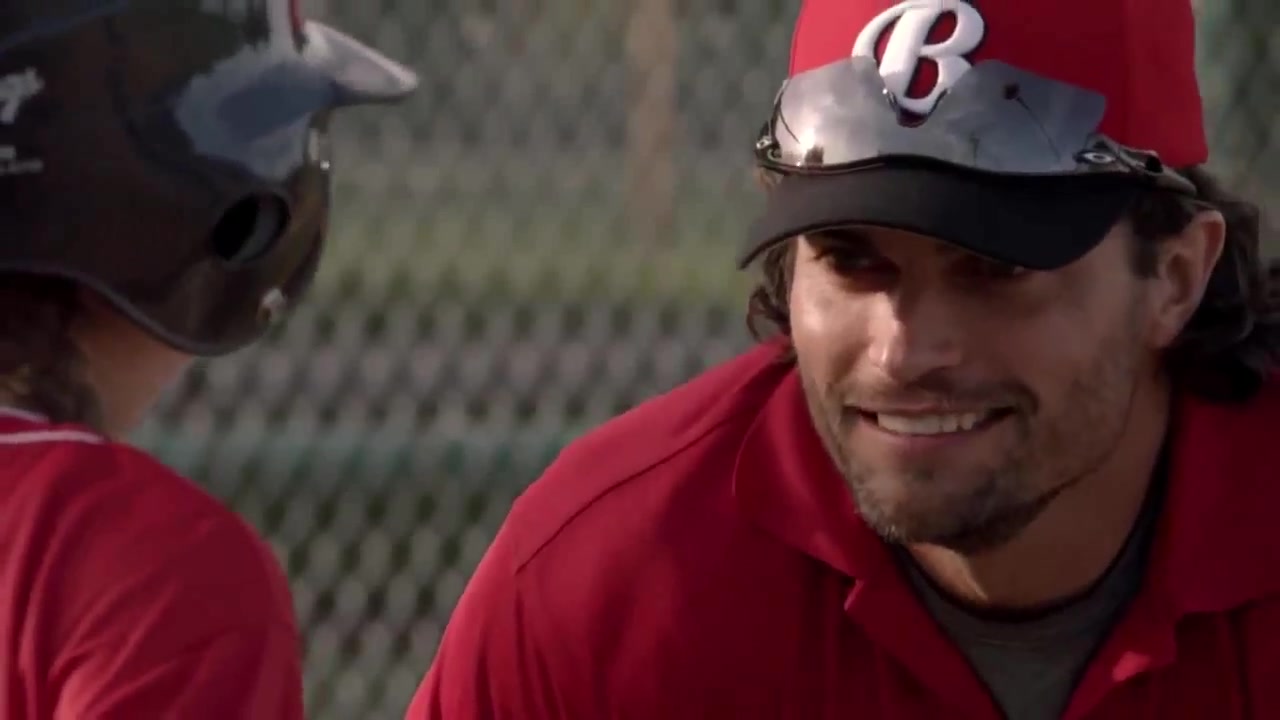}
\fullex{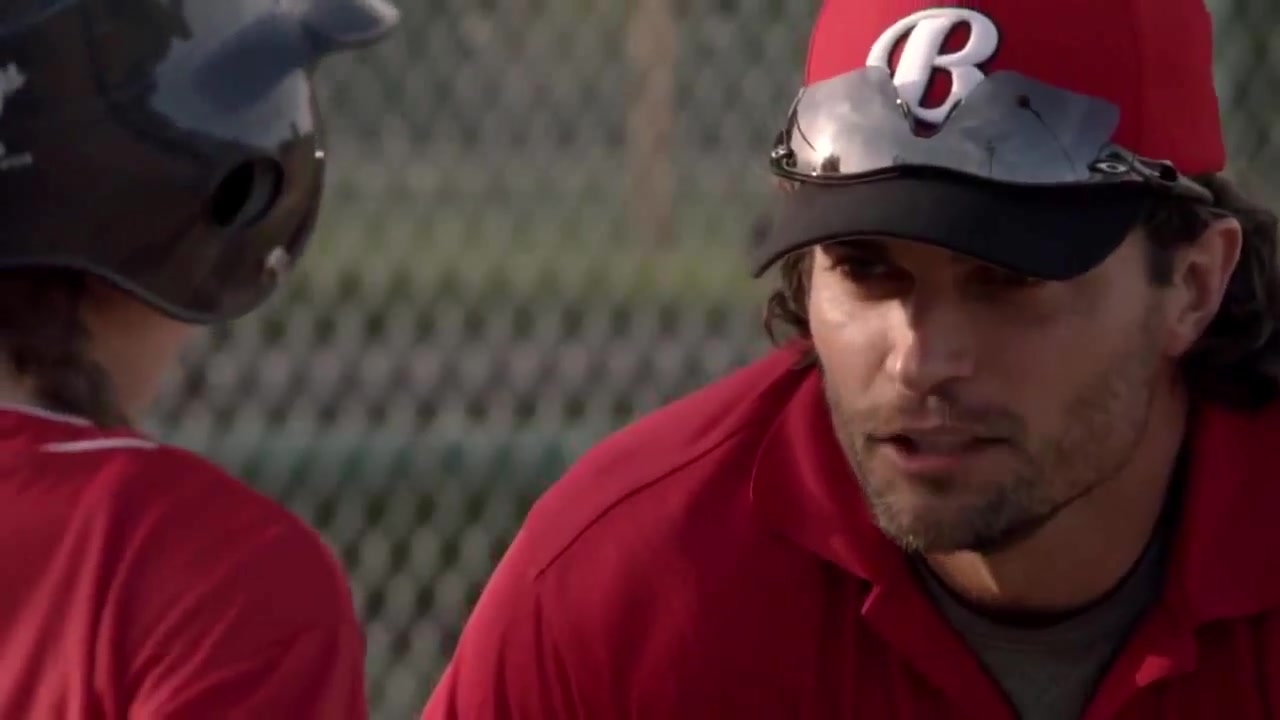}
\fullex{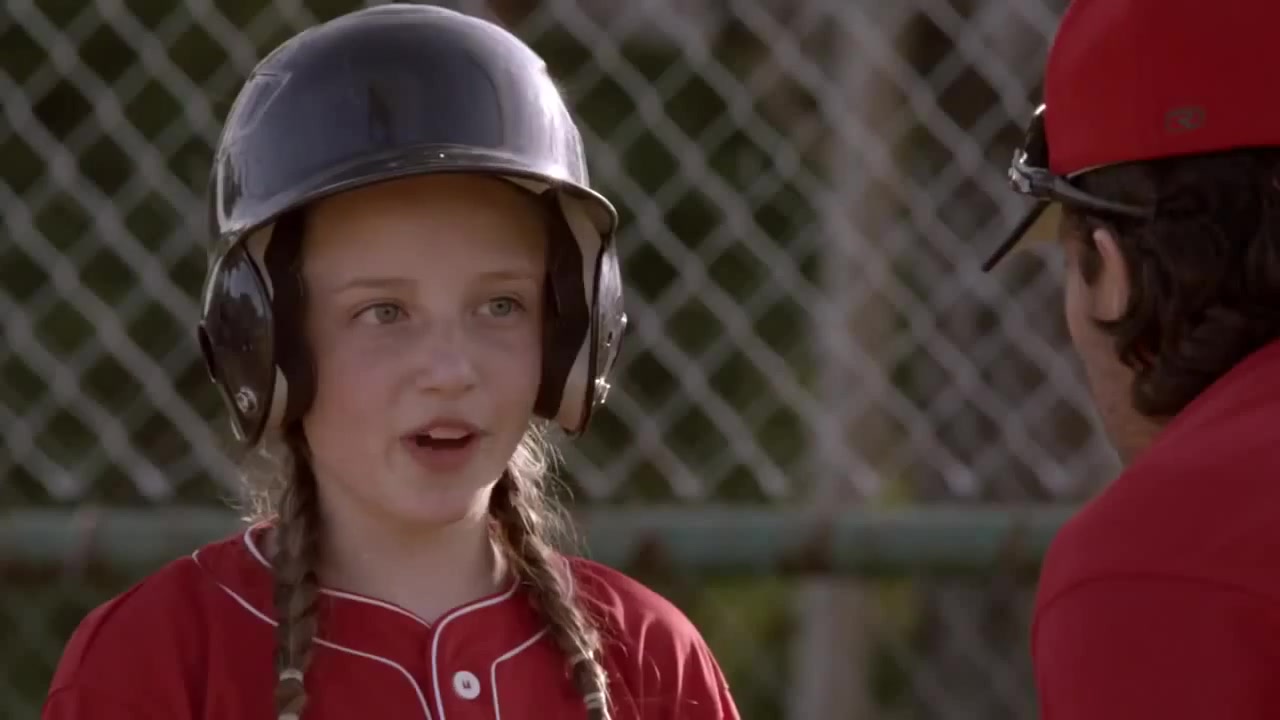}
\fullex{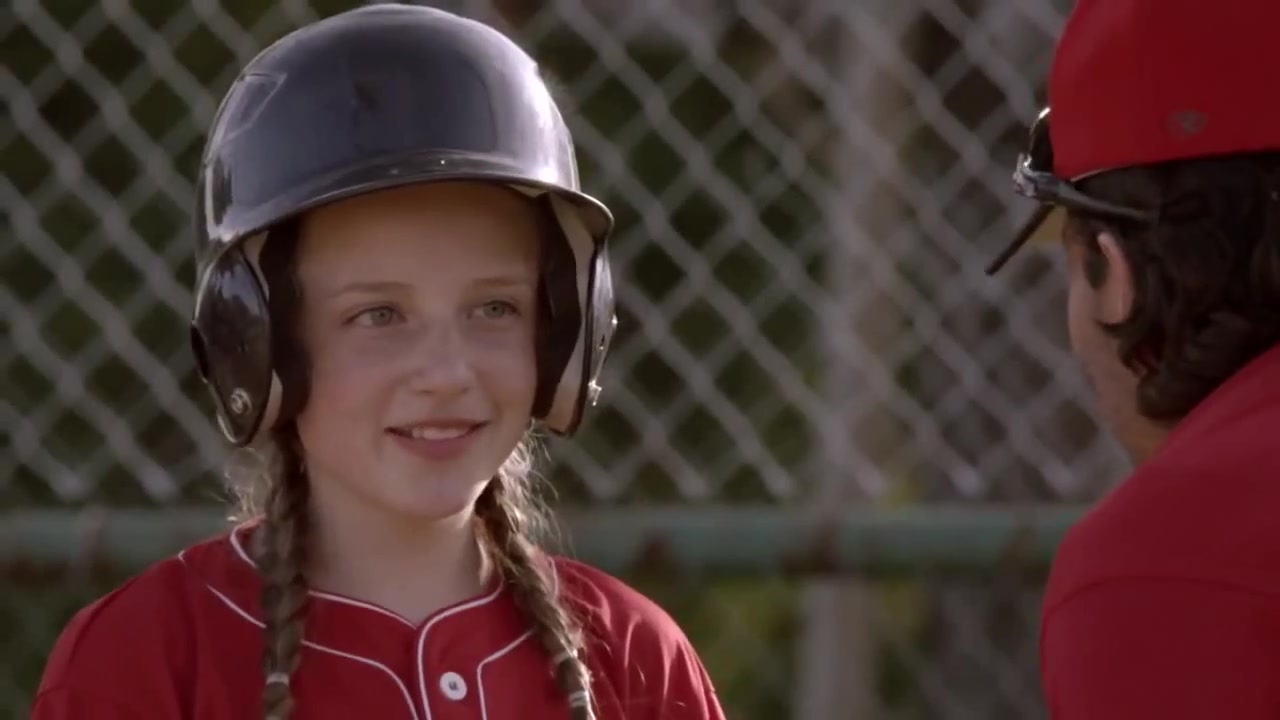}
\fullex{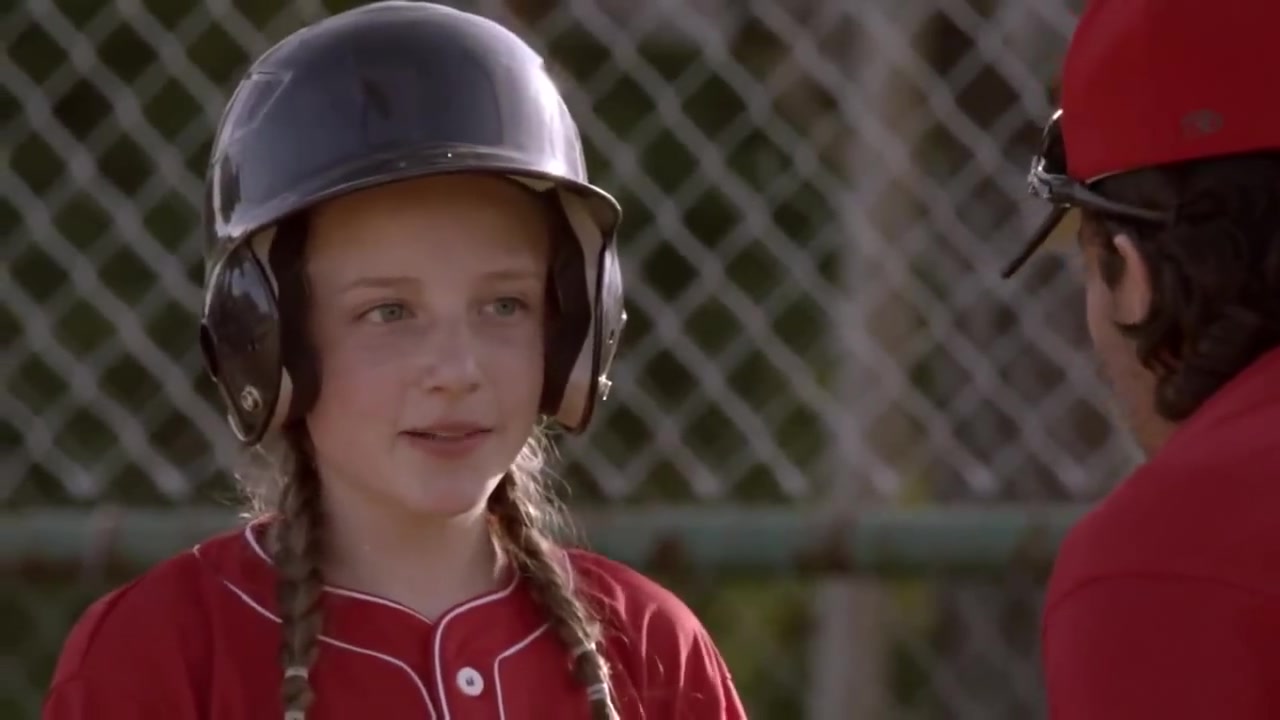}
\fullex{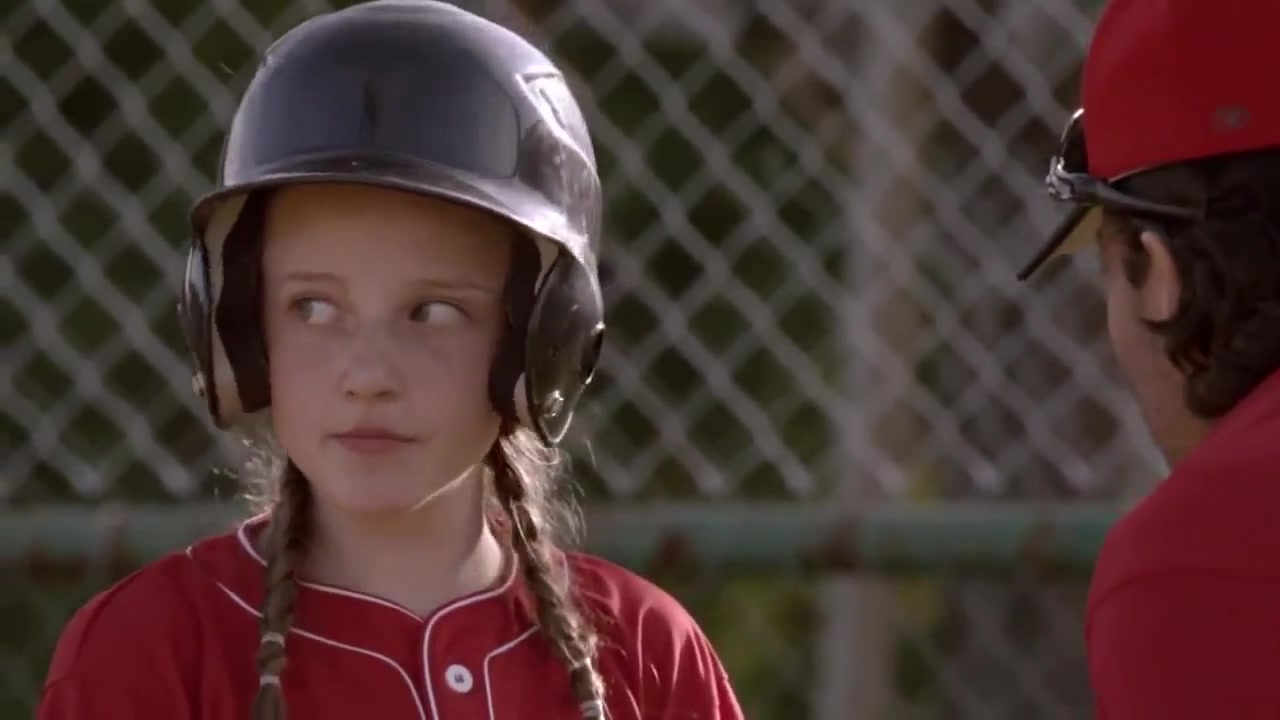}
\fullex{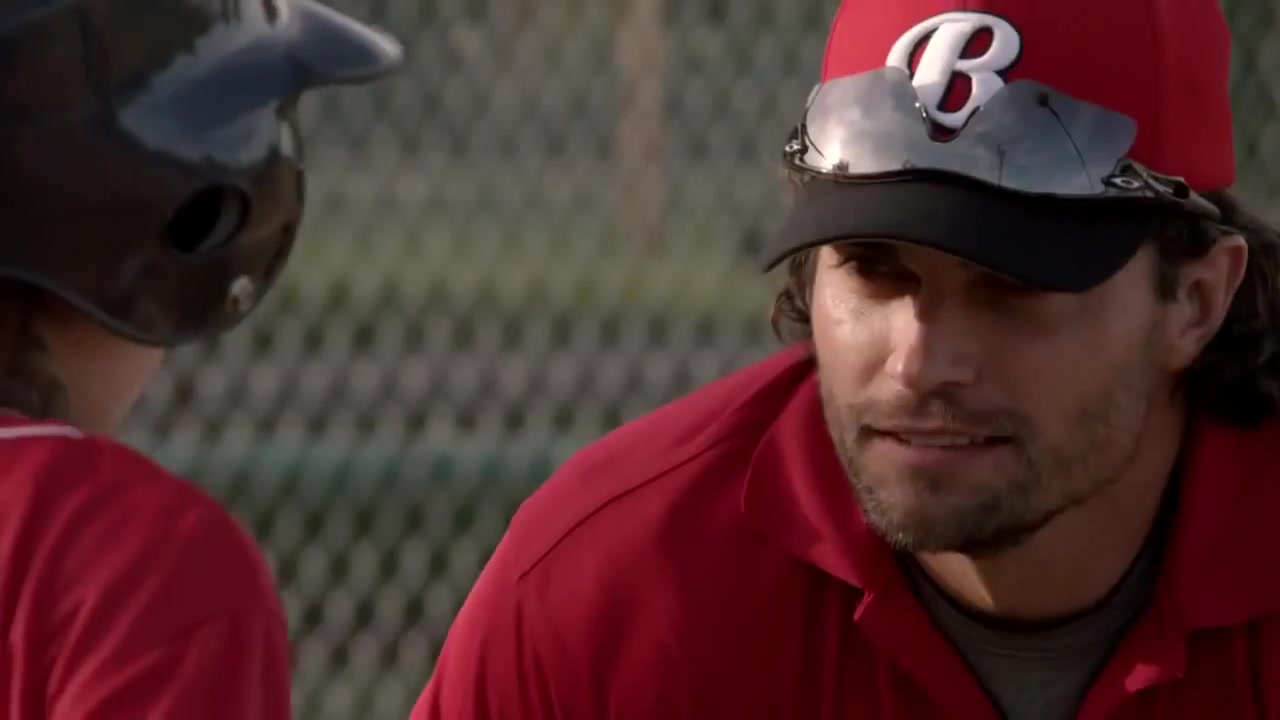}
\vspace{0.7mm}\\
\fullex{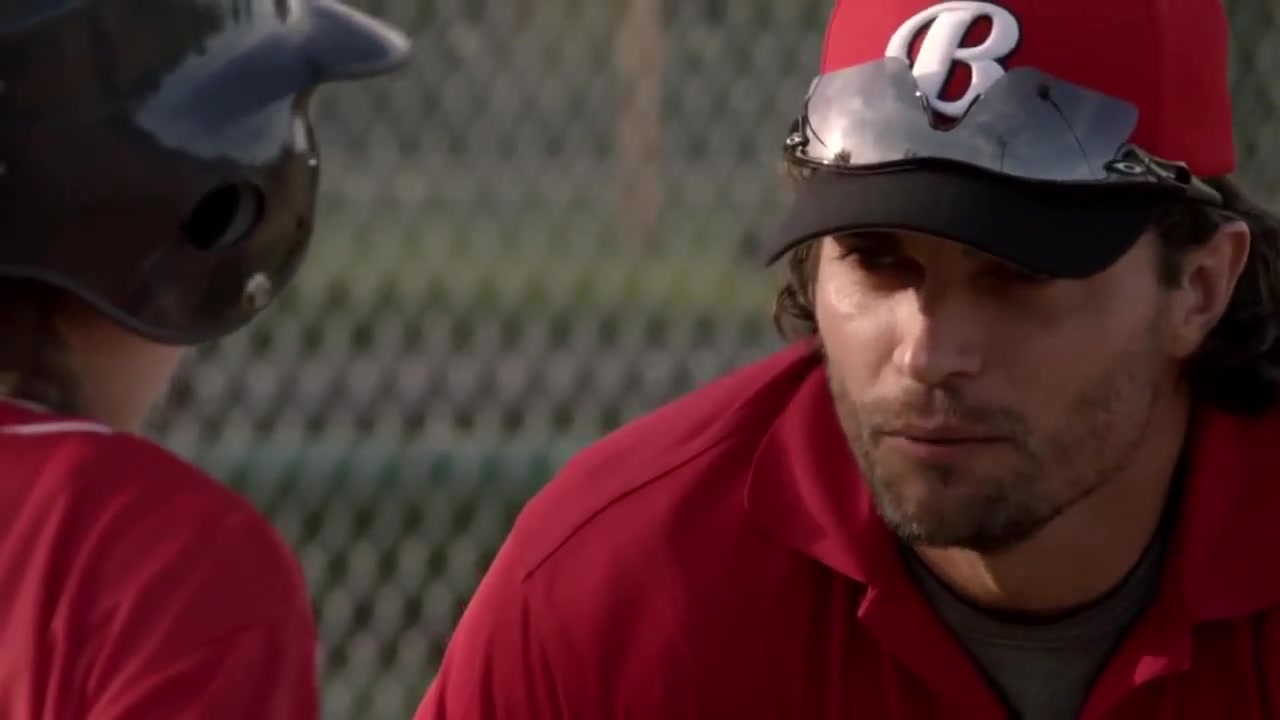}
\fullex{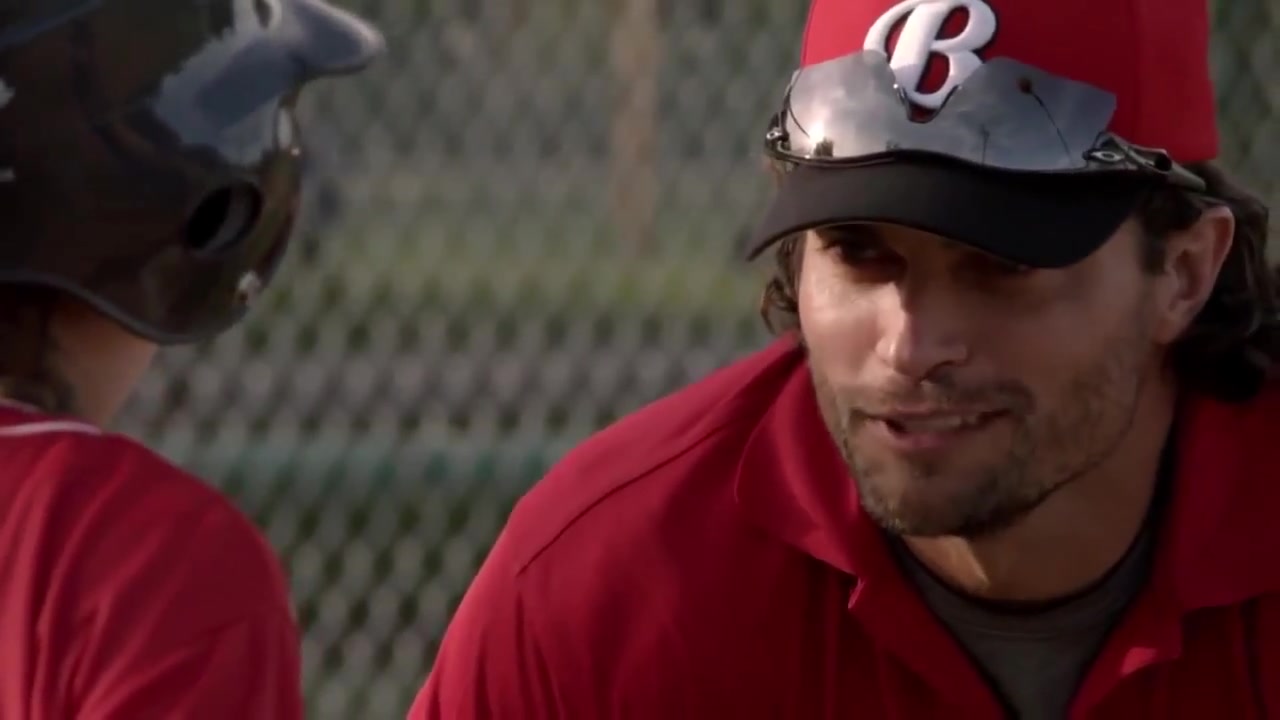}
\fullex{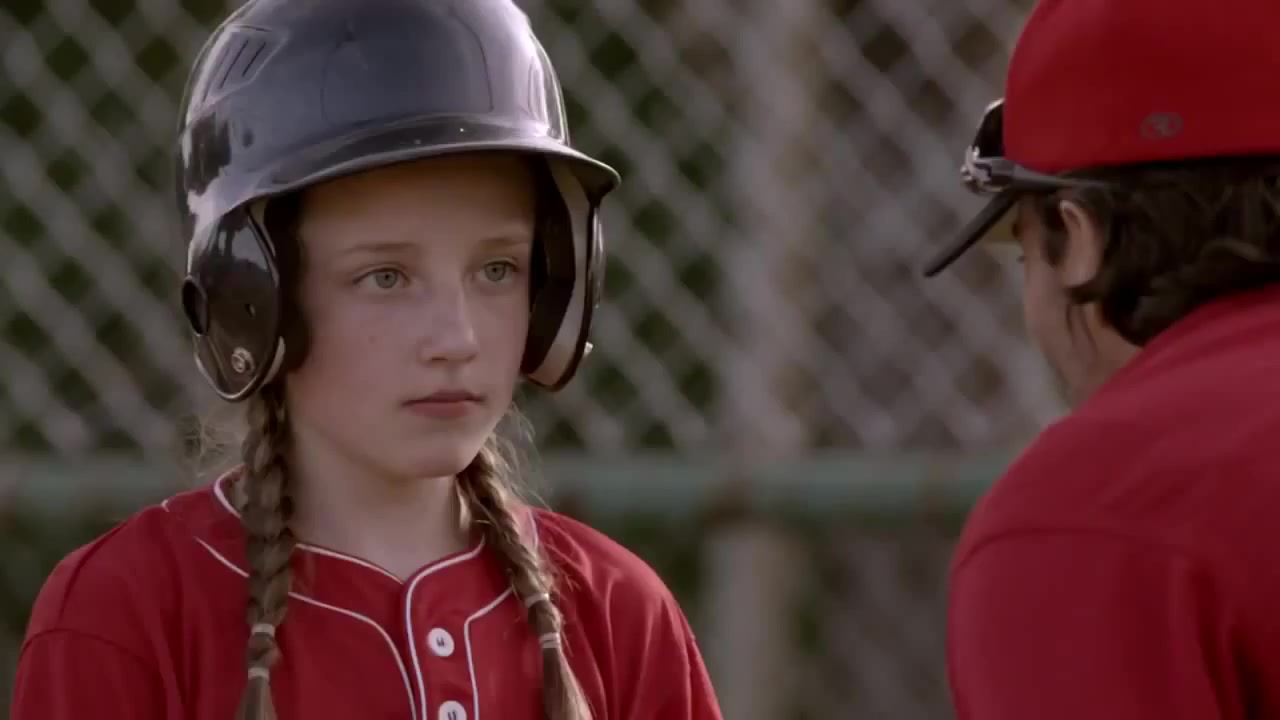}
\fullex{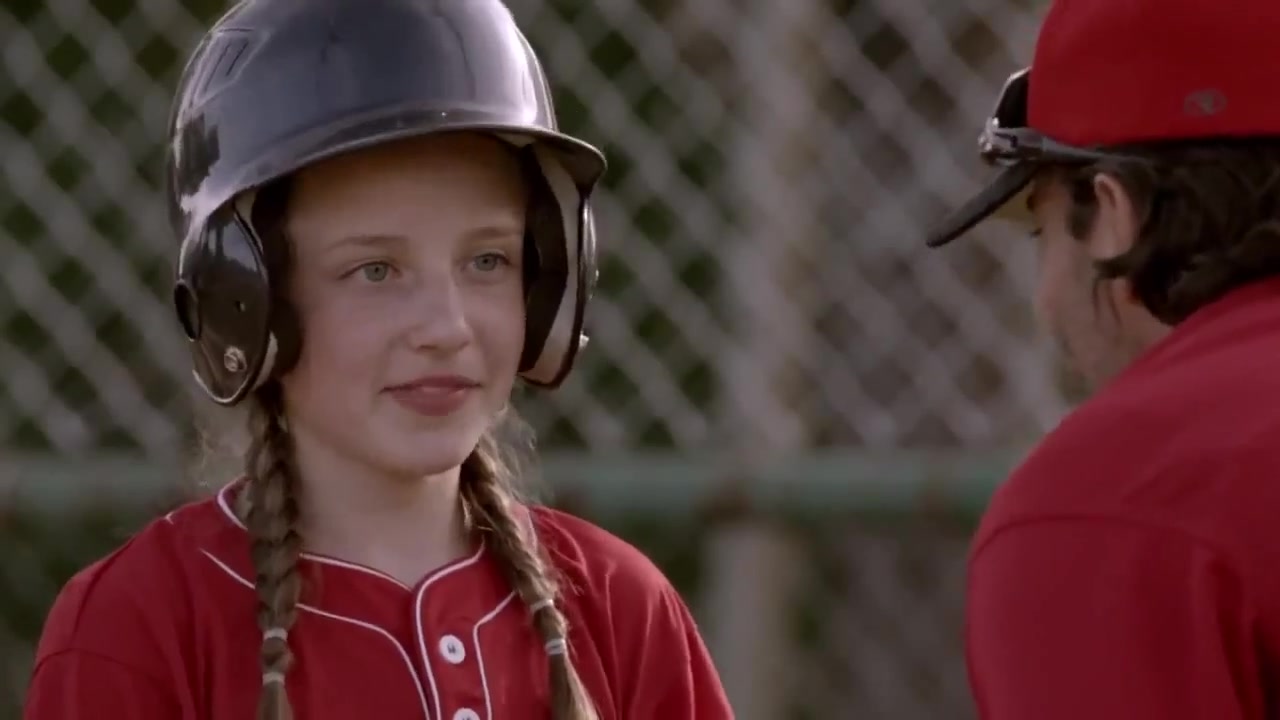}
\fullex{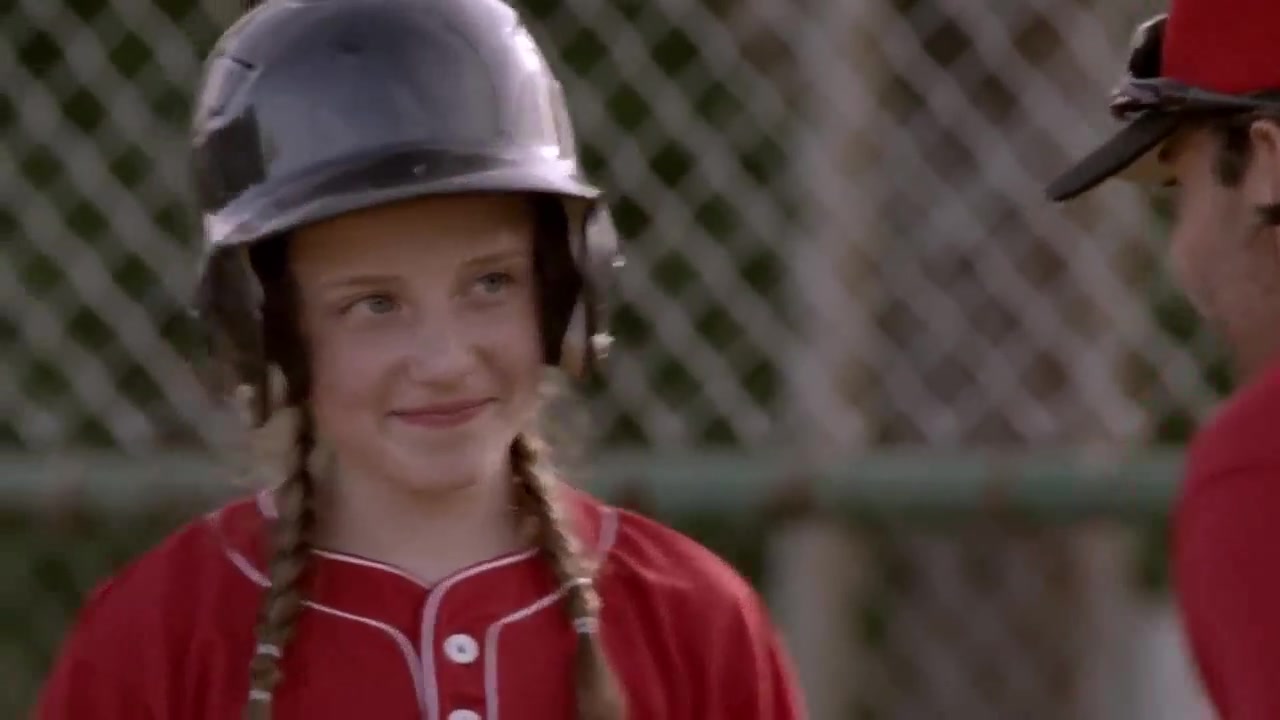}
\fullex{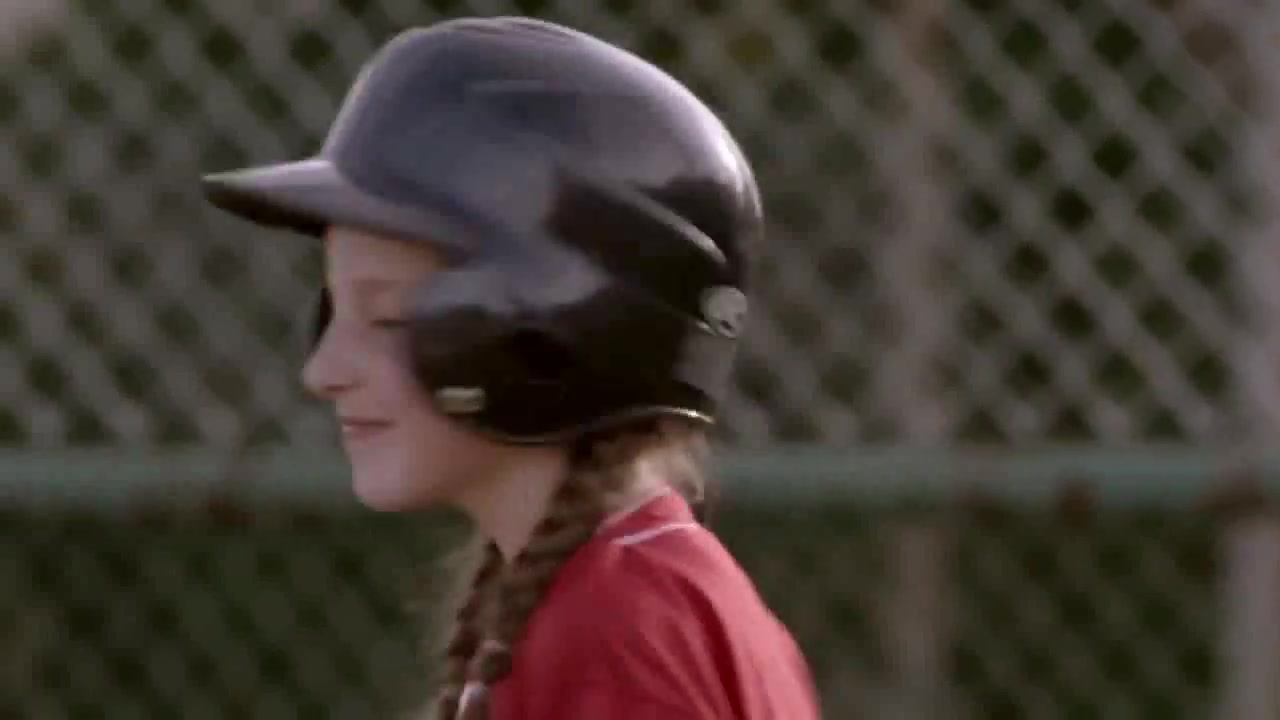}
\fullex{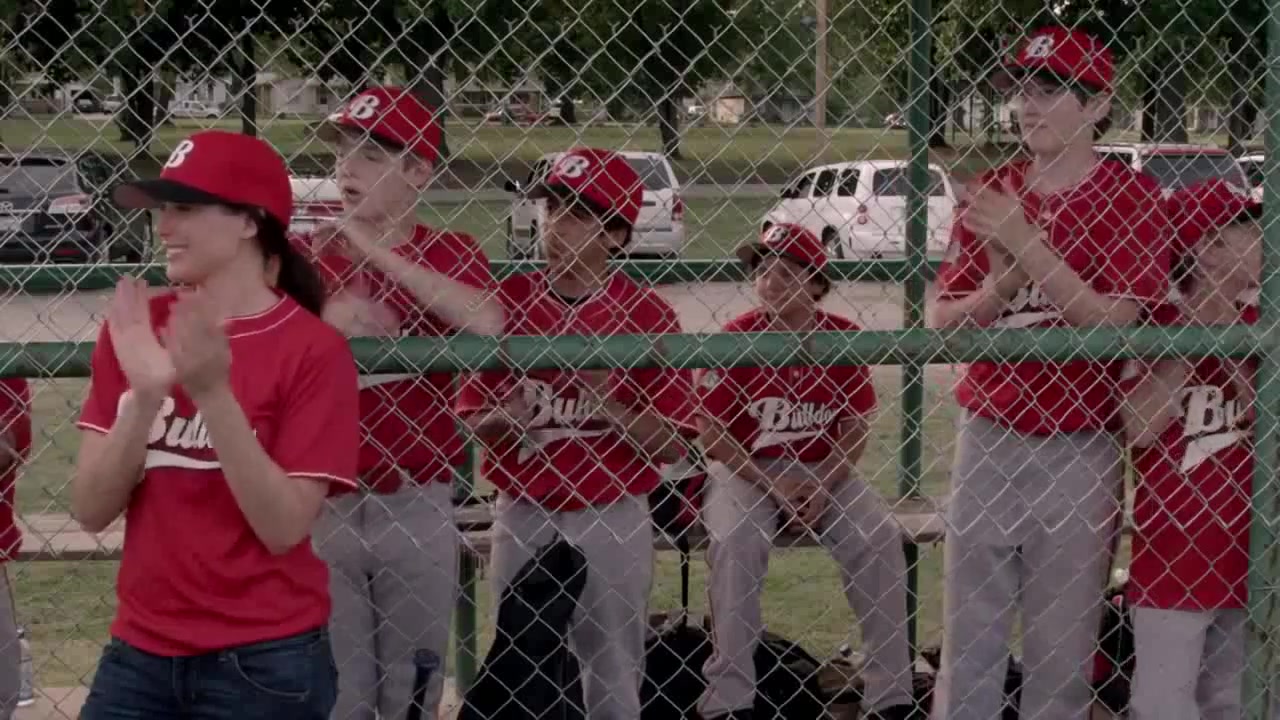}
\fullex{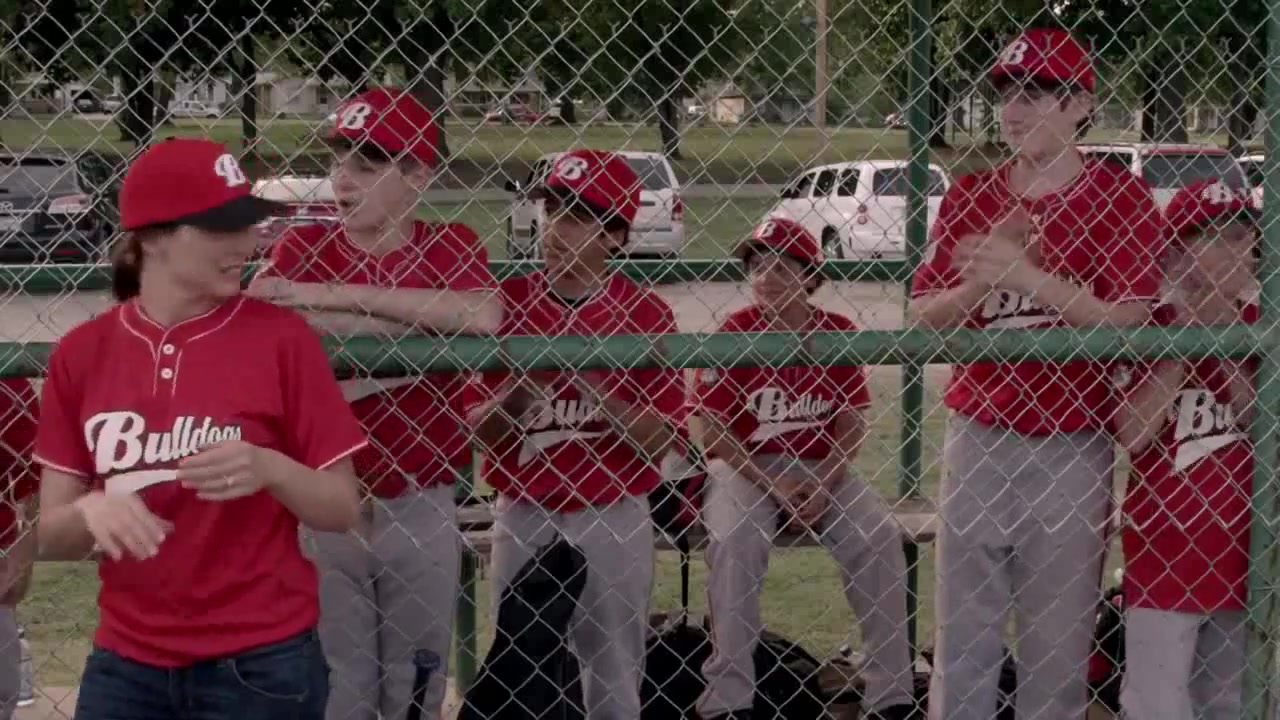}
\fullex{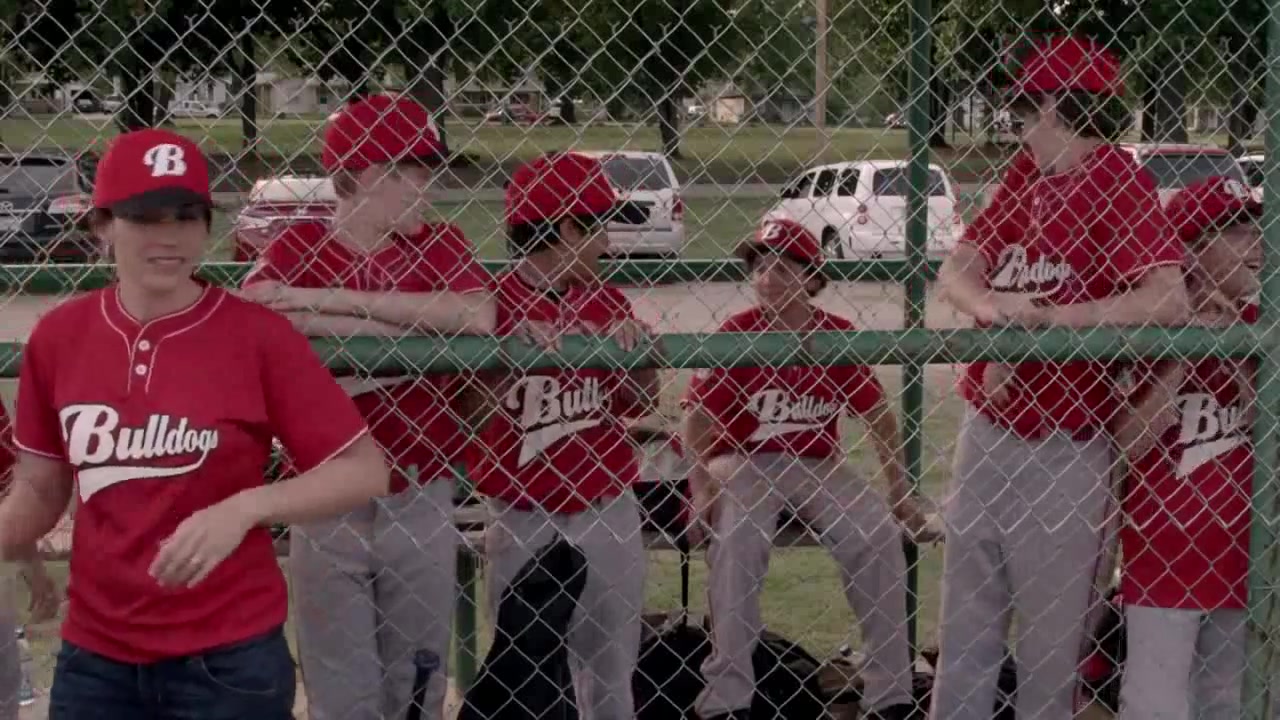}
\fullex{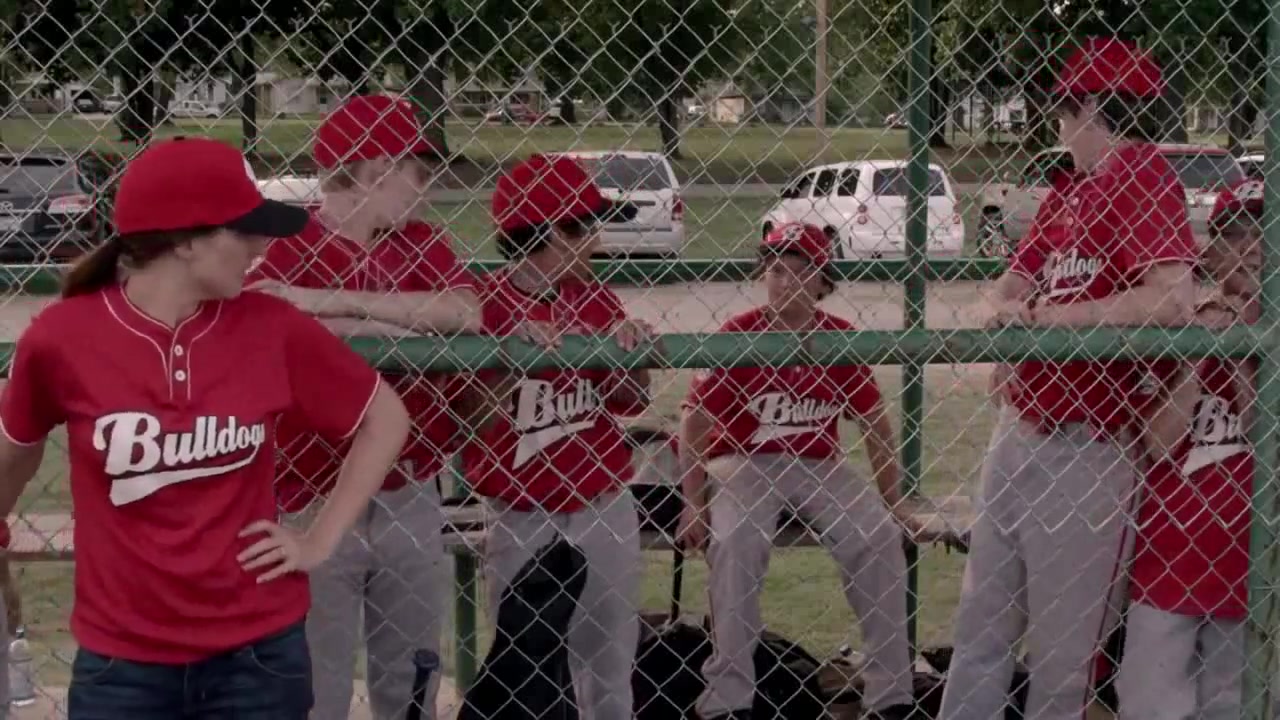}
\vspace{0.7mm}\\
\fullex{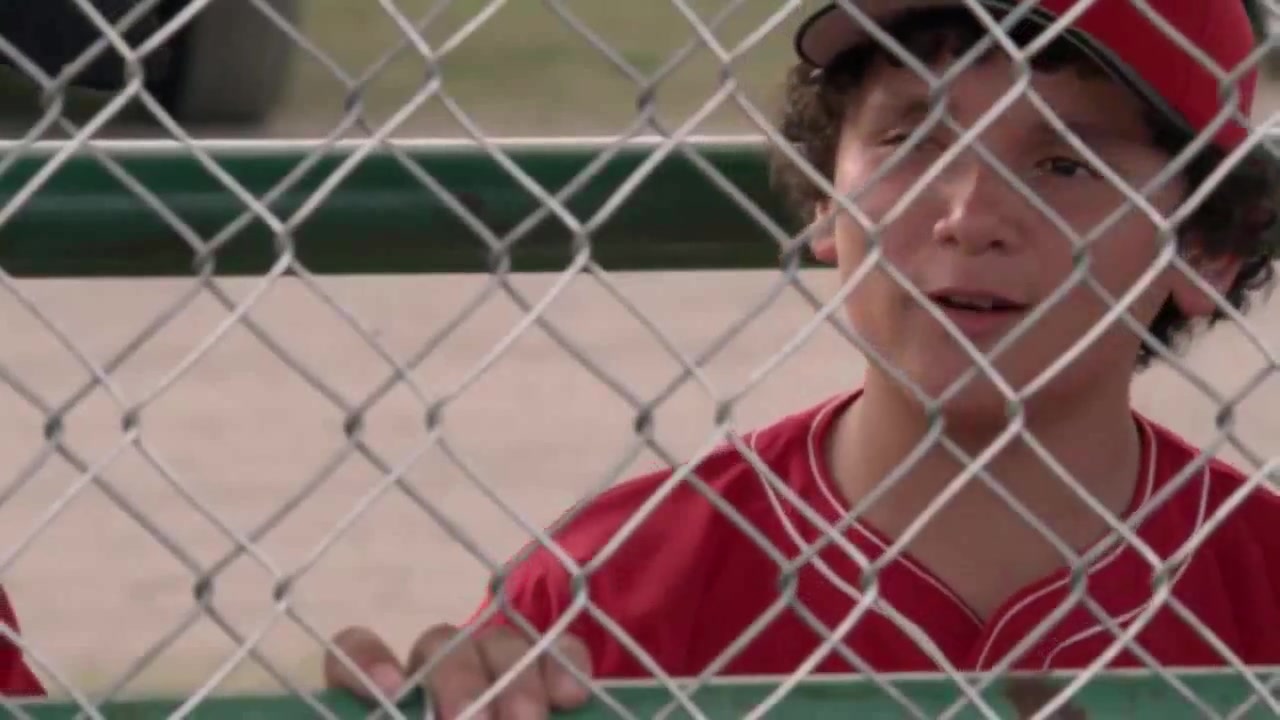}
\fullex{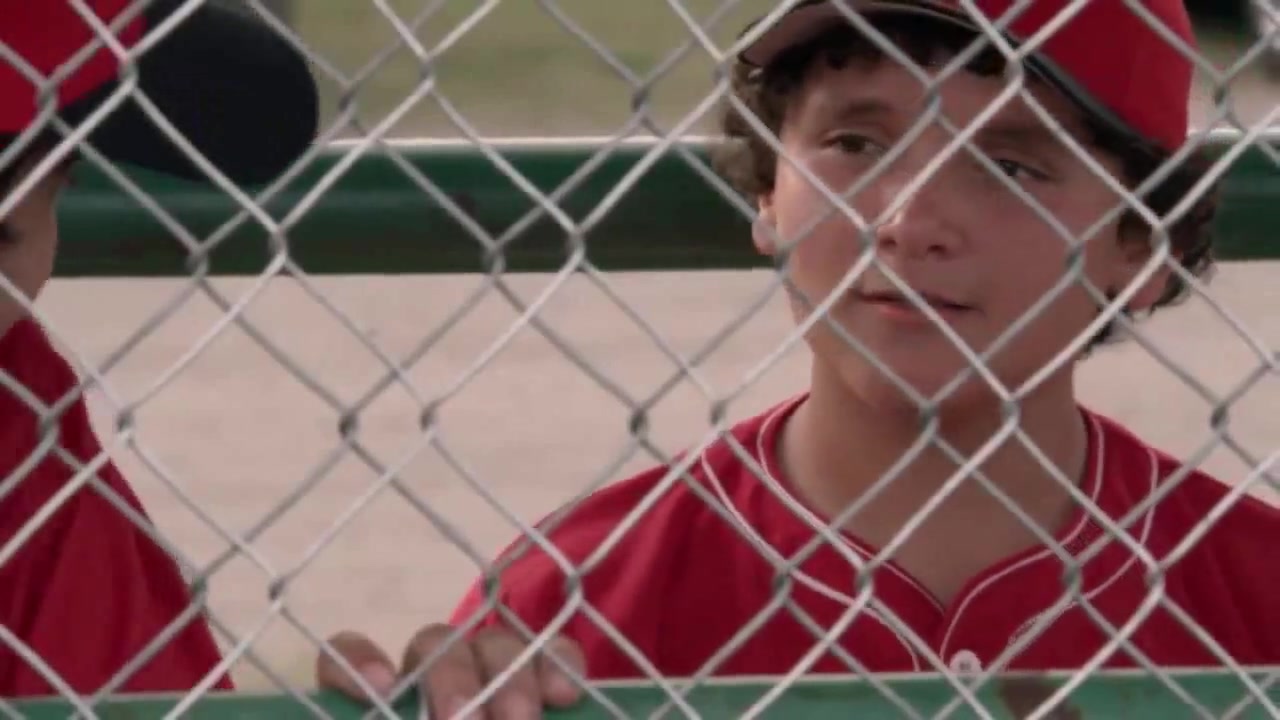}
\fullex{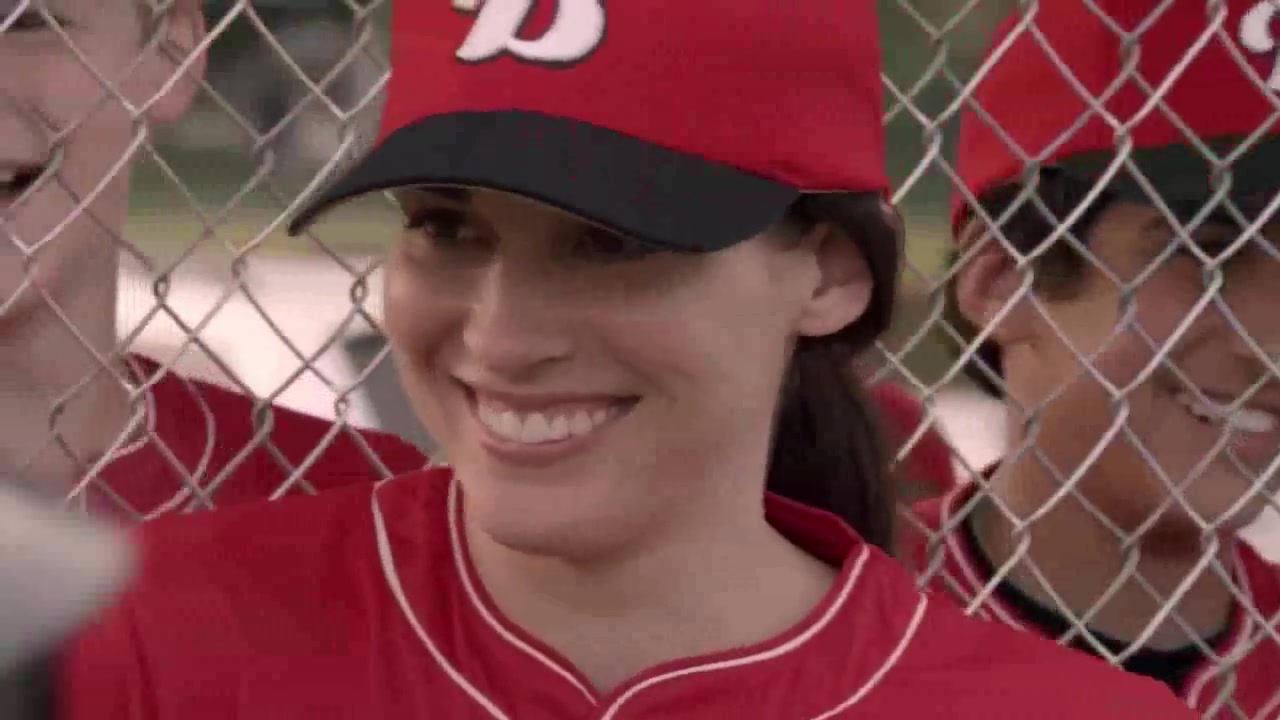}
\fullex{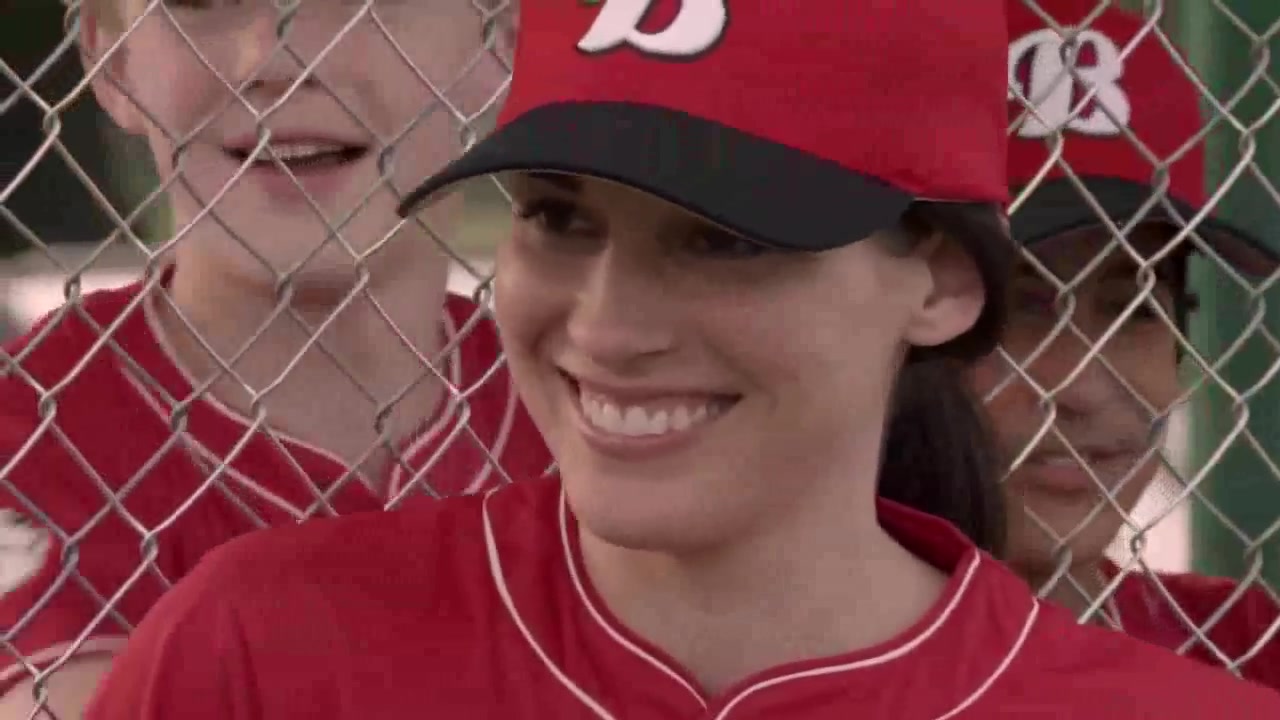}
\fullex{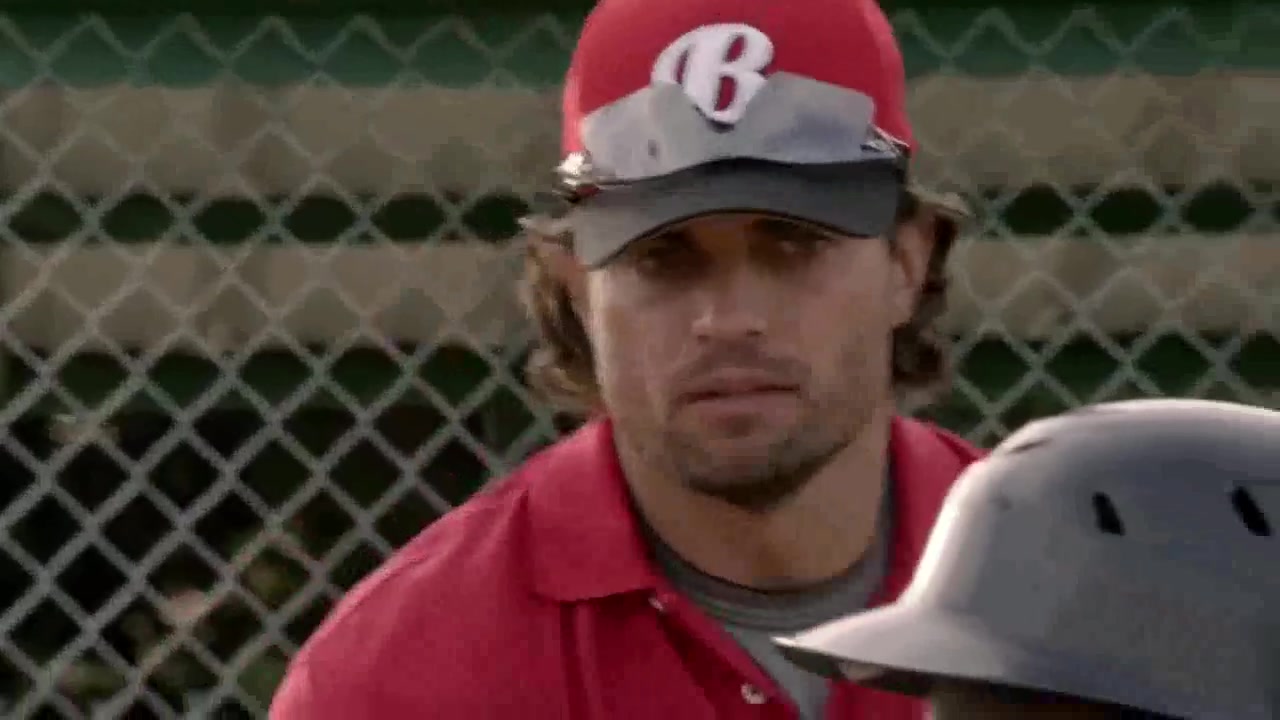}
\fullex{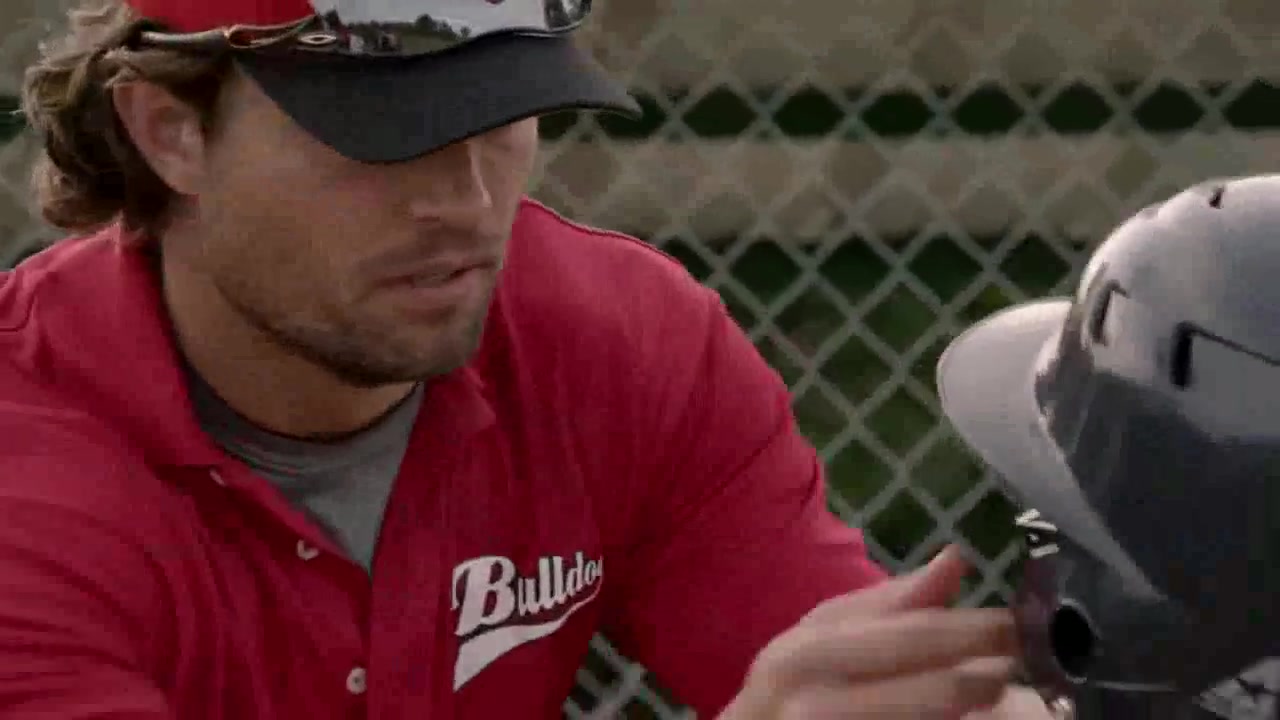}
\fullex{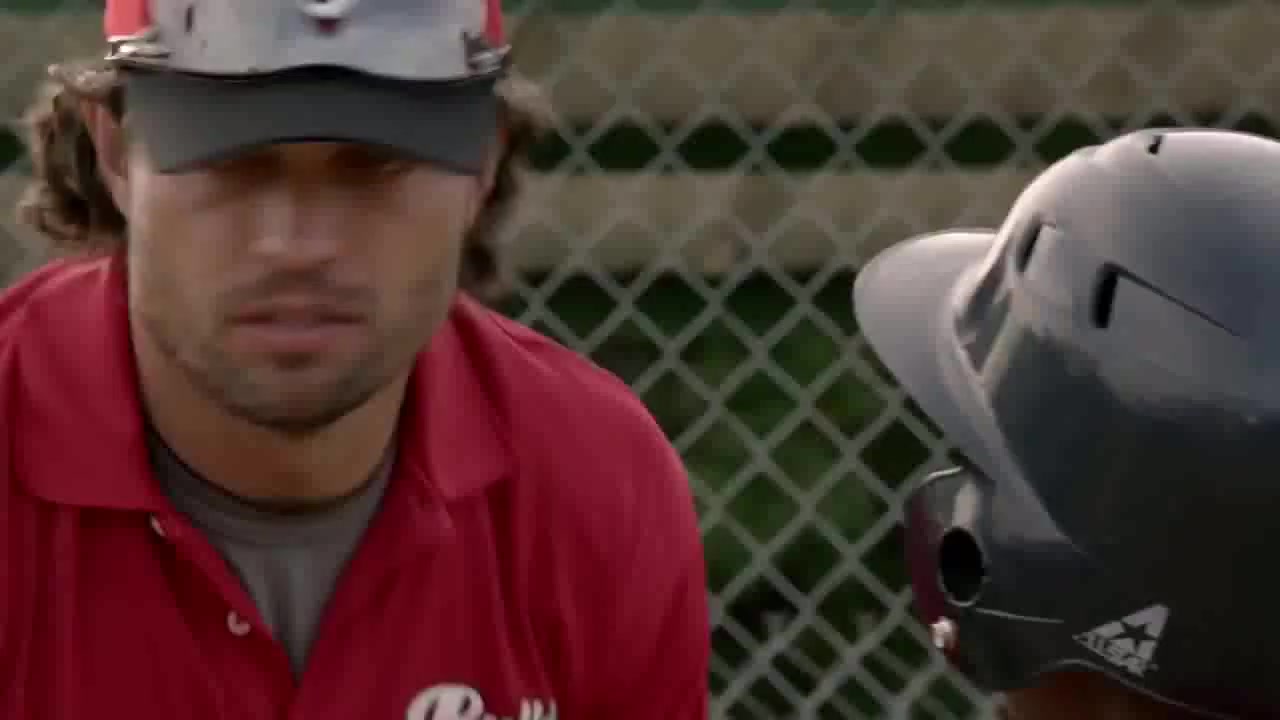}
\fullex{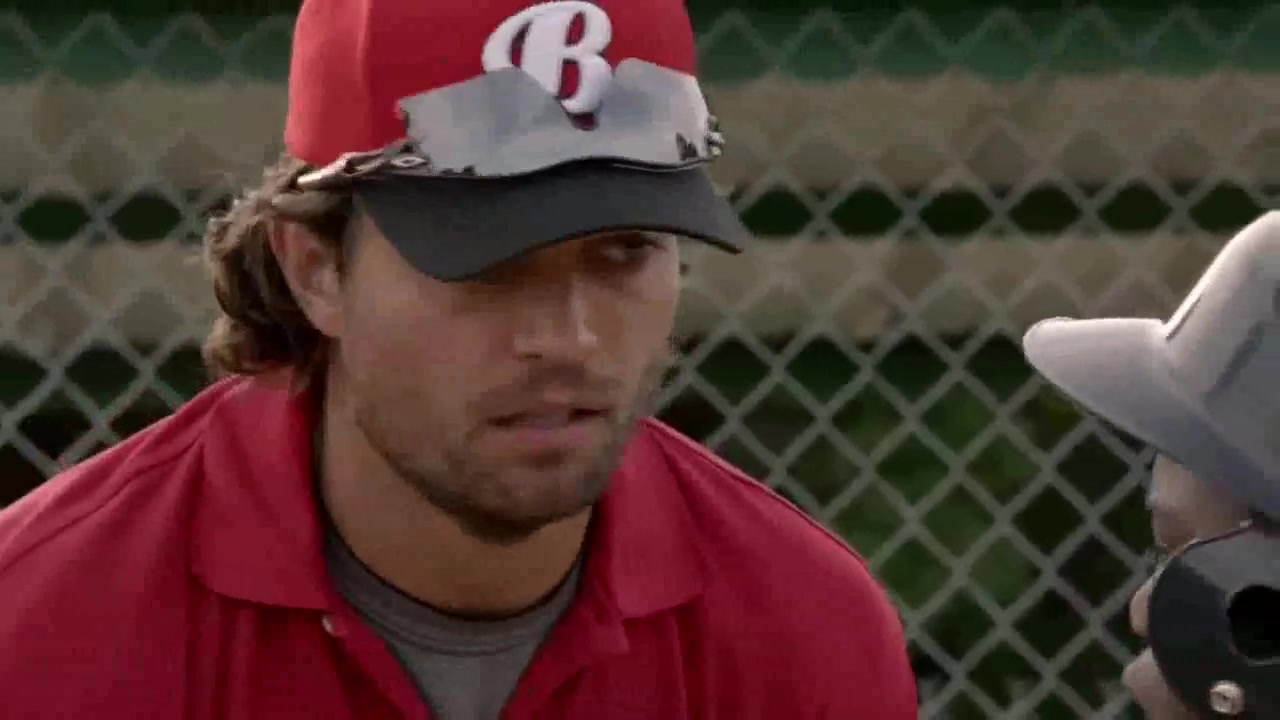}
\fullex{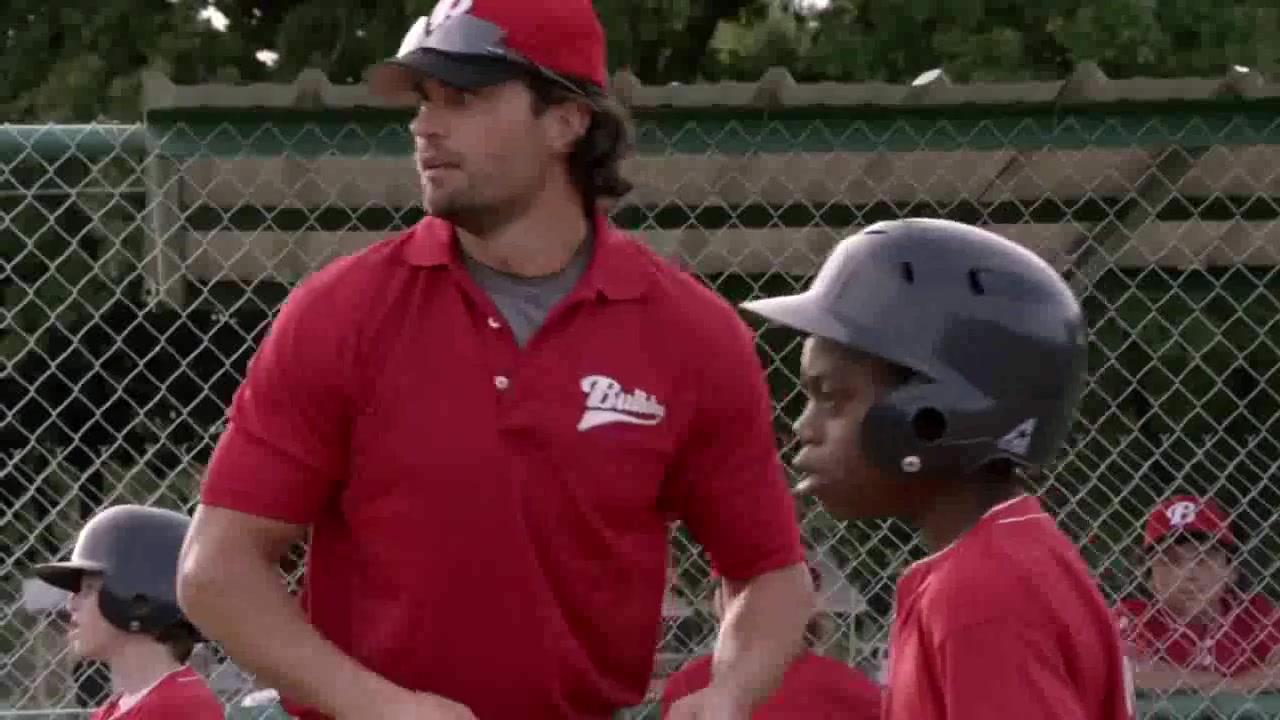}
\fullex{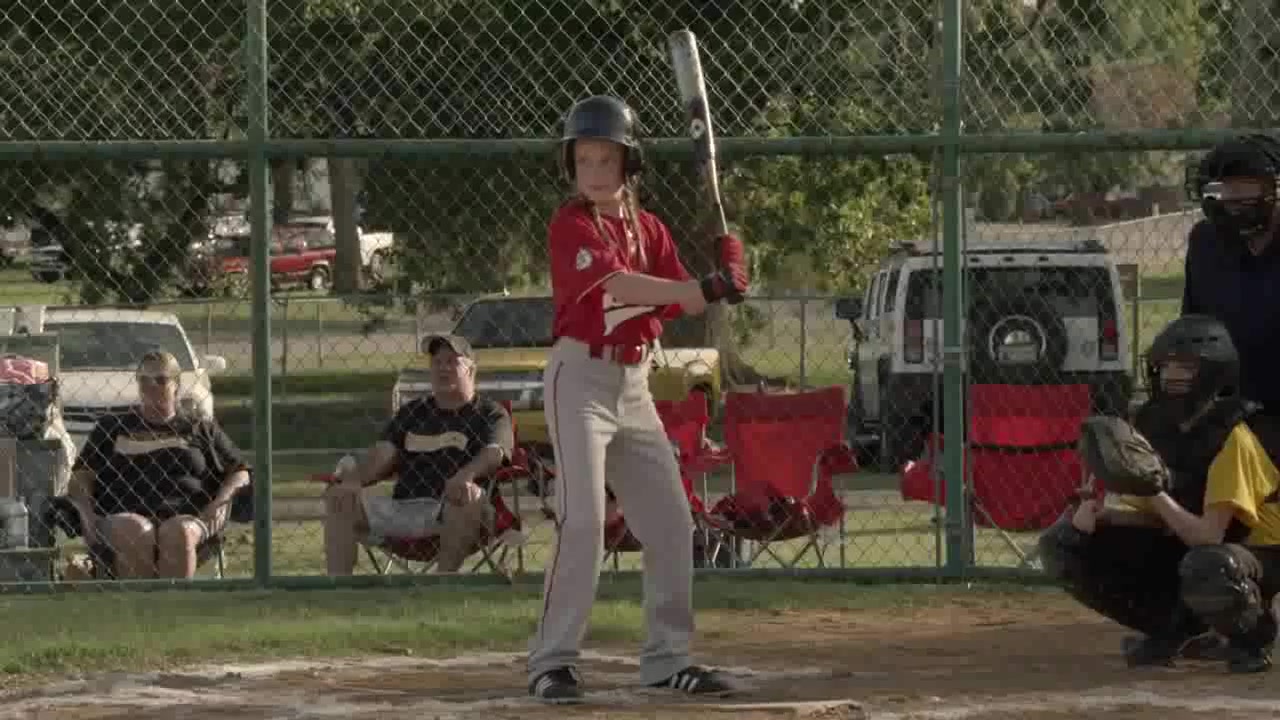}
\vspace{0.7mm}\\
\fullex{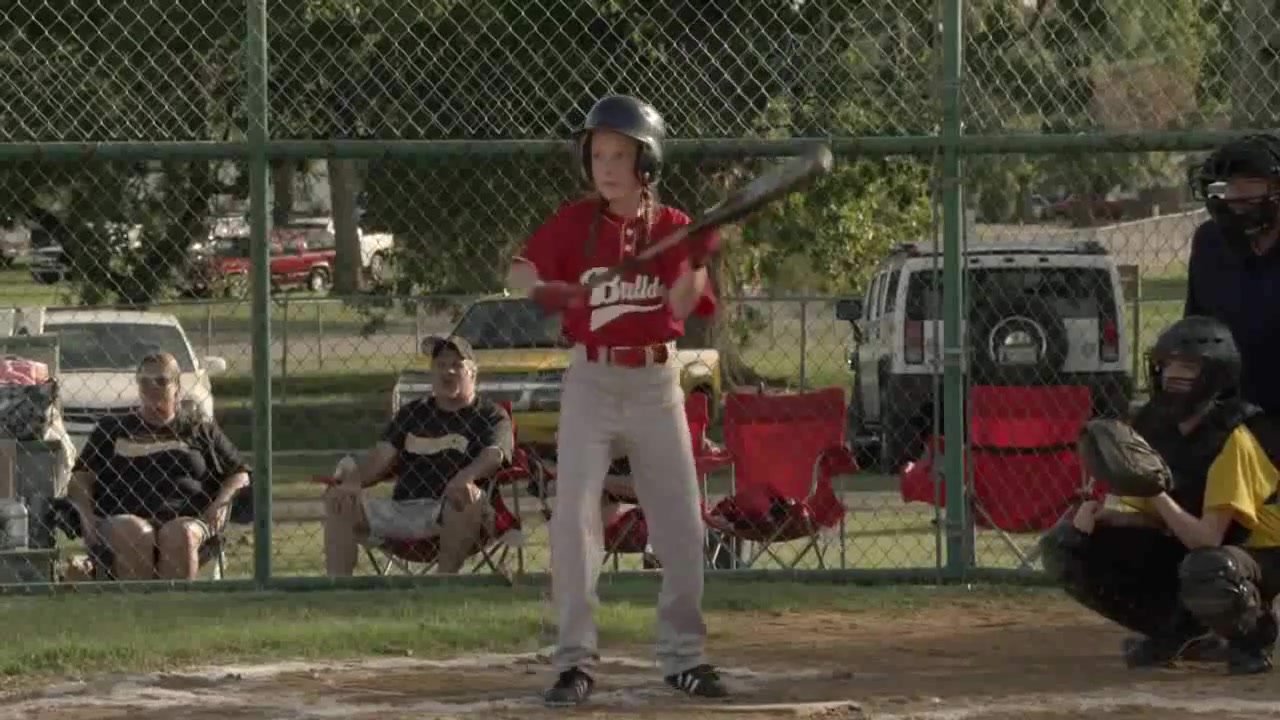}
\fullex{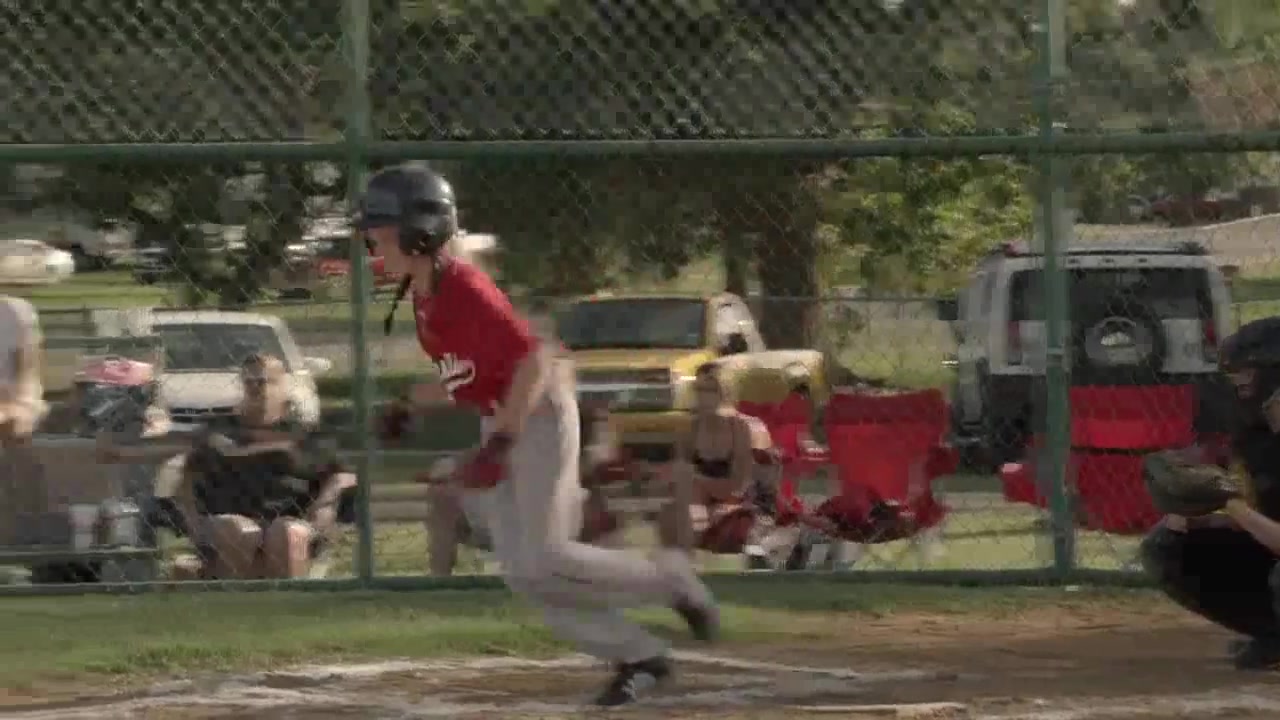}
\fullex{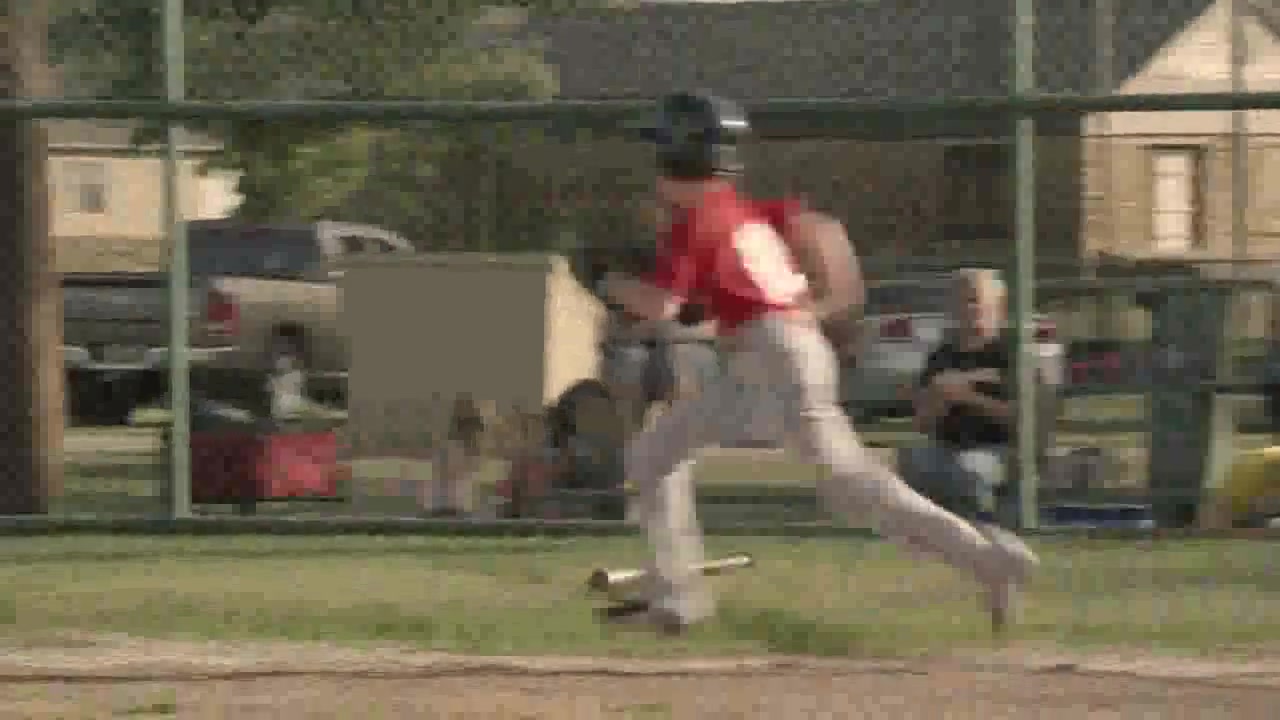}
\fullex{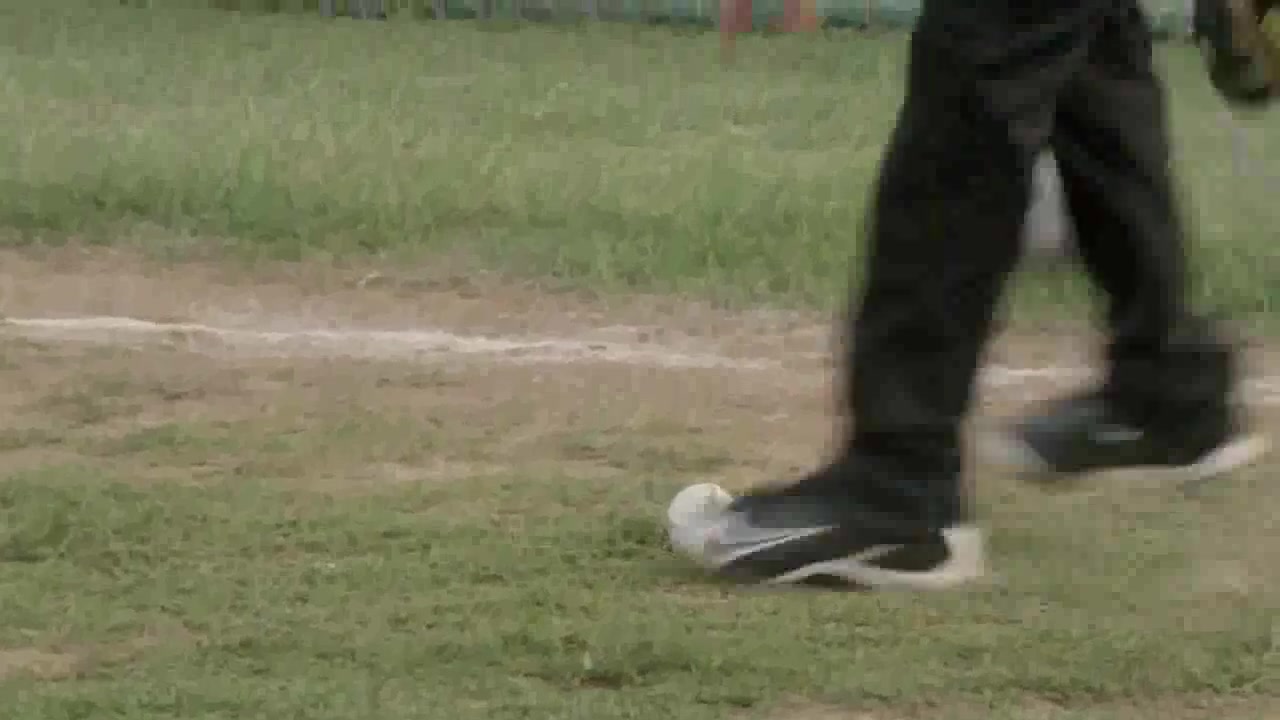}
\fullex{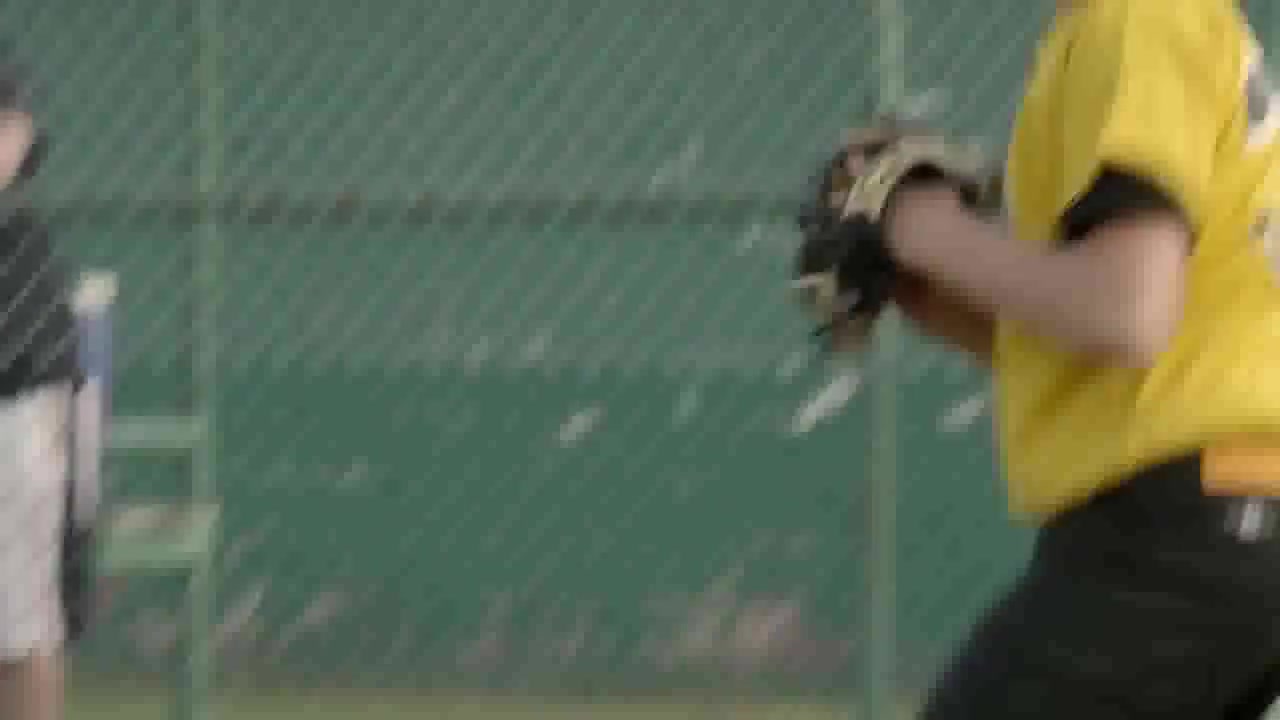}
\fullex{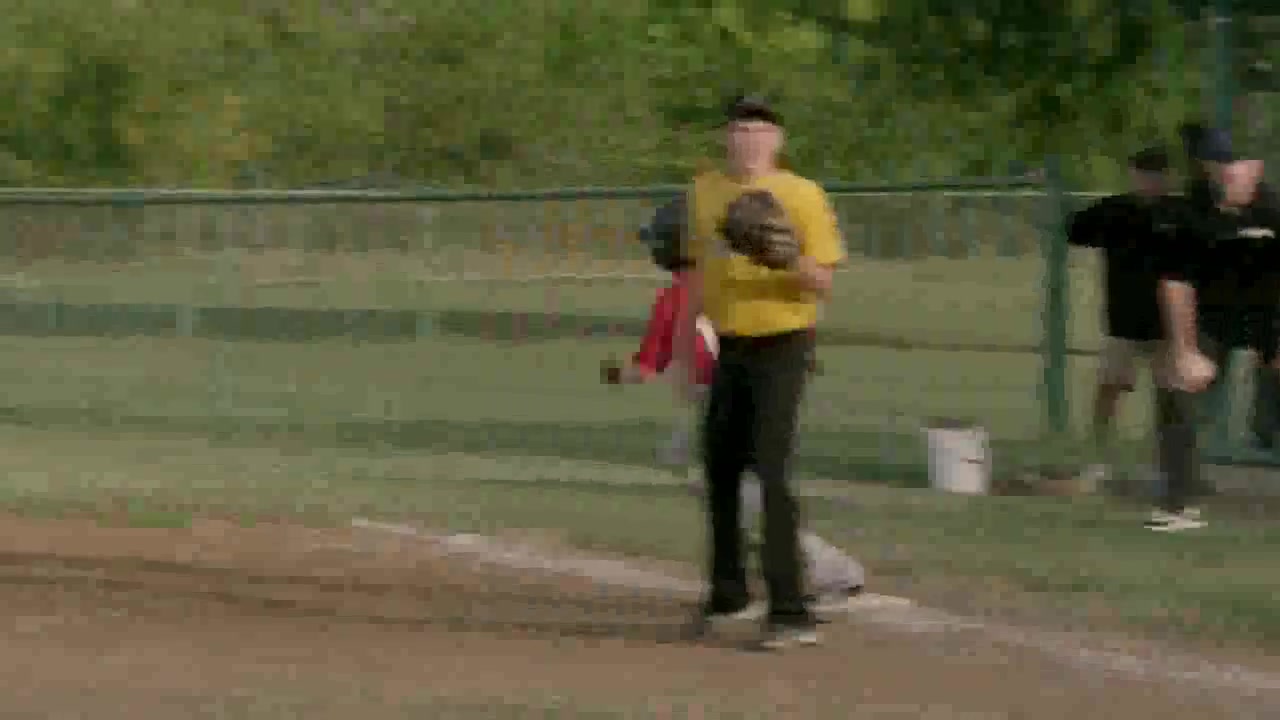}
\fullex{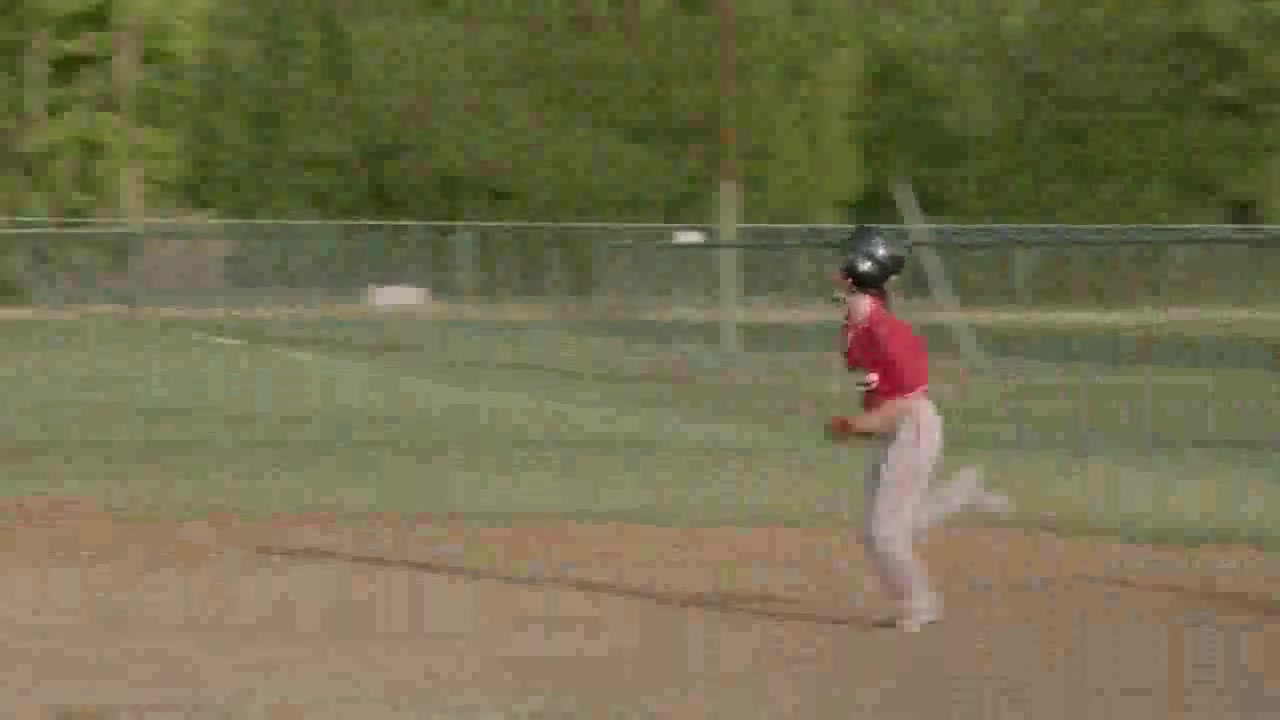}
\fullex{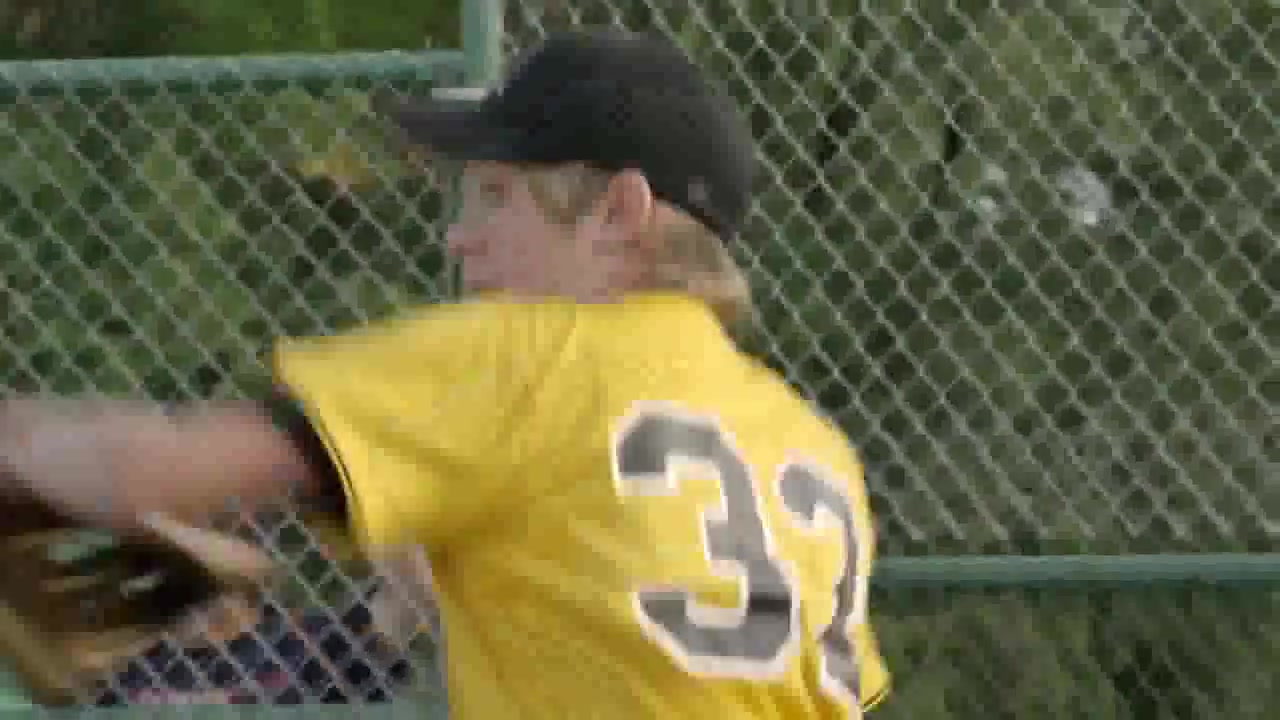}
\fullex{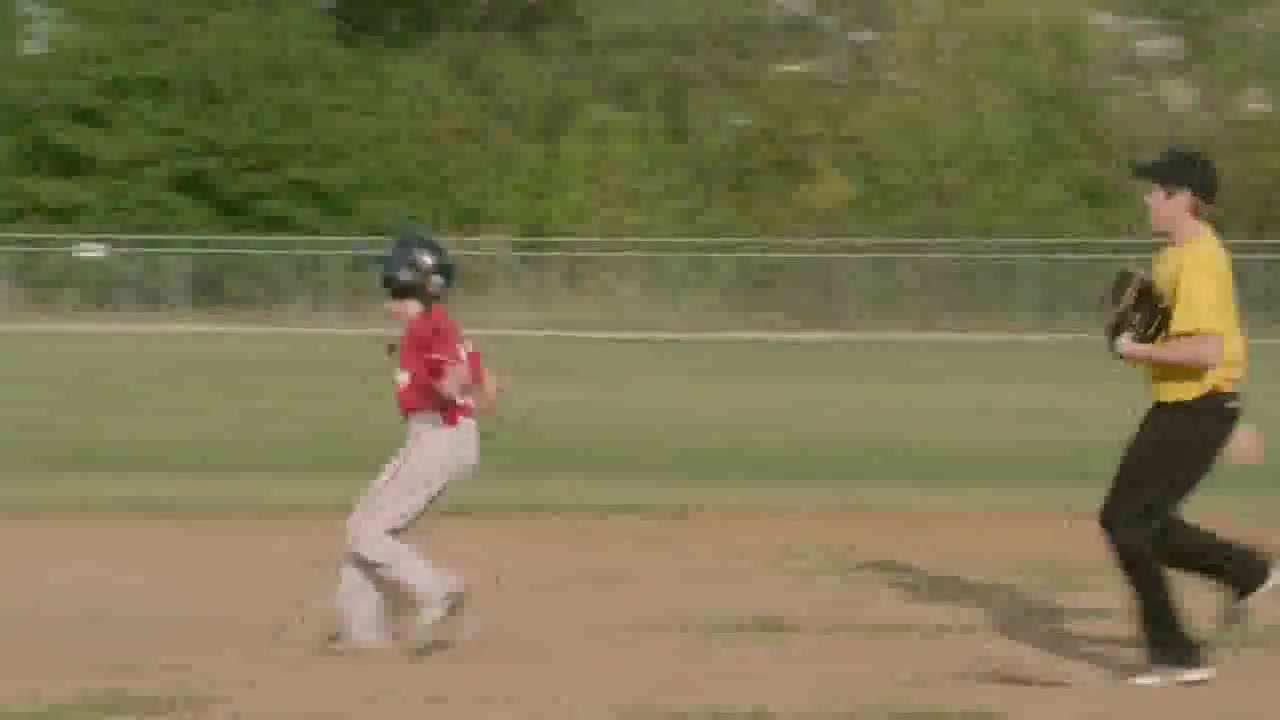}
\fullex{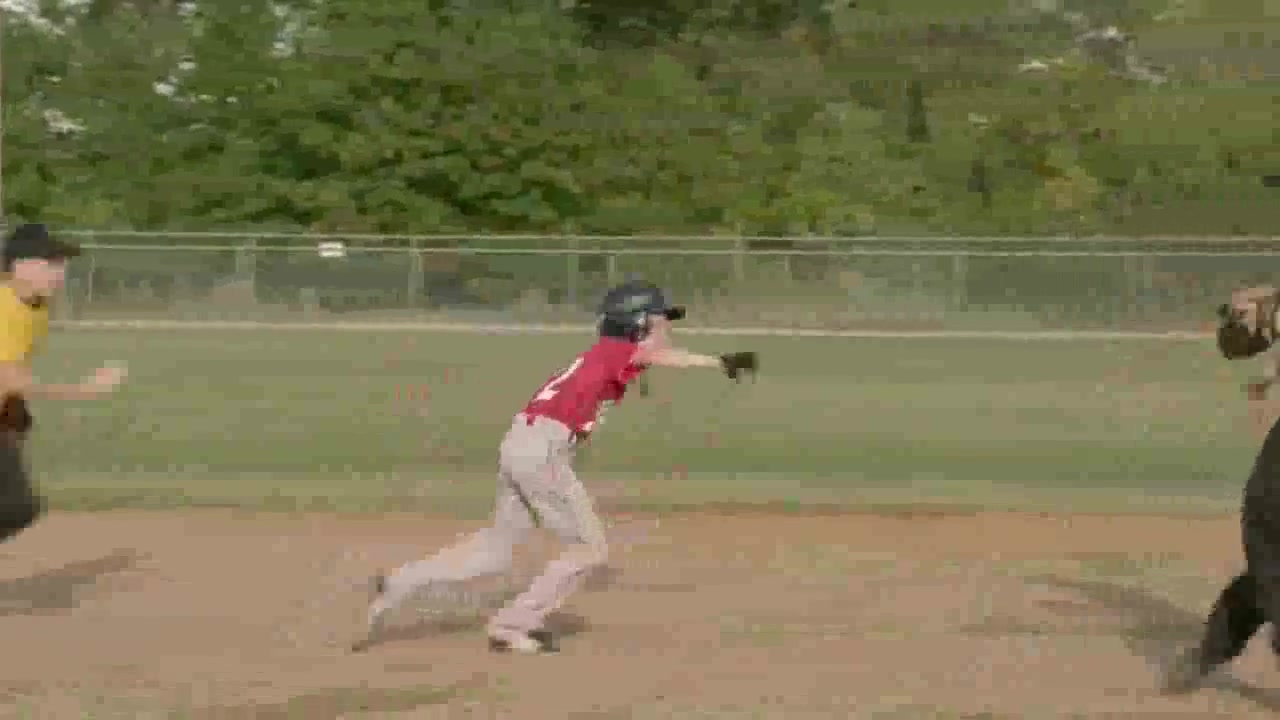}
\vspace{3mm}\\
\fullex{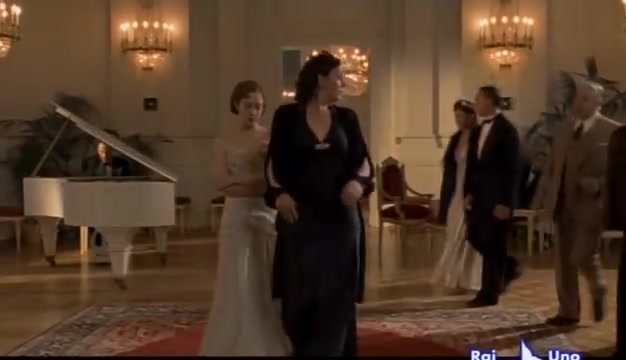}
\fullex{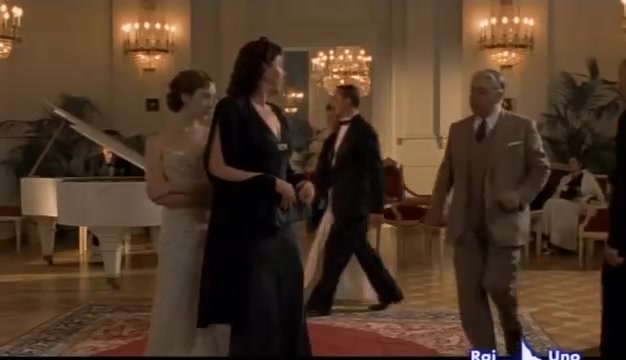}
\fullex{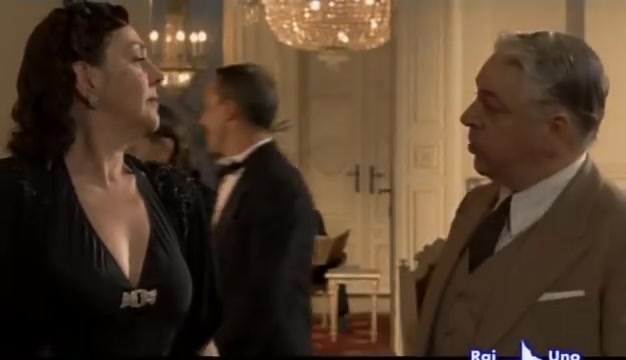}
\fullex{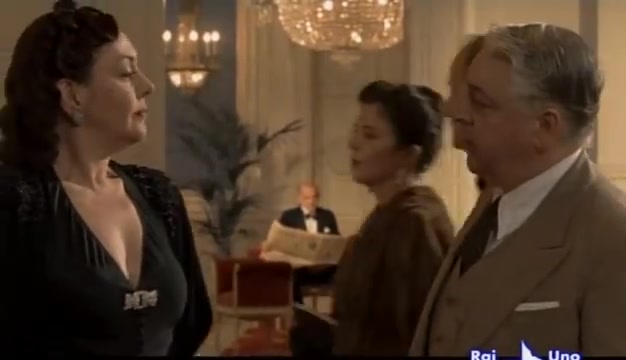}
\fullex{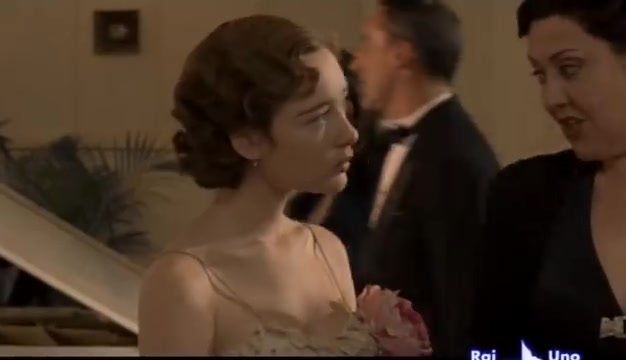}
\fullex{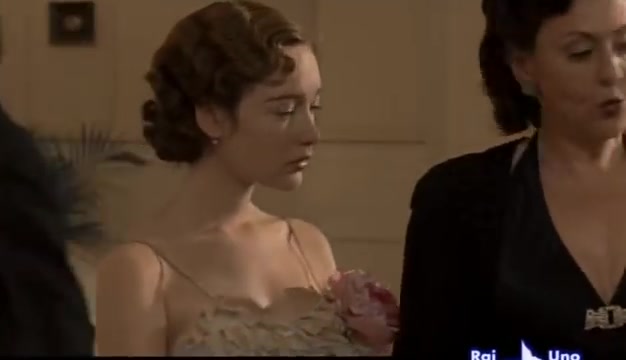}
\fullex{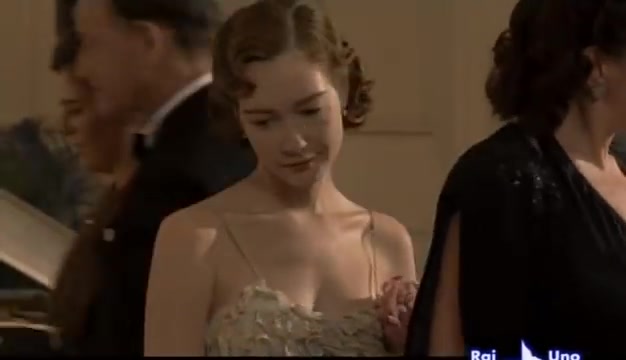}
\fullex{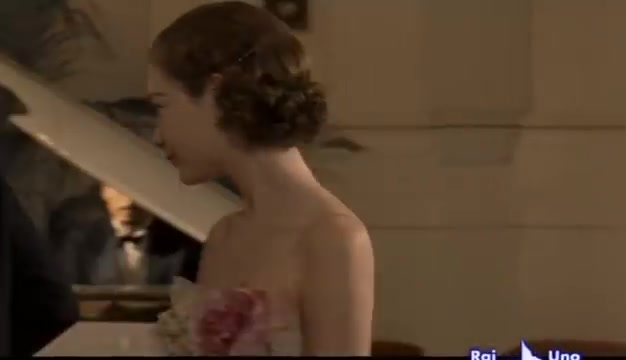}
\fullex{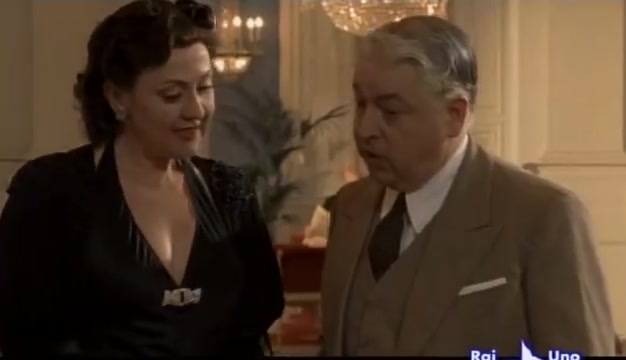}
\fullex{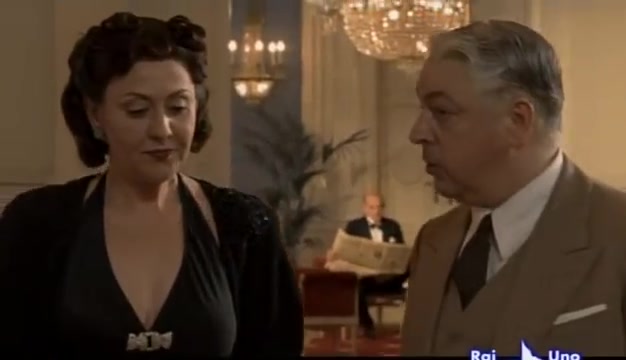}
\vspace{0.7mm}\\
\fullex{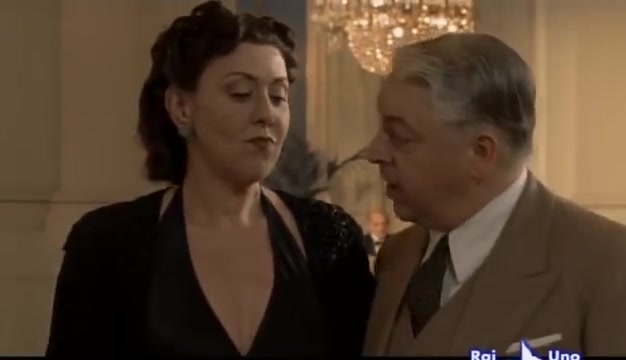}
\fullex{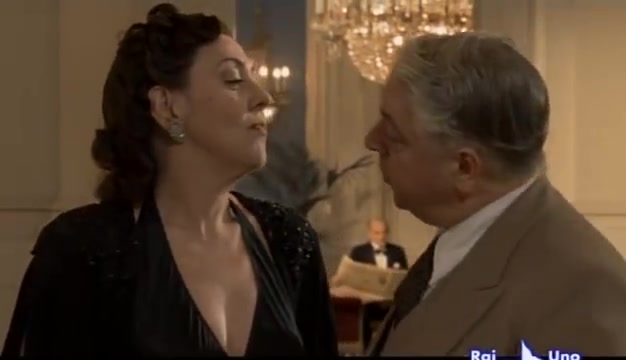}
\fullex{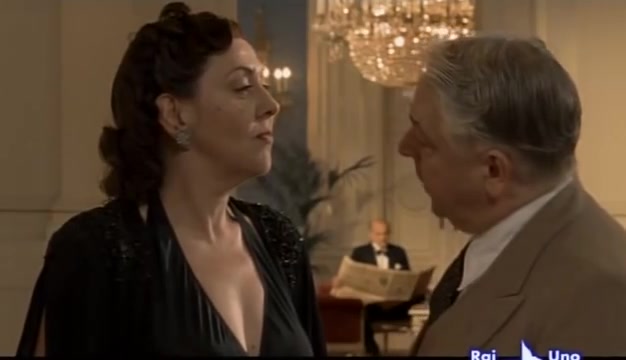}
\fullex{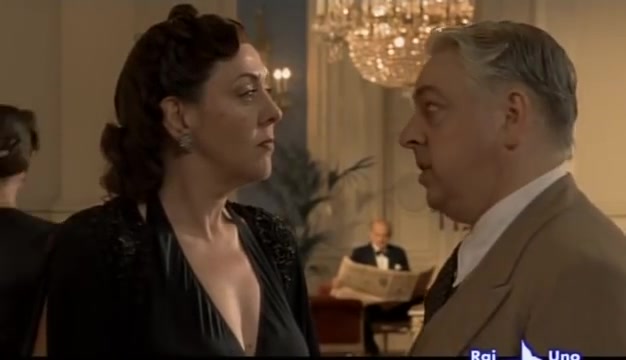}
\fullex{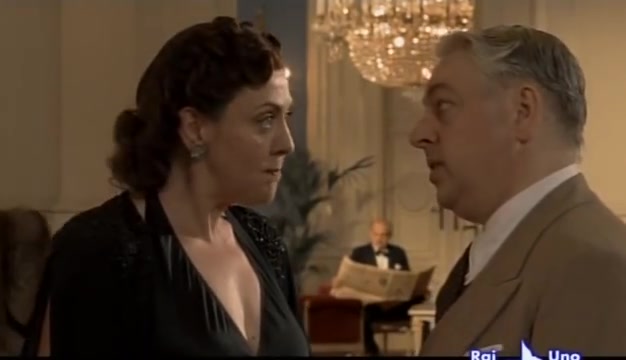}
\fullex{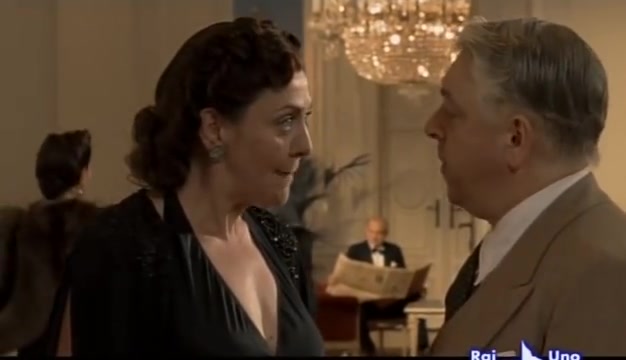}
\fullex{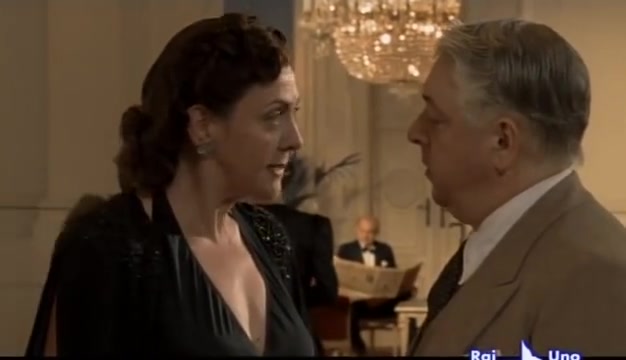}
\fullex{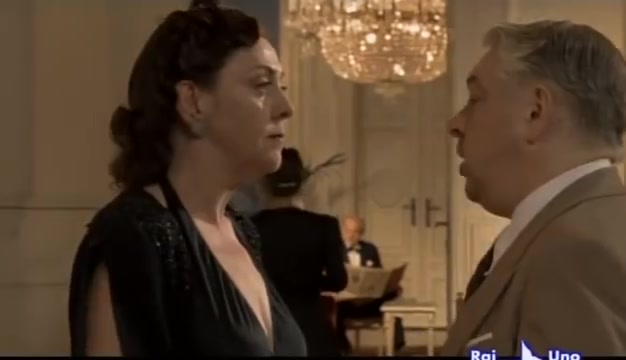}
\fullex{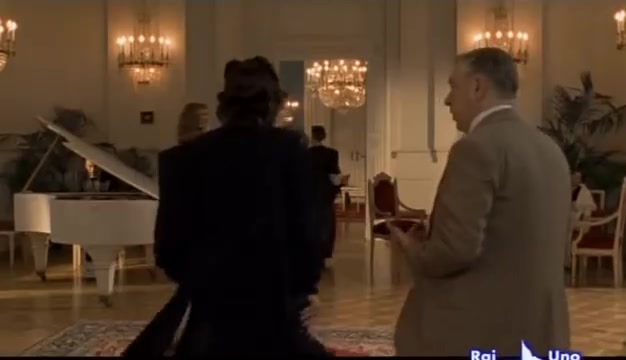}
\fullex{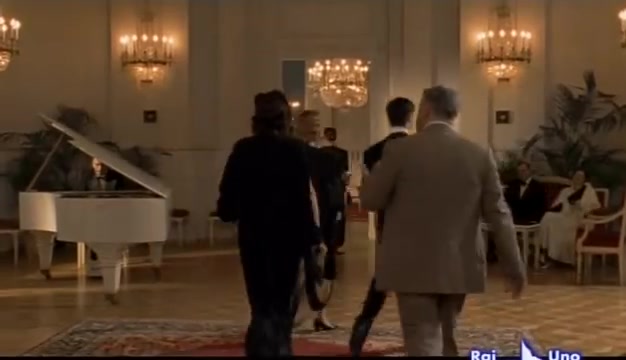}
\vspace{0.7mm}\\
\fullex{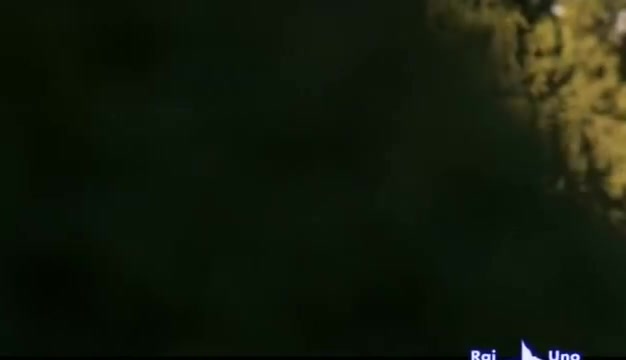}
\fullex{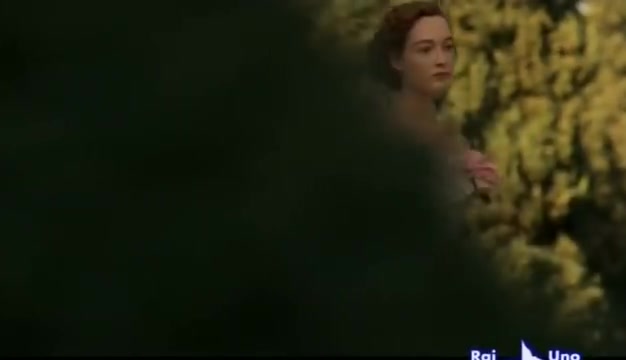}
\fullex{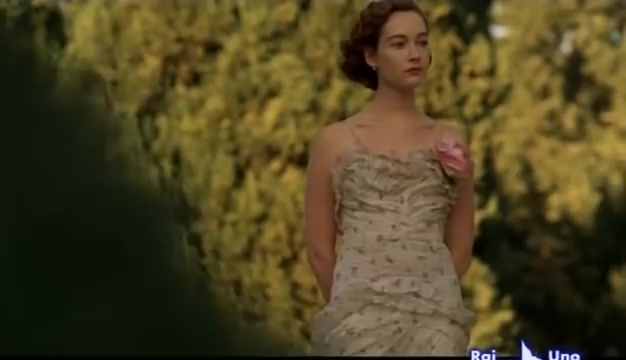}
\fullex{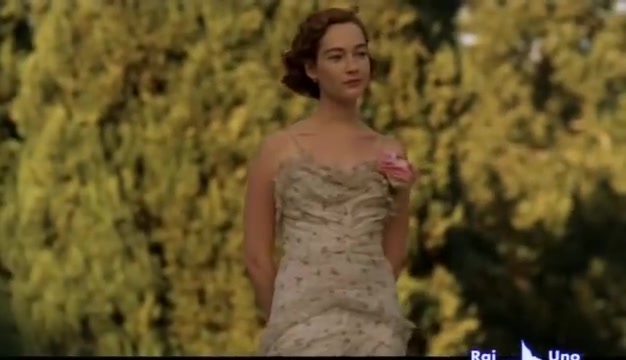}
\fullex{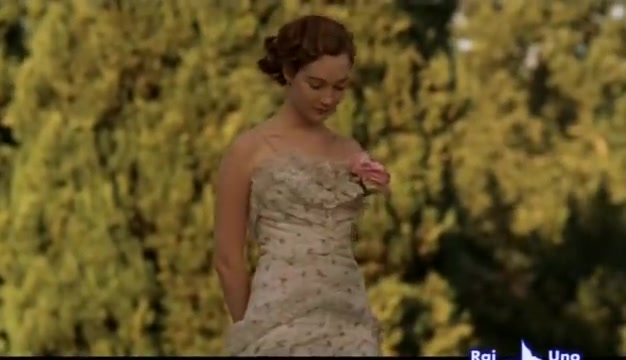}
\fullex{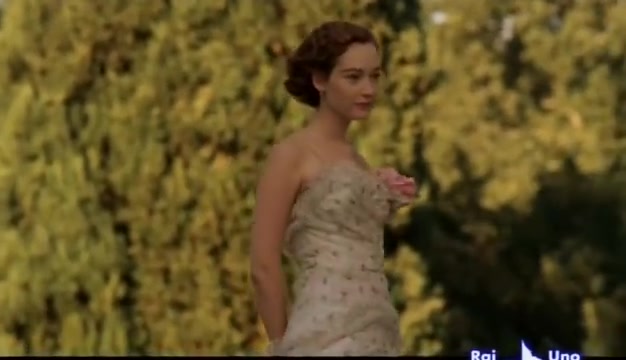}
\fullex{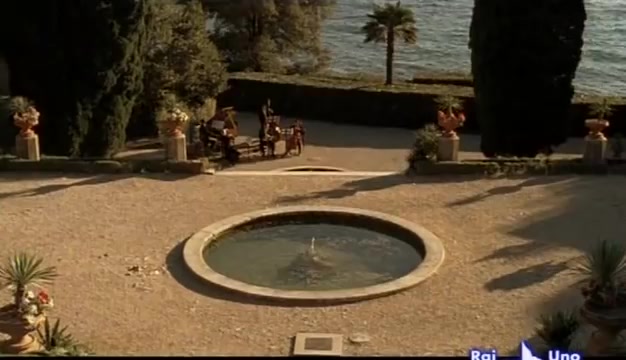}
\fullex{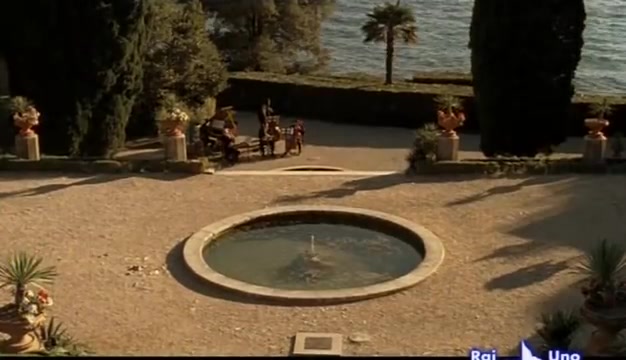}
\fullex{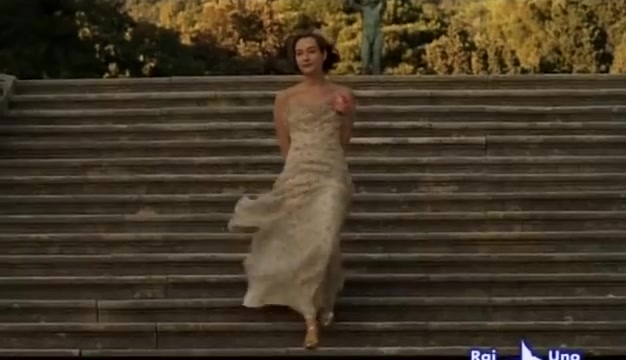}
\fullex{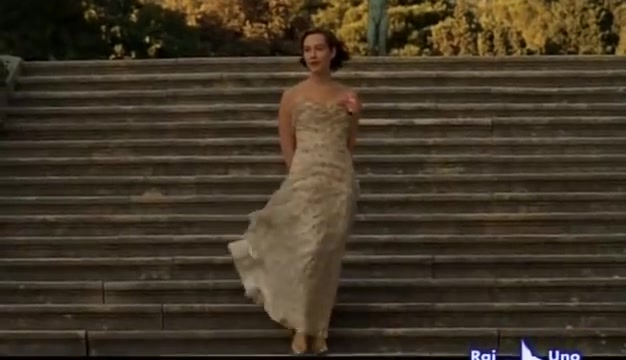}
\vspace{0.7mm}\\
\fullex{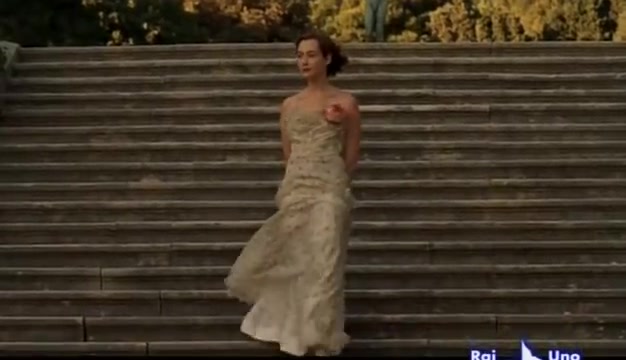}
\fullex{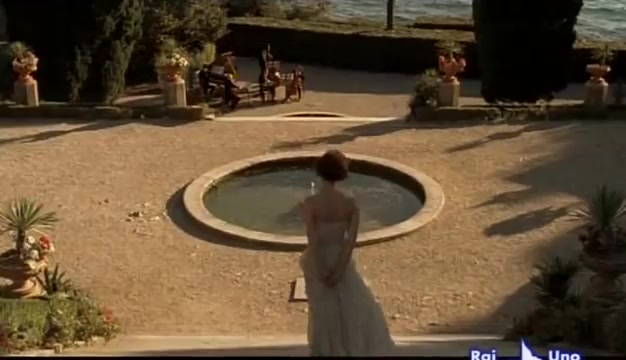}
\fullex{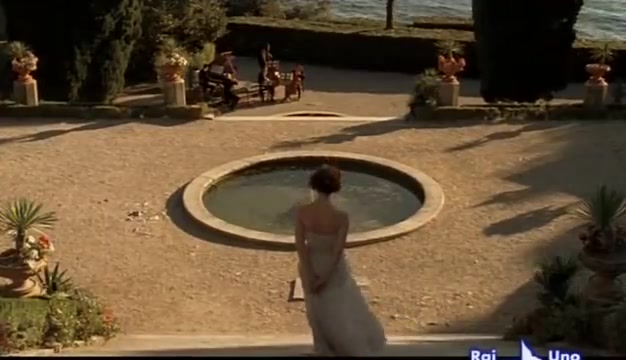}
\fullex{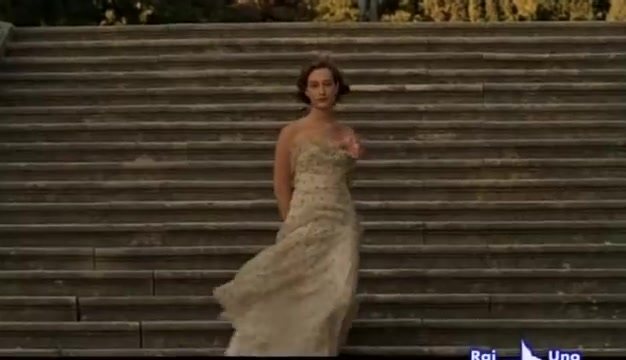}
\fullex{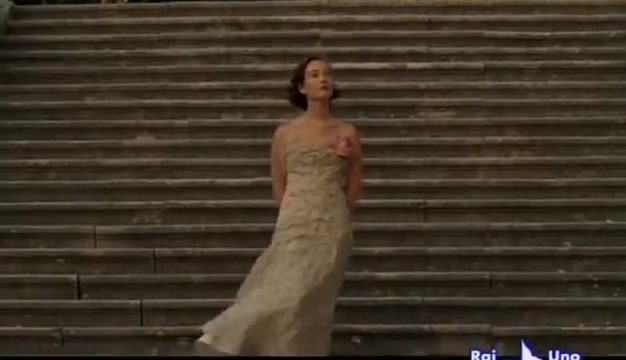}
\fullex{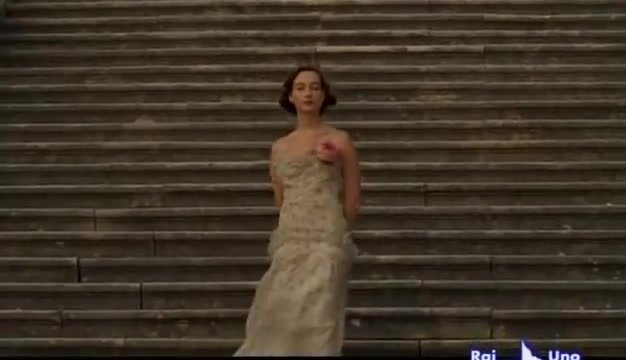}
\fullex{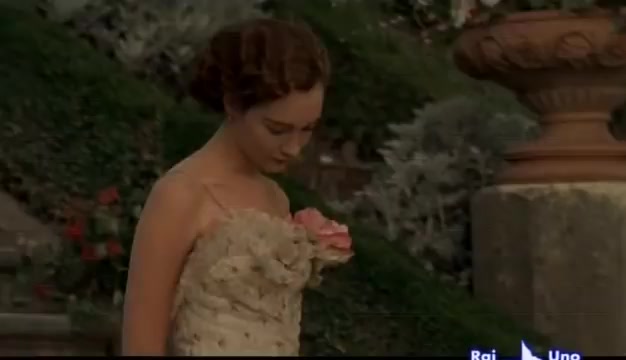}
\fullex{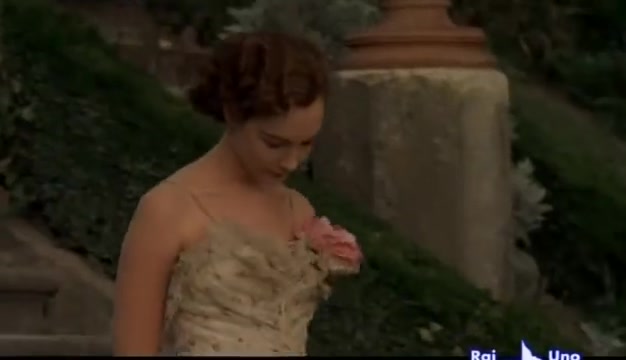}
\fullex{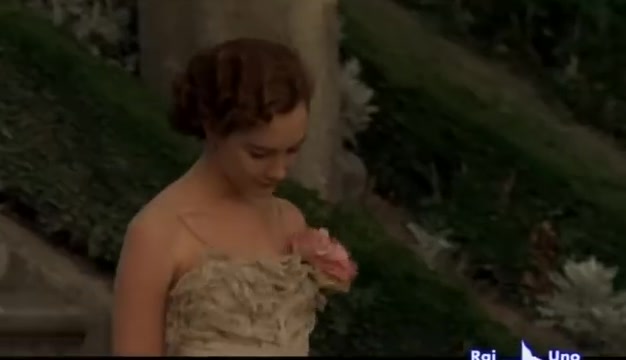}
\fullex{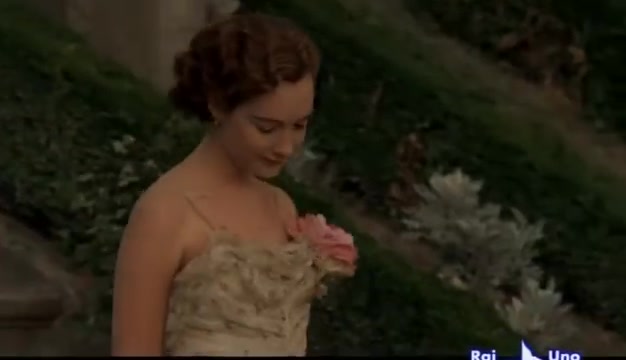}
\vspace{0.7mm}\\
\fullex{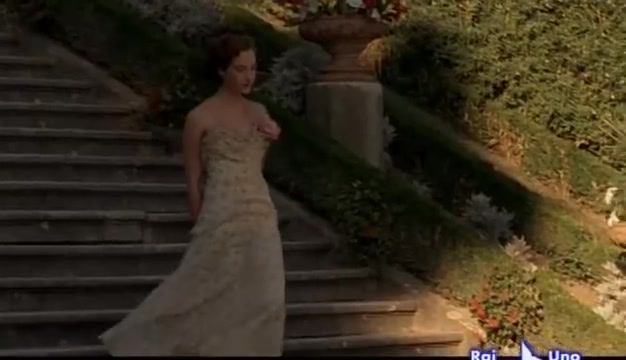}
\fullex{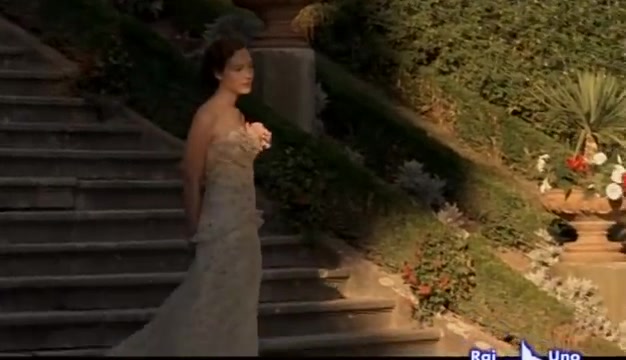}
\fullex{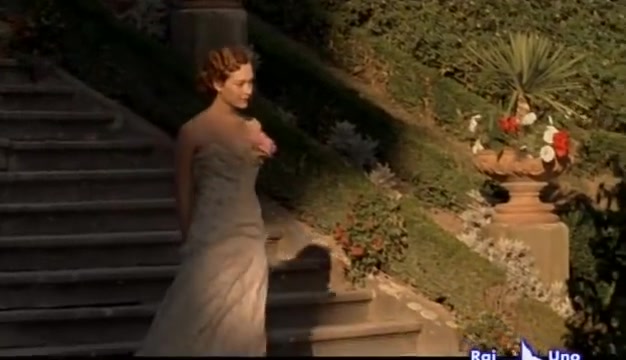}
\fullex{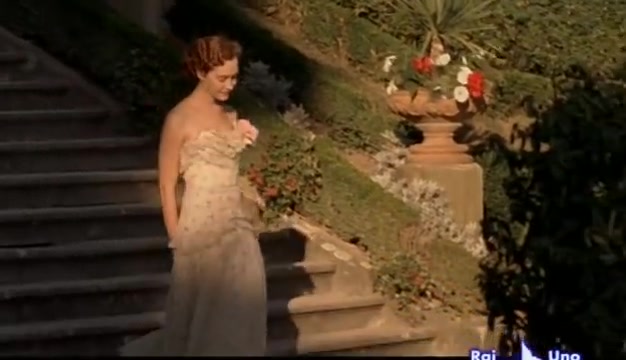}
\fullex{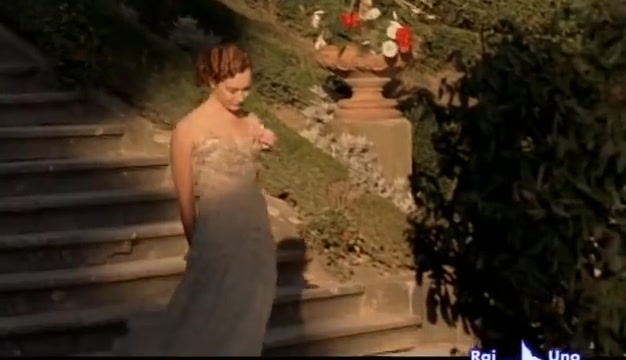}
\fullex{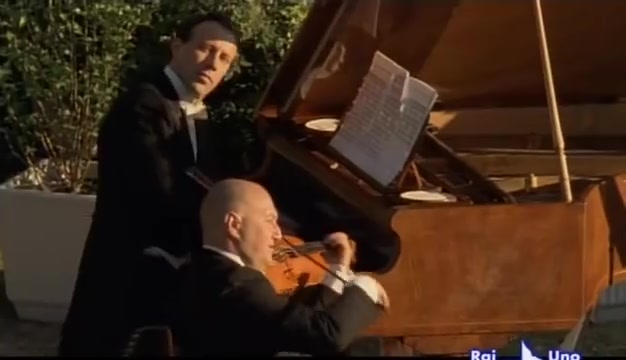}
\fullex{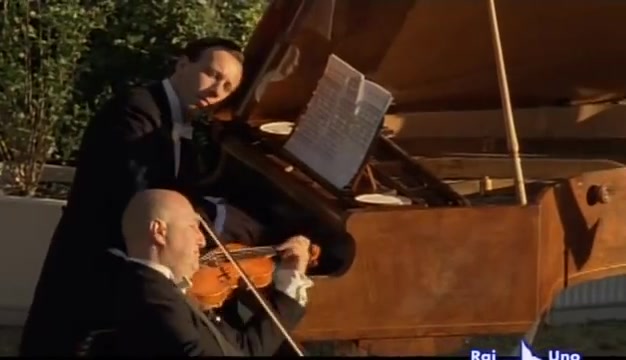}
\fullex{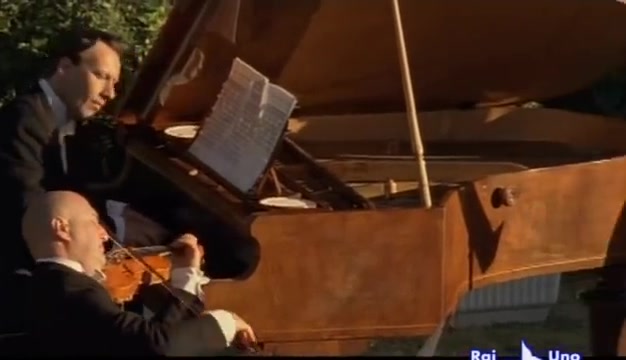}
\fullex{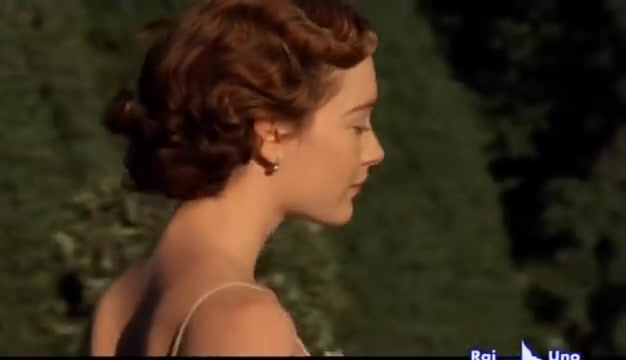}
\fullex{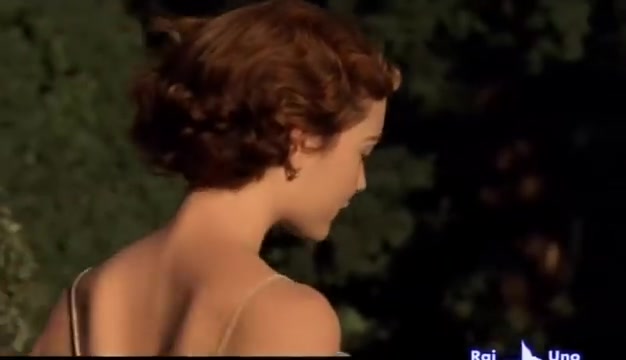}
\vspace{0.7mm}\\
\fullex{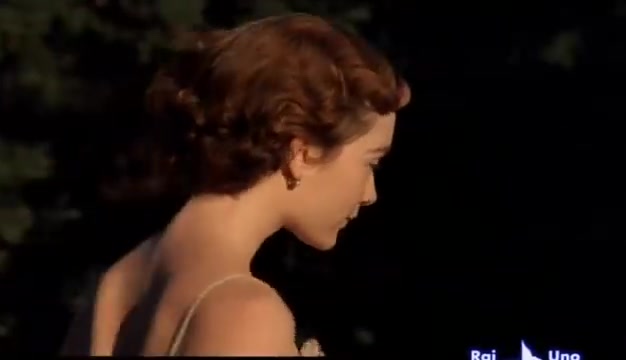}
\fullex{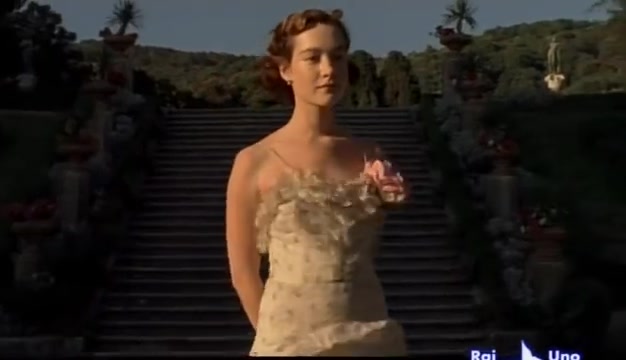}
\fullex{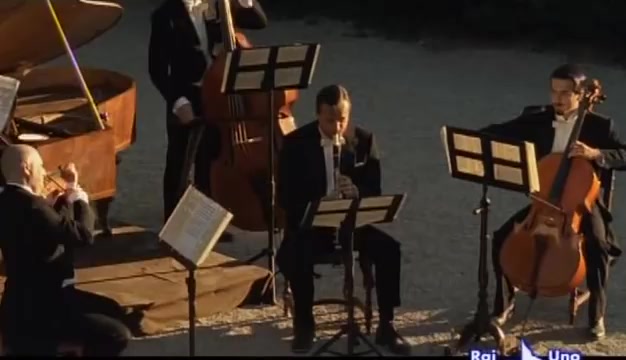}
\fullex{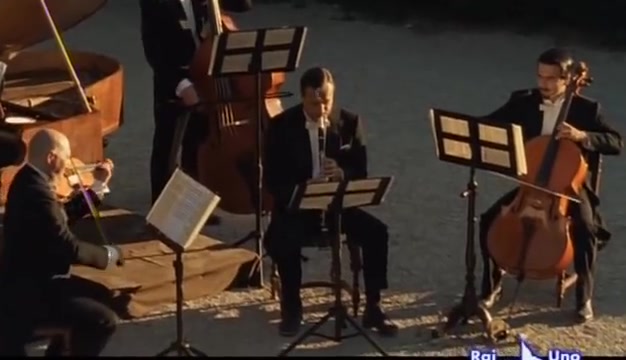}
\fullex{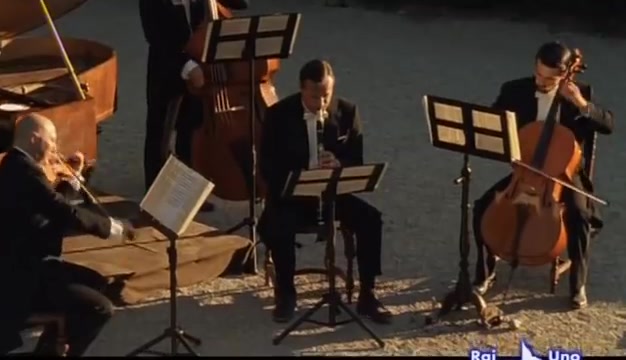}
\fullex{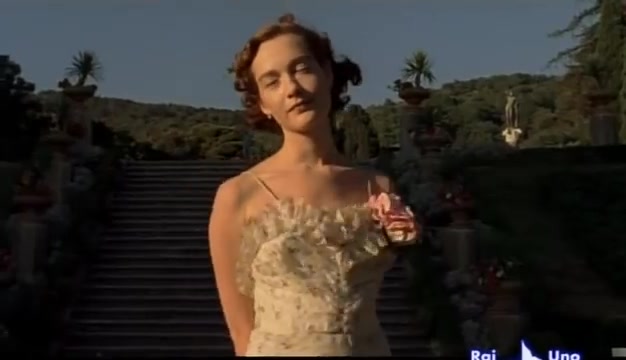}
\fullex{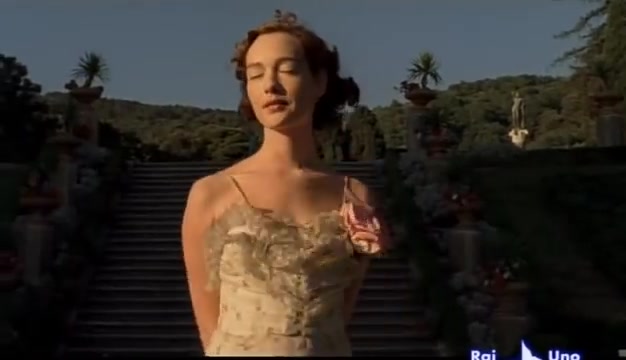}
\fullex{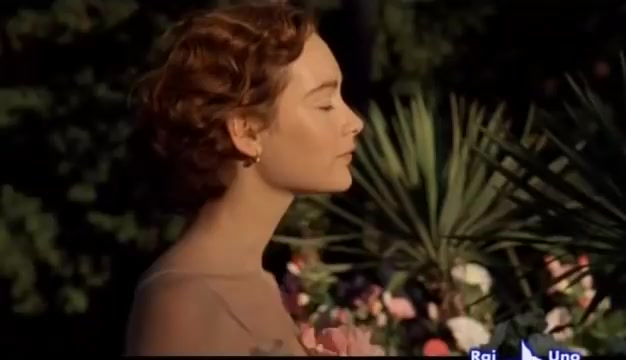}
\fullex{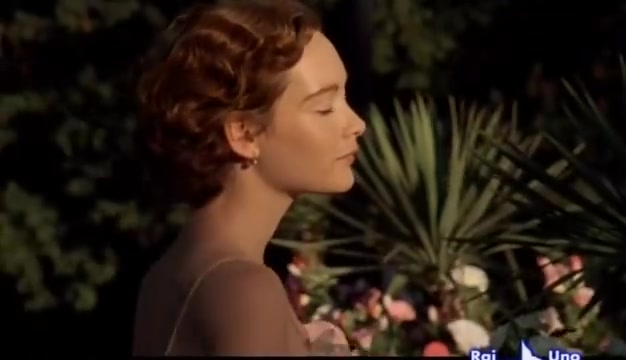}
\fullex{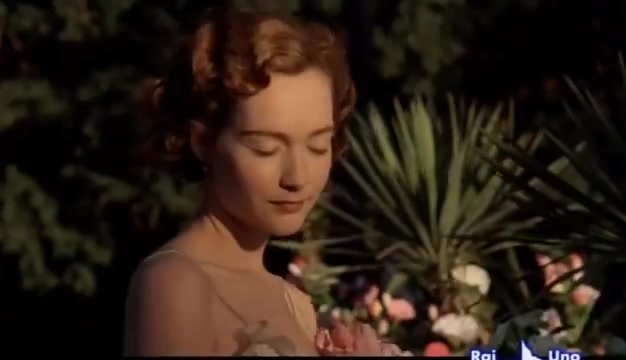}
\vspace{0.0mm}
\end{tabular}}
\vspace{0mm}
\caption{\textbf{Qualitative Evaluation Full Sequences.}
Here we present the full sequences for the three examples in Fig.\ 5 of the main paper.
From left to right and then top to bottom of each group are a full 60-second sequence.}\label{fig:fullexamples}
\end{figure*}

\end{subappendices}

\end{document}